\pdfoutput=1

\documentclass[11pt]{article}

\usepackage[preprint]{acl}

\usepackage{times}
\usepackage{latexsym}

\usepackage{booktabs}

\usepackage[T1]{fontenc}

\usepackage[utf8]{inputenc}

\usepackage{microtype}

\usepackage{inconsolata}

\usepackage{graphicx}

\usepackage{subcaption}
\usepackage{mwe}

\usepackage{multirow}
\usepackage{microtype}

\usepackage{amsfonts}
\usepackage{amsmath}
\usepackage{amssymb}

\usepackage{multirow}

\usepackage{float}

\usepackage{svg}

\usepackage{tcolorbox}

\usepackage{booktabs}
\usepackage{enumitem}
\usepackage{subcaption}
\usepackage[colorinlistoftodos]{todonotes}

\title{When Tokenization is Secretly Output Supervision}

\newcommand{\affilsup}[1]{\rlap{\textsuperscript{\normalfont#1}}}

\author{
    Tanja Baeumel\affilsup{1,2,3}
    \qquad
    Josef van Genabith\affilsup{1,2}
    \qquad
    Simon Ostermann\affilsup{1,3}
    \\ \\
    \small{
        $^1$German Research Center for AI (DFKI) \quad $^2$ Saarland University
    }
    \\
    \small{
        $^3$Center for European Research in Trusted AI (CERTAIN)
        } \\
    \footnotesize{\texttt{\href{mailto:tanja.baeumel@dfki.de}{tanja.baeumel@dfki.de}}}
}

\begin{document}
\maketitle
 
\begin{abstract}
Tokenization in language models is treated by default as an input preprocessing decision.
We argue that this framing is incomplete: in autoregressive models, tokenizer granularity determines what the model must resolve in a single forward pass, and therefore the supervision signal it receives. This affects both the difficulty of the learning problem and the representations that emerge inside the model. 
We test this in a controlled experiment on numeric reasoning with a novel decoupling of input and output tokenization. 
As the output supervision view predicts, differences in task performance, training dynamics, and model internals are induced by \textit{output} tokenization and largely invariant to \textit{input} tokenization.
This may matter in practice, because models with different tokenization strategies differ not only in input representation but in the task they were trained on. Comparisons between models may thus partly reflect task definition rather than ability. A survey of 120 recent *CL papers on numeric reasoning confirms that this is rarely acknowledged: only about 10\% report the numeric tokenization of the models they evaluate, while 69\% compare across tokenization, and thus supervision, regimes without reporting it.
While prior work documents that tokenization consistently affects model performance, there is no principled account of \textit{why}. 
We argue that framing tokenization as output supervision provides that account. 
\end{abstract}


\section{Introduction}

Tokenization is by default treated as an input preprocessing decision in autoregressive language models: a function that maps raw text to discrete tokens before the model processes them. In this position paper, we argue that this framing is incomplete and introduce a new perspective on tokenization as output supervision. We argue that the granularity of output tokens determines what a model must resolve in a single forward pass, and therefore the unit over which the training loss is computed.

The reason is straightforward: In autoregressive training, the model produces a distribution over the next token at each step, and the training loss is computed over exactly that token. 
Output token boundaries therefore define the atomic unit of supervision, and the granularity of the tokenizer determines the granularity of the learning problem the model is trained to solve. A model that must emit a coarser output token faces a fundamentally different learning problem compared to one that emits the same output at finer token granularity, even when the untokenized ground truth is shared between both learning problems. 

\begin{figure}
    \centering
    \includegraphics[width=0.9\linewidth]{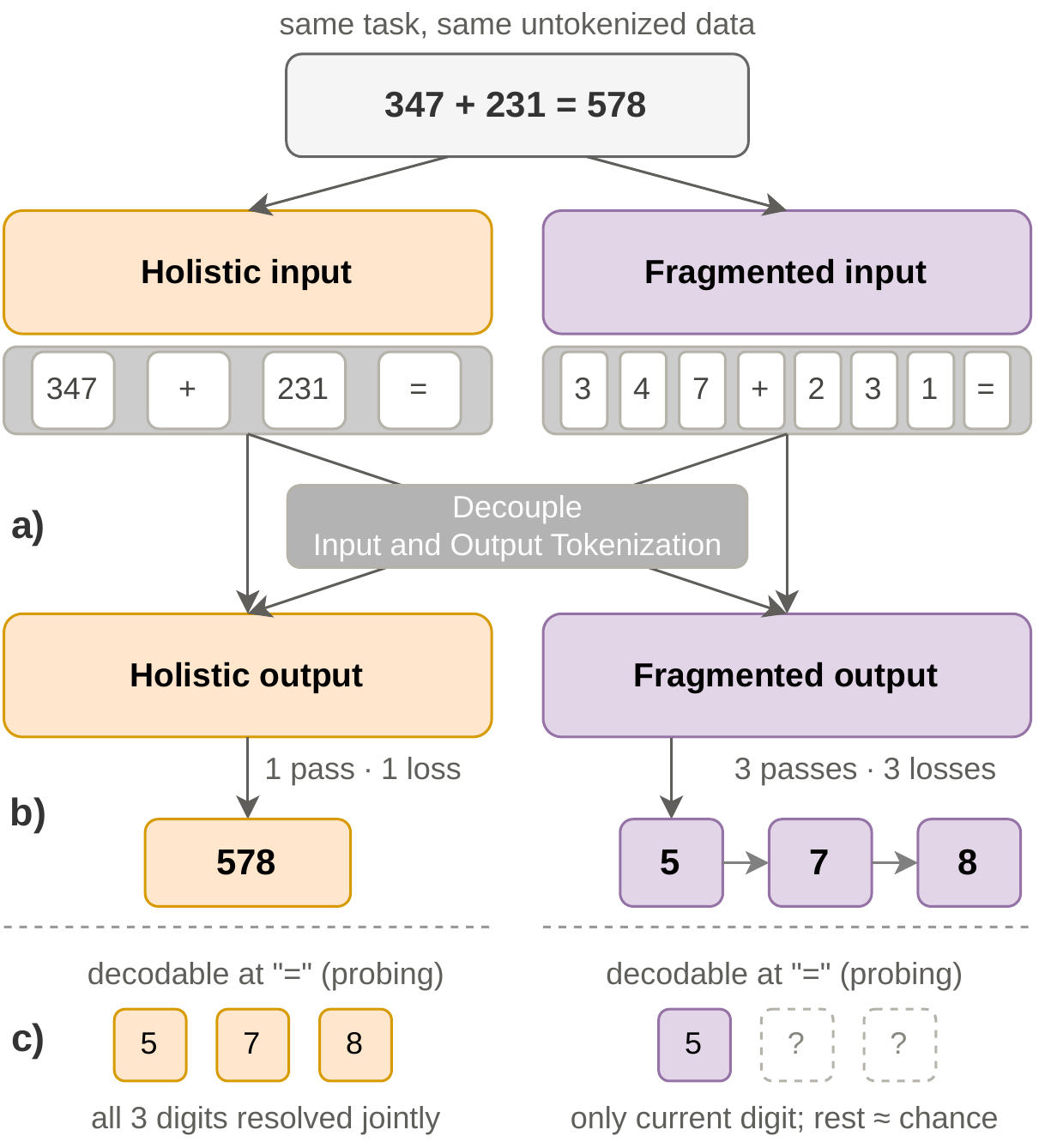}
    \caption{ We argue that tokenization in transformer models provides a supervision signal, because it determines the output granularity and thus the task to be solved per forward pass. 
    We support our argument with experimental evidence where we (a) decouple input from output tokenization to demonstrate that differences in (b) learning task difficulty and (c) model internal representations are due to the \textit{output} tokenization scheme.} 
    \label{fig:Fig1}
\end{figure}

We argue that this output supervision view is an important missing piece to understand why tokenization consistently affects model performance, which has been empirically observed across arithmetic \citep{singh2024tokenizationcountsimpacttokenization, zhou-etal-2024-scaling, zhang_tokenization_2025}, temporal reasoning \citep{bhatia-etal-2025-date}, code analysis \citep{mostafa_how_2025}, 
genomics \citep{lindsey2025impact}, and is actively discussed for morphology \citep{garcia-etal-2025-exploring, dang-etal-2025-tokenization, arnett-bergen-2025-language}.
Throughout, tokenization is largely viewed as a matter of input representation, and the supervision perspective has not been examined directly.

Interestingly, in supervised learning, the granularity of the prediction target is uncontroversially an important task design decision: a model trained to predict word-level labels and one trained to predict character-level labels are not solving the same problem \cite{sogaard-goldberg-2016-deep}, and models trained on coarser targets learn coarser, less discriminative representations than those trained on fine-grained ones \cite{chen-etal-2019-label}.

In this paper we develop a conceptual framework for tokenization as output supervision in autoregressive decoder models, and show that tokenizer granularity shapes task performance, training dynamics, and model-internal representations. We build the argument in three steps, using arithmetic as a controlled testbed. \textbf{First}, we replicate the well-established finding that tokenization affects task performance and training dynamics, and show, via a novel decoupling of input and output tokenization, that this effect is driven by the output side. Models sharing an output tokenization regime behave alike regardless of how their inputs are encoded. 
\textbf{Second}, we ask whether output supervision also shapes what models internally represent. We propose the minimal computation hypothesis: models resolve what their output supervision requires, and are under no direct pressure to resolve more.
Probing confirms this, but also exposes a tension. A task can require future information that the loss at that position never rewards, and we find models partially resolve such information anyway. We discuss whether this gap between what training rewards and what a task requires may be part of why some learning problems are much harder than they look.
\textbf{Third}, we examine what this means in practice. Production LLMs differ in tokenization granularity, so comparing them means comparing models trained under different output supervision regimes. As a case study we survey 120 recent *CL papers on numeric reasoning and find that fewer than 10\% report numeric tokenization granularity, while 69\% compare models across regimes without noting it. Conclusions attributed to numeric reasoning ability may thus partly reflect differences in task definition.
\section{A New Perspective: Tokenization Defines the Task to Solve}

We develop our argument through the lens of tokenization of integers, which provides a well-defined testbed. The theoretical argument generalizes to any domain where tokens have internal compositional structure, as we discuss in Section \ref{sec:discussion}.

We distinguish two idealised tokenization regimes. Under \textit{holistic} tokenization,  a multi-component unit (such as a multi-digit integer $578$ (or a morphologically complex word like \texttt{unbelievable}\footnote{The same holistic/fragmented distinction applies wherever tokens have internal compositional structure, e.g., date expressions (\texttt{2024-03-15}), code identifiers (\texttt{getUserById}), genomic k-mers (\texttt{ATCGG}), and agglutinative morphology (Turkish \texttt{evlerinizden}, "from your houses"). In each case a tokenizer may treat the unit as a single token or decompose it into sub-components. We discuss this in Section \ref{sec:discussion}.})) is represented as a single token ([578]), while under \textit{fragmented} tokenization, the same multi-component unit is decomposed into multiple tokens ([5, 7, 8]), according to its sub-components (here, digits, or in the case of [un, believ, able] into morphemes)\footnote{
By fragmented tokenization we mean dividing compositional structure into sub-components, not a synonym for character-level tokenization.}.
In practice, \textit{holistic} tokenization is an idealization, and what matters is the granularity of fragmentation, i.e., how coarsely a unit is segmented relative to its compositional structure.
 
\subsection{Output Tokens as Units of Computation}
\label{subsec:concept}
Most language models are trained autoregressively, producing a distribution over the next token position at each step, i.e., the supervision signal at one time step is exactly one token. Whatever computation is necessary to produce the correct output must be completed within a single forward pass.

\begin{tcolorbox}[
  colframe=black,
  coltitle=white,
  colbacktitle=black,
  boxrule=0.8pt,
]
\textbf{Key claim}: In autoregressive models, output tokenization determines supervision granularity and thereby shapes training dynamics and internal representations.
\end{tcolorbox}

Consider a running example of three-digit addition: $347+231=578$. A model with \emph{holistic} numeric tokenization, as exemplified by Llama~3, Pythia, and OlMo~2 \cite{grattafiori2024llama3herdmodels, biderman2023pythiasuiteanalyzinglarge, olmo20242olmo2furious}, which encode multi-digit integers as single tokens, receives four input tokens ([347, +, 231, =]) and must emit one output token ([578]). A model with \emph{fragmented} numeric tokenization, exemplified by Qwen~2, Mistral, and Gemma~2 \cite{yang2024qwen2technicalreport, jiang2023mistral7b, gemmateam2024gemma2improvingopen}, which tokenize numbers digit-by-digit, receives eight input tokens ([3, 4, 7, +, 2, 3, 1, =]) and emits three output tokens across three successive forward passes ([5, 7, 8]). While the abstract task is identical, the learning problem that is to be solved is fundamentally different.

Under \textit{holistic} tokenization, the model must jointly determine all three result digits in a single forward pass. 
The learning signal thus rewards correctness of the complete token $578$ and penalizes any error in it, with no signal about which digit contributed to the failure.
Under \textit{fragmented} tokenization, the model only needs to output the first result digit on its first pass; the remaining 
result digits can be deferred, and computed with previously generated digits available as context. 
This encourages a step-wise computation, building the output one position at a time with position-specific learning signals at each step\footnote{For our experiment, we adopt little-endian digit order (see Section \ref{subsec:setup} for justification).}.

This difference in granularity has a direct consequence for the structure of the supervision, while not affecting the overall probability of producing the correct result. Restricted to valid result tokens, a random baseline scores $1/10$ per forward pass under fragmented tokenization versus $1/1000$ under holistic tokenization. End-to-end the two are matched: emitting the full result at random succeeds with probability $(1/10)^3 = 1/1000$ in both regimes. What differs is how that difficulty is distributed across the supervision signal. Holistic supervision computes a single loss over the atomic token $578$, so an error carries no information about which digit was responsible and awards no partial credit for the digits that were correct. Fragmented supervision decomposes the same target into three 10-way predictions, each with its own loss, giving position-specific credit assignment and an objective that can be satisfied one digit at a time. The two conditions are therefore not two presentations of the same problem, but structurally different learning problems that happen to share a ground truth in the untokenized data.

We do not claim that output tokenization is the sole determinant of what a model learns. Architecture, training data, optimizer, and model scale all shape learned representations. Our claim is narrower: output token granularity defines a constraint on what the model must resolve per forward pass, and this constraint leaves a detectable signature in task performance and internal representations that has been systematically overlooked.

\subsection{The Minimal Computation Hypothesis}
\label{subsec:mincompH}
The output supervision view of tokenization makes a concrete prediction about internal representations. At a given position, the loss supervises exactly one token. Under fragmented supervision, it therefore rewards the current output sub-component and nothing beyond it. 
For example, for an input [3, 4, 7, +, 2, 3, 1, =] producing [5] is rewarded, while there is no gradient reward (nor penalty) for already `knowing' subsequent result digits $7$ and $8$. 
We hypothesize that models resolve what their output supervision demands and are under no direct pressure to resolve more, and call this the \textit{minimal computation hypothesis}.

This hypothesis is a claim about what training rewards, and not necessarily about what a model is capable of representing. Where the task itself demands more than the loss rewards at a given position, an interesting tension arises; we return to this in Section~\ref{sec:gradient-vs-task}.

\paragraph{Predictions.} The minimal computation hypothesis makes two concrete, falsifiable predictions about internal representations of autoregressive decoder models.
First, under fragmented supervision, only the currently required output sub-component (result digit $5$ in our running example) should be `internally generated' by the model and therefore decodable from intermediate representations; subsequent sub-components (result digits $7$ and $8$) should remain at chance level when probed. Second, under holistic supervision, all sub-components (result digits $5$, $7$, and $8$ in our running example) must be resolved simultaneously as a single token \texttt{578} because all of them are required for the token to be generated, and should therefore all be decodable from the residual stream. 
\section{Controlled Evidence for the Output Supervision View}

We present experimental evidence for our claim that tokenization is output supervision in three steps: Using a minimal example, we show that tokenization affects task performance and training dynamics (Section~\ref{subsec:experiment}). We then show through a novel decoupling of input and output tokenization that this effect is driven primarily by the output side (Section~\ref{sec:experiment}). 
Finally, we use probing classifiers to test the minimal computation hypothesis (Section~\ref{sec:outputtokenizationshapes}).

\begin{figure*}[t]
    \centering
    \begin{tabular}{c cc}
        & \textbf{Train Loss} & \textbf{Eval Accuracy}\\[0.5em]
        \parbox[c]{1em}{{\rotatebox{90}{\textbf{$\mathrm{M_F}$}}}} &
        \begin{subfigure}[c]{0.38\linewidth}
            \includegraphics[width=\linewidth]{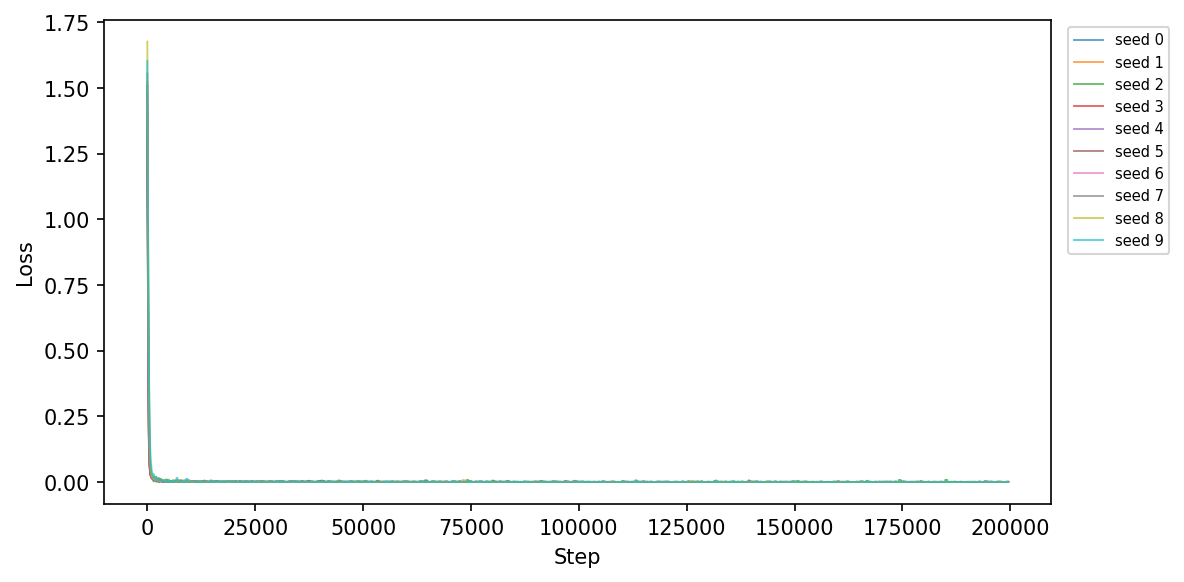}
        \end{subfigure}
        &
        \begin{subfigure}[c]{0.38\linewidth}
            \includegraphics[width=\linewidth]{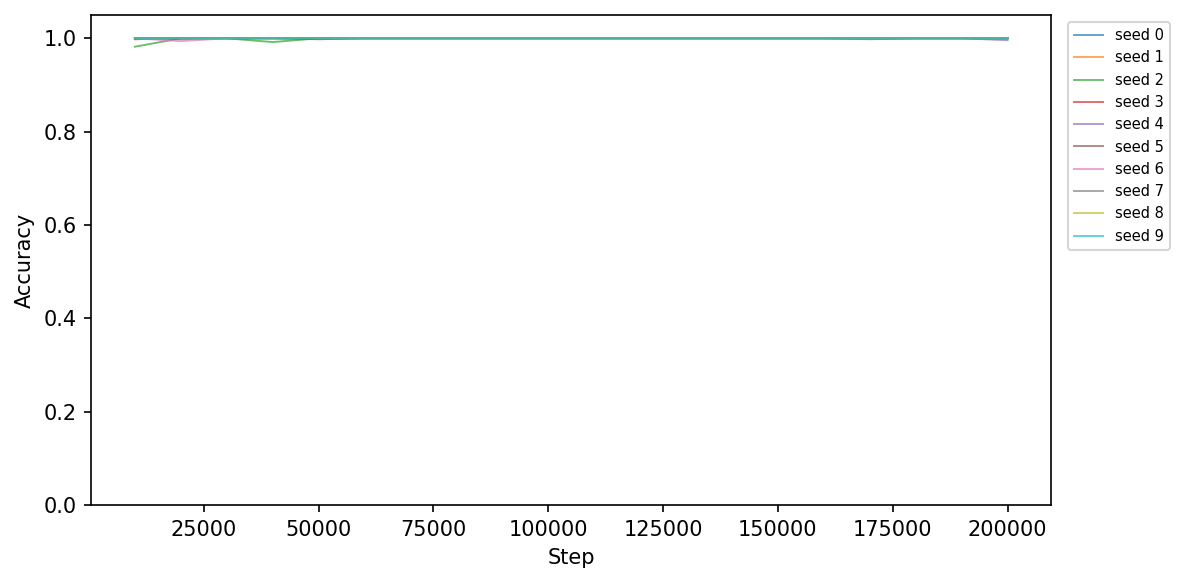}
        \end{subfigure}
        \\[1em]
        \parbox[c]{1em}{{\rotatebox{90}{\textbf{$\mathrm{M_H}$}}}} &
        \begin{subfigure}[c]{0.38\linewidth}
            \includegraphics[width=\linewidth]{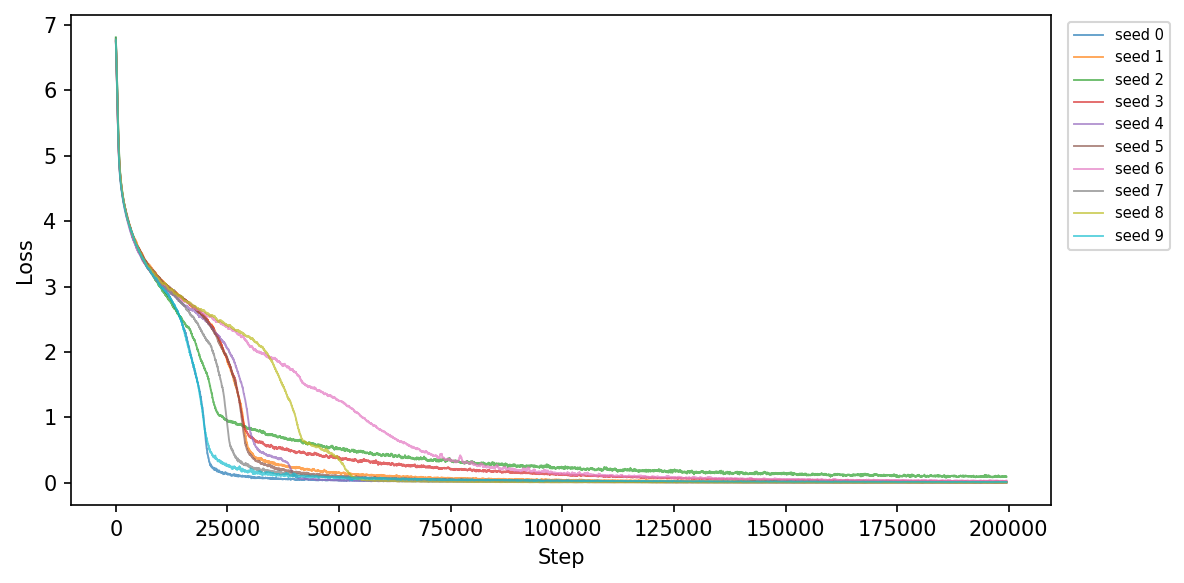}
        \end{subfigure}
        &
        \begin{subfigure}[c]{0.38\linewidth}
            \includegraphics[width=\linewidth]{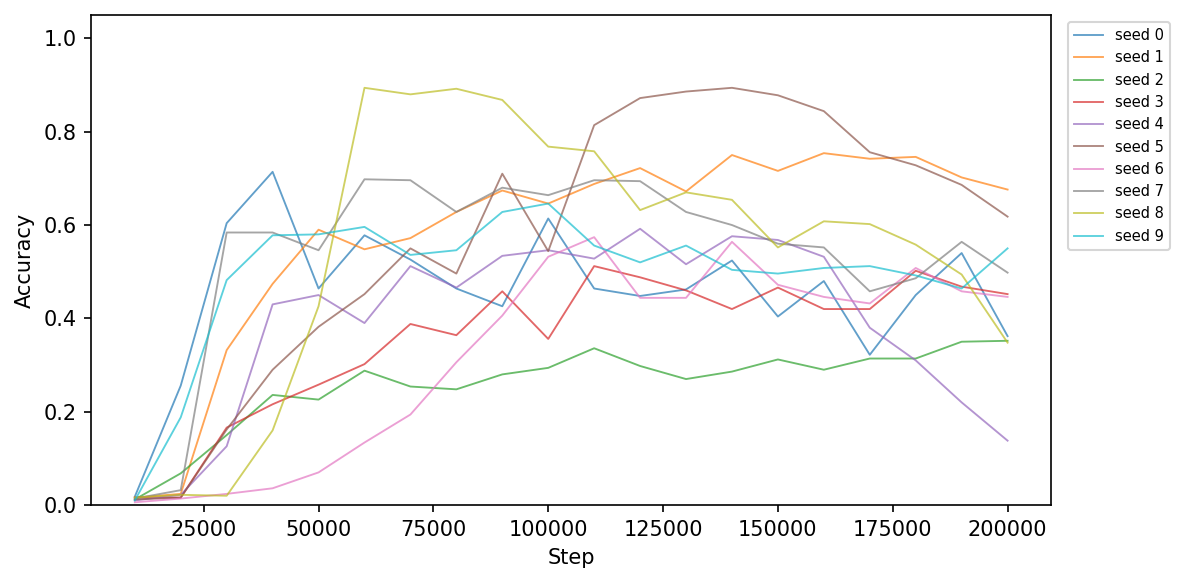}
        \end{subfigure}
        \\[1em]
    \end{tabular}
    \caption{Training Loss and Evaluation Accuracy across seeds for the best hyperparameter setting ($\mathrm{lr:=0.0001, wd:=0.01}$) of $\mathrm{M_F}$ and $\mathrm{M_H}$. $\mathrm{M_F}$ reliably reaches ceiling accuracy and converges fast. $\mathrm{M_H}$ does not converge reliably across seeds and reaches lower accuracy. Results for all hyperparameter settings in Appendix \ref{app:loss}.}
    \label{fig:loss_Mf_Mh}
\end{figure*}

\subsection{Experimental Setup, Architecture, Data and Training}
\label{subsec:setup}
We train small decoder-only transformer models on 3-digit integer addition (e.g., $347 + 231 = 578$) from scratch, 
identical in architecture (4~layers, $d_{\text{model}} = 256$, 4~attention heads), training data, and optimizer, differing only in how numbers are tokenized. All models share a joint vocabulary of 1003 tokens: the integers 0–999, plus +, =, and PAD. The tokenization conditions differ only in which subset of integer tokens they use: fragmented models represent each digit as its own token, drawing only from {0,...,9}; holistic models represent each three-digit number as a single token, drawing from the full range {0,...,999}\footnote{This shared vocabulary is what makes the decoupled conditions in Section \ref{sec:experiment} possible: Input and output tokenization can be varied independently without any architectural changes.}.
Models are trained for 200{,}000 steps with a batch size of~256, across 10~random seeds and a grid of learning rates and weight decays. Full architecture and training details are provided in Appendix~\ref{app:experiment}.

We adopt little-endian digit order, encoding numbers least-significant digit first (e.g., $347 + 231 = 578$ is reversed and becomes $743 + 132 = 875$). 
This makes addition strictly left-to-right, so each output digit can be computed without needing information from future result digits. Gradient reward and task requirement thus coincide, which makes this the cleanest case for the argument we develop.

\subsection{Tokenization Affects Task Performance}
\label{subsec:experiment}
We train two models $\mathrm{M_F}$ with \textit{fragmented} tokenization and $\mathrm{M_H}$ with \textit{holistic} tokenization. 
Figure~\ref{fig:loss_Mf_Mh} shows training loss and evaluation accuracy for the best hyperparameter setting 
across seeds. $\mathrm{M_F}$ converges reliably and fast across all seeds; $\mathrm{M_H}$ does not converge reliably and achieves substantially lower accuracy: Predicting the whole number in one go is apparently a much harder task. Results across all hyperparameter settings are provided in Appendix~\ref{app:loss}.

This replicates the well-established finding that tokenization affects task performance. Viewed through the lens of output supervision, the reason is already visible in the task structure: as predicted in Section \ref{subsec:concept}, the two tokenization regimes differ dramatically in the difficulty of the task they introduce, as generating the whole 3-digit integer at once is much harder than generating it digit by digit. Importantly, we do not take this as evidence that fragmented tokenization is superior; our goal is to isolate what drives the difference.

\subsection{\textit{Output} Tokenization Affects Task Performance}
\label{sec:experiment}

\begin{figure*}[t]
    \centering
    \begin{tabular}{c cc}
        & \textbf{Train Loss} & \textbf{Eval Accuracy}\\[0.5em]
        \parbox[c]{1em}{{\rotatebox{90}{\textbf{$\mathrm{M_{H_{\mathrm{in}},\, F_{\mathrm{out}}}}$}}}} &
        \begin{subfigure}[c]{0.38\linewidth}
            \includegraphics[width=\linewidth]{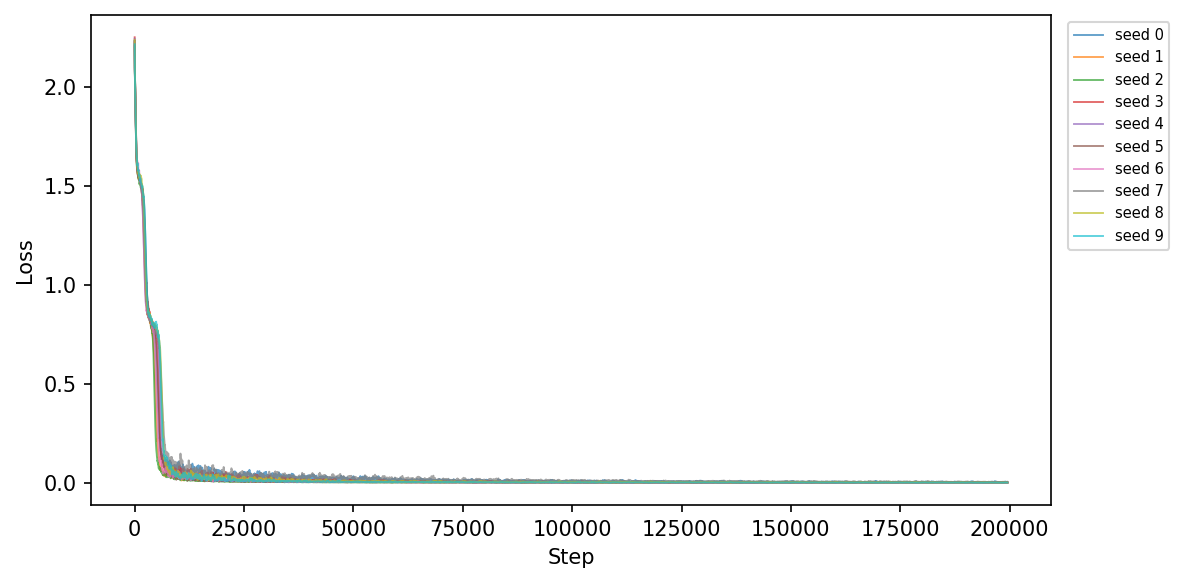}
        \end{subfigure}
        &
        \begin{subfigure}[c]{0.38\linewidth}
            \includegraphics[width=\linewidth]{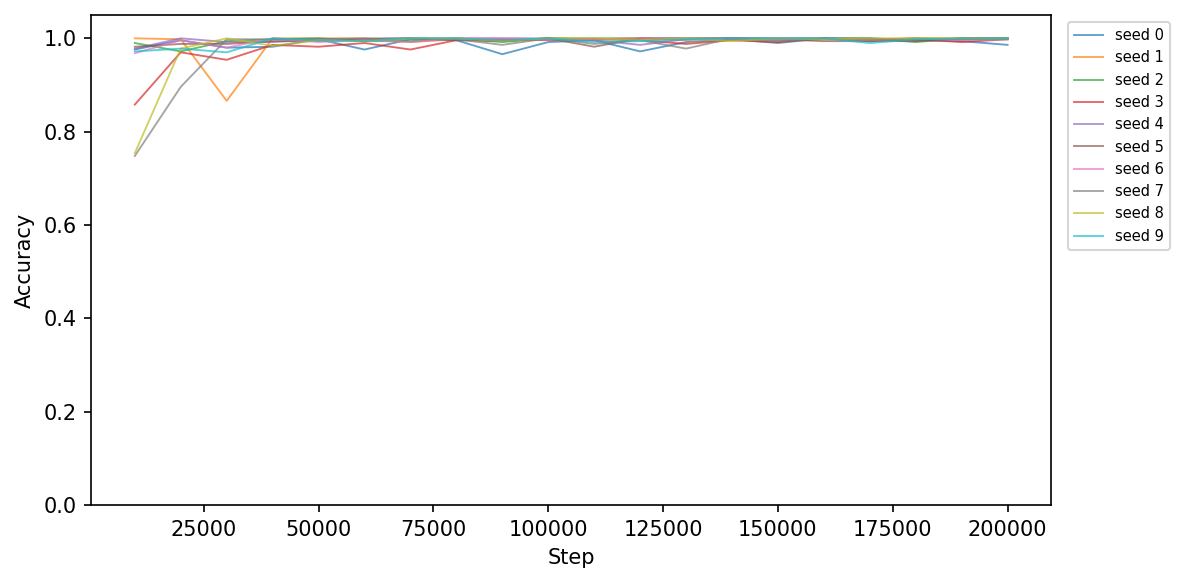}
        \end{subfigure}
        \\[1em]
        \parbox[c]{1em}{{\rotatebox{90}{\textbf{$\mathrm{M_{F_{\mathrm{in}},\, H_{\mathrm{out}}}}$}}}} &
        \begin{subfigure}[c]{0.38\linewidth}
            \includegraphics[width=\linewidth]{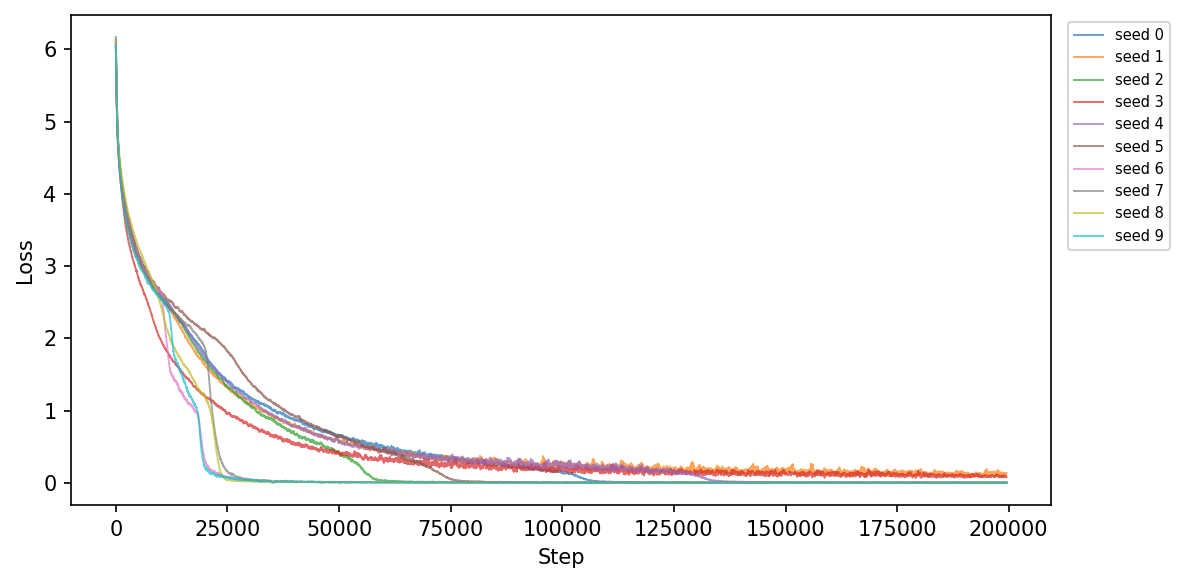}
        \end{subfigure}
        &
        \begin{subfigure}[c]{0.38\linewidth}
            \includegraphics[width=\linewidth]{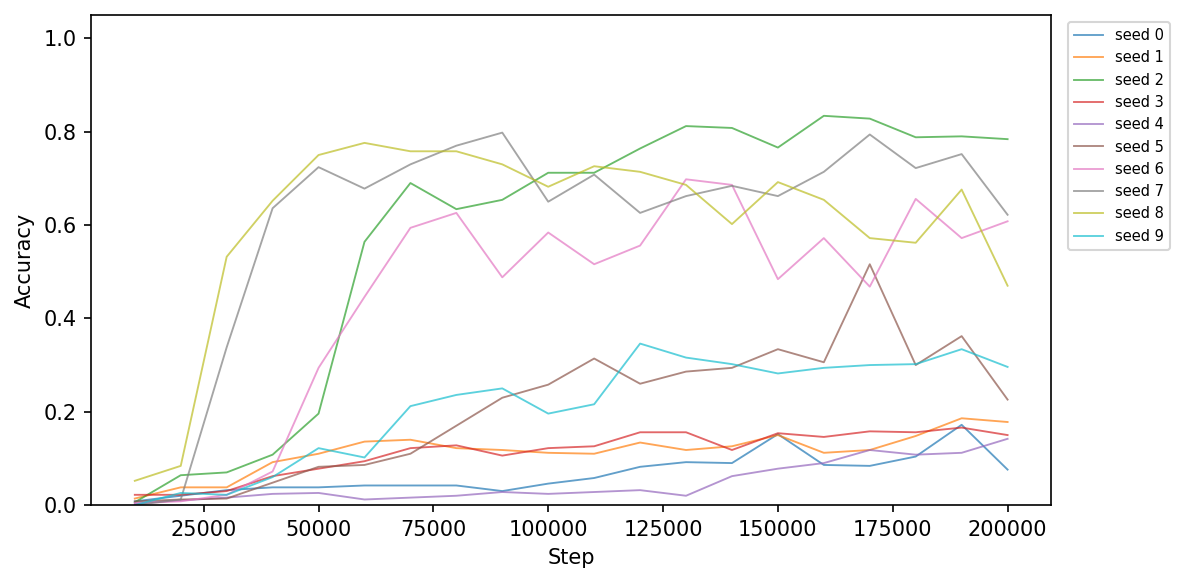}
        \end{subfigure}
        \\[1em]
    \end{tabular}
    \caption{Training Loss and Evaluation Accuracy across seeds for the best hyperparameter setting ($\mathrm{lr:=0.0001, wd:=0.01}$) of $\mathrm{M_{H_{\mathrm{in}},\, F_{\mathrm{out}}}}$ and $\mathrm{M_{F_{\mathrm{in}},\, H_{\mathrm{out}}}}$. Training loss trajectory and evaluation accuracy are comparable between $\mathrm{M_{F_{\mathrm{in}},\, H_{\mathrm{out}}}}$ and $\mathrm{M_H}$, and between $\mathrm{M_{H_{\mathrm{in}},\, F_{\mathrm{out}}}}$ and $\mathrm{M_F}$, confirming that \textit{output} tokenization affects task performance. Results for all hyperparameter settings in Appendix \ref{app:loss}.}
    \label{fig:loss_Mfh_Mhf}
\end{figure*}

To isolate whether the performance difference is due to input representation or output supervision, we decouple input and output tokenization using a factorial design.
For that we train two additional models, $\mathrm{M_{F_{\mathrm{in}},\, H_{\mathrm{out}}}}$ with \textit{fragmented input} and \textit{holistic output} tokenization ([3,4,7,+,2,3,1,=,578]), and $\mathrm{M_{H_{\mathrm{in}},\, F_{\mathrm{out}}}}$ with \textit{holistic input} and \textit{fragmented output} tokenization ([347,+,231,=,5,7,8]).
By comparing $\mathrm{M_{F_{\mathrm{in}},\, H_{\mathrm{out}}}}$ and $\mathrm{M_{H_{\mathrm{in}},\, F_{\mathrm{out}}}}$ to $\mathrm{M_F}$ and $\mathrm{M_H}$ with respect to task performance and training dynamics, we are able to observe whether models behave more similarly if they share \textit{output} or \textit{input }tokenization regime. 
If output tokenization is the main driver for task performance, as we predict, conditions sharing an output tokenization regime should show similar task performance and training dynamics (denoted by $\simeq$) regardless of input encoding (i.e., we expect: $\mathrm{M_F} \simeq \mathrm{M_{H_{\mathrm{in}},\, F_{\mathrm{out}}}}$ and $\mathrm{M_H} \simeq \mathrm{M_{F_{\mathrm{in}},\, H_{\mathrm{out}}}}$).

The tokenization decoupling is implemented at the data level, with input operands and output results tokenized by separate functions, drawing from the shared vocabulary introduced in Section \ref{subsec:setup}. Cross-entropy loss is computed only over output tokens (i.e., all tokens after the \texttt{=} sign) so the gradient signal received during training is determined entirely by output tokenization.

Figure~\ref{fig:loss_Mfh_Mhf} shows training loss and evaluation accuracy for the best hyperparameter setting 
across seeds.
As predicted, models sharing an output tokenization regime have similar task performance and training dynamics, i.e., indeed $\mathrm{M_{F_{\mathrm{in}},\, H_{\mathrm{out}}}} \simeq \mathrm{M_H}$, and $\mathrm{M_{H_{\mathrm{in}},\, F_{\mathrm{out}}}} \simeq \mathrm{M_F}$. 
The granularity of output tokenization, regardless of how the inputs are encoded, thus determines training dynamics and task performance.

\begin{figure*}[t]
    \centering
    \begin{tabular}{c cc}
        & \textbf{Fragmented Output} & \textbf{Holistic Output} \\[0.5em]
        \parbox[c]{1em}{\rotatebox{90}{\textbf{Fragmented Input}}} &
        \begin{subfigure}[c]{0.35\linewidth}
            \includegraphics[width=\linewidth]{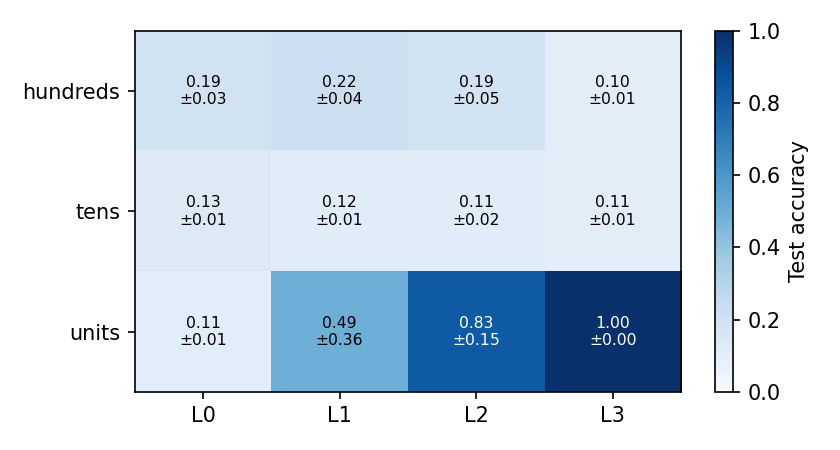 
            }
            \caption{$\mathrm{M_F}$}
        \end{subfigure}
        &
        \begin{subfigure}[c]{0.35\linewidth}
            \includegraphics[width=\linewidth]{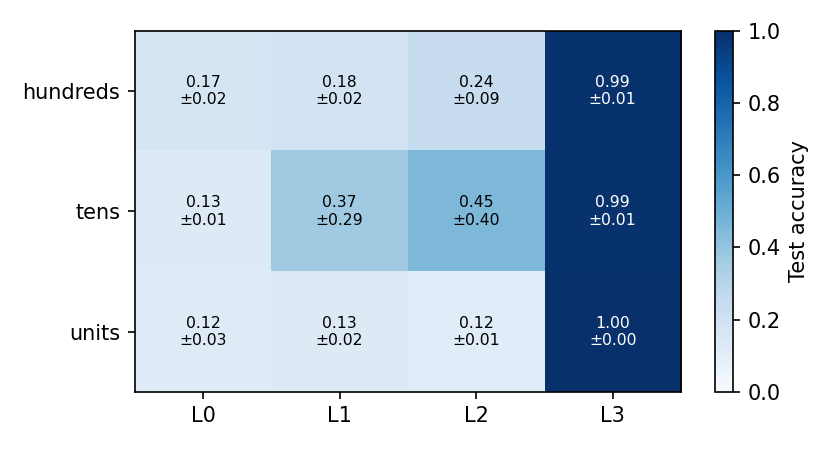 
            }
            \caption{$\mathrm{M_{F_{\mathrm{in}},\, H_{\mathrm{out}}}}$}
        \end{subfigure}
        \\[1em]
        \parbox[c]{1em}{\rotatebox{90}{\textbf{Holistic Input}}} &
        \begin{subfigure}[c]{0.35\linewidth}
            \includegraphics[width=\linewidth]{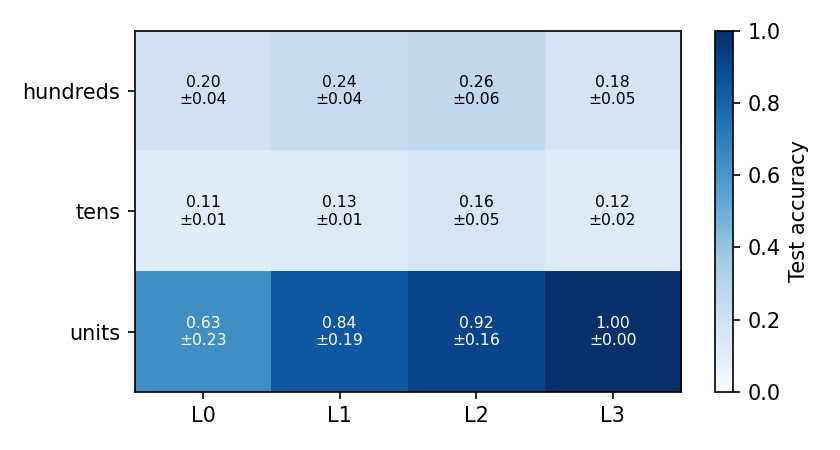 
            }
            \caption{$\mathrm{M_{H_{\mathrm{in}},\, F_{\mathrm{out}}}}$}
        \end{subfigure}
        &
        \begin{subfigure}[c]{0.35\linewidth}
            \includegraphics[width=\linewidth]{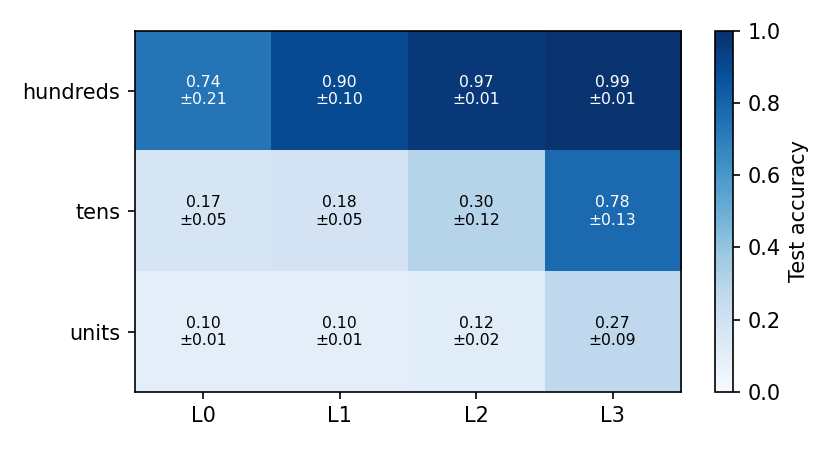 
            }
            \caption{$\mathrm{M_H}$}
        \end{subfigure}
    \end{tabular}
    \caption{Linear probe accuracy for individual result digits across layers. Values are mean probe accuracy $\pm$ standard deviation across the five probes trained per condition. At Layer~3, 
    models with holistic output tokenization simultaneously encode all three digit positions\protect\footnotemark; 
    models with fragmented output tokenization encode the first digit (unit digit, due to little endian convention), with subsequent digit positions remaining closer to chance across layers.}
    \label{fig:probing}
\end{figure*}

\subsection{\textit{Output} Tokenization Shapes Model Internals}
\label{sec:outputtokenizationshapes}

Having established that output tokenization drives task performance, we ask whether output tokenization also shapes model internal representations, as predicted by the minimal computation hypothesis. 
Specifically, a model with fragmented output tokenization is predicted to have no gradient pressure to represent future result digits in the little-endian number convention. For example, for an input "$347 + 231 =$" (irrespective of whether it is tokenized in a fragmented way [3, 4, 7, +, 2, 3, 1, =] or tokenized in a holistic way [347, +, 231, =]) the minimal computation hypothesis predicts that the model has only resolved the first result digit $5$, but not subsequent result digits $7$ and $8$. If this is the case only the result digit $5$ should be represented within the model.
In contrast, a model trained with holistic output tokenization has to generate the whole number $578$, and thus the individual result digits should be represented within the model. 

We test this prediction of result-digit availability with probing classifiers. 
Linear probes are trained on residual stream activations at the \texttt{=} token position, i.e., the point at which the model must have assembled all information necessary to begin generating output. 
For each condition we use the five checkpoints with highest task accuracy (Table~\ref{tab:eval_results}), and train separate probes for each digit position of the result (units, tens, hundreds) at each layer.
Details and results of MLP probes and circular probes \cite{nanda2023progress}, which tell the same qualitative story, as well as details on the selected model checkpoints are provided in Appendix~\ref{app:probes}.

\begin{table}[t]
\centering
\small
\begin{tabular}{lcccc}
\toprule
Condition & Exact Match & Hundreds & Tens & Units \\
\midrule
$\mathrm{M_F}$ & 1.000 & 1.000 & 1.000 & 1.000 \\
 & \tiny{$\pm$.000} & \tiny{$\pm$.000} & \tiny{$\pm$.000} & \tiny{$\pm$.000} \\
$\mathrm{M_{F_{in},H_{out}}}$ & 0.989 & 0.997 & 0.993 & 0.998 \\
 & \tiny{$\pm$.012} & \tiny{$\pm$.004} & \tiny{$\pm$.012} & \tiny{$\pm$.002} \\
$\mathrm{M_{H_{in},F_{out}}}$ & 0.986 & 0.999 & 0.994 & 0.992 \\
 & \tiny{$\pm$.009} & \tiny{$\pm$.001} & \tiny{$\pm$.006} & \tiny{$\pm$.007} \\
$\mathrm{M_H}$ & 0.853 & 0.964 & 0.922 & 0.857 \\
 & \tiny{$\pm$.050} & \tiny{$\pm$.012} & \tiny{$\pm$.018} & \tiny{$\pm$.050} \\
\bottomrule
\end{tabular}
\caption{Evaluation accuracy on a held-out random test set (500 samples),
mean $\pm$ standard deviation across the model checkpoints probes are trained on. Per-digit accuracy reports the fraction of examples where each digit position of the result is correct.}
\label{tab:eval_results}
\end{table}

Figure~\ref{fig:probing} shows linear probe accuracy for each digit position across layers.
Under fragmented output tokenization ($\mathrm{M_F}$, $\mathrm{M_{H_{\mathrm{in}},\, F_{\mathrm{out}}}}$), only the first result digit (units) is decodable. The tens and hundreds digits remain near chance at every layer. This is what the minimal computation hypothesis predicts: the model does not develop representations for output sub-components it is not yet supervised to produce. Under holistic output tokenization ($\mathrm{M_H}$, $\mathrm{M_{F_{\mathrm{in}},\, H_{\mathrm{out}}}}$), all three result digits become available by the last layer. In one fell swoop, the model is forced to jointly resolve the full result in a single forward pass, and its internal representations reflect this from the earliest layers.
Crucially, this pattern tracks the \emph{output} axis of the factorial design, not the input axis ($\mathrm{M_F} \simeq \mathrm{M_{H_{\mathrm{in}},\, F_{\mathrm{out}}}}$, $\mathrm{M_H} \simeq \mathrm{M_{F_{\mathrm{in}},\, H_{\mathrm{out}}}}$), i.e., what the model internally resolves is determined by output tokenization. Fragmented output tokenization induces a genuinely sequential task. The model does not merely emit digits one at a time, it computes them one at a time, not representing future output components until the supervision signal demands them. Holistic output tokenization induces the opposite, a parallel task in which all sub-components must be jointly resolved from the start. 
\section{Tokenization Goes Unreported: A Survey of Reporting Practice}
\label{sec:survey}

\subsection{Motivation}
We have established that tokenization defines the supervision regime a model is trained under. Production LLMs differ systematically in numeric tokenization granularity.
Comparing such models on arithmetic benchmarks therefore inevitably means comparing models trained under different output supervision regimes even if they were trained on comparable (prior to tokenization) data. This section documents that this is widely unacknowledged in the *CL community.

\subsection{Method}
We extract all papers published to the main or findings tracks of ACL, EMNLP, NAACL, and EACL in 2024 and 2025 whose titles contain \textit{math}, \textit{numer} (such as in \textit{numer}\texttt{ic}), \textit{number}, or \textit{arithmetic}, manually filtering for topical fit (excluding e.g.\ task arithmetic, social numeracy). This retrieves 120 papers on mathematical or numeric reasoning. For each, we search the full text for the substring \textit{token} and manually determine whether the distinction between digit-level and multi-digit numeric tokenization is explicitly mentioned and whether cross-model comparisons account for it. Full methodology and inclusion criteria are provided in Appendix~\ref{app:survey}.
 
\subsection{Results}
Fewer than 1 in 10 papers (11/120, 9.2\%) explicitly mention the numeric tokenization strategy of the models they evaluate. Of the remaining 109, 83 actively compare models across fragmented and holistic tokenization without flagging it or controlling for supervision regime; 20 evaluate only one type without justifying the restriction; 6 are unresolvable due to closed-source or undisclosed tokenizers. 
Notably, even among the 11 papers that do mention tokenization, none frames the distinction in terms of output supervision, i.e., the consistent framing, when tokenization appears at all, is as an input representation choice.
The most consequential figure is the 83 papers that compare models across incompatible supervision regimes without flagging it, risking conflation of differences in task definition with differences in reasoning ability.

\paragraph{Implications.}
When a paper reports that model A outperforms model B on an arithmetic benchmark, and A uses holistic output tokenization while B uses fragmented output tokenization, the comparison does not support a conclusion about reasoning ability alone, but conflates reasoning ability with supervision regime.
This is not a flaw in the analysis of any individual paper, but reflects a systematic gap in reporting practice. The practical impact of this omission will vary, in some comparisons the supervision regime difference may be small, in others decisive, but it cannot be assessed without reporting. We thus propose that tokenization strategy should be reported as a standard model descriptor alongside architecture, parameter count, and training data.
More broadly, this case illustrates that benchmark comparisons between models can conflate task performance with task definition when models differ along dimensions that restructure the learning problem. Claims about progress that do not account for such structural differences risk measuring the difficulty of the learning problem as much as the capability of the model.

\footnotetext{Limited linear decodability of the units digit in $\mathrm{M_H}$ is likely a combination of limited accuracy of the models on the units digit (compare Table~\ref{tab:eval_results}) and a possible non-linearity in representation as suggested by significantly higher MLP compared to linear probe accuracy (compare Figure~\ref{fig:probing_mlp}).}
\section{Related Work}
\label{sec:related}

\paragraph{Tokenization and Task Performance.}
Subword tokenization is standard in NLP \citep{sennrich-etal-2016-neural, kudo-richardson-2018-sentencepiece} and is treated almost universally as an input preprocessing decision, including in work on what makes tokenizers effective, which examines compression \citep{goldman-etal-2024-unpacking, schmidt-etal-2024-tokenization} and vocabulary design \citep{ali_tokenizer_2024, lotz-etal-2025-beyond, chizhov-etal-2024-bpe}. \citet{he-etal-2020-dynamic} is a notable exception, treating source and target segmentation as distinct problems in machine translation, but motivates this by translation quality rather than by claims about what models learn. \citet{schmidt-etal-2024-tokenization} explicitly ask what makes tokenization effective beyond compression but do not look at the output side.

Prior work further documents that tokenization affects outcomes across arithmetic \citep{singh2024tokenizationcountsimpacttokenization, zhou-etal-2024-scaling, zhang_tokenization_2025}, temporal reasoning \citep{bhatia-etal-2025-date}, code analysis \citep{mostafa_how_2025}, genomics \citep{lindsey2025impact}, and against tokenization bias more broadly \citep{phan_understanding_2024}. 
Morphology is the contested case. Whether morphologically complex languages are harder to model is itself disputed \citep{cotterell-etal-2018-languages, gerz-etal-2018-language, park-etal-2021-morphology}, and results on tokenization's role are mixed: \citet{garcia-etal-2025-exploring} report gains from morphology-aware tokenization for Spanish, while \citet{dang-etal-2025-tokenization} find comparable average performance between subword and character-level tokenization across 17 languages, and \citet{arnett-bergen-2025-language} find no support for morphological alignment as an explanation for cross-linguistic performance gaps. \citet{poelman-etal-2025-confounding} attribute the conflicting evidence to confounds in experimental setup, a diagnosis that parallels our own. We therefore treat morphology as motivating the generality of our framing rather than as evidence for it.
 
\paragraph{Supervision Granularity in Non-Autoregressive Settings.} In supervised learning, prediction target granularity shapes learned representations: coarser labels yield coarser features \citep{chen-etal-2019-label}, and in multi-task NLP, auxiliary supervision granularity determines what representations emerge and at what depth \citep{sogaard-goldberg-2016-deep}.

\paragraph{Internal Mechanisms of Arithmetic.}
Recent work has investigated numeric tokenization effects on numeric representations and reasoning mechanisms in LLMs \cite{levy_language_2025, baeumel-etal-2025-modular, baeumel-etal-2025-lookahead}. \citet{baeumel-etal-2025-modular} find that models with coarse numeric tokenization develop three digit-position-specific processing pathways while those with digit-wise tokenization develop only one. We hypothesize that this pattern is a consequence of output supervision regime. 
\section{Discussion}
\label{sec:discussion}

\subsection{Gradient Reward versus Task Requirement: The Big-endian Case}
\label{sec:gradient-vs-task}
A natural objection to the minimal computation hypothesis is that autoregressive models are already known to represent information about tokens beyond the immediate next one \cite{NEURIPS2024_8936fa16, wu2024languagemodelsplanahead, pal-etal-2023-future}.
This is not a contradiction, but it does require a distinction about whether a task requires more than its loss rewards.
\emph{Gradient reward} is what the loss actually supervises at a given position, i.e., exactly the current output token. \emph{Task requirement} is what must in fact be resolved to produce that token correctly.

To separate gradient reward and task requirement, we repeat our experiments in big-endian digit order, where carry information propagates right to left, so the first result digit cannot be reliably predicted without resolving whether a carry arrives from digits that are supervised only later. 
Gradient reward is unchanged (one token at a time) while the task requirement now extends beyond it. We therefore expect above-chance decodability of future result digits in the fragmented output conditions ($\mathrm{M_F}$, $\mathrm{M_{H_{\mathrm{in}},\,F_{\mathrm{out}}}}$), despite the absence of gradient pressure to represent them. Training and probing details and results are in Appendix~\ref{sec:big-endian}.
Our results support this hypothesis: In fragmented output models that solve the big-endian task
with high accuracy (Table~\ref{tab:checkpoints_bigendian}), future result digits become decodable well above chance.
Relative to little-endian, final-layer linear probe accuracy for the second result digit rises from $0.11 \rightarrow 0.29$ ($\mathrm{M_F}$) and $0.12 \rightarrow 0.69$ ($\mathrm{M_{H_{\mathrm{in}},\,F_{\mathrm{out}}}}$), and for the third result digit from $0.10 \rightarrow 0.13$ and $0.18 \rightarrow 0.56$. Decodability is consistently higher for the adjacent digit than for the more distant one, consistent with a carry from the adjacent position affecting the current digit more often.

Objection and hypothesis thus simultaneously hold: gradient reward tracks only the current token, and task-forced lookahead is represented anyway. 
The pattern is in line with the \textit{breadcrumbs} hypothesis \citep{wu2024languagemodelsplanahead} which predicts that future information is represented because it is already beneficial to the current generation step. Carry information is decodable not because the model set out to represent ``the tens digit'', but because the currently-due digit cannot be computed without it.

This leaves a gap between what training rewards and what a task requires.
Anything falling into it must be acquired as a byproduct of getting the currently-rewarded token right, with no gradient insisting it be done completely or robustly. This gap is not unique to addition; any output tokenization whose granularity is misaligned with the dependencies the task requires (e.g., long-range syntactic dependencies) will face the same problem. Tokenization therefore determines not only what a model is trained to output, but which parts of the task it is ever directly accountable for. Whether the size of this gap predicts where lookahead-dependent tasks become brittle - with the systematic lookahead limitation reported for LLMs \citep{baeumel-etal-2025-lookahead} as one candidate - is a testable consequence of reading tokenization as supervision that we leave to future work.

\subsection{Does the Output Supervision View Hold at Production Scale?}
Our controlled experiment requires training from scratch, as we vary input and output tokenization independently. Production-scale LLMs share a single tokenizer across input and output and cannot be manipulated in this way. The core argument is scale-independent: autoregressive training computes loss over output tokens regardless of model size, so the supervision impact of output token granularity holds at any scale.
There is indirect evidence of tokenization granularity affecting learned task granularity in LLMs. \citet{baeumel-etal-2025-modular} find that production-scale LLMs with holistic numeric tokenization develop three digit-position-specific arithmetic circuits while those with fragmented numeric tokenization develop only one. The evidence is indirect because these models share a single tokenizer across input and output, so output supervision cannot be isolated from input representation. Direct evidence at scale remains future work.
More broadly, a substantial body of work documents that tokenization affects arithmetic performance in frontier LLMs \citep{singh2024tokenizationcountsimpacttokenization, zhou-etal-2024-scaling, zhang_tokenization_2025}. These studies treat tokenization as an input-side variable and do not isolate output supervision as the responsible factor. Our framework offers an explanation: performance differences between models with holistic and fragmented tokenization may reflect differences in the task each model was trained to solve, not just differences in input representation.

\subsection{Does the Output Supervision View Extend Beyond Arithmetic?}
Arithmetic serves as our proof of concept, but we expect the output supervision view of tokenization to extend to domains where output tokens have internal compositional structure. A model that emits a morphologically complex word as a single token faces a different learning problem than one that emits its morphemes. The same holds for date expressions in temporal reasoning and for identifiers and literals in code generation.
Prior work investigates tokenization effects in exactly these domains: date expressions as a bottleneck for temporal reasoning \citep{bhatia-etal-2025-date} and code analysis \citep{mostafa_how_2025}, where the \textit{that} is established but the \textit{why} is not. For morphology the picture is less settled (Section~\ref{sec:related}), and the studies closest to our framing evaluate encoder-only \citep{garcia-etal-2025-exploring} or encoder-decoder \citep{dang-etal-2025-tokenization} models rather than decoder-only ones.
Our framework supplies a candidate explanation: performance differences reflect differences in output supervision regime, which induce structurally different learning problems. Whether this explanation accounts for the documented effects, and whether the representational signatures we observe in arithmetic replicate in these domains, remains an open empirical question for future work.
\section{Conclusion}
Tokenization is typically treated as a preprocessing decision made once, upstream of everything else that matters. 
We argue that because autoregressive language models are trained to predict output tokens, tokenizer granularity determines the unit of supervision during training. Tokenization thus shapes not only how inputs are represented, but also what problem the model is trained to solve.

Using arithmetic as a controlled testbed, our 2x2 factorial experiment decoupling input and output tokenization shows that \textit{output} tokenization drives differences in task performance and training dynamics. 
We develop the minimal computation hypothesis, confirm it via probing, and use it to expose a tension between what training rewards and what a task requires when the task demands lookahead.
Our survey of 120 recent publications documents how widely the potential impact of tokenization regimes goes unnoticed: fewer than 10\% of papers on mathematical reasoning in LLMs report tokenization strategies, while the majority draw conclusions about reasoning ability from comparisons that risk conflating task definition with task performance.

We draw three concrete consequences for practice. \textbf{First}, output tokenization regime on task-relevant tokens (e.g., numerical tokenization for math, morphological tokenization for morphology) should be reported as a standard model descriptor alongside architecture, parameter count, and training data, echoing recent calls to treat tokenization as a modeling decision \citep{alqahtani-etal-2026-stop}. 
\textbf{Second}, capability comparisons should where possible be stratified by output supervision regime. 
\textbf{Third}, mechanistic claims about how models internally represent structured outputs are claims about a model under a specific supervision regime and should be qualified accordingly. 

The output supervision view extends rather than replaces existing perspectives on tokenization, which have focused on input representation, compression, vocabulary construction, and linguistic structure. 
Recognizing it is a prerequisite for valid cross-model comparison wherever output tokens have internal compositional structure. Given the field's reliance on benchmark comparisons, this is one place where the validity of those comparisons has been silently assumed rather than established.


\section*{Limitations}

Our experiment is limited to three-digit addition with small models trained from
scratch. Though we argue that the core argument is scale-independent, we cannot predict the magnitude of the effect of different tokenization strategies on model performance in practice. In the works studied in our survey the confound introduced by cross-tokenizer comparison may be very small in practice in some cases, and meaningful in other work. This is not something we can estimate.

Our argument also concerns standard next-token-prediction training, where the loss
is computed over exactly one output token per position. Post-training objectives
that supply intermediate or sequence-level signal (process supervision, preference
optimization over full solutions, RL with outcome rewards), and inference
protocols that let a model externalize intermediate results before committing to
an answer, change what a model is effectively supervised to resolve per forward
pass. How output token granularity interacts with these regimes is beyond the
scope of our experiment.

Probing classifiers measure decodability, not presence or absence of information.
Our results establish that future digit positions are not geometrically accessible
in fragmented models, not that the information is absent. We report MLP probes
alongside linear probes to partially address this: if a non-linear probe cannot
decode a signal, it is less likely to be present than if only a linear probe
fails.

Finally, the survey relies on keyword search over paper titles and manual annotation by the authors, both of which may introduce errors.

\section*{Acknowledgements}
We thank Gerrit Großmann and Christian Schuler for their helpful feedback on the paper draft. We further thank Amelie Seyfried for assisting with the tokenization annotation in the literature survey.
This research was supported by the German Federal Ministry of Research, Technology and Space (BMFTR) as part of the project TRAILS (01IW24005). 
AI assistance was used to improve the clarity and fluency of the writing, to help refine phrasing and structure, and to support exploratory literature search and organization. All scientific claims, interpretations, and conclusions remain the responsibility of the authors.
We further thank the anonymous reviewers for their helpful comments.

\bibliography{NumericTokenization} 

\appendix
\section{Experimental Setup}
\label{app:experiment}

\subsection{Task and Data}
We train models on three-digit integer, two-operand addition: given operands $x, y \in [0, 999]$ with $x + y \leq 999$, predict the result $z = x + y$. Training problems are sampled by drawing $z \sim \text{Uniform}(0, 999)$, then $x \sim \text{Uniform}(0, z)$, with $y = z - x$, ensuring uniform coverage of the result space. A fixed held-out evaluation set of 500 problems randomly sampled from the distribution once is excluded from training throughout. All integers are zero-padded to three digits before tokenization.

\subsection{Tokenization Conditions}
We vary input and output tokenization independently, yielding four conditions in a $2 \times 2$ factorial design. Under \textit{fragmented} number tokenization (F), each digit is a separate token from $\{0, \ldots, 9\}$. Under \textit{holistic} number tokenization (H), each operand or result is encoded as a single token from $\{0, \ldots, 999\}$. This yields four conditions: FF (fragmented-in, fragmented-out), FH (fragmented-in, holistic-out), HF (holistic-in, fragmented-out), and HH (holistic-in, holistic-out). All conditions share a joint vocabulary of 1003 tokens: integers 0--999 plus special tokens for \texttt{+} (index 1000), \texttt{=} (index 1001), and \texttt{PAD} (index 1002).

\subsection{Model Architecture}
All models are decoder-only transformers trained from scratch. The architecture is identical across all four conditions: 4 layers, model dimension $d_\text{model} = 256$, 4 attention heads ($d_\text{head} = 64$), MLP dimension $d_\text{mlp} = 1024$, ReLU activations, no layer normalization. The embedding and unembedding matrices share the joint vocabulary of 1003 tokens. This yields approximately 3.4M parameters.
Models are trained with teacher forcing; cross-entropy loss is computed only over output token positions using a binary loss mask.

\subsection{Optimization and Compute}
All models are trained with AdamW \citep{loshchilov-adam} with $\beta_1 = 0.9$, $\beta_2 = 0.98$, batch size 256, and a linear learning rate warmup over the first 10 steps. Training runs for up to 200{,}000 steps. 
We train 10 seeds per condition and report results averaged across seeds. The seeds are $193$, $237$, $378$, $416$, $598$, $623$, $705$, $891$, $974$, $273$.

We train 4 input-output tokenizer conditions x 3 learning rates x 6 weight decays x 10 seeds = 720 models in total. Each run trains for 200,000 steps and takes approximately 30 minutes, giving a total training budget of roughly 360 GPU hours. Experiments were run on Nvidia RTX 3090, RTX A6000, and H100 GPUs.

\section{Training Loss and Evaluation Curves}
\label{app:loss}

\begin{figure*}[t]
    \centering
    \begin{tabular}{c ccc}
        & \textbf{$lr=0.0001$} & \textbf{$lr=0.001$} & \textbf{$lr=0.003$}\\[0.5em]
        \parbox[c]{1em}{\rotatebox{90}{\textbf{$wd=0.01$}}} &
        \begin{subfigure}[c]{0.29\linewidth}
            \includegraphics[width=\linewidth]{Fig/loss_le/HH/HH_lr0.0001_wd0.01_train_loss_linear.png}
        \end{subfigure}
        &
        \begin{subfigure}[c]{0.29\linewidth}
            \includegraphics[width=\linewidth]{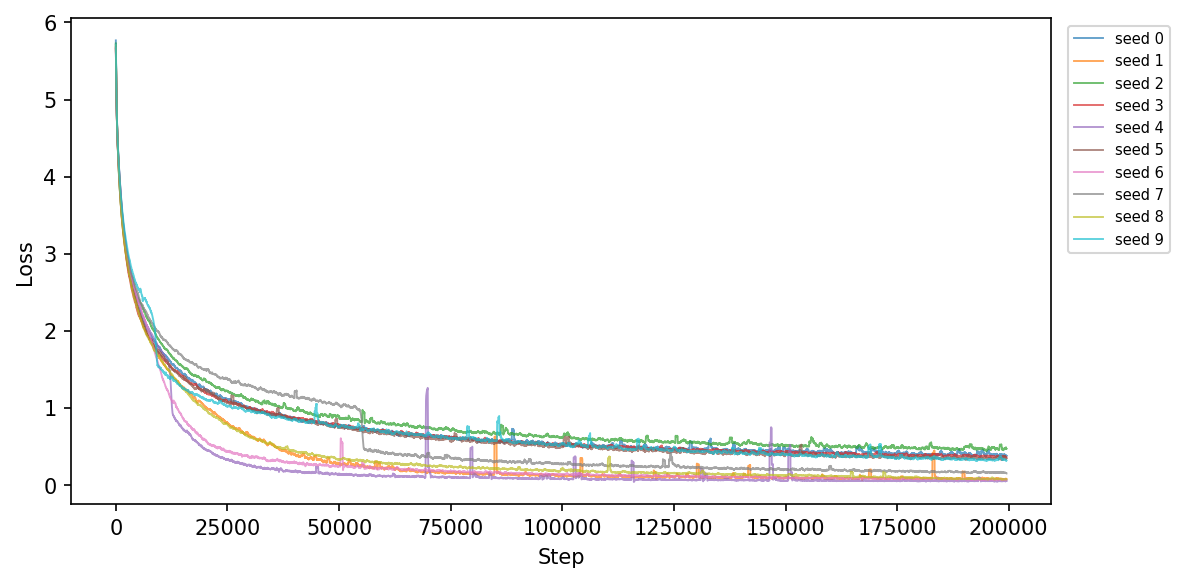}
        \end{subfigure}
        &
        \begin{subfigure}[c]{0.29\linewidth}
            \includegraphics[width=\linewidth]{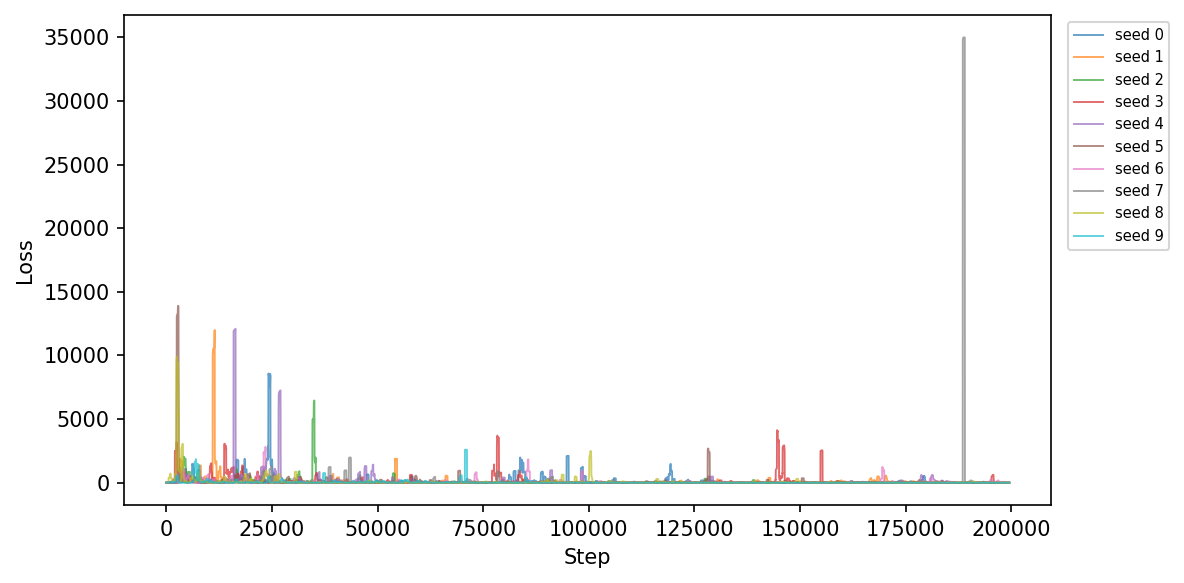}
        \end{subfigure}
        \\[1em]
        \parbox[c]{1em}{\rotatebox{90}{\textbf{$wd=0.1$}}} &
        \begin{subfigure}[c]{0.29\linewidth}
            \includegraphics[width=\linewidth]{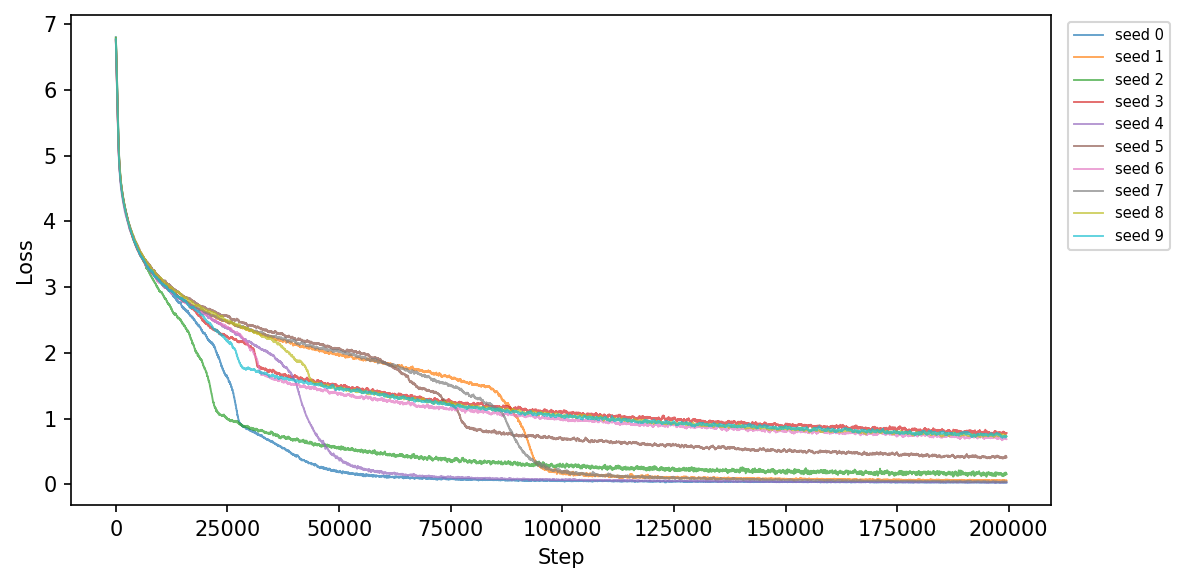}
        \end{subfigure}
        &
        \begin{subfigure}[c]{0.29\linewidth}
            \includegraphics[width=\linewidth]{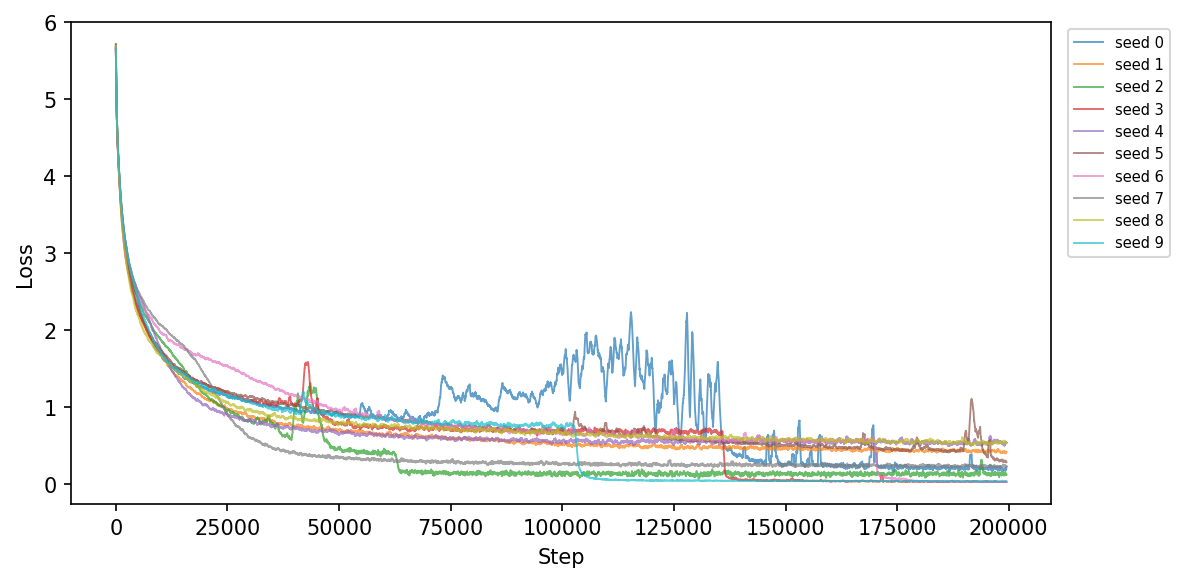}
        \end{subfigure}
        &
        \begin{subfigure}[c]{0.29\linewidth}
            \includegraphics[width=\linewidth]{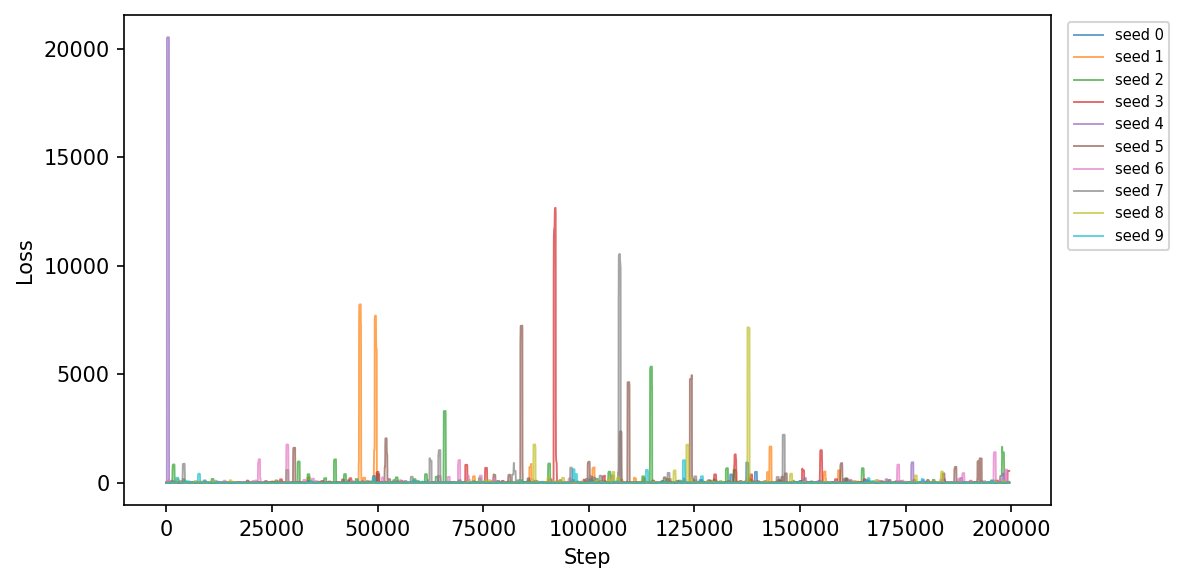}
        \end{subfigure}
        \\[1em]
        \parbox[c]{1em}{\rotatebox{90}{\textbf{$wd=0.2$}}} &
        \begin{subfigure}[c]{0.29\linewidth}
            \includegraphics[width=\linewidth]{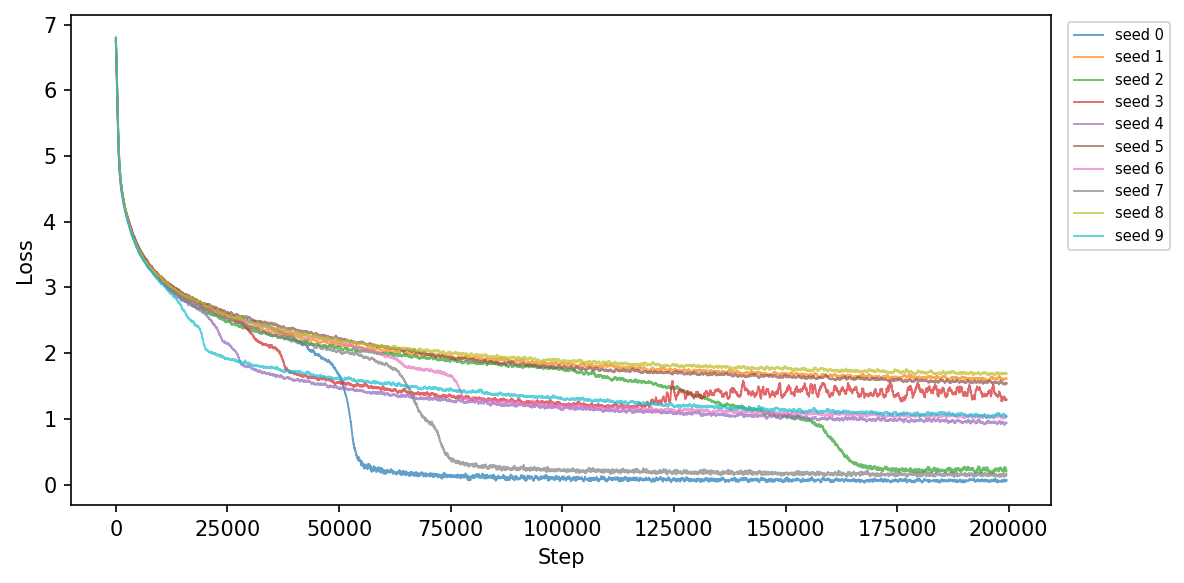}
        \end{subfigure}
        &
        \begin{subfigure}[c]{0.29\linewidth}
            \includegraphics[width=\linewidth]{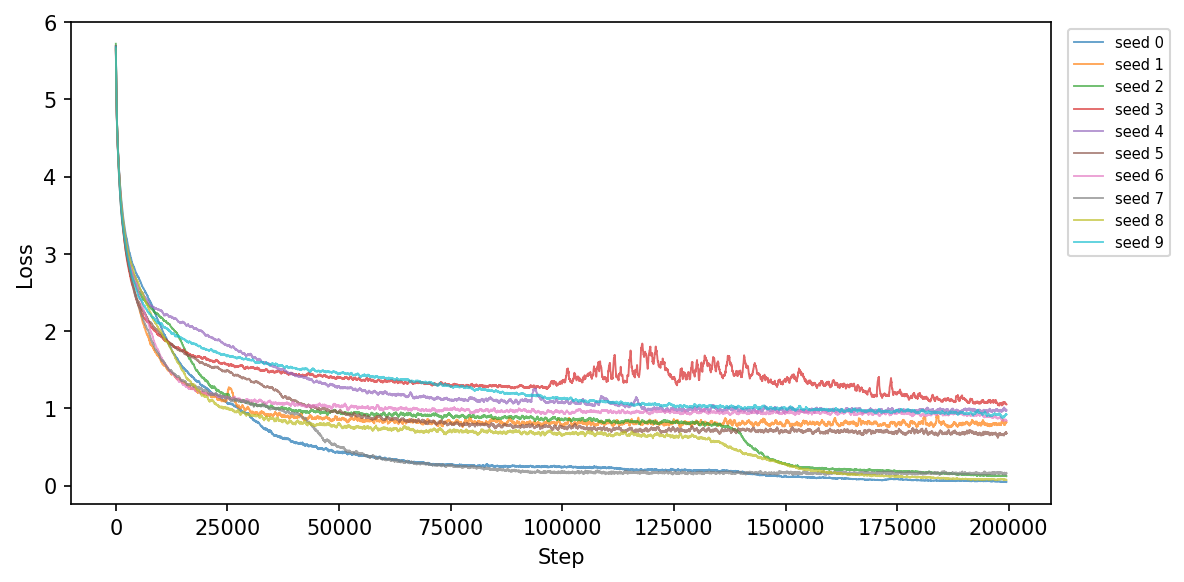}
        \end{subfigure}
        &
        \begin{subfigure}[c]{0.29\linewidth}
            \includegraphics[width=\linewidth]{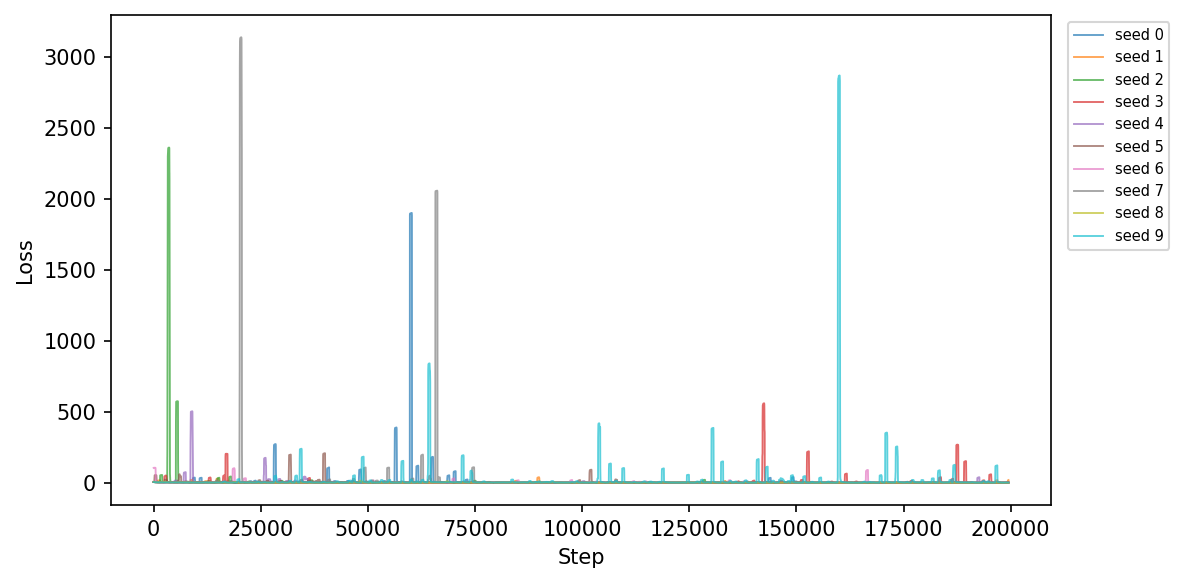}
        \end{subfigure}
        \\[1em]
        \parbox[c]{1em}{\rotatebox{90}{\textbf{$wd=0.3$}}} &
        \begin{subfigure}[c]{0.29\linewidth}
            \includegraphics[width=\linewidth]{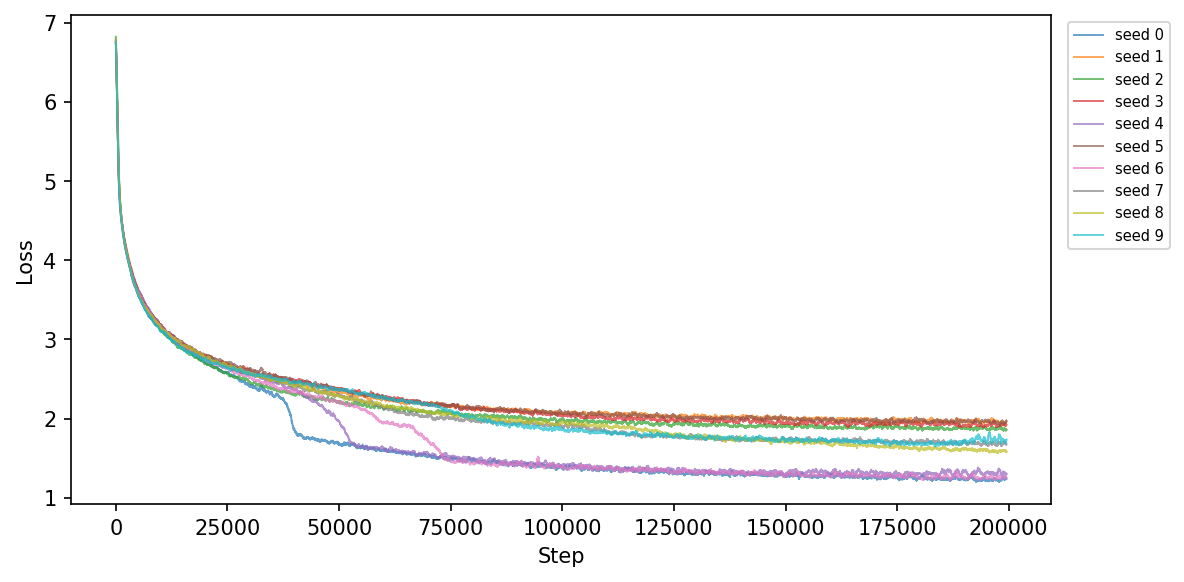}
        \end{subfigure}
        &
        \begin{subfigure}[c]{0.29\linewidth}
            \includegraphics[width=\linewidth]{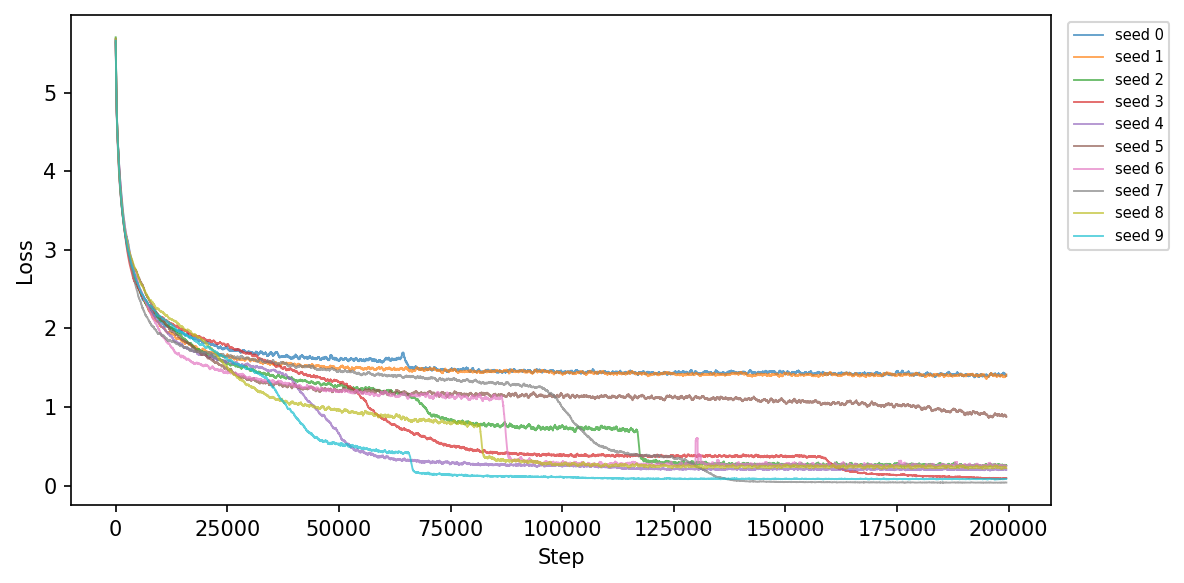}
        \end{subfigure}
        &
        \begin{subfigure}[c]{0.29\linewidth}
            \includegraphics[width=\linewidth]{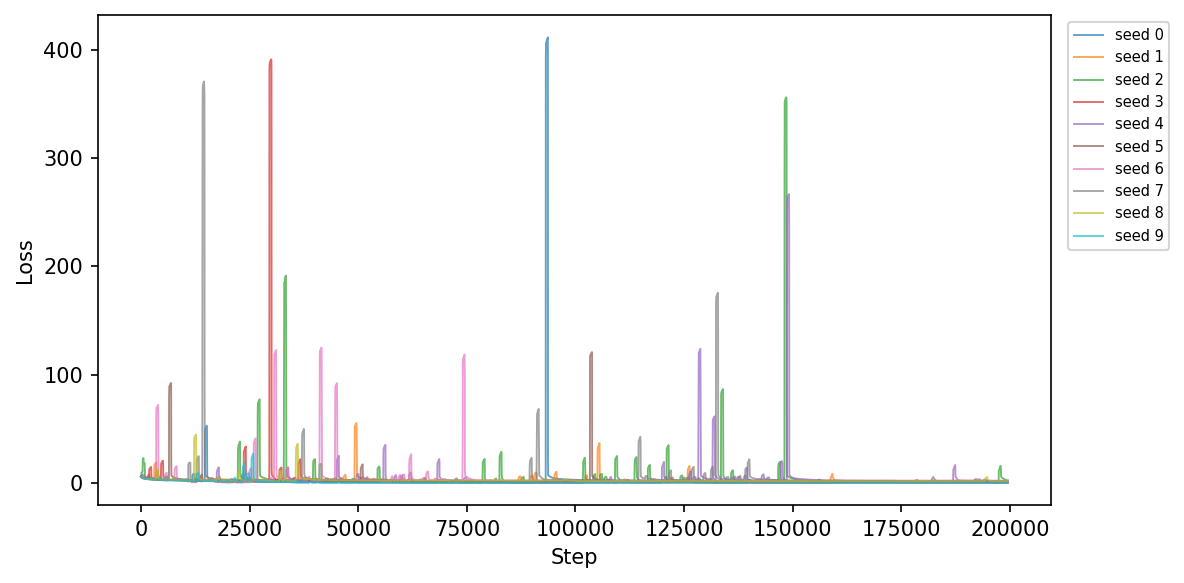}
        \end{subfigure}
        \\[1em]
        \parbox[c]{1em}{\rotatebox{90}{\textbf{$wd=0.4$}}} &
        \begin{subfigure}[c]{0.29\linewidth}
            \includegraphics[width=\linewidth]{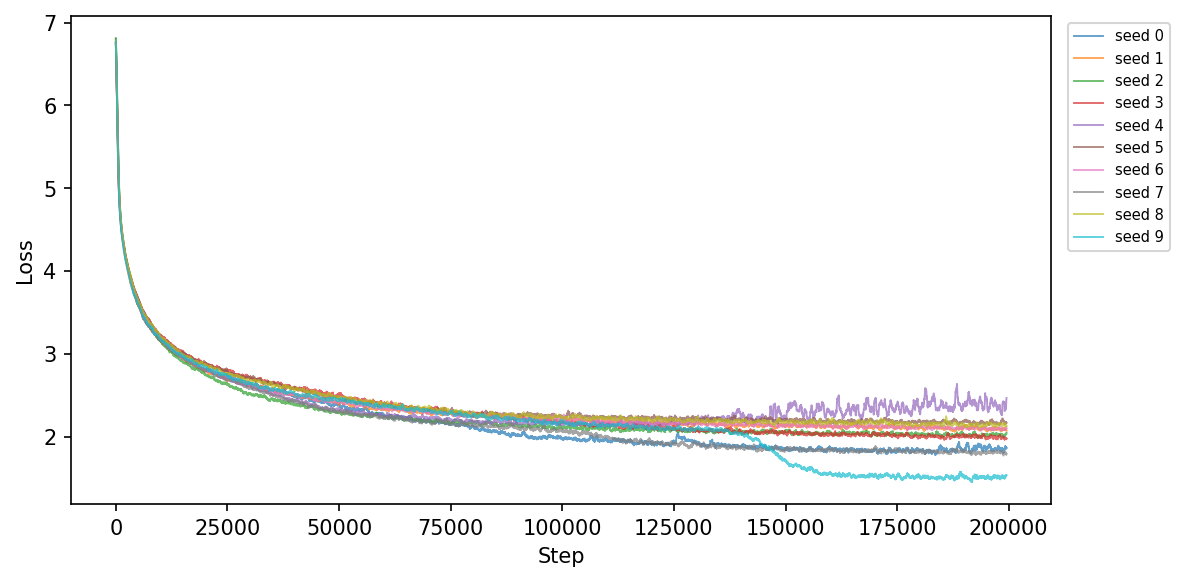}
        \end{subfigure}
        &
        \begin{subfigure}[c]{0.29\linewidth}
            \includegraphics[width=\linewidth]{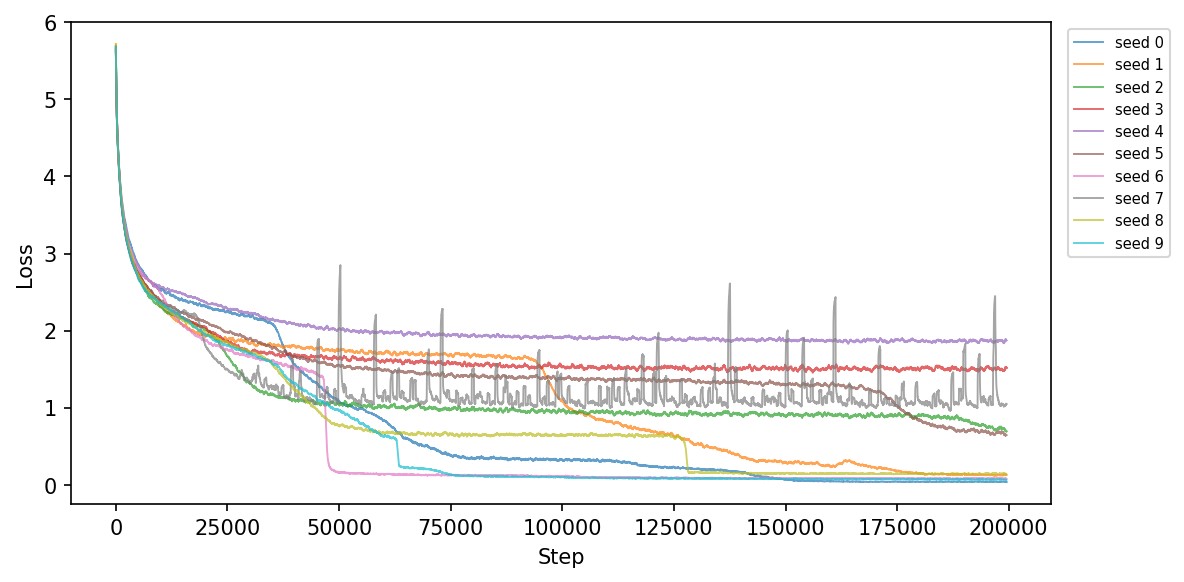}
        \end{subfigure}
        &
        \begin{subfigure}[c]{0.29\linewidth}
            \includegraphics[width=\linewidth]{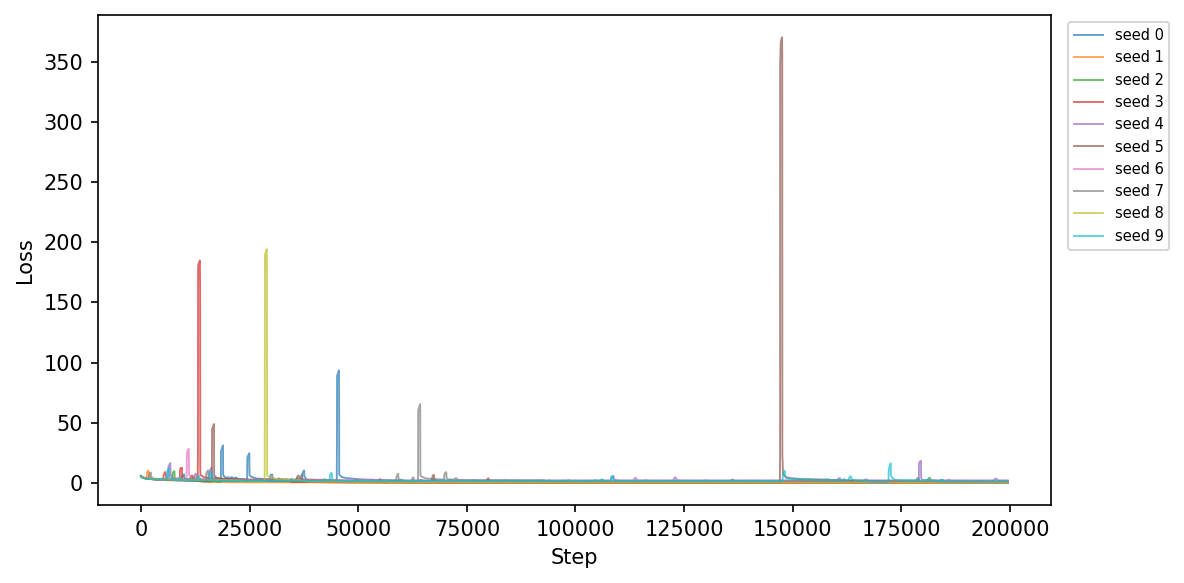}
        \end{subfigure}
        \\[1em]
        \parbox[c]{1em}{\rotatebox{90}{\textbf{$wd=0.5$}}} &
        \begin{subfigure}[c]{0.29\linewidth}
            \includegraphics[width=\linewidth]{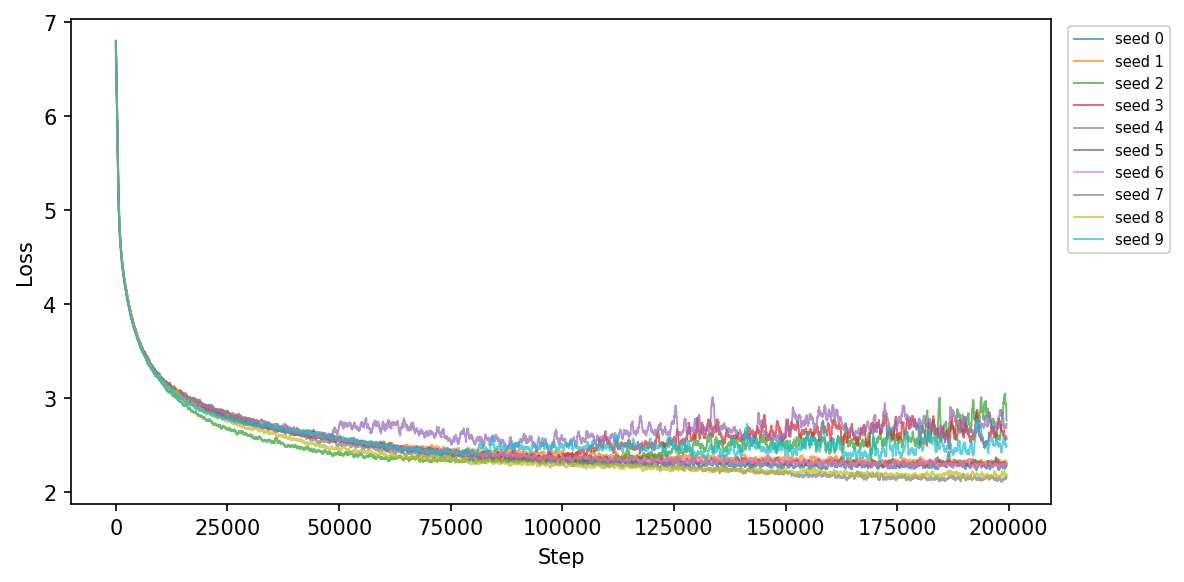}
        \end{subfigure}
        &
        \begin{subfigure}[c]{0.29\linewidth}
            \includegraphics[width=\linewidth]{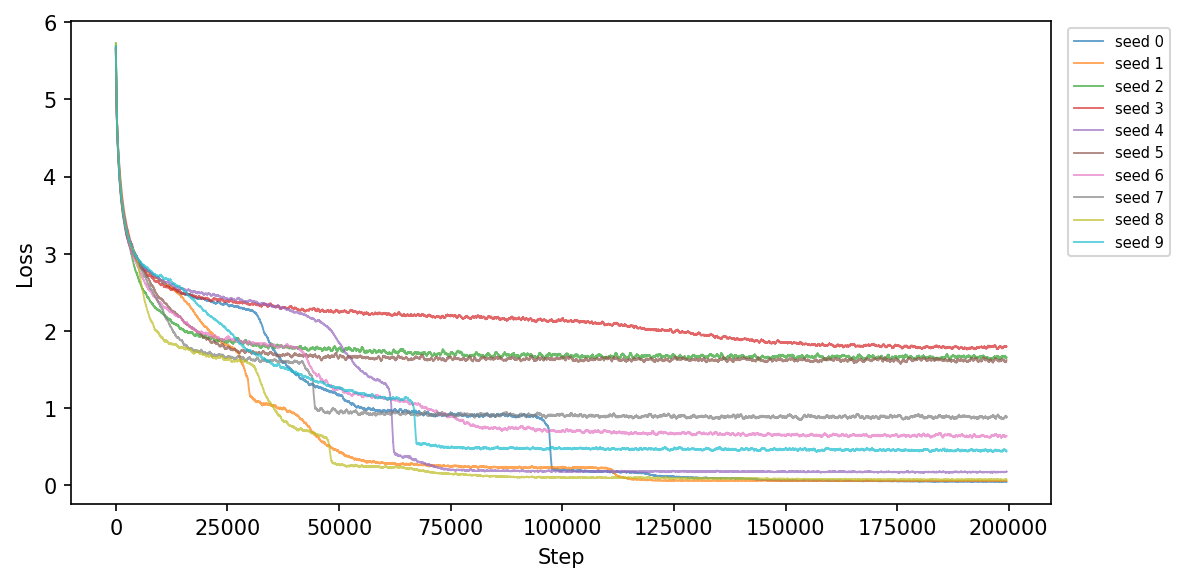}
        \end{subfigure}
        &
        \begin{subfigure}[c]{0.29\linewidth}
            \includegraphics[width=\linewidth]{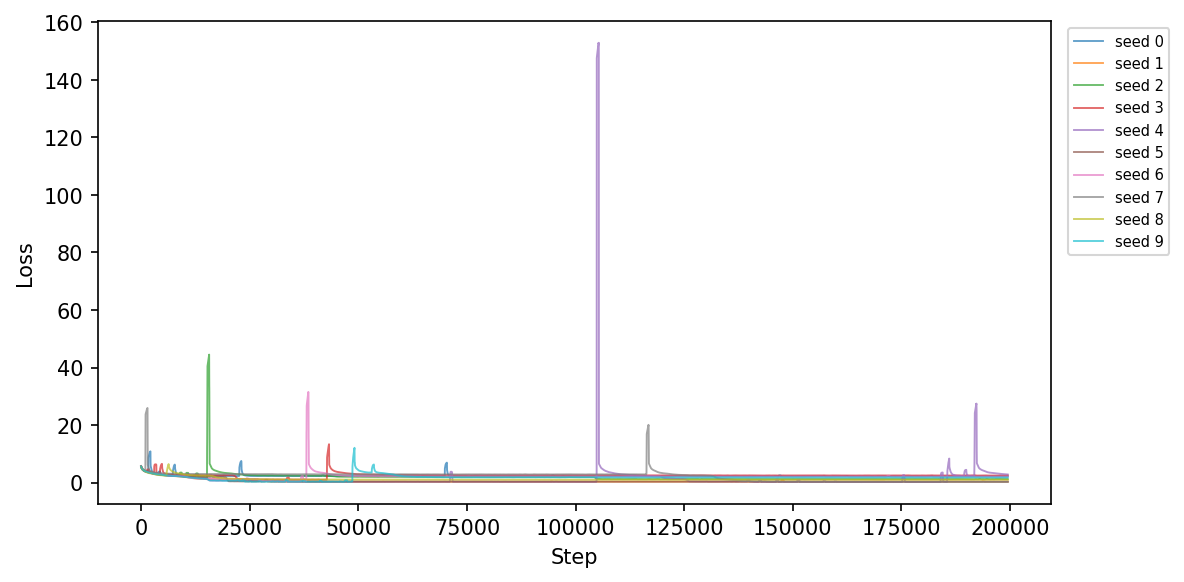}
        \end{subfigure}
        \\[1em]
    \end{tabular}
    \caption{$\mathrm{M_H}$ - Train Loss across epochs, on different hyperparameters}
    \label{fig:loss_HH}
\end{figure*}

\begin{figure*}[t]
    \centering
    \begin{tabular}{c ccc}
        & \textbf{$lr=0.0001$} & \textbf{$lr=0.001$} & \textbf{$lr=0.003$}\\[0.5em]
        \parbox[c]{1em}{\rotatebox{90}{\textbf{$wd=0.01$}}} &
        \begin{subfigure}[c]{0.29\linewidth}
            \includegraphics[width=\linewidth]{Fig/loss_le/HH/HH_lr0.0001_wd0.01_random_eval_acc.png}
        \end{subfigure}
        &
        \begin{subfigure}[c]{0.29\linewidth}
            \includegraphics[width=\linewidth]{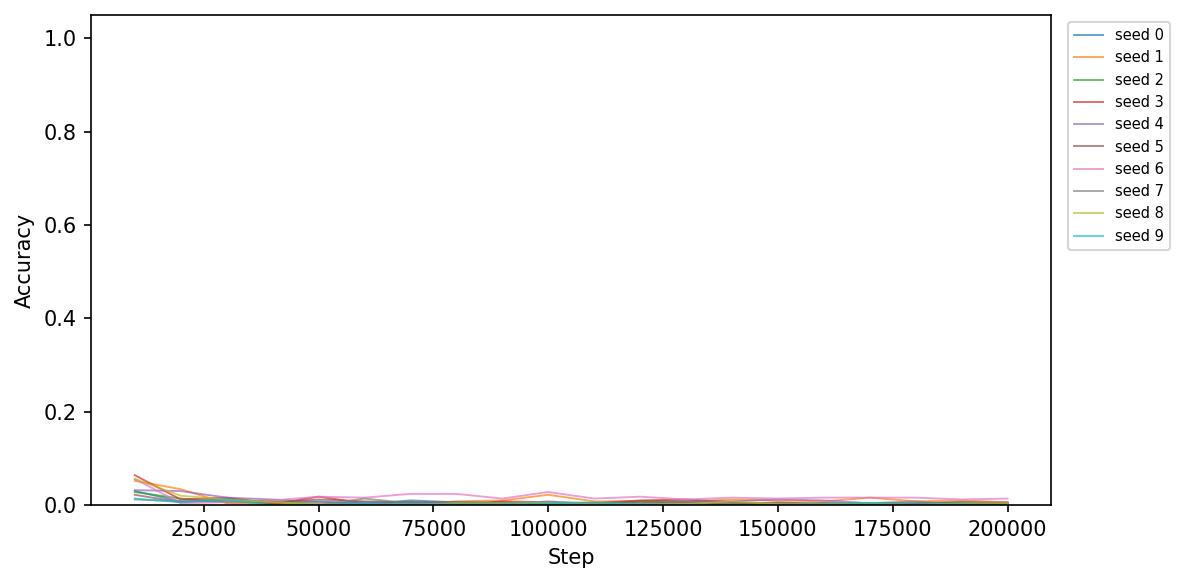}
        \end{subfigure}
        &
        \begin{subfigure}[c]{0.29\linewidth}
            \includegraphics[width=\linewidth]{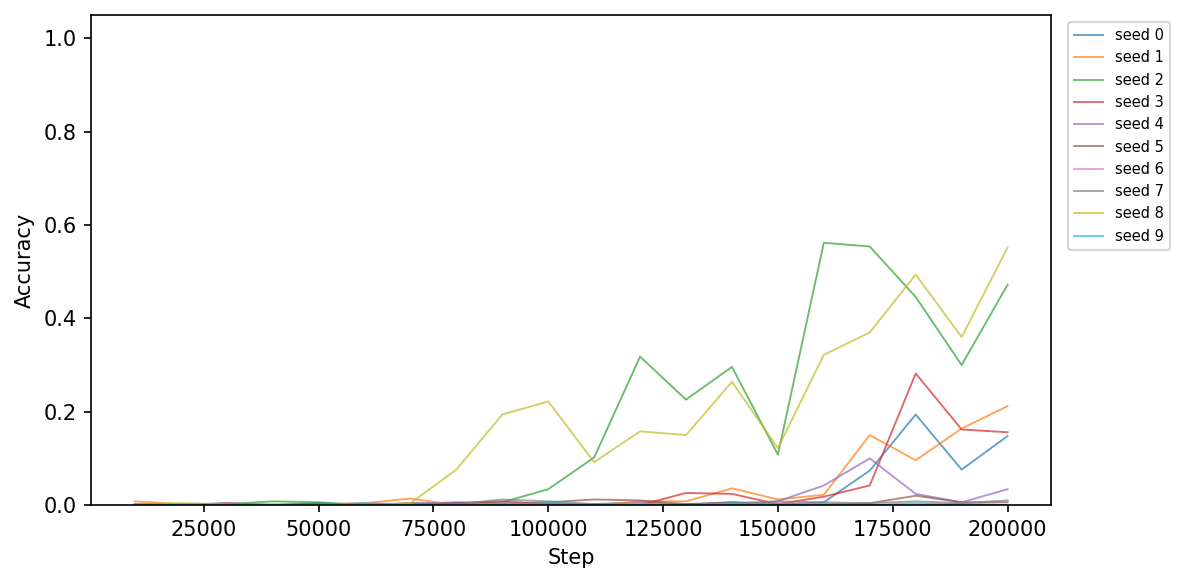}
        \end{subfigure}
        \\[1em]
        \parbox[c]{1em}{\rotatebox{90}{\textbf{$wd=0.1$}}} &
        \begin{subfigure}[c]{0.29\linewidth}
            \includegraphics[width=\linewidth]{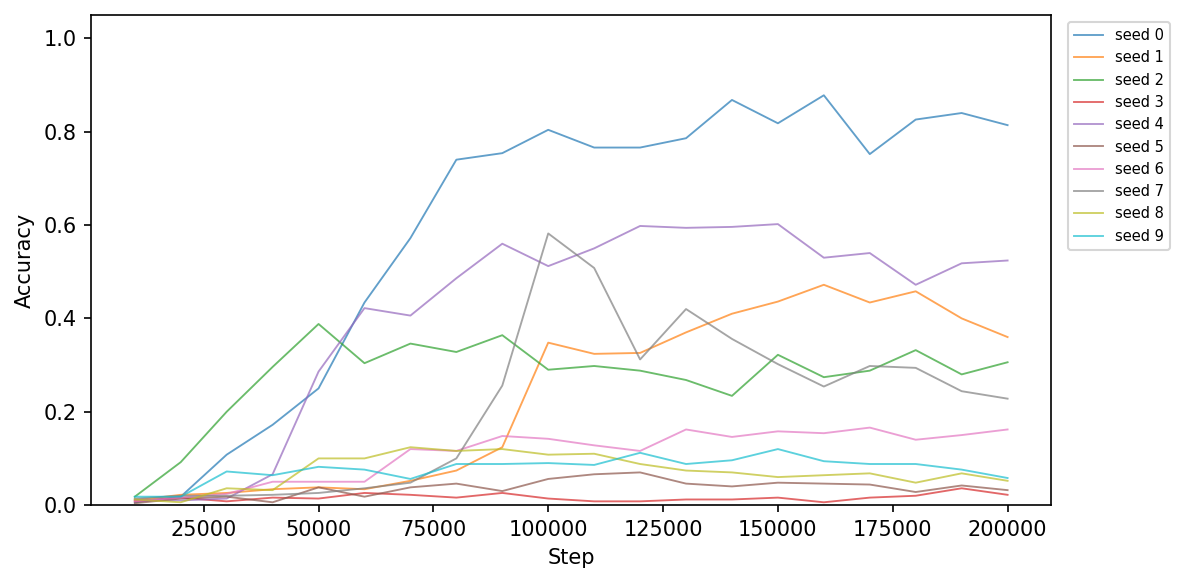}
        \end{subfigure}
        &
        \begin{subfigure}[c]{0.29\linewidth}
            \includegraphics[width=\linewidth]{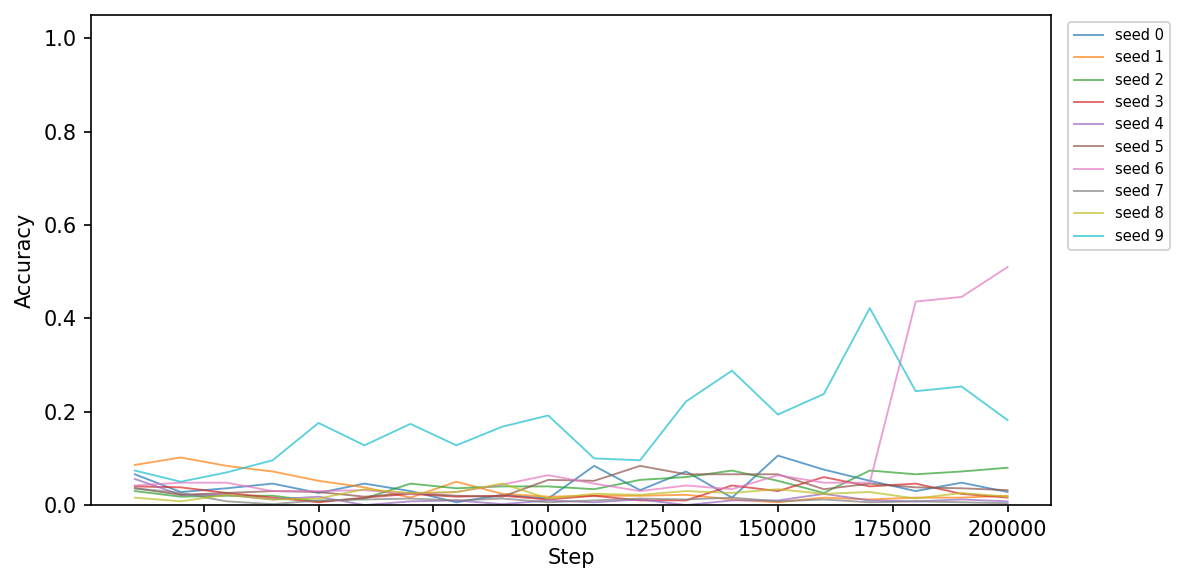}
        \end{subfigure}
        &
        \begin{subfigure}[c]{0.29\linewidth}
            \includegraphics[width=\linewidth]{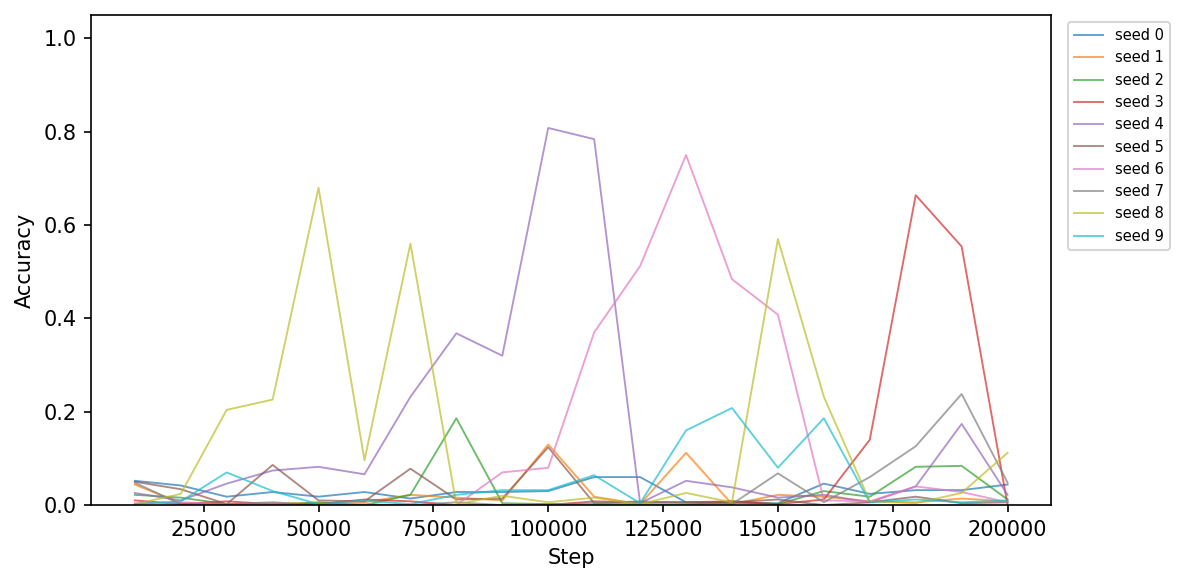}
        \end{subfigure}
        \\[1em]
        \parbox[c]{1em}{\rotatebox{90}{\textbf{$wd=0.2$}}} &
        \begin{subfigure}[c]{0.29\linewidth}
            \includegraphics[width=\linewidth]{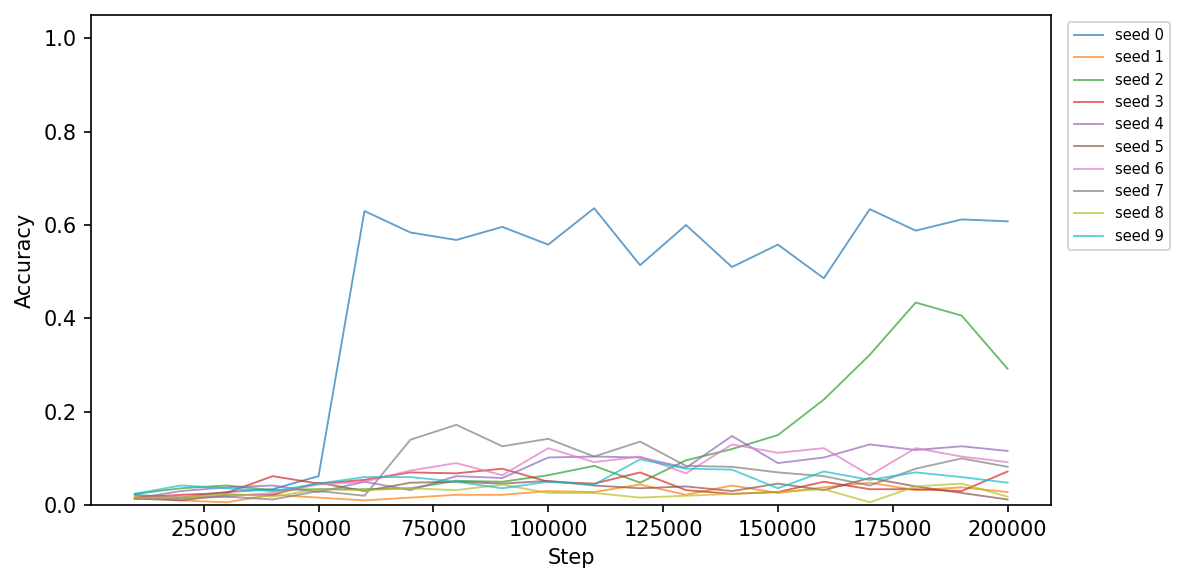}
        \end{subfigure}
        &
        \begin{subfigure}[c]{0.29\linewidth}
            \includegraphics[width=\linewidth]{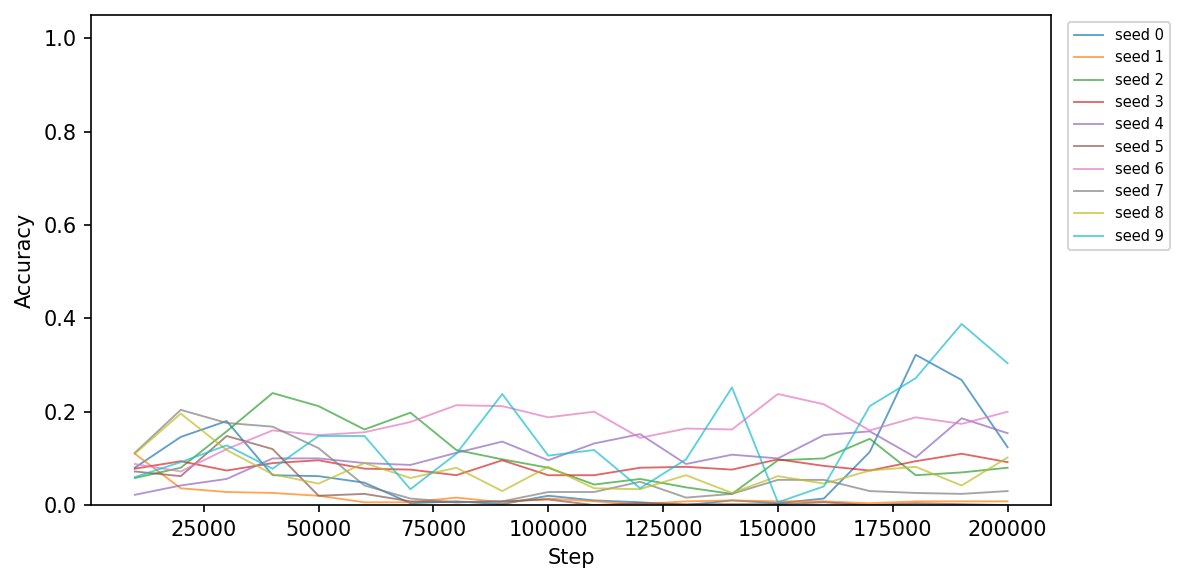}
        \end{subfigure}
        &
        \begin{subfigure}[c]{0.29\linewidth}
            \includegraphics[width=\linewidth]{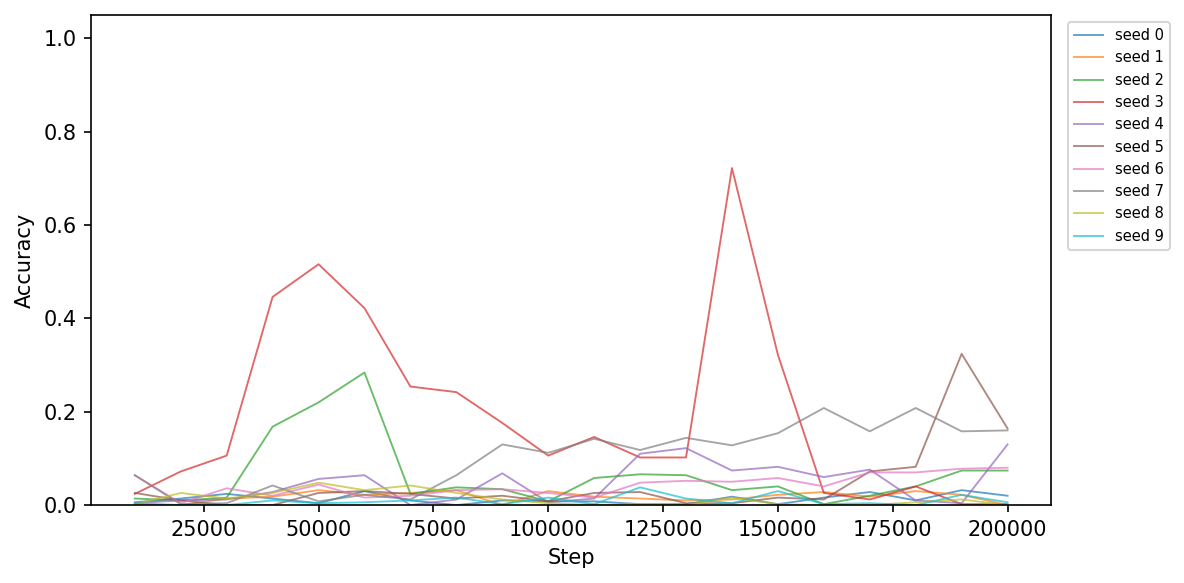}
        \end{subfigure}
        \\[1em]
        \parbox[c]{1em}{\rotatebox{90}{\textbf{$wd=0.3$}}} &
        \begin{subfigure}[c]{0.29\linewidth}
            \includegraphics[width=\linewidth]{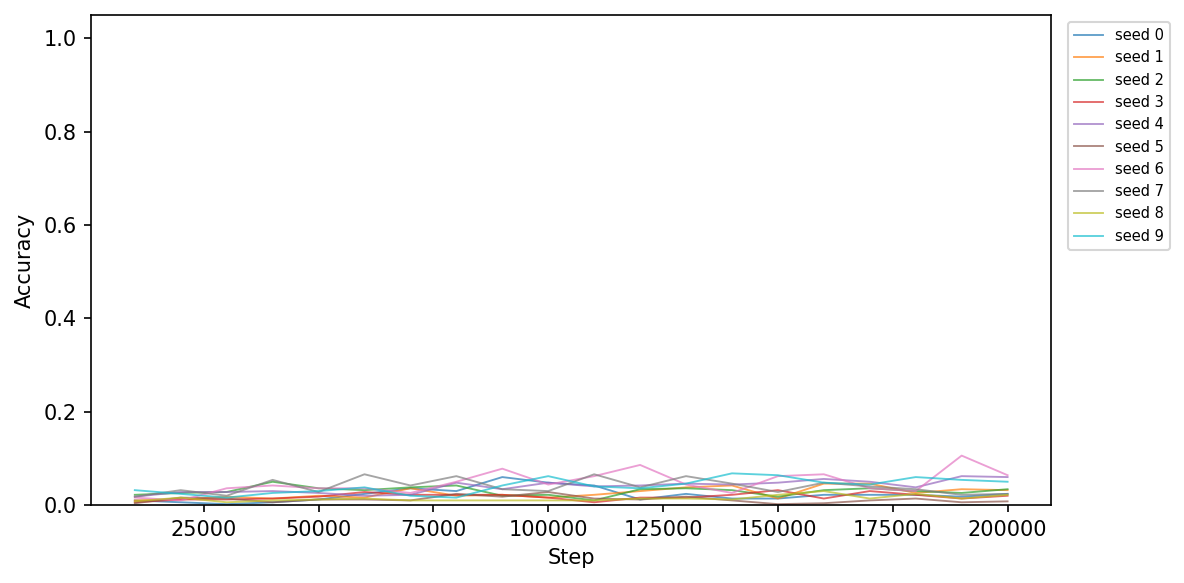}
        \end{subfigure}
        &
        \begin{subfigure}[c]{0.29\linewidth}
            \includegraphics[width=\linewidth]{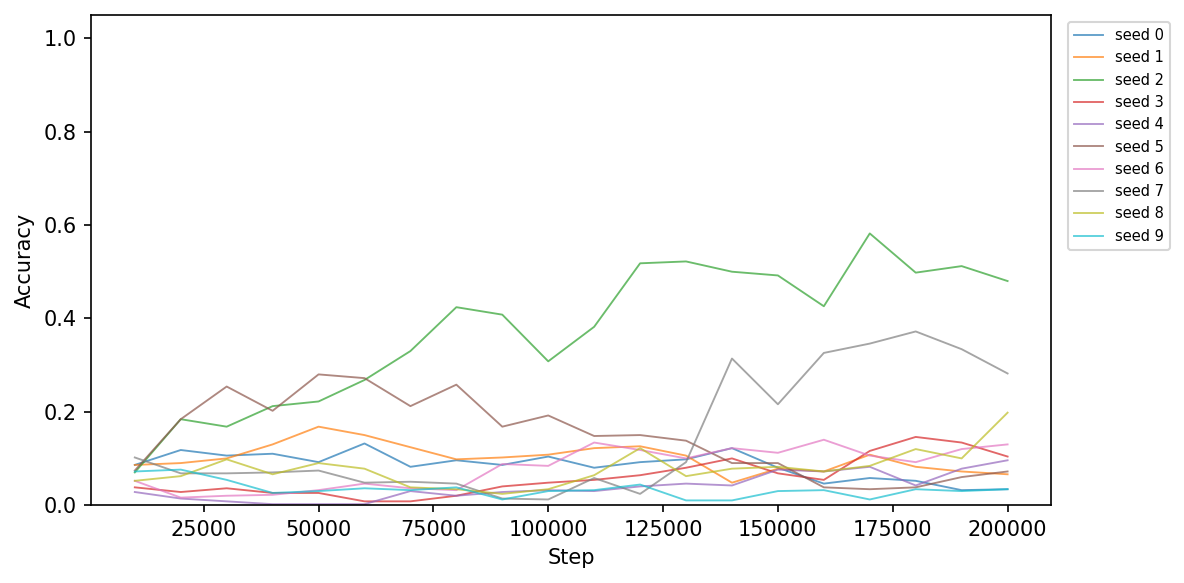}
        \end{subfigure}
        &
        \begin{subfigure}[c]{0.29\linewidth}
            \includegraphics[width=\linewidth]{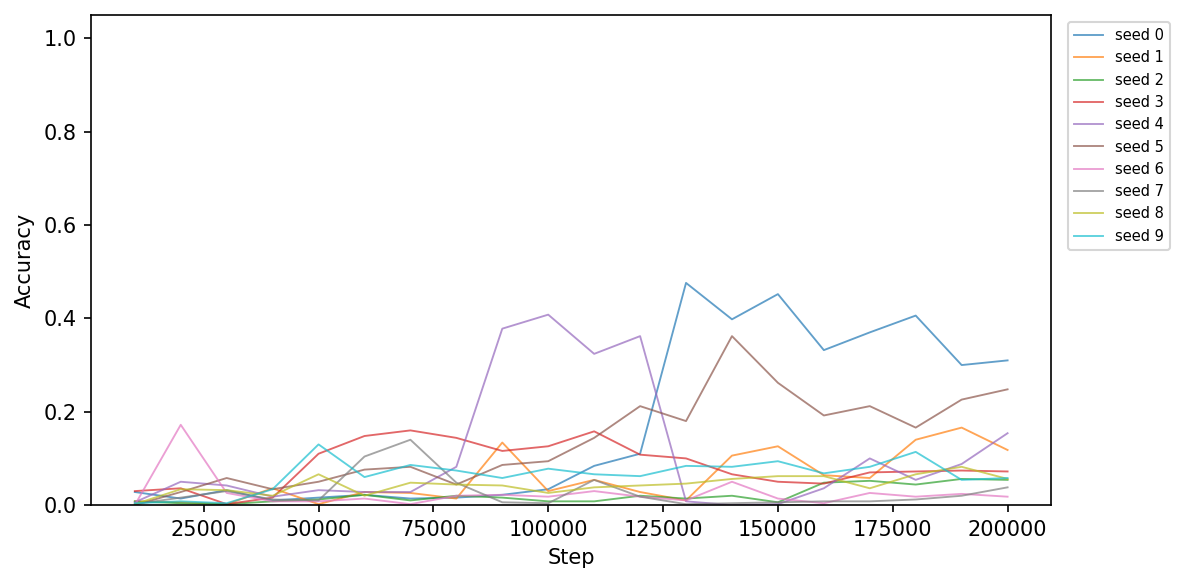}
        \end{subfigure}
        \\[1em]
        \parbox[c]{1em}{\rotatebox{90}{\textbf{$wd=0.4$}}} &
        \begin{subfigure}[c]{0.29\linewidth}
            \includegraphics[width=\linewidth]{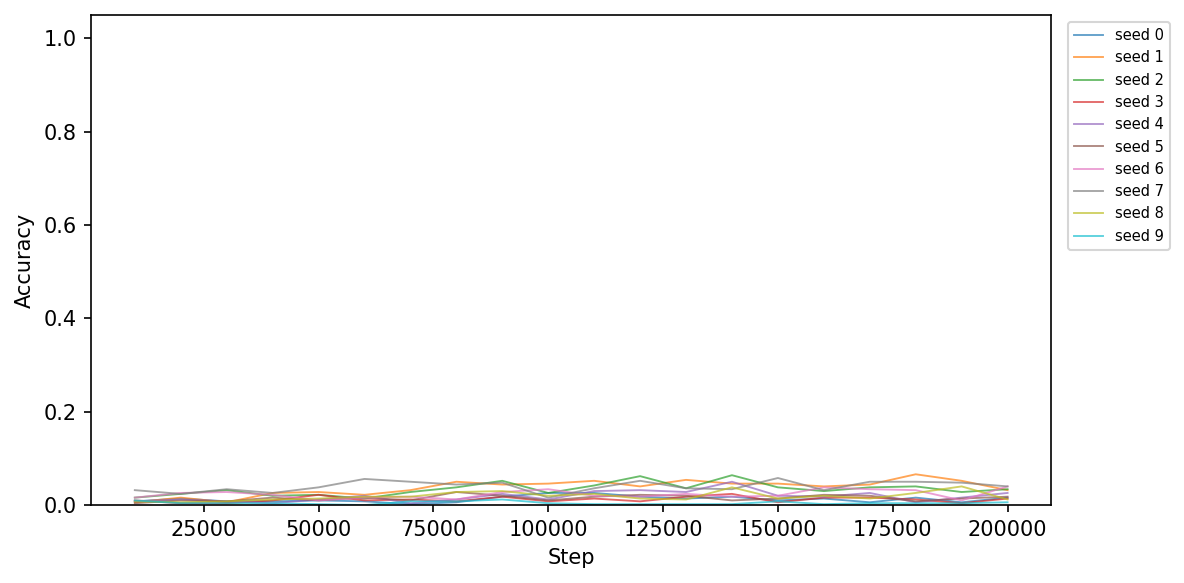}
        \end{subfigure}
        &
        \begin{subfigure}[c]{0.29\linewidth}
            \includegraphics[width=\linewidth]{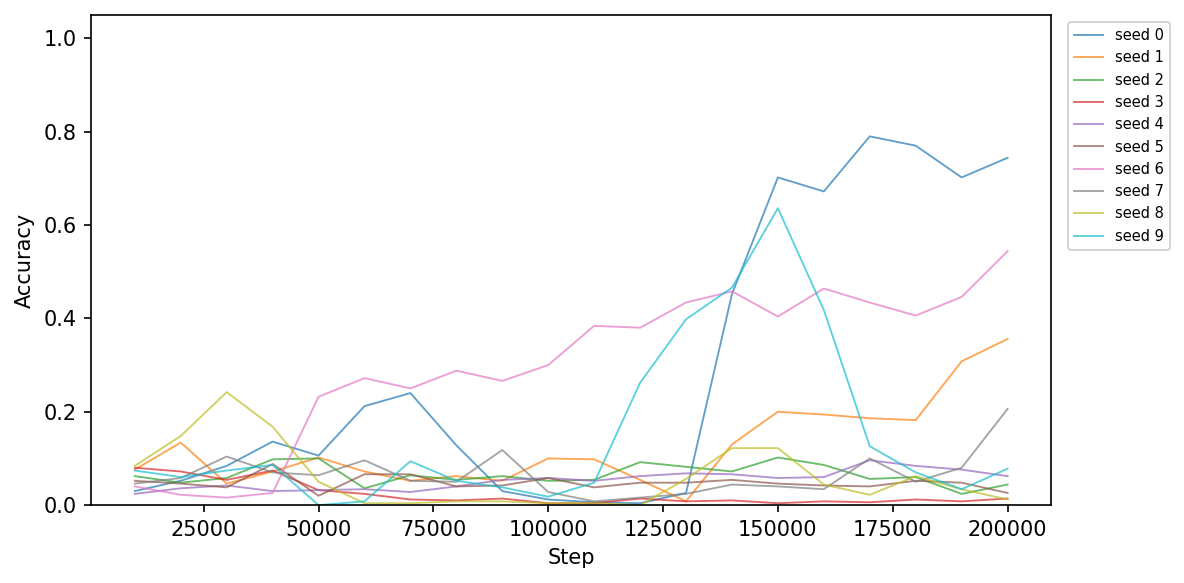}
        \end{subfigure}
        &
        \begin{subfigure}[c]{0.29\linewidth}
            \includegraphics[width=\linewidth]{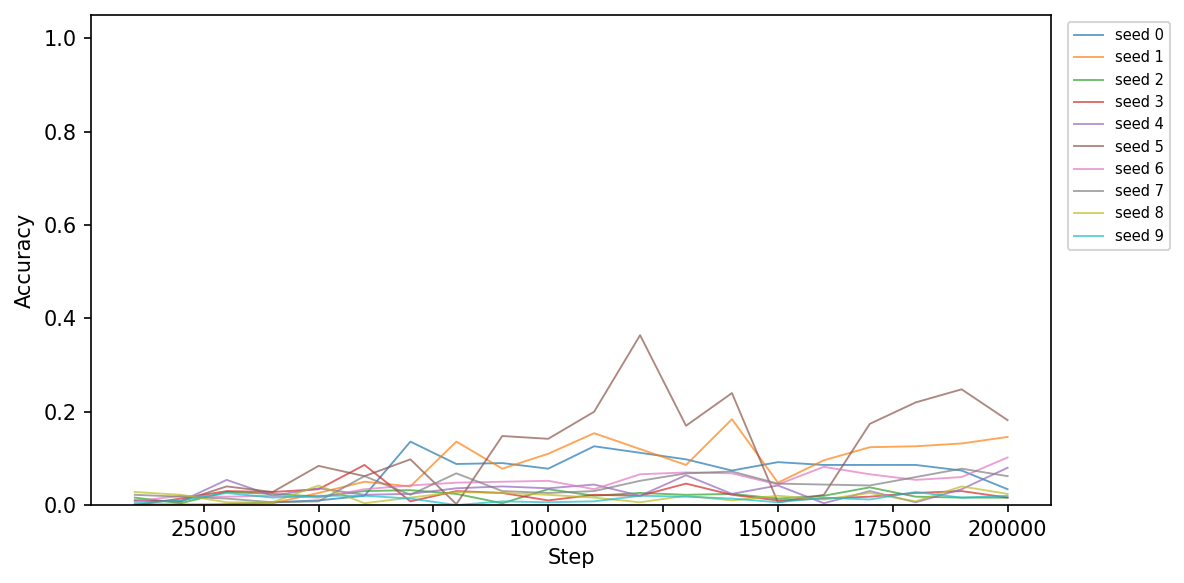}
        \end{subfigure}
        \\[1em]
        \parbox[c]{1em}{\rotatebox{90}{\textbf{$wd=0.5$}}} &
        \begin{subfigure}[c]{0.29\linewidth}
            \includegraphics[width=\linewidth]{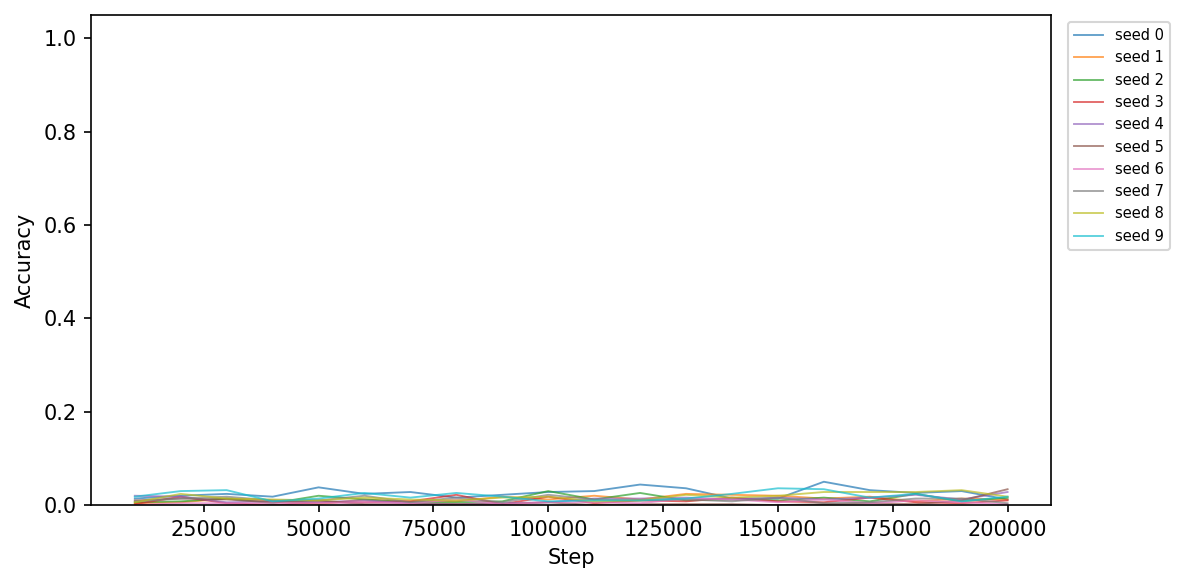}
        \end{subfigure}
        &
        \begin{subfigure}[c]{0.29\linewidth}
            \includegraphics[width=\linewidth]{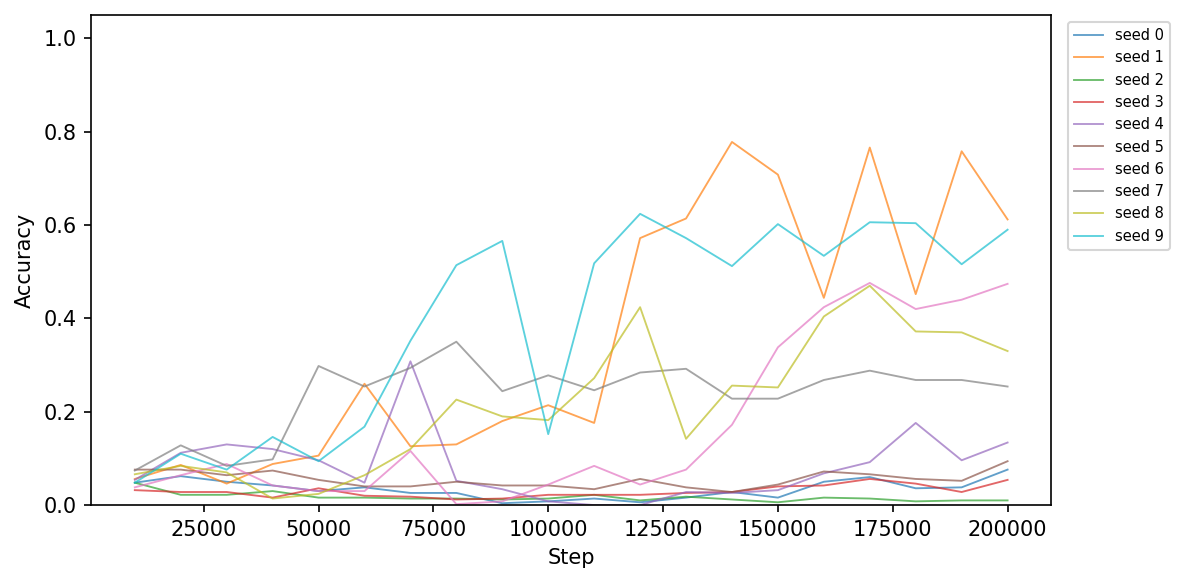}
        \end{subfigure}
        &
        \begin{subfigure}[c]{0.29\linewidth}
            \includegraphics[width=\linewidth]{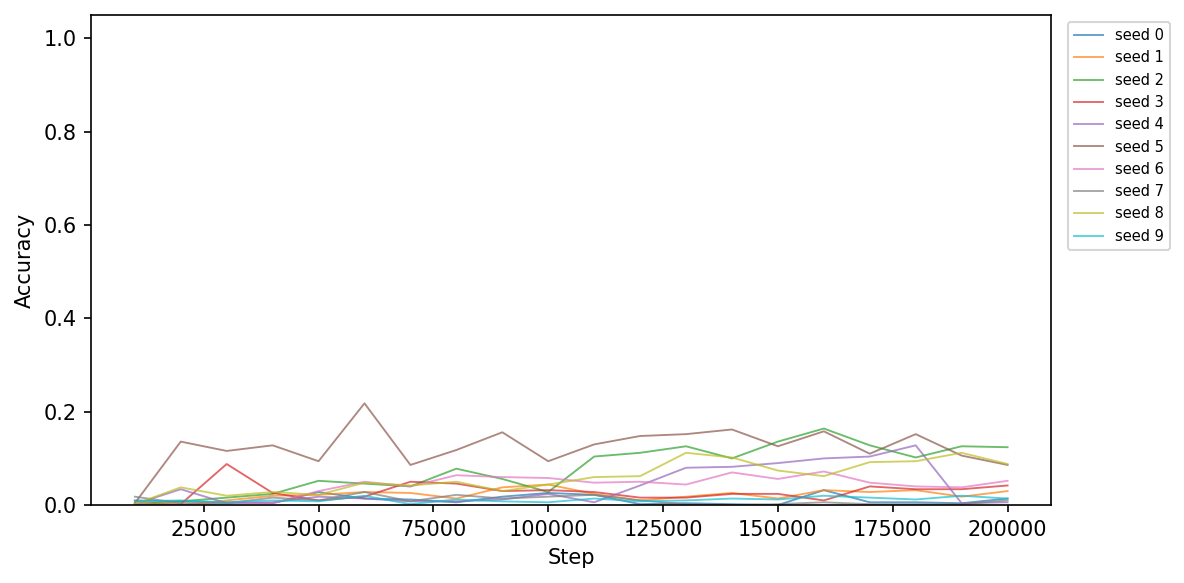}
        \end{subfigure}
        \\[1em]
    \end{tabular}
    \caption{$\mathrm{M_H}$ - Evaluation Accuracy across epochs, on different hyperparameters}
    \label{fig:eval_HH}
\end{figure*}

\begin{figure*}[t]
    \centering
    \begin{tabular}{c ccc}
        & \textbf{$lr=0.0001$} & \textbf{$lr=0.001$} & \textbf{$lr=0.003$}\\[0.5em]
        \parbox[c]{1em}{\rotatebox{90}{\textbf{$wd=0.01$}}} &
        \begin{subfigure}[c]{0.29\linewidth}
            \includegraphics[width=\linewidth]{Fig/loss_le/FF/FF_lr0.0001_wd0.01_train_loss_linear.png}
        \end{subfigure}
        &
        \begin{subfigure}[c]{0.29\linewidth}
            \includegraphics[width=\linewidth]{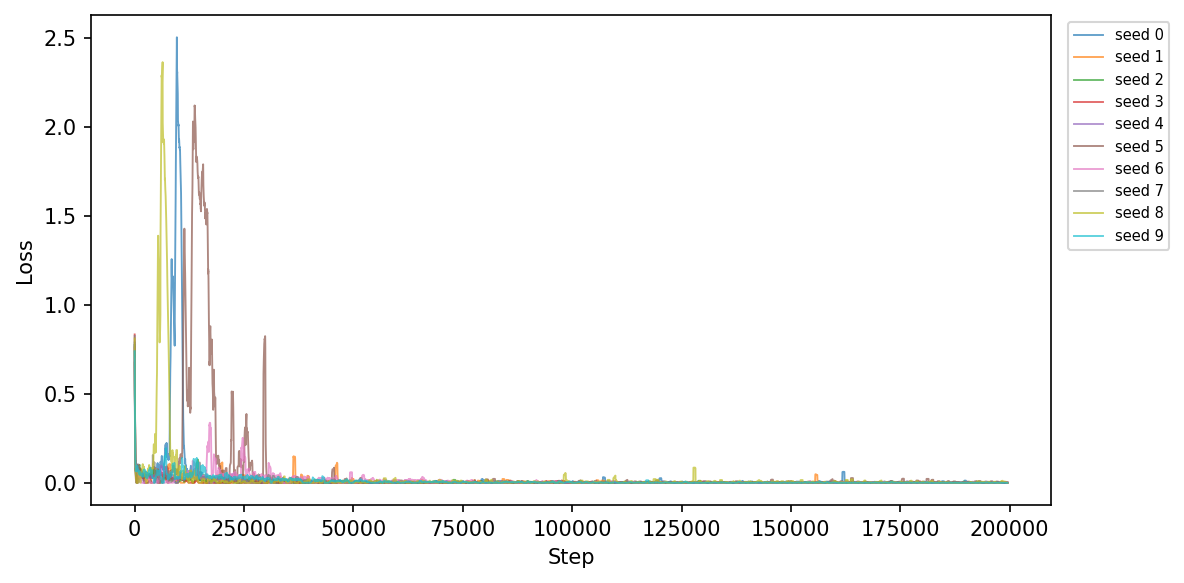}
        \end{subfigure}
        &
        \begin{subfigure}[c]{0.29\linewidth}
            \includegraphics[width=\linewidth]{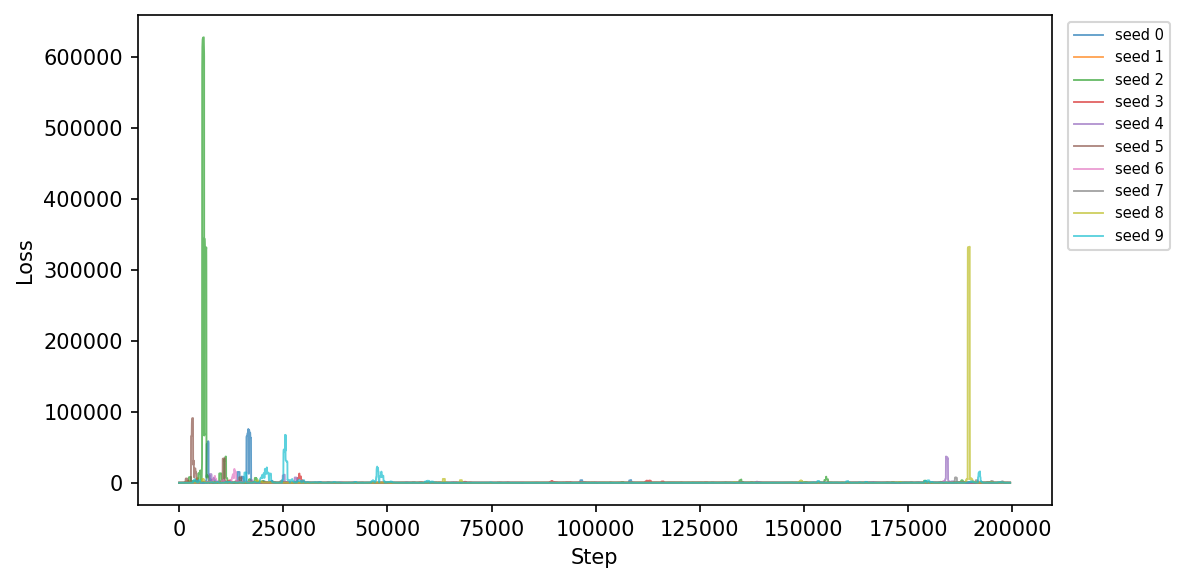}
        \end{subfigure}
        \\[1em]
        \parbox[c]{1em}{\rotatebox{90}{\textbf{$wd=0.1$}}} &
        \begin{subfigure}[c]{0.29\linewidth}
            \includegraphics[width=\linewidth]{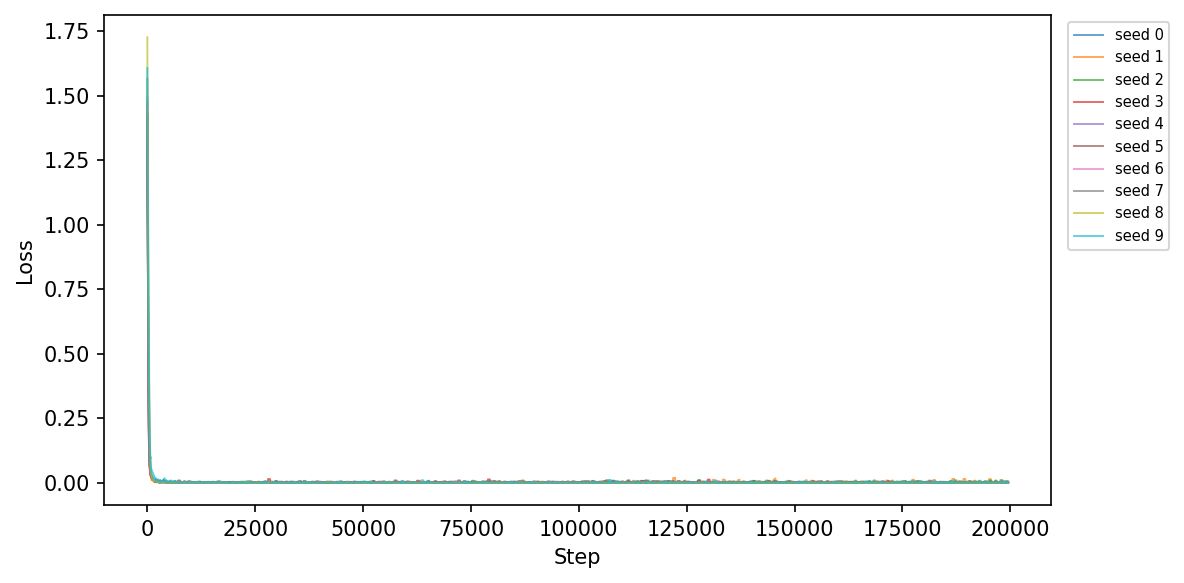}
        \end{subfigure}
        &
        \begin{subfigure}[c]{0.29\linewidth}
            \includegraphics[width=\linewidth]{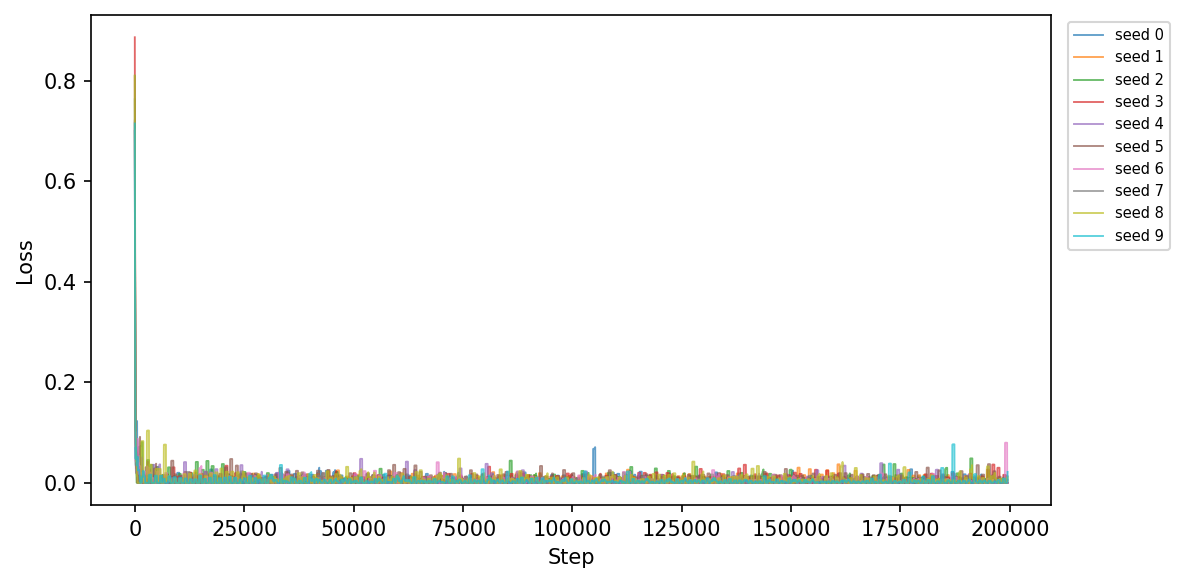}
        \end{subfigure}
        &
        \begin{subfigure}[c]{0.29\linewidth}
            \includegraphics[width=\linewidth]{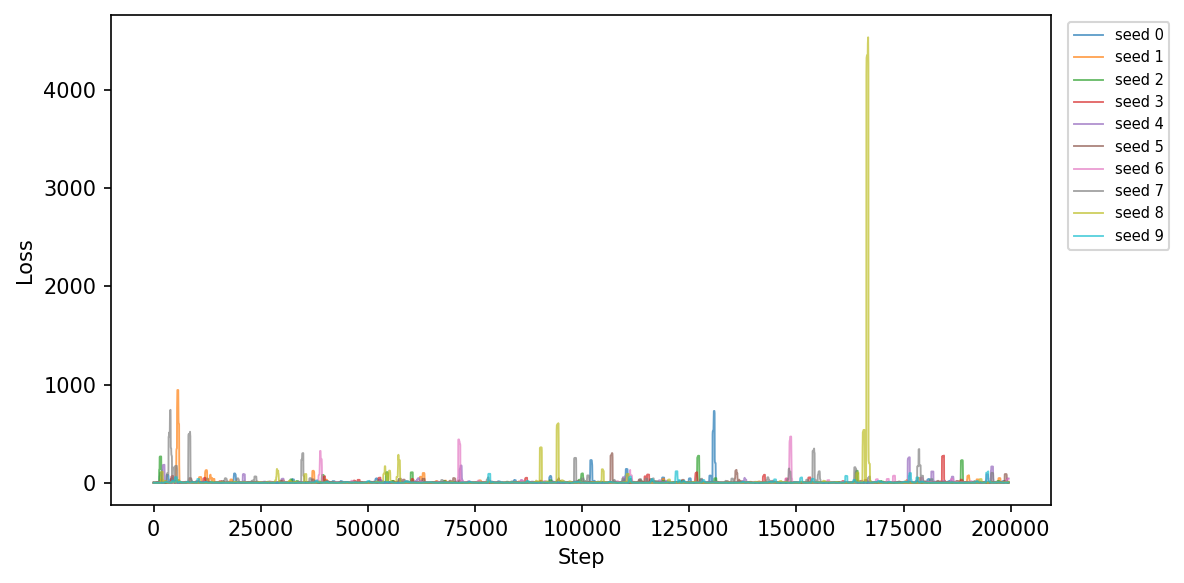}
        \end{subfigure}
        \\[1em]
        \parbox[c]{1em}{\rotatebox{90}{\textbf{$wd=0.2$}}} &
        \begin{subfigure}[c]{0.29\linewidth}
            \includegraphics[width=\linewidth]{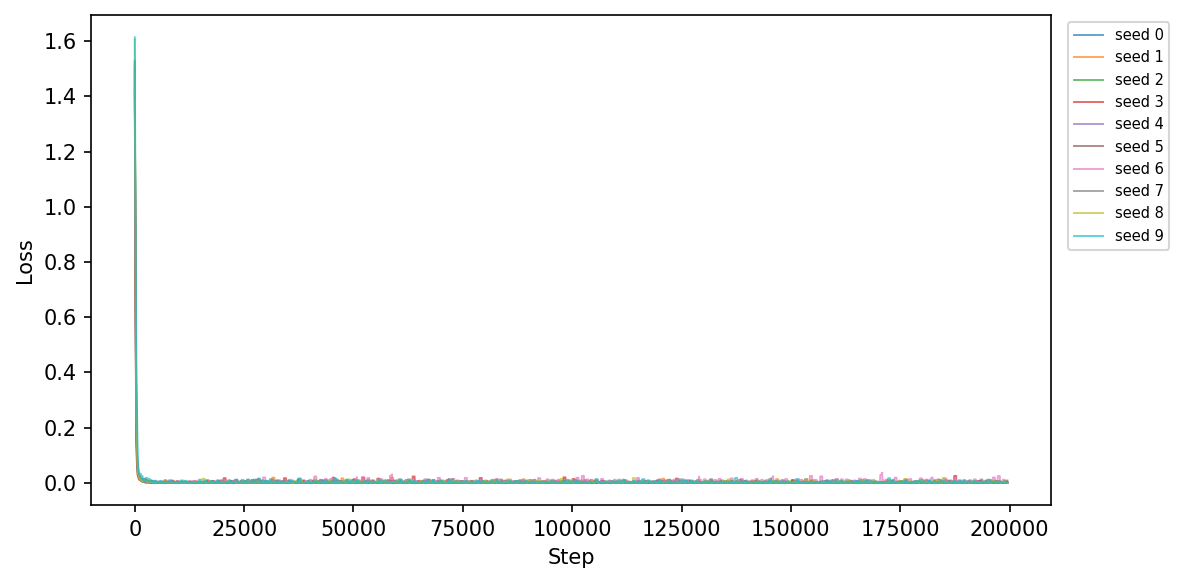}
        \end{subfigure}
        &
        \begin{subfigure}[c]{0.29\linewidth}
            \includegraphics[width=\linewidth]{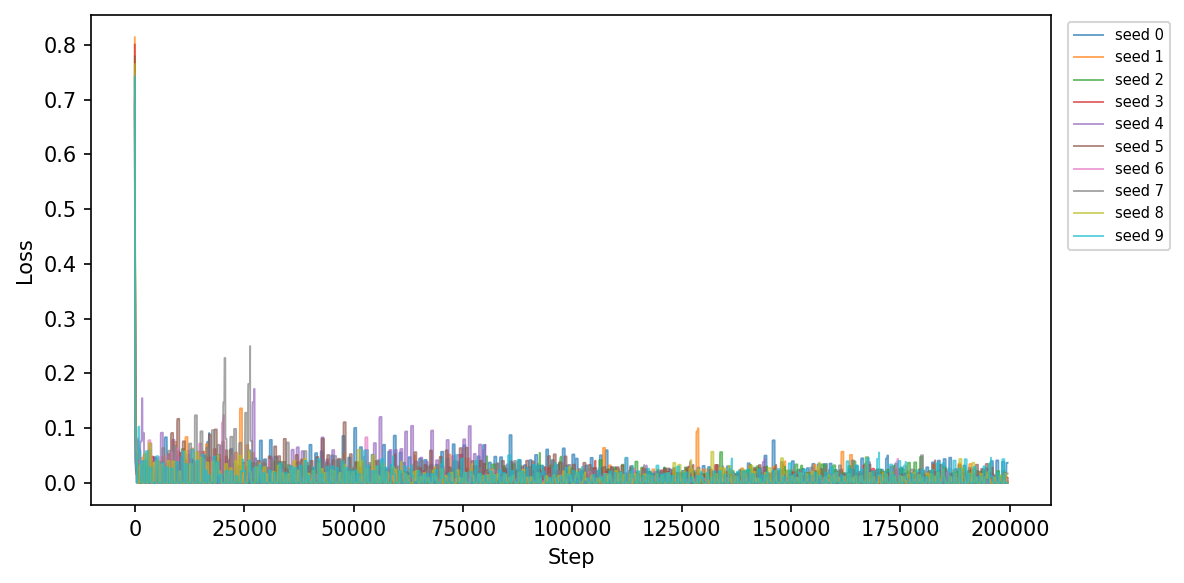}
        \end{subfigure}
        &
        \begin{subfigure}[c]{0.29\linewidth}
            \includegraphics[width=\linewidth]{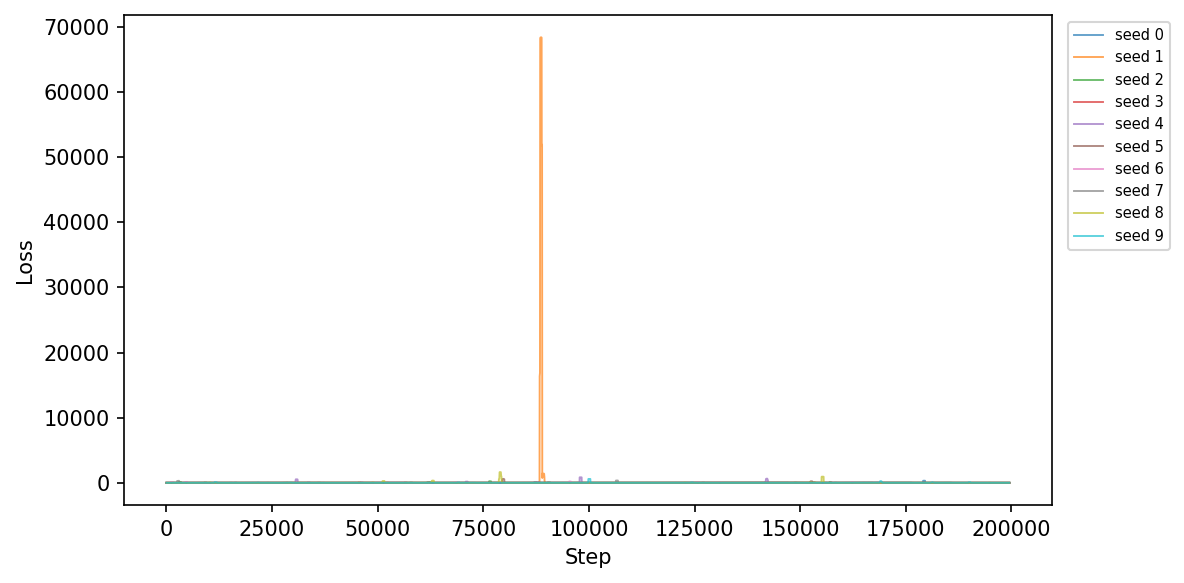}
        \end{subfigure}
        \\[1em]
        \parbox[c]{1em}{\rotatebox{90}{\textbf{$wd=0.3$}}} &
        \begin{subfigure}[c]{0.29\linewidth}
            \includegraphics[width=\linewidth]{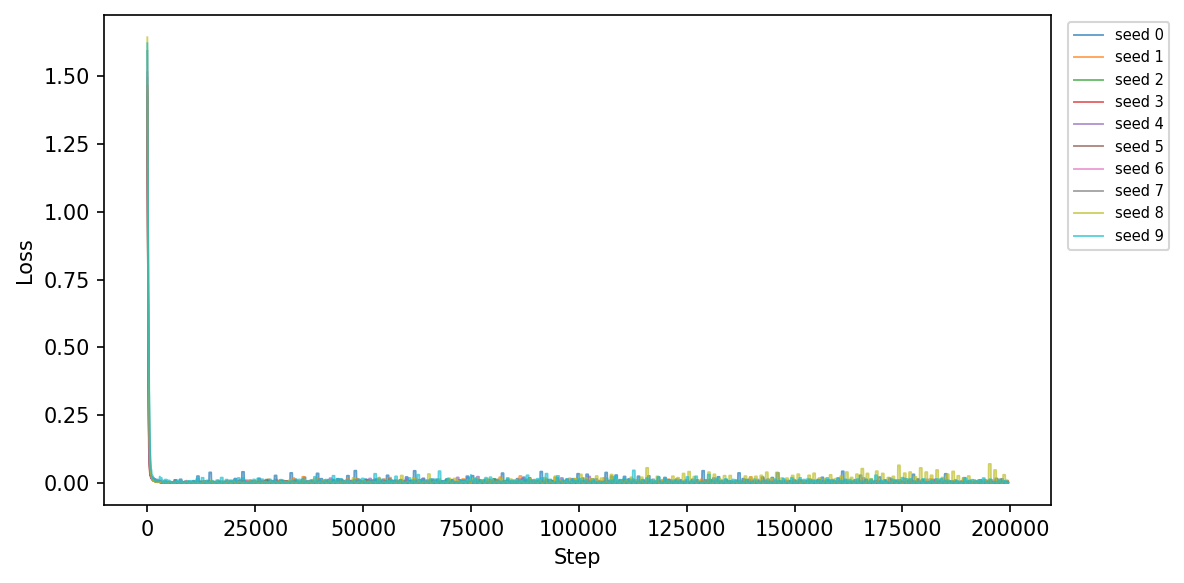}
        \end{subfigure}
        &
        \begin{subfigure}[c]{0.29\linewidth}
            \includegraphics[width=\linewidth]{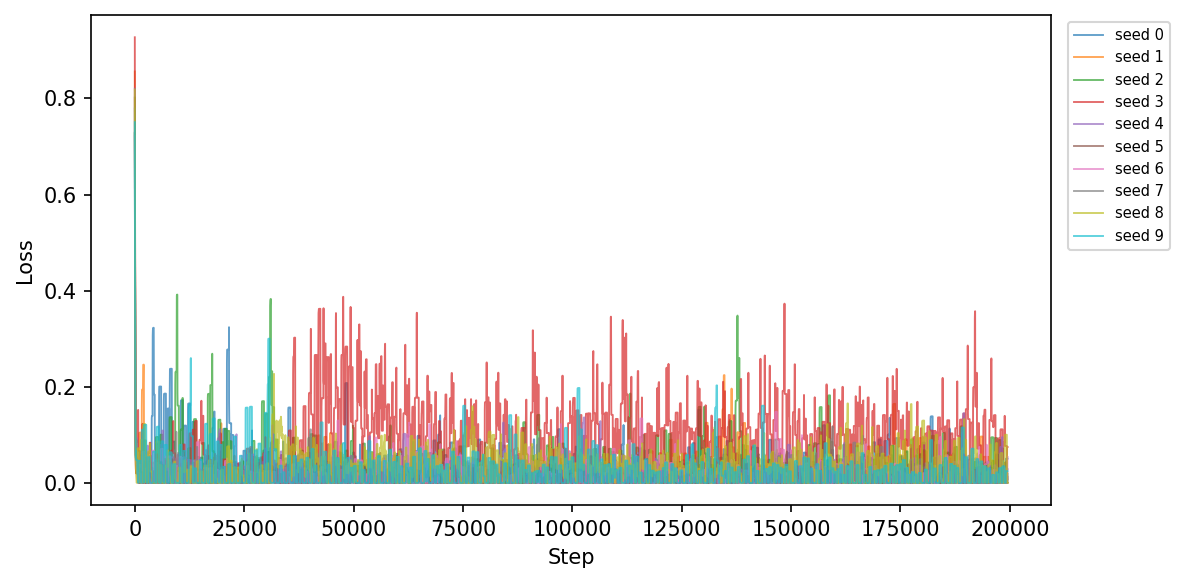}
        \end{subfigure}
        &
        \begin{subfigure}[c]{0.29\linewidth}
            \includegraphics[width=\linewidth]{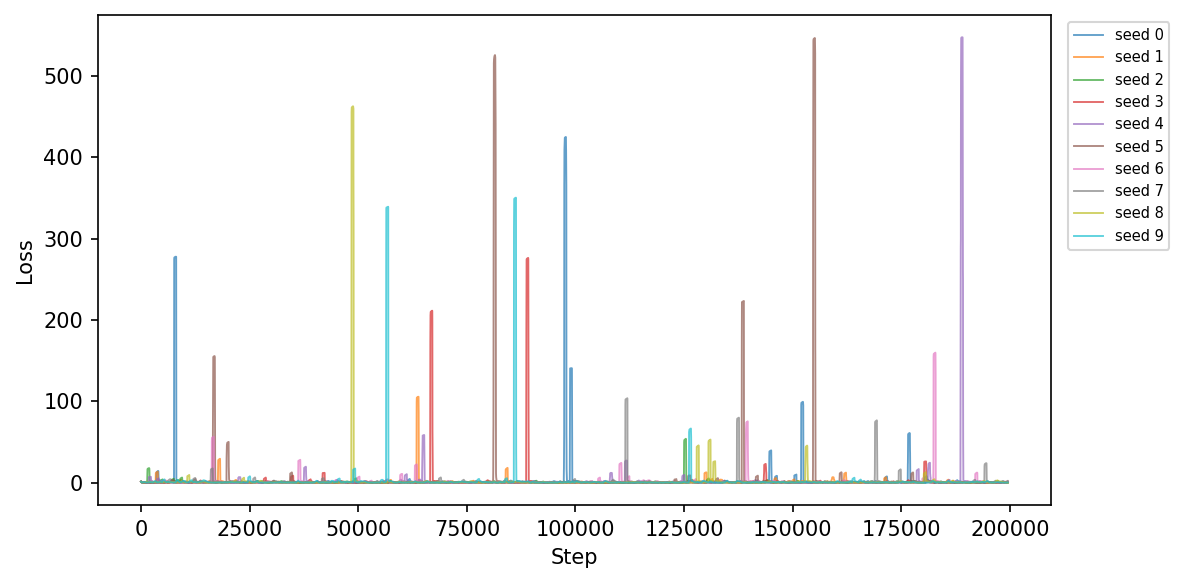}
        \end{subfigure}
        \\[1em]
        \parbox[c]{1em}{\rotatebox{90}{\textbf{$wd=0.4$}}} &
        \begin{subfigure}[c]{0.29\linewidth}
            \includegraphics[width=\linewidth]{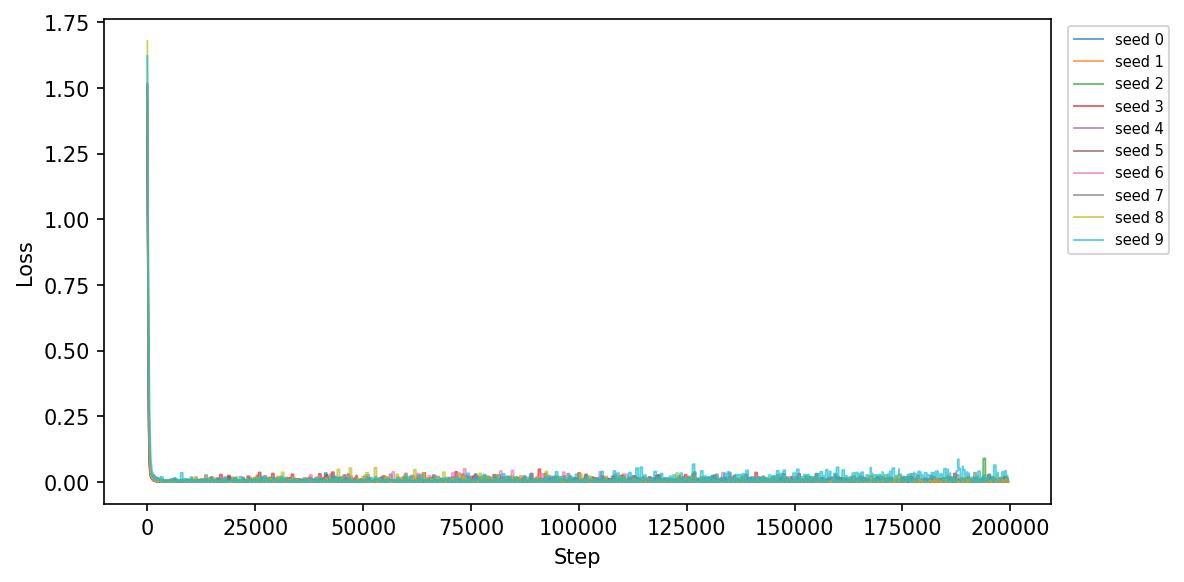}
        \end{subfigure}
        &
        \begin{subfigure}[c]{0.29\linewidth}
            \includegraphics[width=\linewidth]{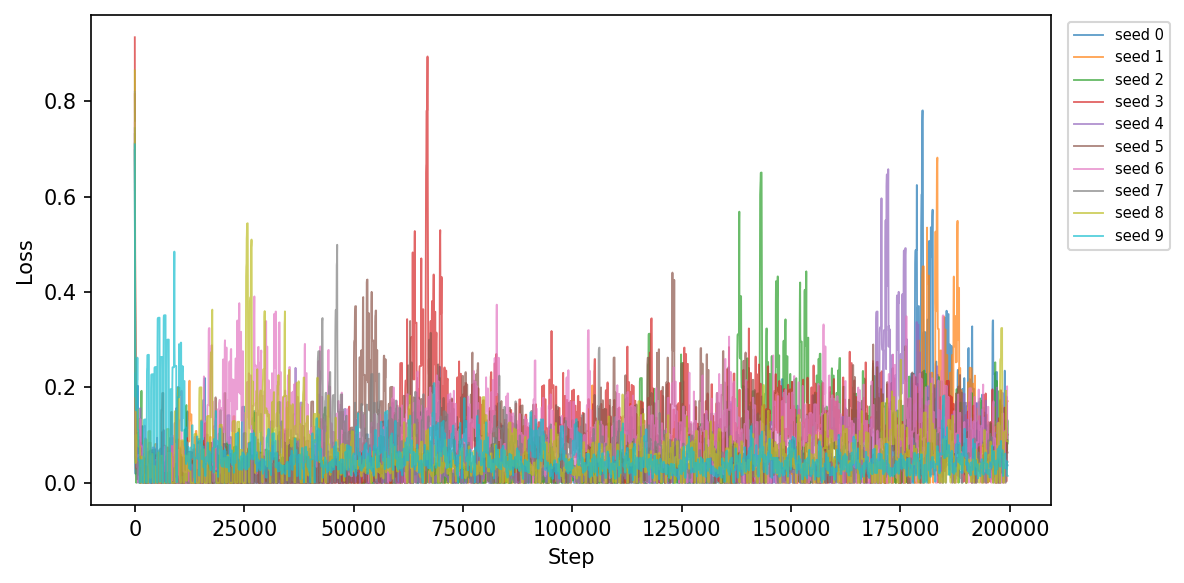}
        \end{subfigure}
        &
        \begin{subfigure}[c]{0.29\linewidth}
            \includegraphics[width=\linewidth]{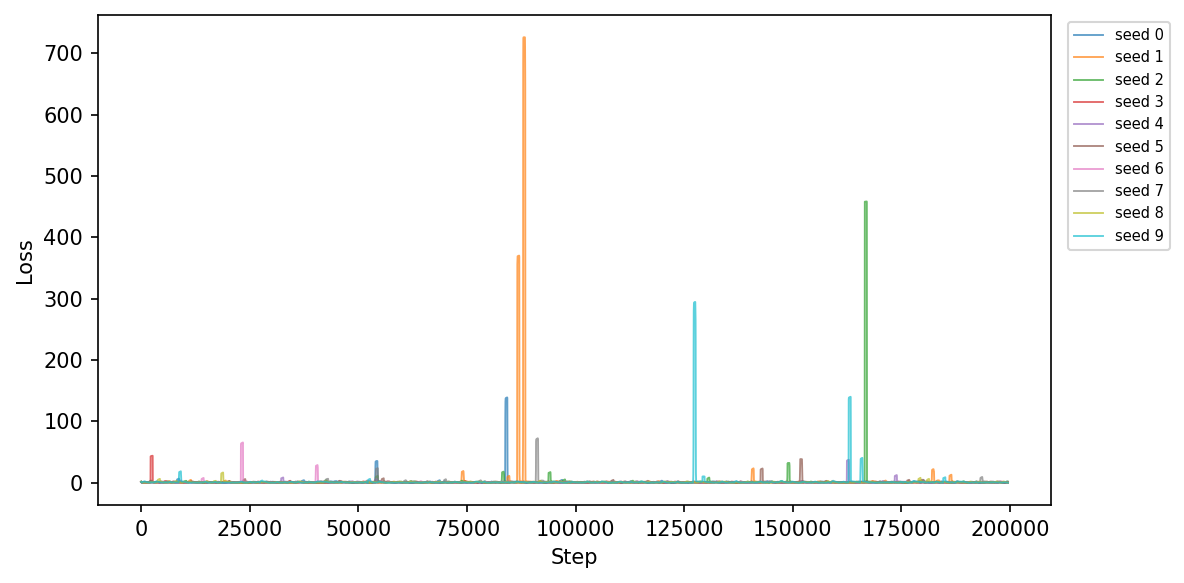}
        \end{subfigure}
        \\[1em]
        \parbox[c]{1em}{\rotatebox{90}{\textbf{$wd=0.5$}}} &
        \begin{subfigure}[c]{0.29\linewidth}
            \includegraphics[width=\linewidth]{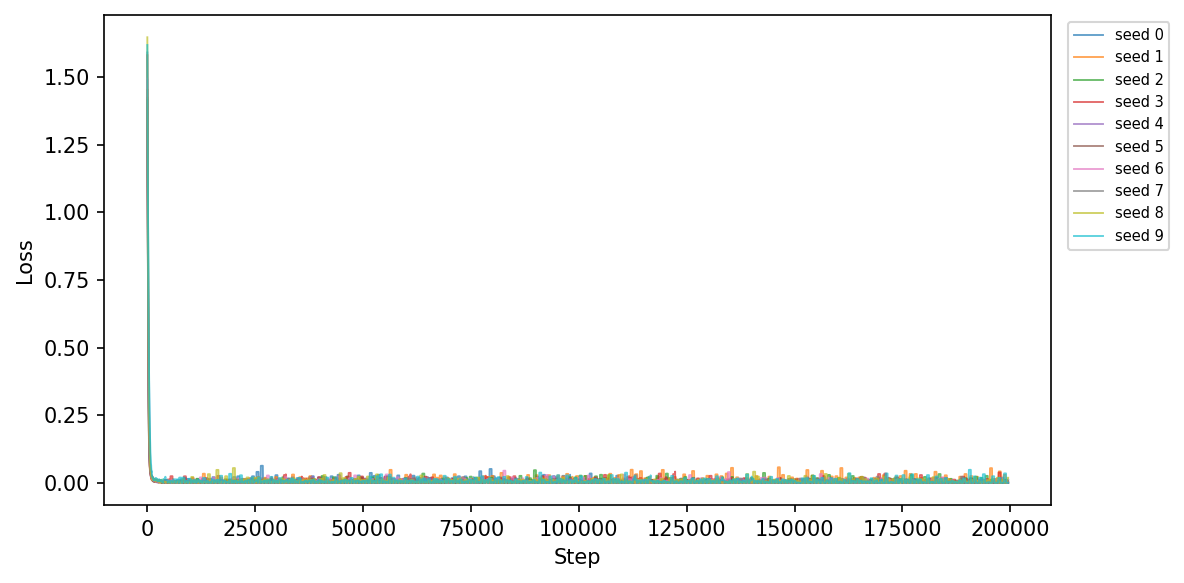}
        \end{subfigure}
        &
        \begin{subfigure}[c]{0.29\linewidth}
            \includegraphics[width=\linewidth]{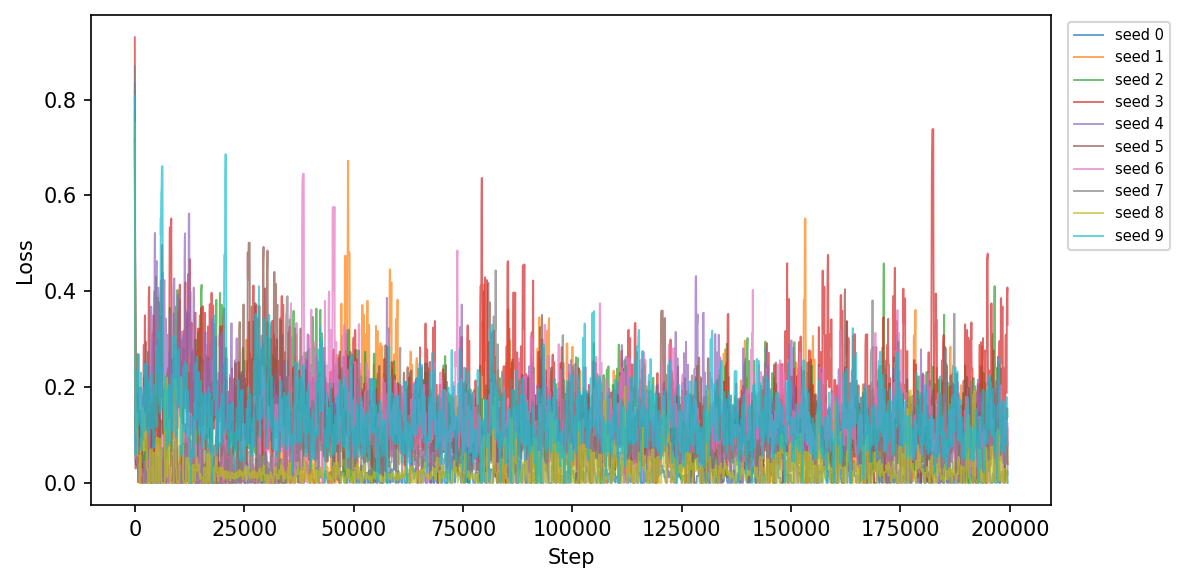}
        \end{subfigure}
        &
        \begin{subfigure}[c]{0.29\linewidth}
            \includegraphics[width=\linewidth]{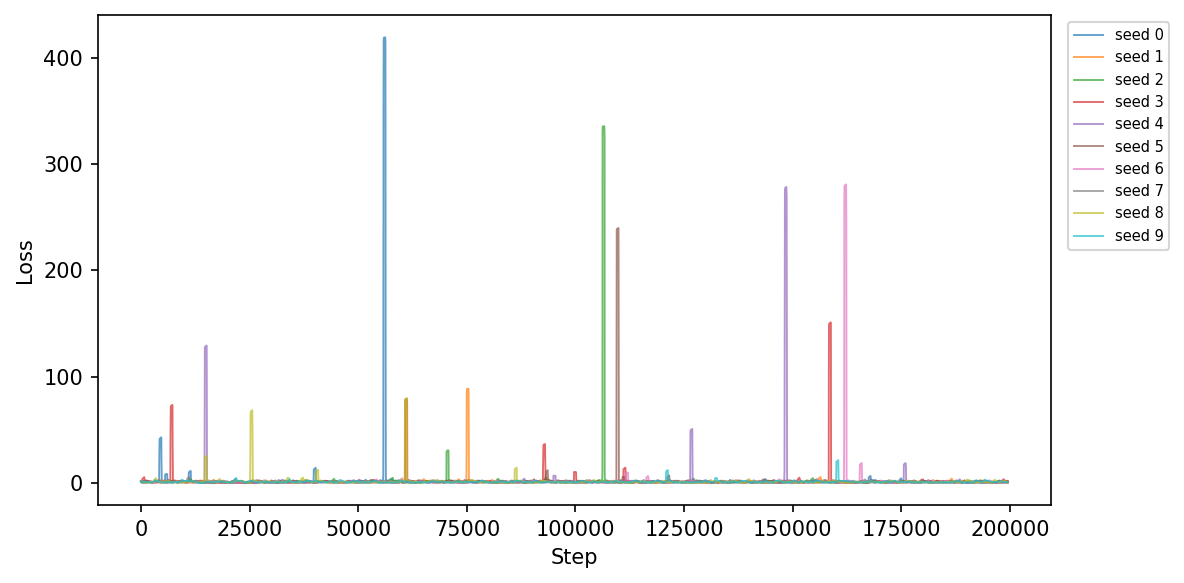}
        \end{subfigure}
        \\[1em]
    \end{tabular}
    \caption{$\mathrm{M_F}$ - Train Loss across epochs, on different hyperparameters}
    \label{fig:loss_FF}
\end{figure*}

\begin{figure*}[t]
    \centering
    \begin{tabular}{c ccc}
        & \textbf{$lr=0.0001$} & \textbf{$lr=0.001$} & \textbf{$lr=0.003$}\\[0.5em]
        \parbox[c]{1em}{\rotatebox{90}{\textbf{$wd=0.01$}}} &
        \begin{subfigure}[c]{0.29\linewidth}
            \includegraphics[width=\linewidth]{Fig/loss_le/FF/FF_lr0.0001_wd0.01_random_eval_acc.png}
        \end{subfigure}
        &
        \begin{subfigure}[c]{0.29\linewidth}
            \includegraphics[width=\linewidth]{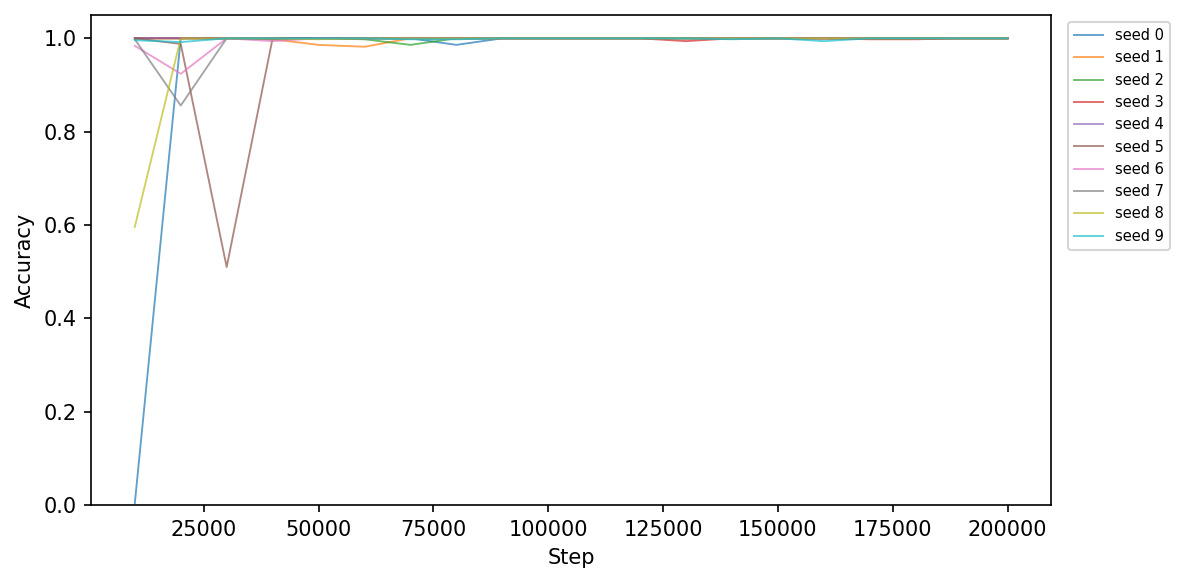}
        \end{subfigure}
        &
        \begin{subfigure}[c]{0.29\linewidth}
            \includegraphics[width=\linewidth]{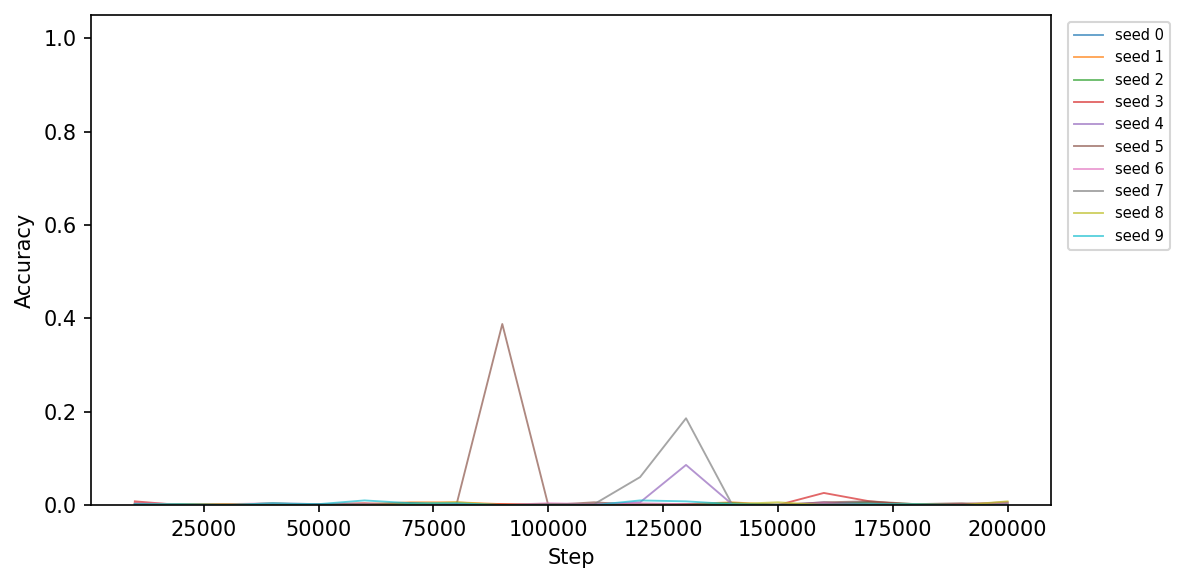}
        \end{subfigure}
        \\[1em]
        \parbox[c]{1em}{\rotatebox{90}{\textbf{$wd=0.1$}}} &
        \begin{subfigure}[c]{0.29\linewidth}
            \includegraphics[width=\linewidth]{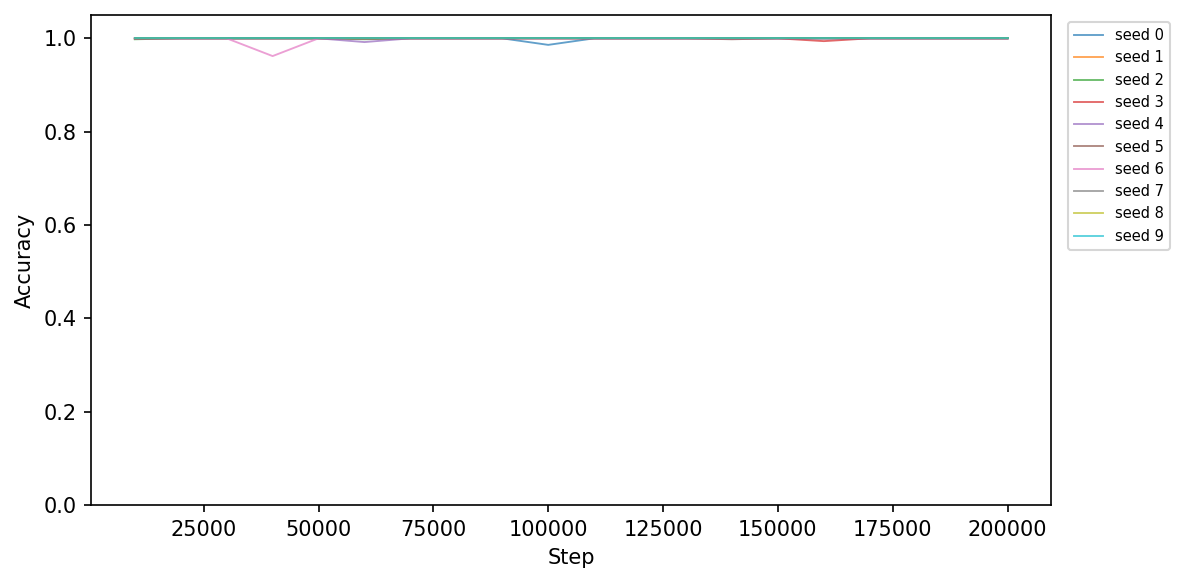}
        \end{subfigure}
        &
        \begin{subfigure}[c]{0.29\linewidth}
            \includegraphics[width=\linewidth]{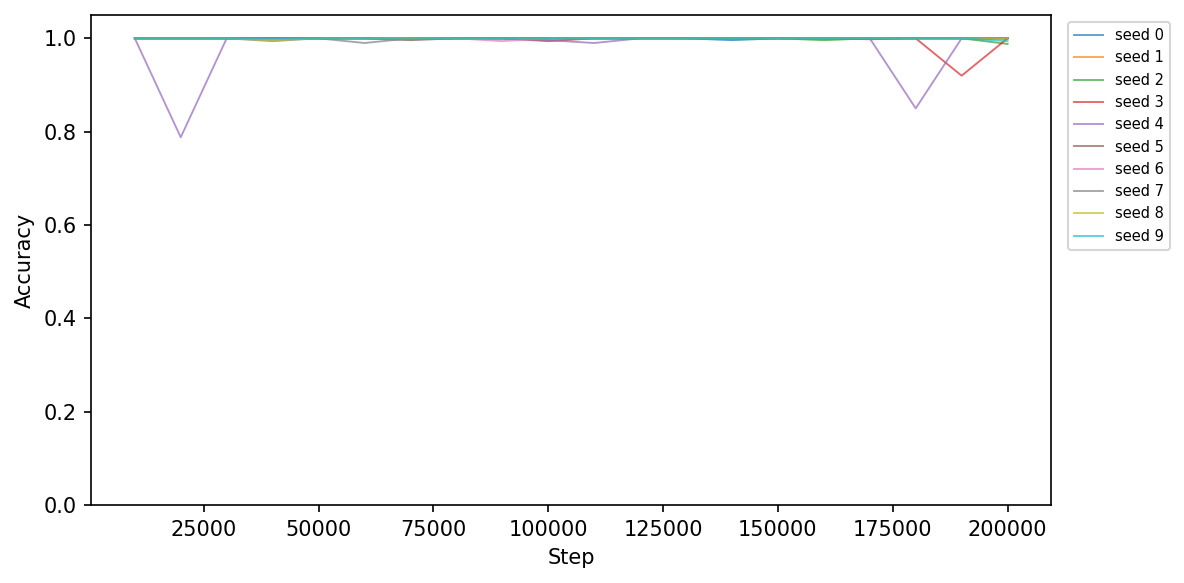}
        \end{subfigure}
        &
        \begin{subfigure}[c]{0.29\linewidth}
            \includegraphics[width=\linewidth]{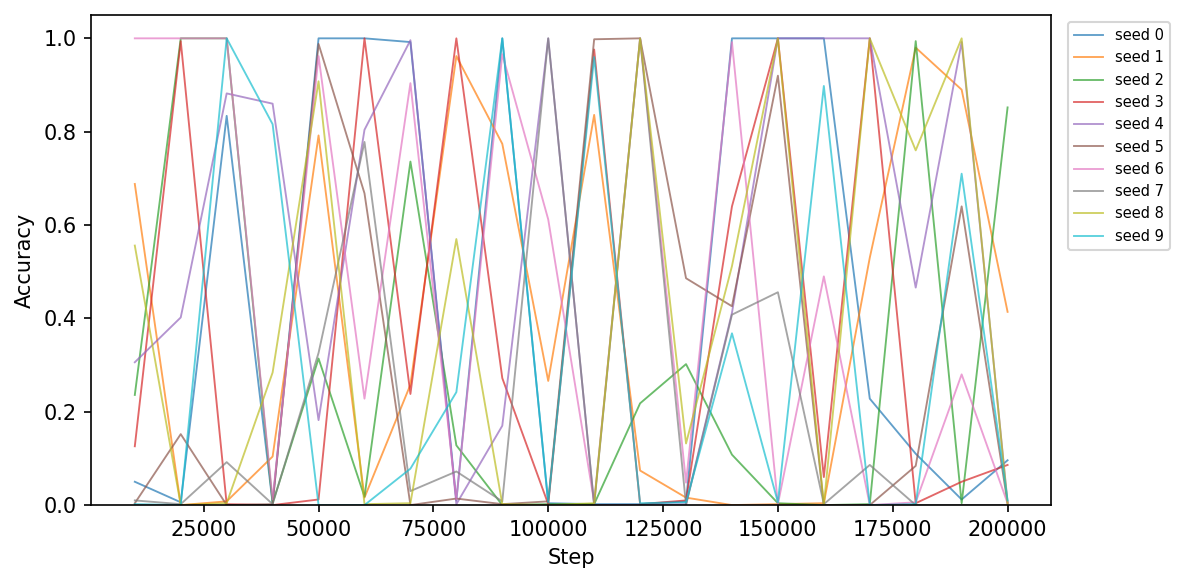}
        \end{subfigure}
        \\[1em]
        \parbox[c]{1em}{\rotatebox{90}{\textbf{$wd=0.2$}}} &
        \begin{subfigure}[c]{0.29\linewidth}
            \includegraphics[width=\linewidth]{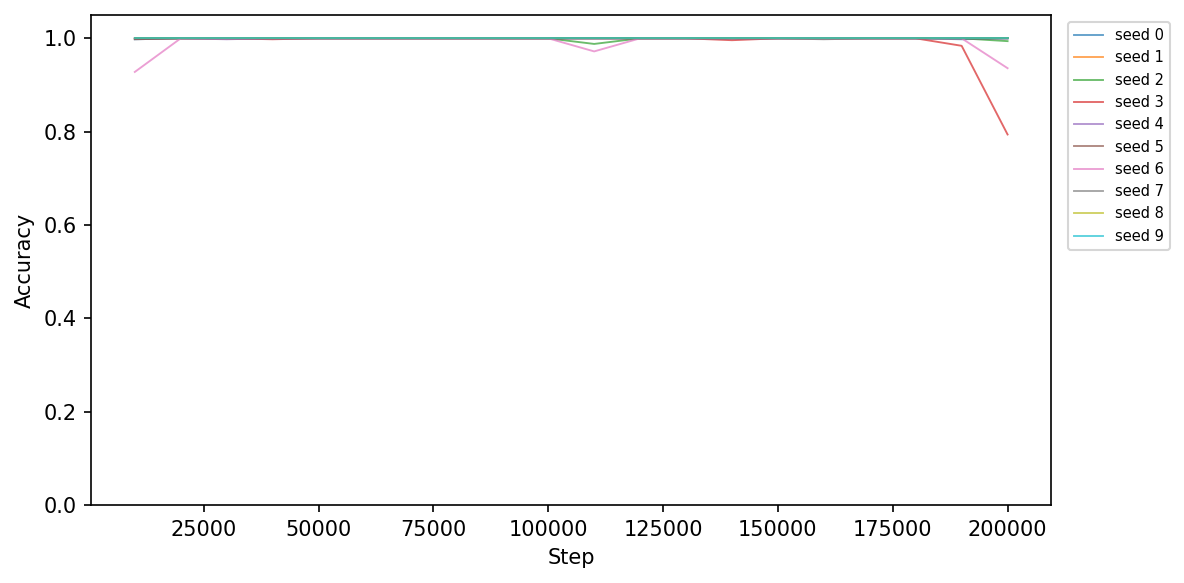}
        \end{subfigure}
        &
        \begin{subfigure}[c]{0.29\linewidth}
            \includegraphics[width=\linewidth]{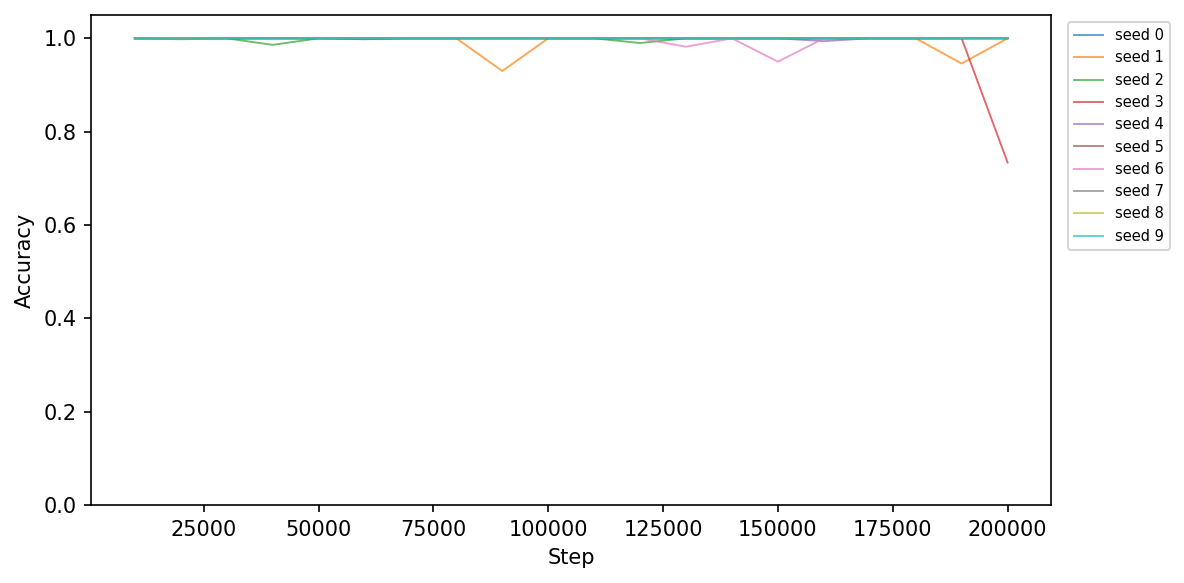}
        \end{subfigure}
        &
        \begin{subfigure}[c]{0.29\linewidth}
            \includegraphics[width=\linewidth]{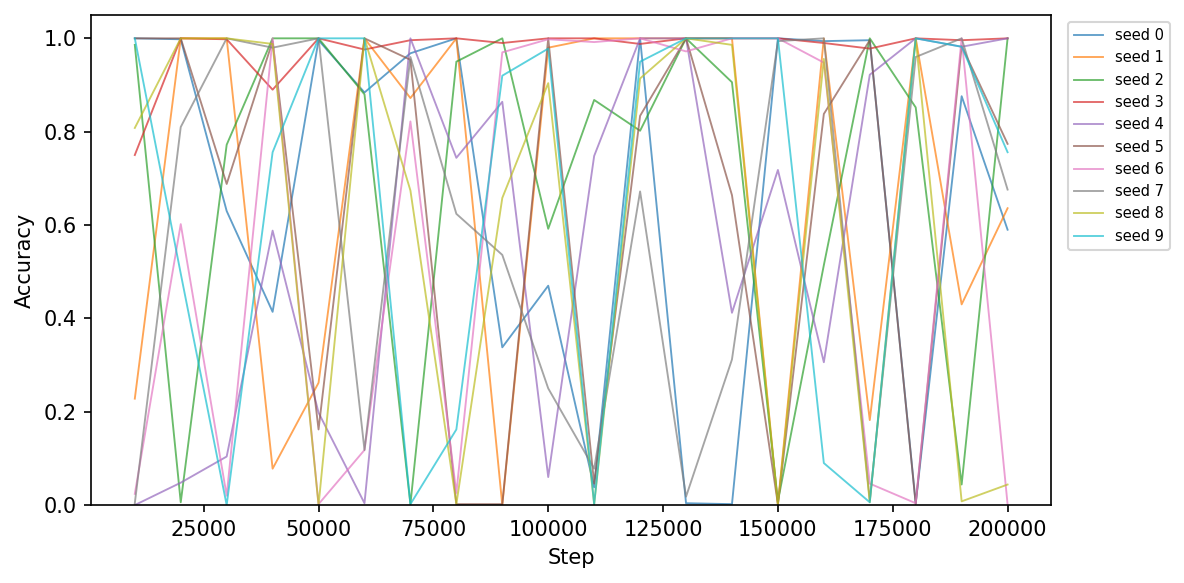}
        \end{subfigure}
        \\[1em]
        \parbox[c]{1em}{\rotatebox{90}{\textbf{$wd=0.3$}}} &
        \begin{subfigure}[c]{0.29\linewidth}
            \includegraphics[width=\linewidth]{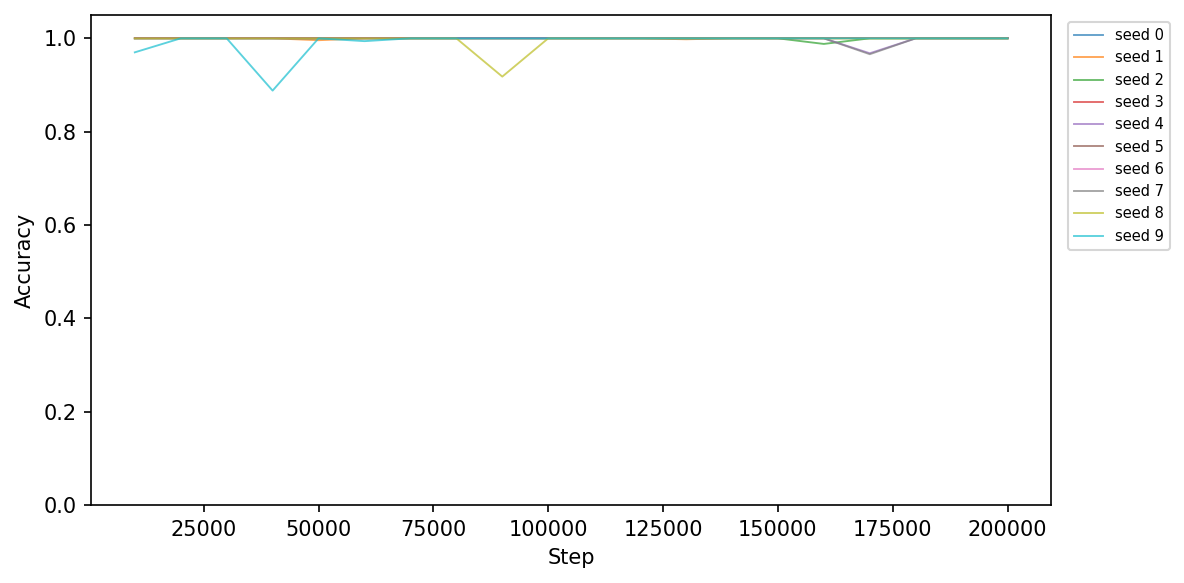}
        \end{subfigure}
        &
        \begin{subfigure}[c]{0.29\linewidth}
            \includegraphics[width=\linewidth]{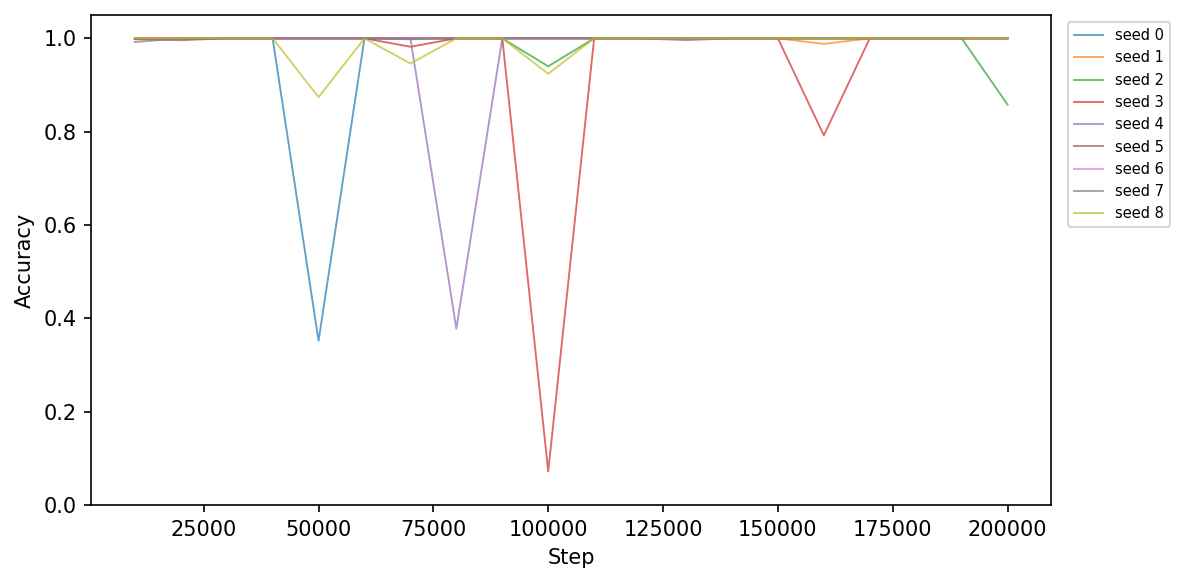}
        \end{subfigure}
        &
        \begin{subfigure}[c]{0.29\linewidth}
            \includegraphics[width=\linewidth]{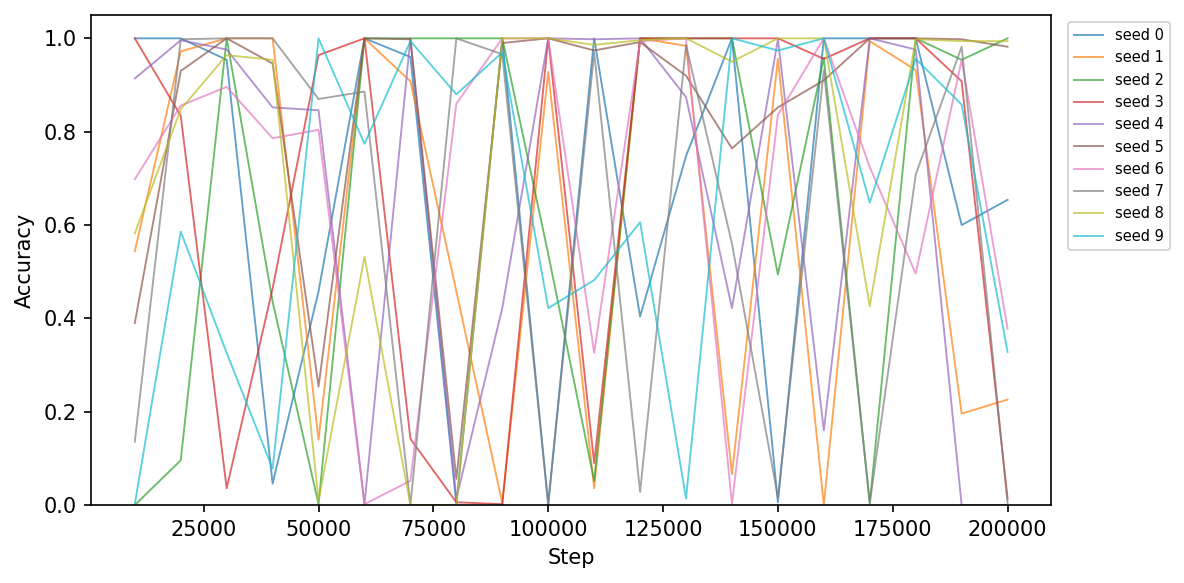}
        \end{subfigure}
        \\[1em]
        \parbox[c]{1em}{\rotatebox{90}{\textbf{$wd=0.4$}}} &
        \begin{subfigure}[c]{0.29\linewidth}
            \includegraphics[width=\linewidth]{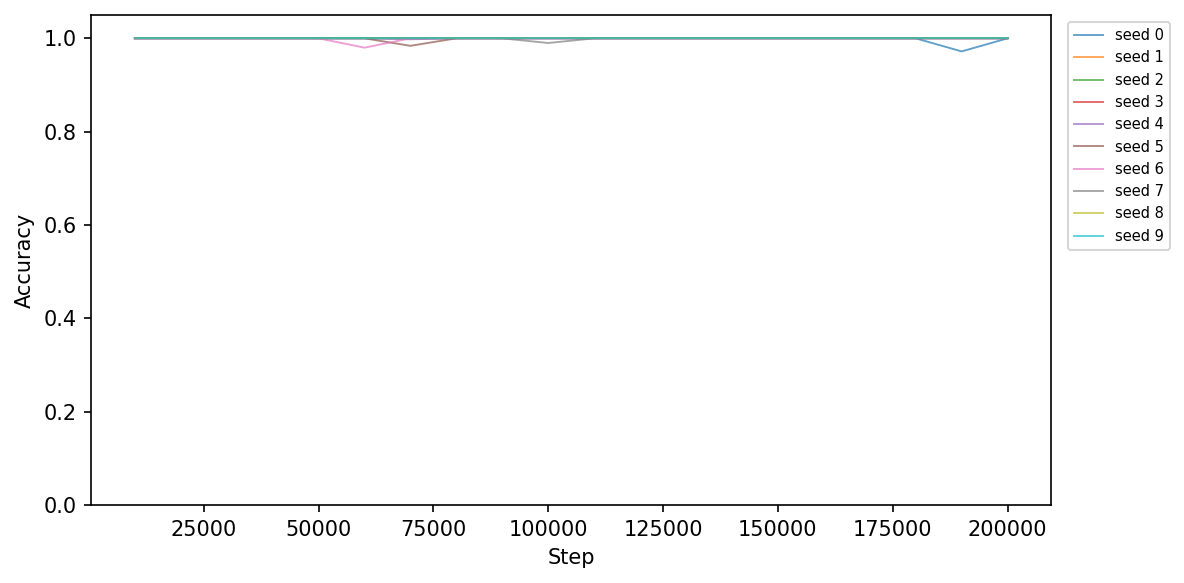}
        \end{subfigure}
        &
        \begin{subfigure}[c]{0.29\linewidth}
            \includegraphics[width=\linewidth]{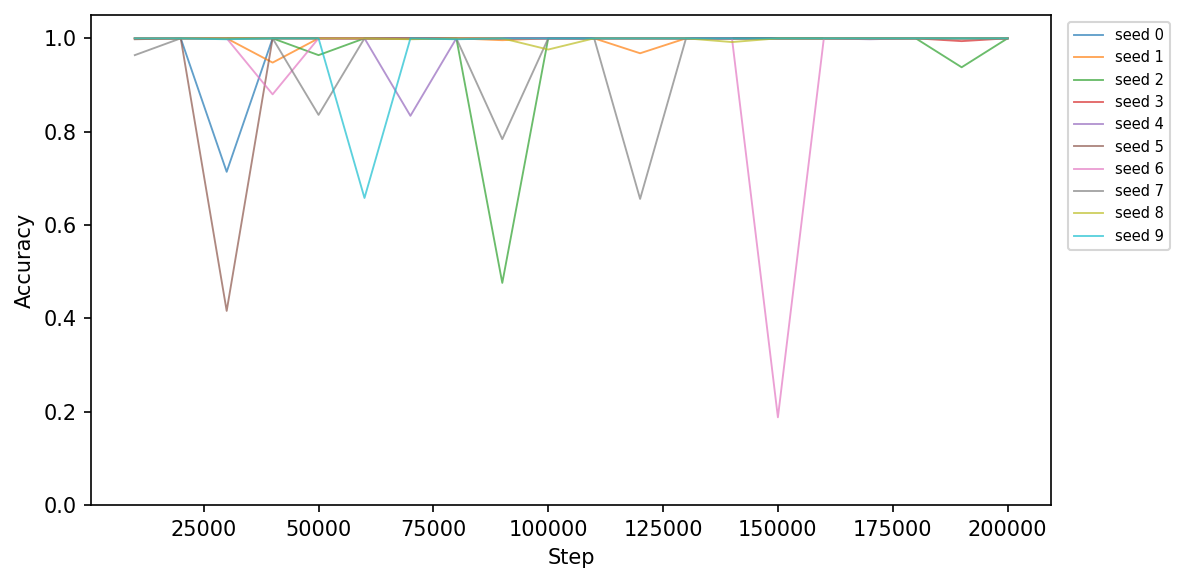}
        \end{subfigure}
        &
        \begin{subfigure}[c]{0.29\linewidth}
            \includegraphics[width=\linewidth]{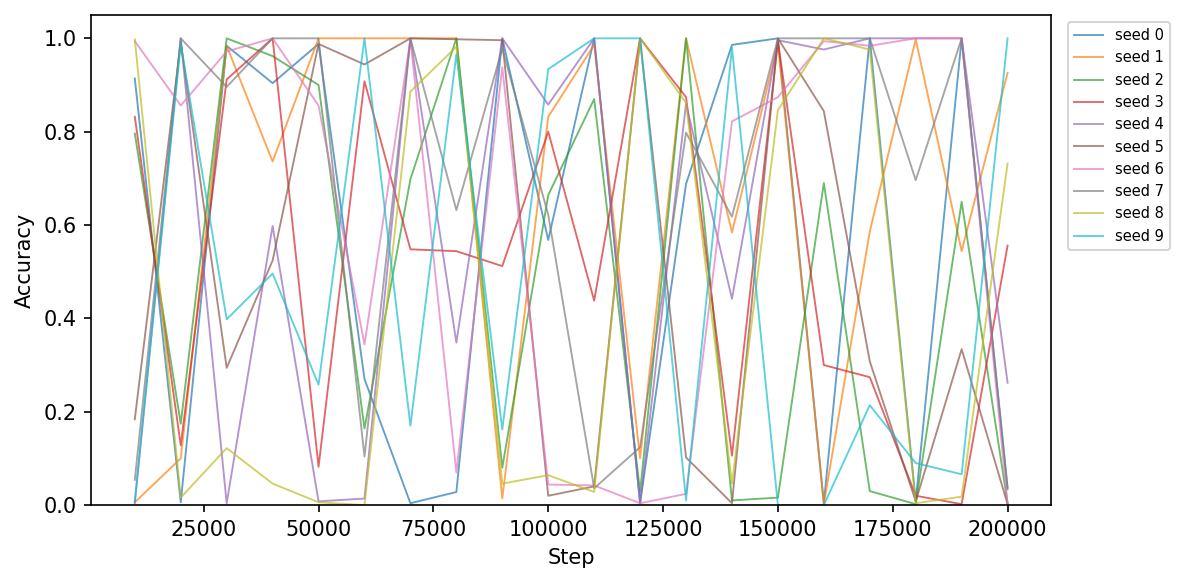}
        \end{subfigure}
        \\[1em]
        \parbox[c]{1em}{\rotatebox{90}{\textbf{$wd=0.5$}}} &
        \begin{subfigure}[c]{0.29\linewidth}
            \includegraphics[width=\linewidth]{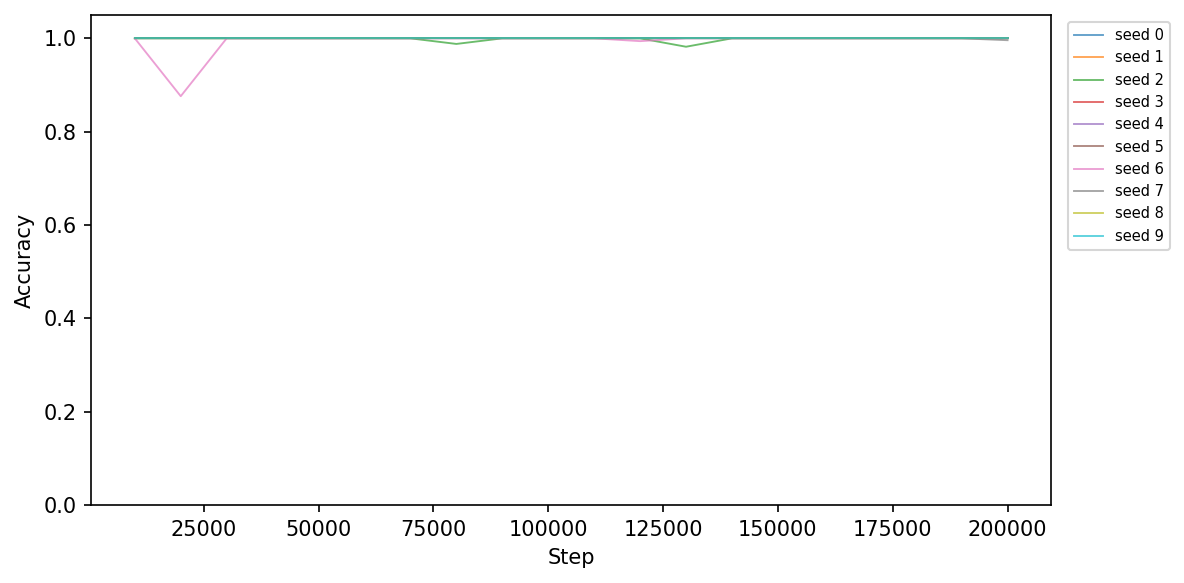}
        \end{subfigure}
        &
        \begin{subfigure}[c]{0.29\linewidth}
            \includegraphics[width=\linewidth]{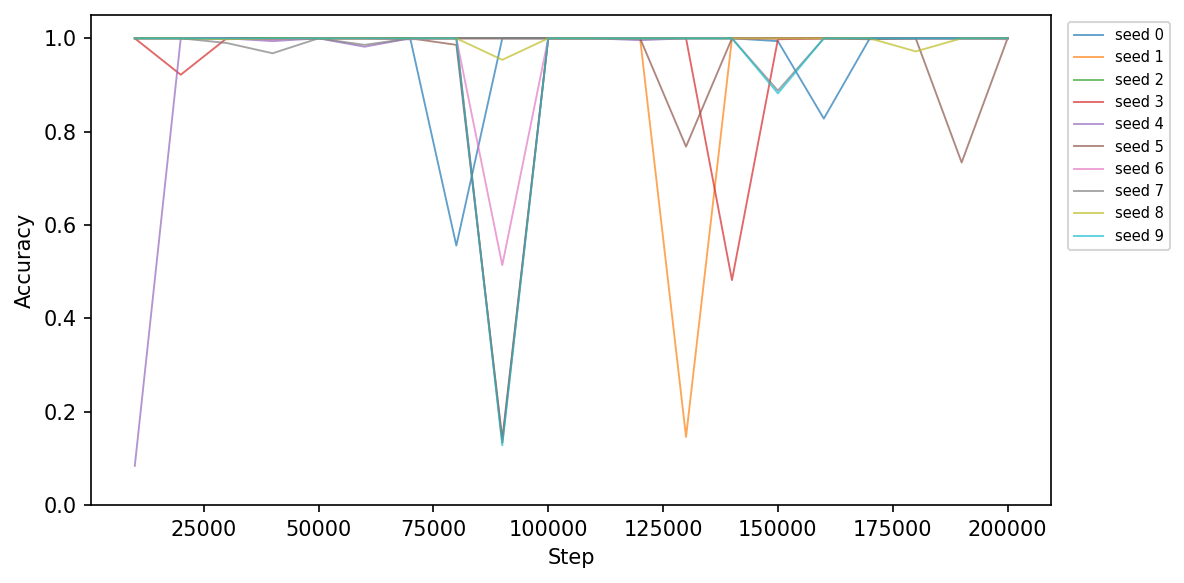}
        \end{subfigure}
        &
        \begin{subfigure}[c]{0.29\linewidth}
            \includegraphics[width=\linewidth]{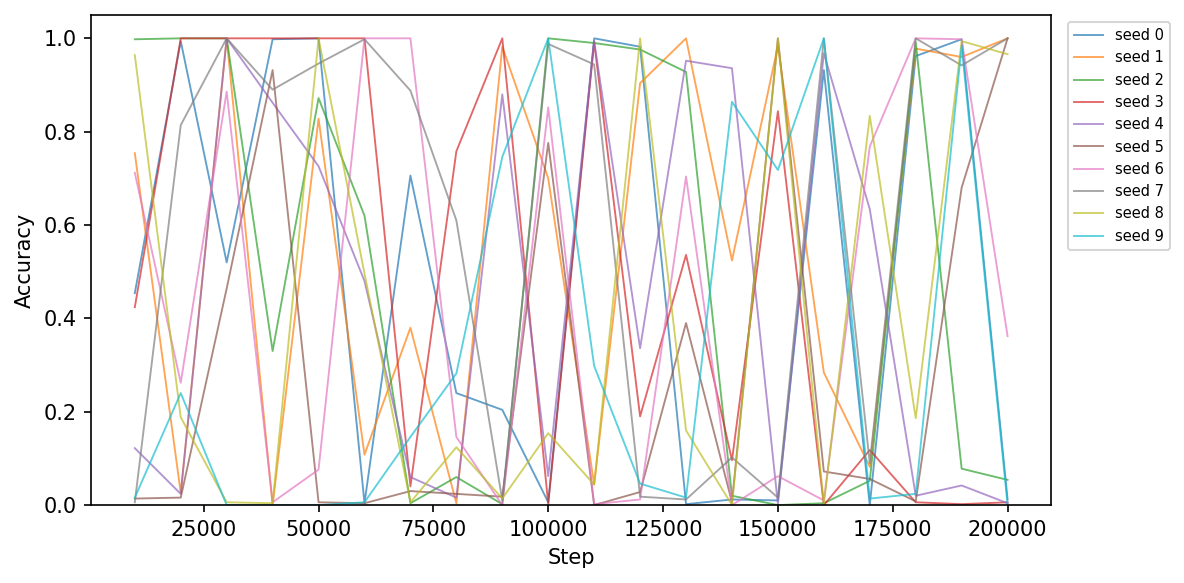}
        \end{subfigure}
        \\[1em]
    \end{tabular}
    \caption{$\mathrm{M_F}$ - Evaluation Accuracy across epochs, on different hyperparameters}
    \label{fig:eval_FF}
\end{figure*}

\begin{figure*}[t]
    \centering
    \begin{tabular}{c ccc}
        & \textbf{$lr=0.0001$} & \textbf{$lr=0.001$} & \textbf{$lr=0.003$}\\[0.5em]
        \parbox[c]{1em}{\rotatebox{90}{\textbf{$wd=0.01$}}} &
        \begin{subfigure}[c]{0.29\linewidth}
            \includegraphics[width=\linewidth]{Fig/loss_le/FH/FH_lr0.0001_wd0.01_train_loss_linear.png}
        \end{subfigure}
        &
        \begin{subfigure}[c]{0.29\linewidth}
            \includegraphics[width=\linewidth]{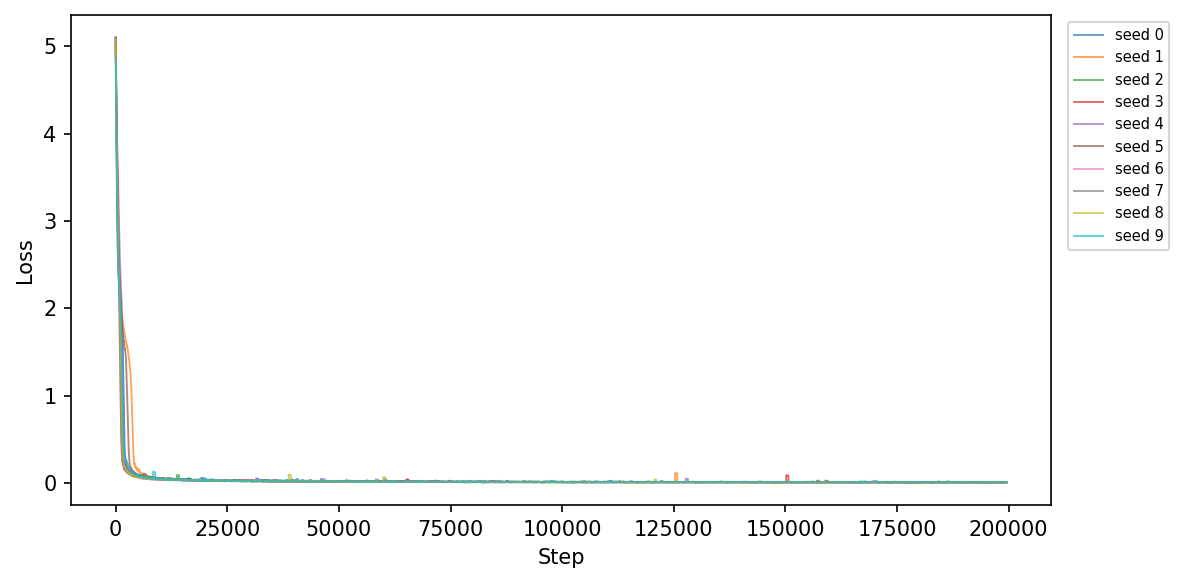}
        \end{subfigure}
        &
        \begin{subfigure}[c]{0.29\linewidth}
            \includegraphics[width=\linewidth]{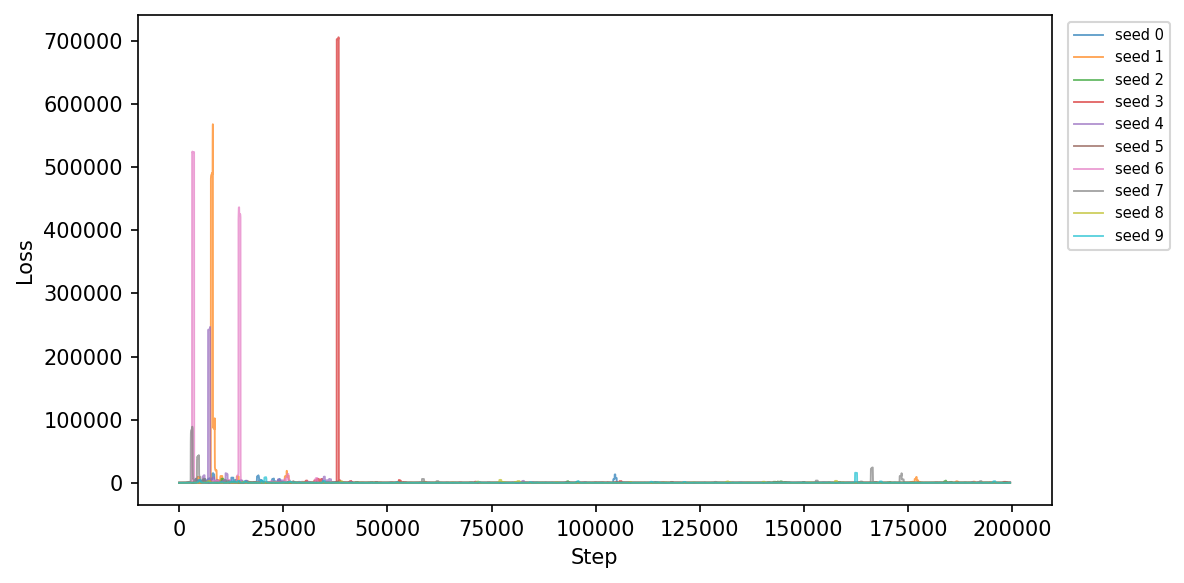}
        \end{subfigure}
        \\[1em]
        \parbox[c]{1em}{\rotatebox{90}{\textbf{$wd=0.1$}}} &
        \begin{subfigure}[c]{0.29\linewidth}
            \includegraphics[width=\linewidth]{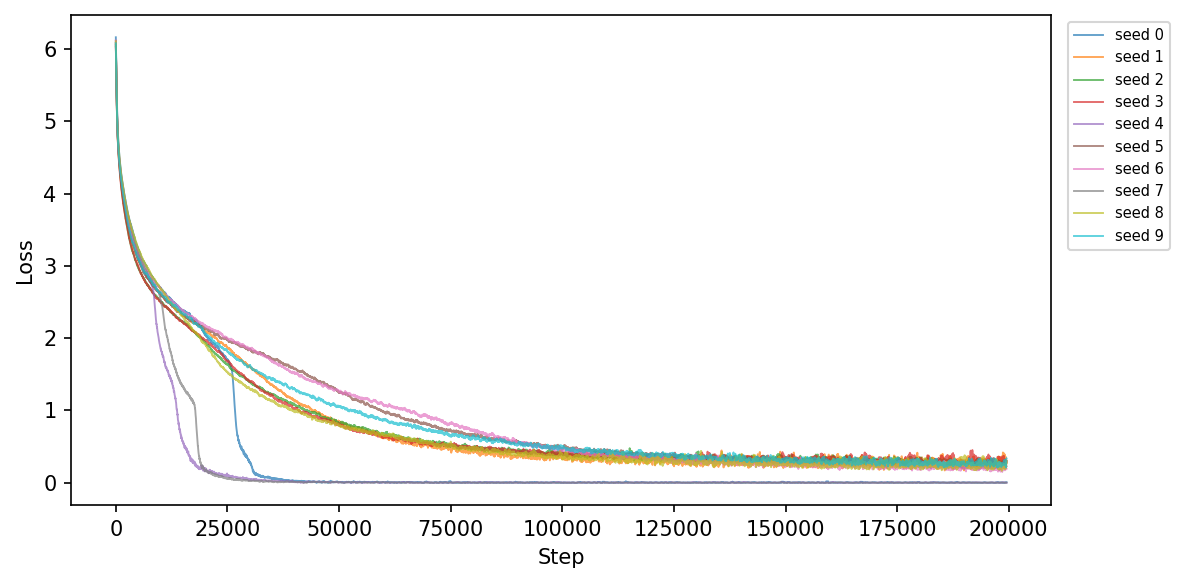}
        \end{subfigure}
        &
        \begin{subfigure}[c]{0.29\linewidth}
            \includegraphics[width=\linewidth]{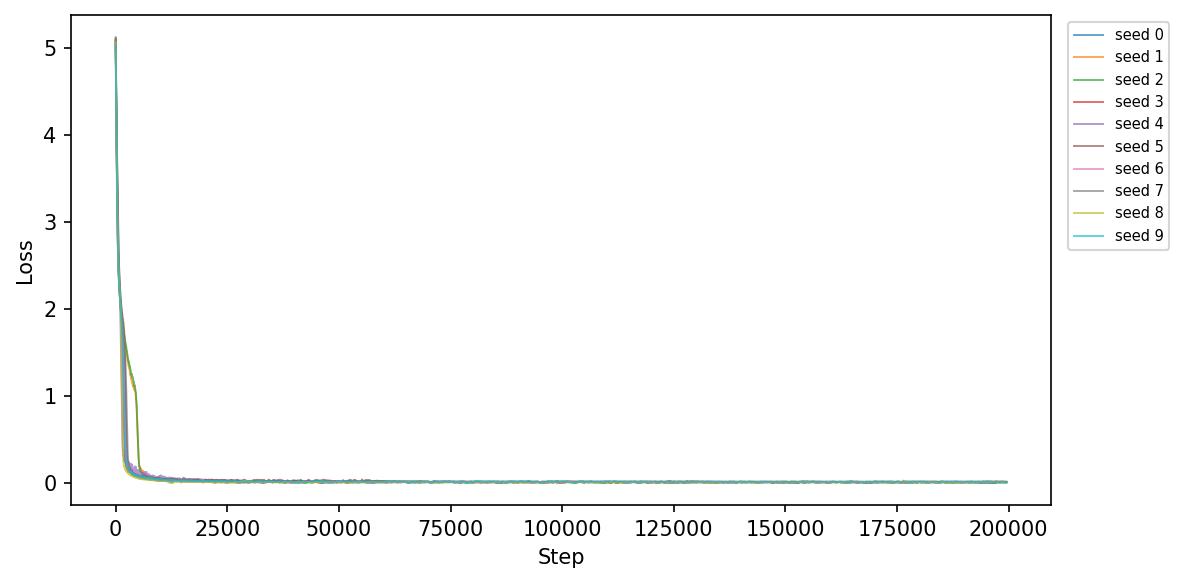}
        \end{subfigure}
        &
        \begin{subfigure}[c]{0.29\linewidth}
            \includegraphics[width=\linewidth]{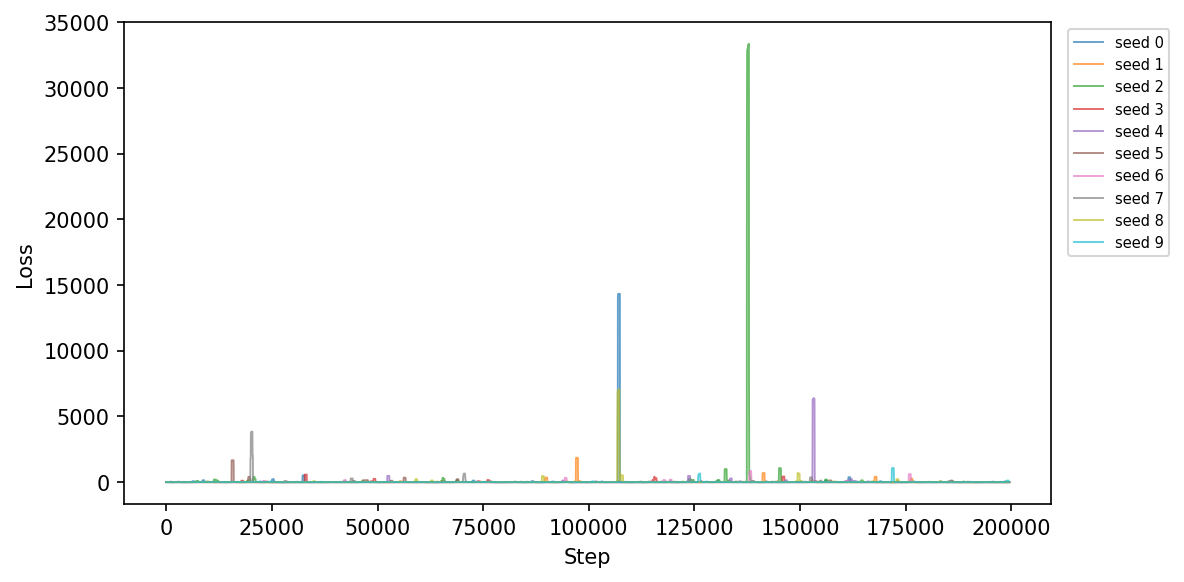}
        \end{subfigure}
        \\[1em]
        \parbox[c]{1em}{\rotatebox{90}{\textbf{$wd=0.2$}}} &
        \begin{subfigure}[c]{0.29\linewidth}
            \includegraphics[width=\linewidth]{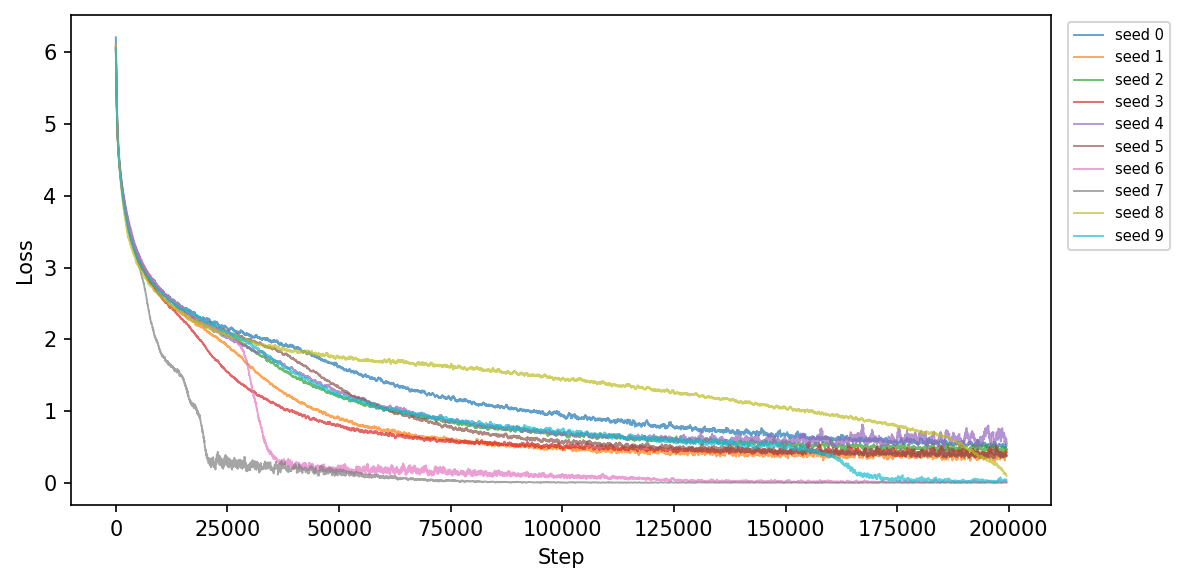}
        \end{subfigure}
        &
        \begin{subfigure}[c]{0.29\linewidth}
            \includegraphics[width=\linewidth]{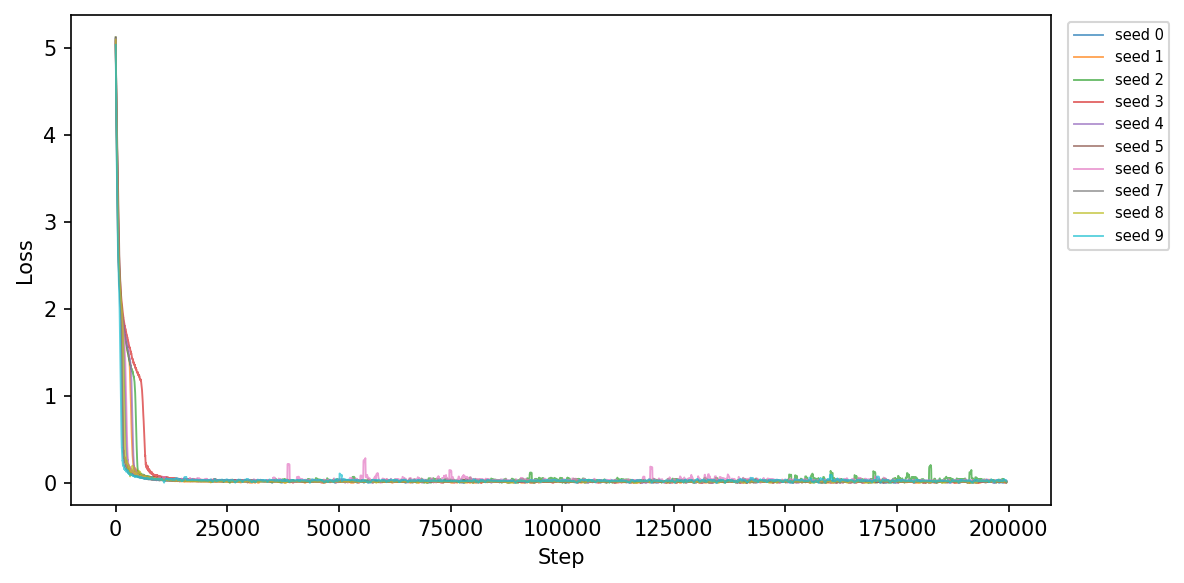}
        \end{subfigure}
        &
        \begin{subfigure}[c]{0.29\linewidth}
            \includegraphics[width=\linewidth]{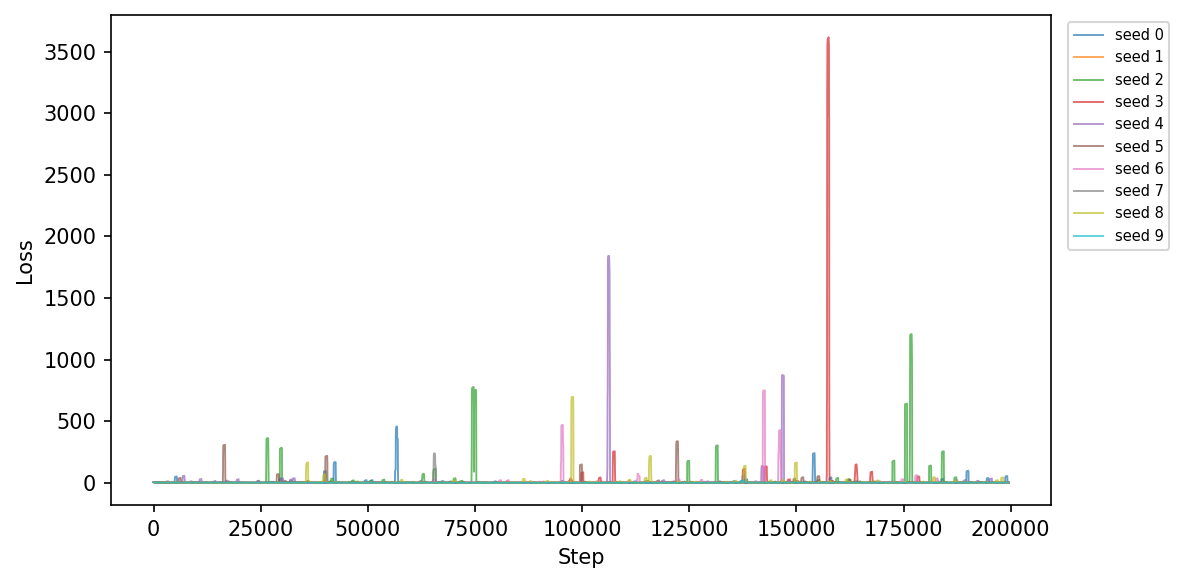}
        \end{subfigure}
        \\[1em]
        \parbox[c]{1em}{\rotatebox{90}{\textbf{$wd=0.3$}}} &
        \begin{subfigure}[c]{0.29\linewidth}
            \includegraphics[width=\linewidth]{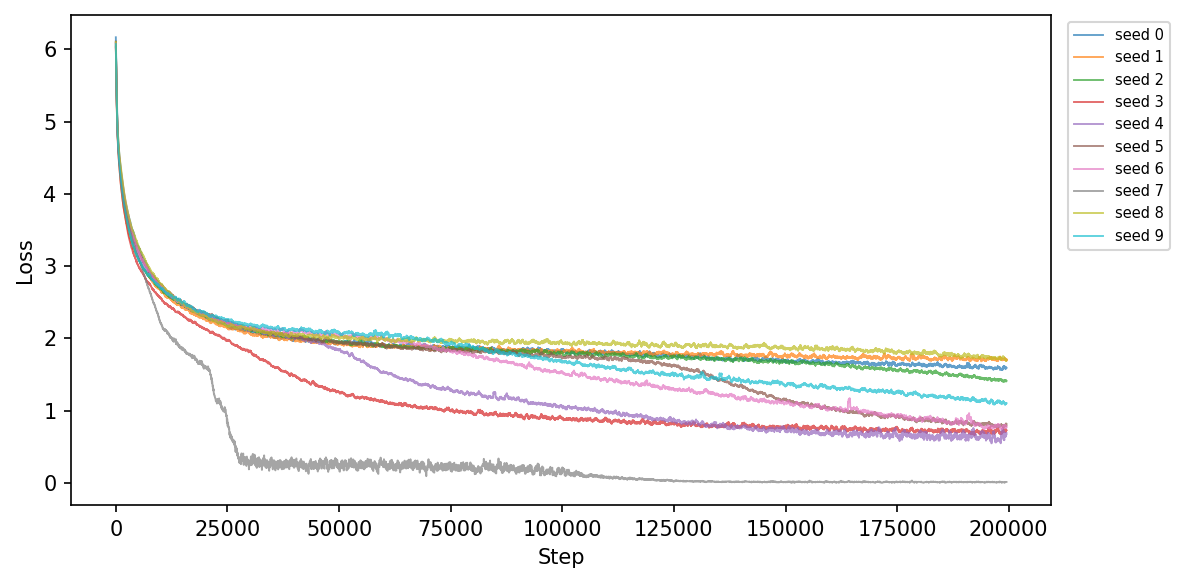}
        \end{subfigure}
        &
        \begin{subfigure}[c]{0.29\linewidth}
            \includegraphics[width=\linewidth]{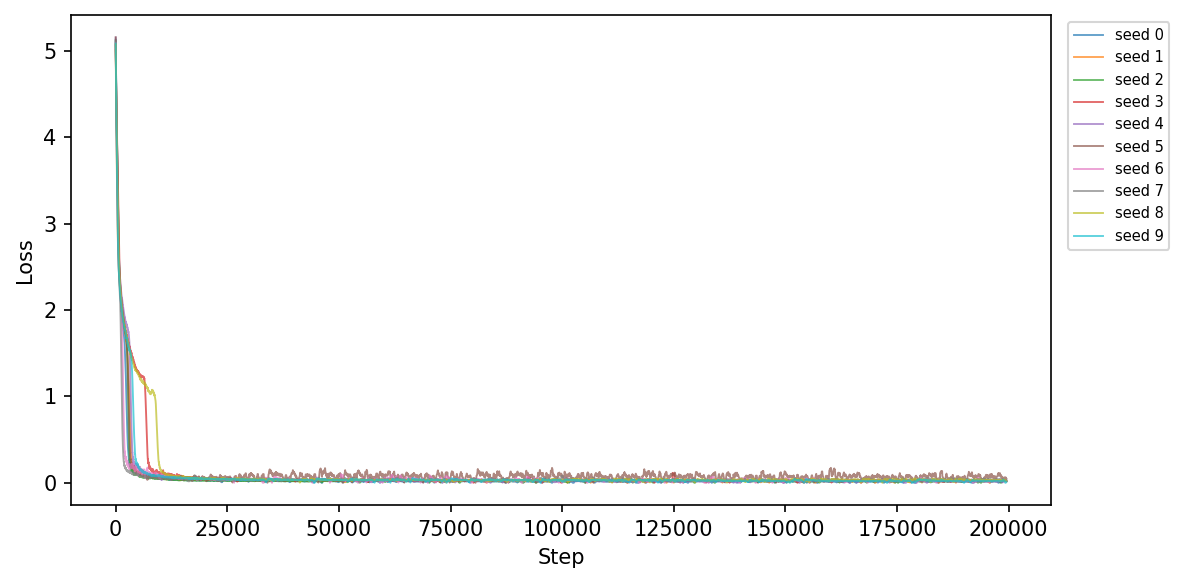}
        \end{subfigure}
        &
        \begin{subfigure}[c]{0.29\linewidth}
            \includegraphics[width=\linewidth]{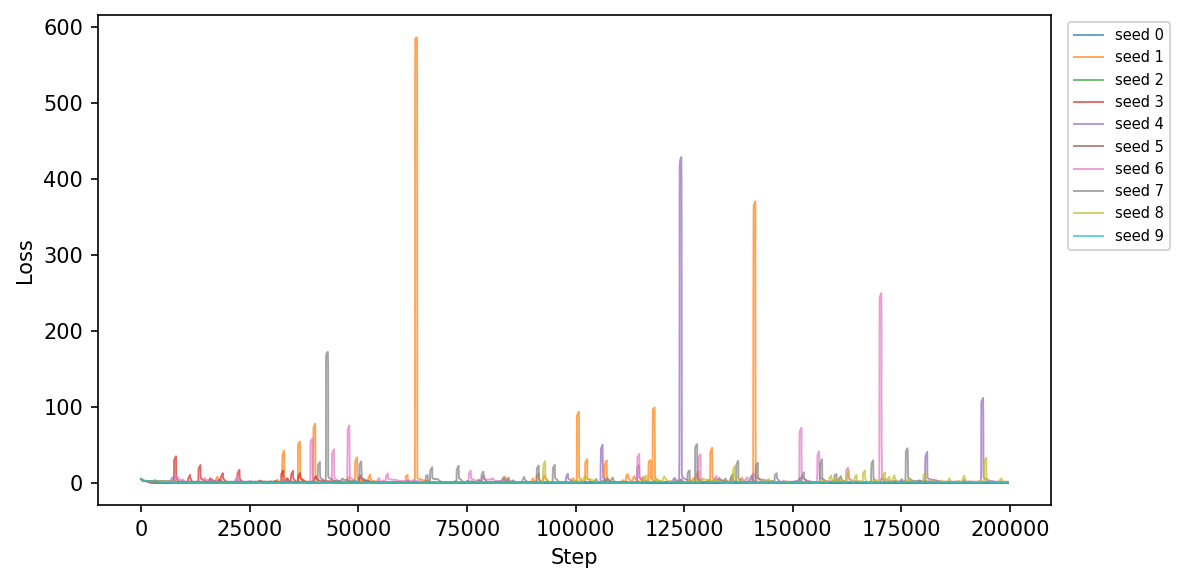}
        \end{subfigure}
        \\[1em]
        \parbox[c]{1em}{\rotatebox{90}{\textbf{$wd=0.4$}}} &
        \begin{subfigure}[c]{0.29\linewidth}
            \includegraphics[width=\linewidth]{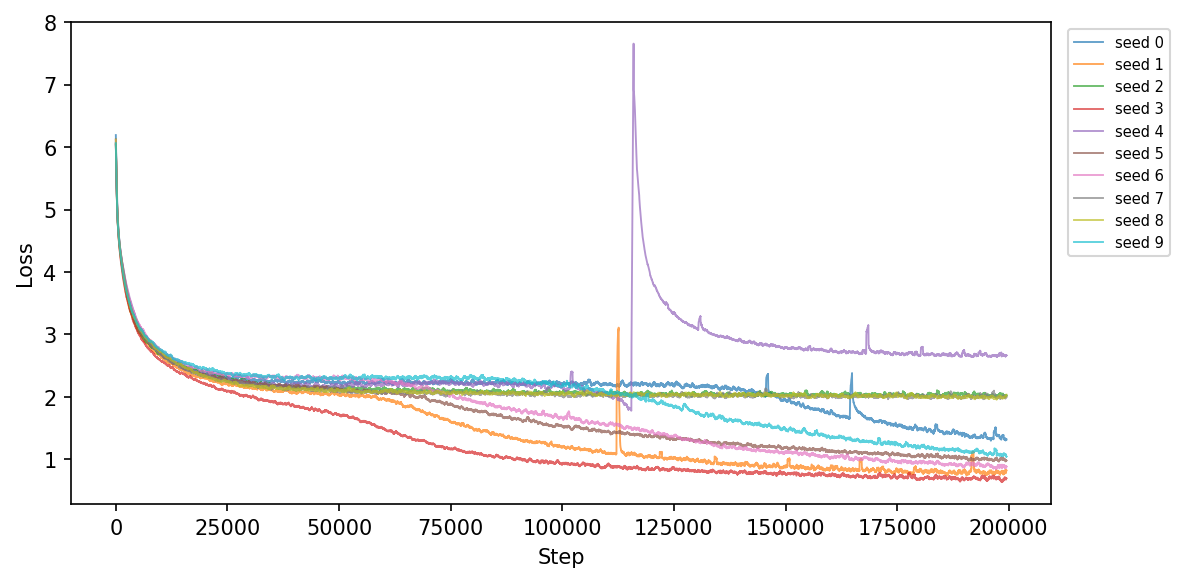}
        \end{subfigure}
        &
        \begin{subfigure}[c]{0.29\linewidth}
            \includegraphics[width=\linewidth]{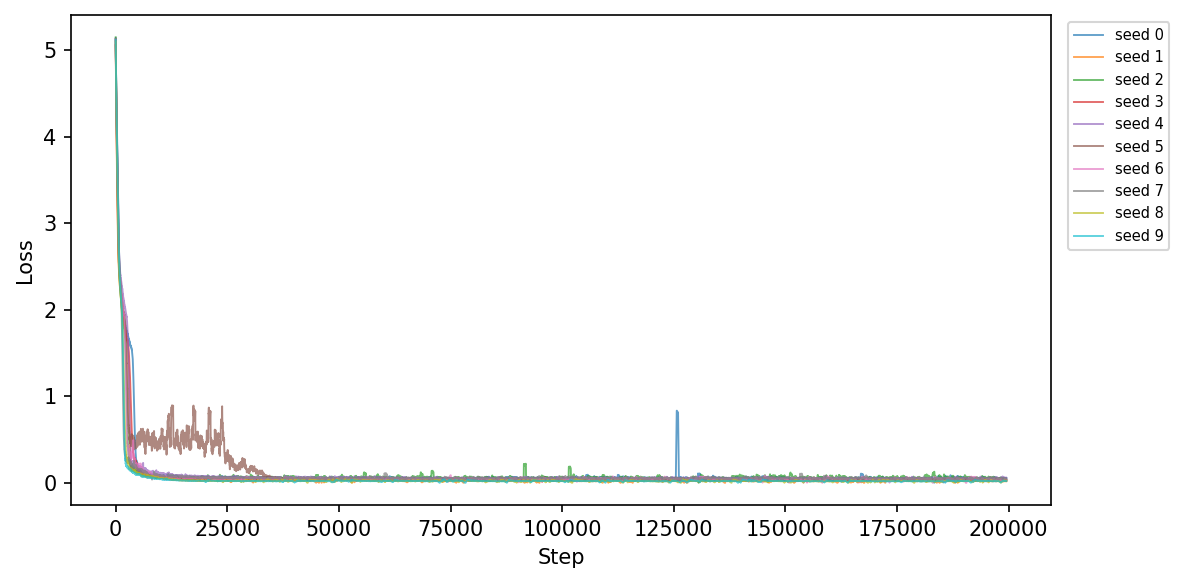}
        \end{subfigure}
        &
        \begin{subfigure}[c]{0.29\linewidth}
            \includegraphics[width=\linewidth]{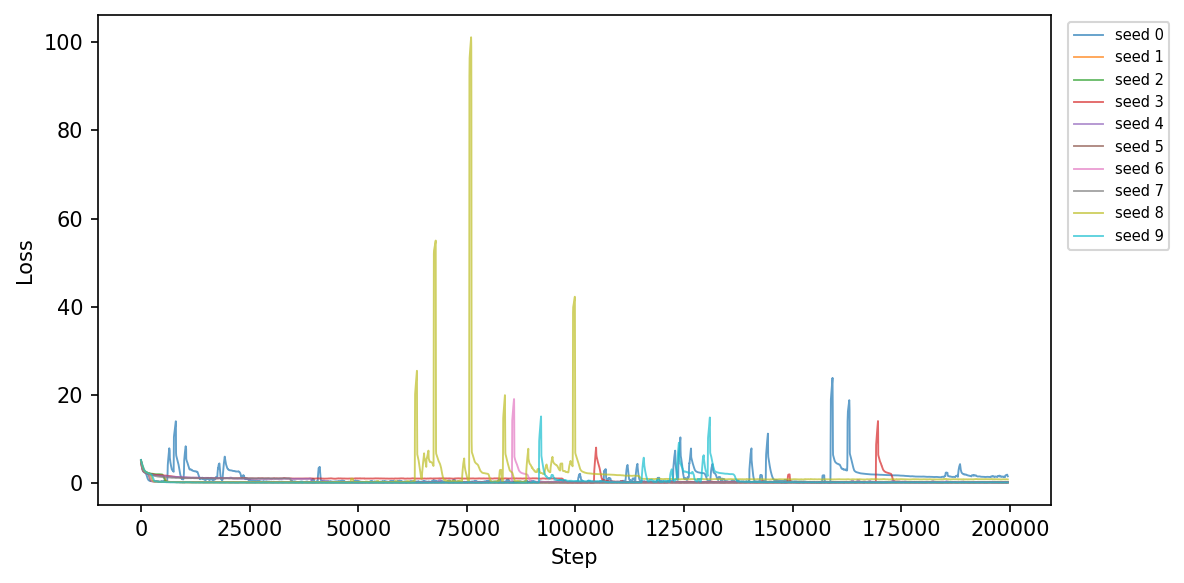}
        \end{subfigure}
        \\[1em]
        \parbox[c]{1em}{\rotatebox{90}{\textbf{$wd=0.5$}}} &
        \begin{subfigure}[c]{0.29\linewidth}
            \includegraphics[width=\linewidth]{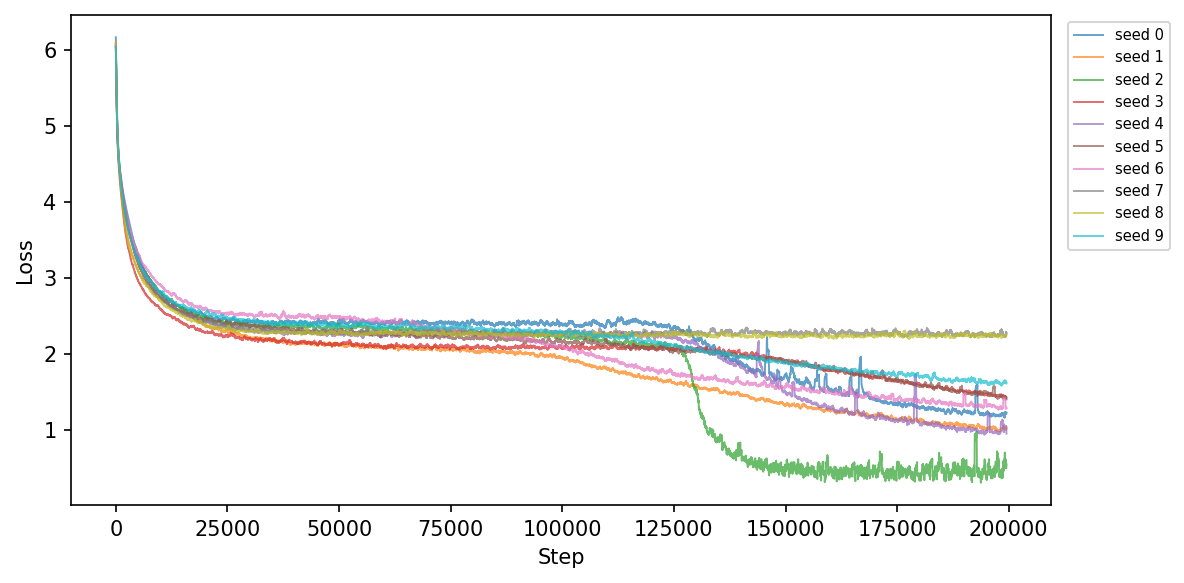}
        \end{subfigure}
        &
        \begin{subfigure}[c]{0.29\linewidth}
            \includegraphics[width=\linewidth]{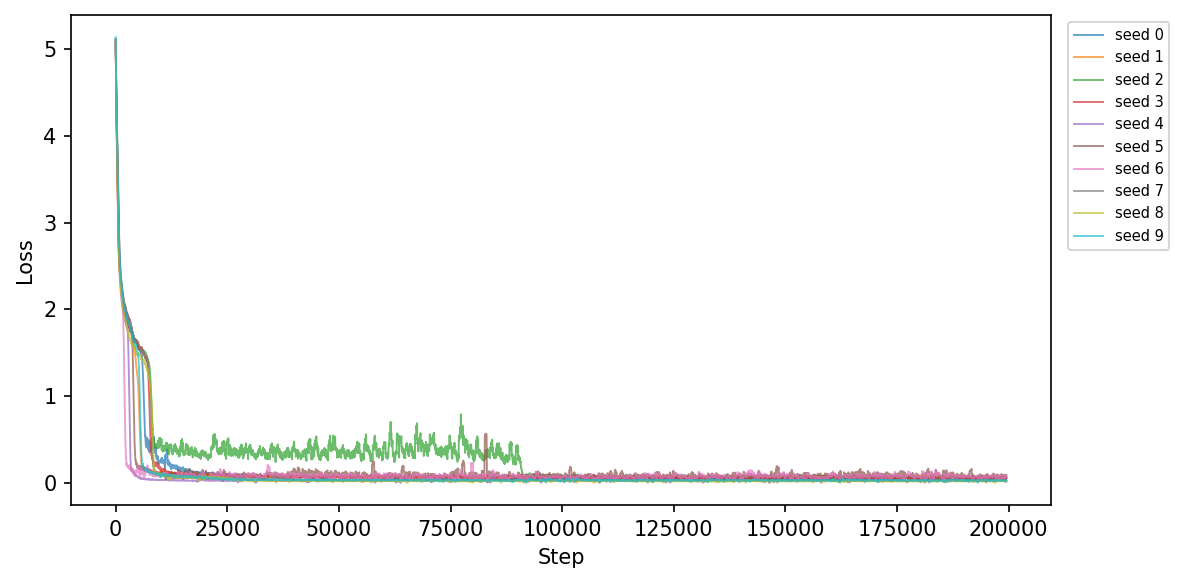}
        \end{subfigure}
        &
        \begin{subfigure}[c]{0.29\linewidth}
            \includegraphics[width=\linewidth]{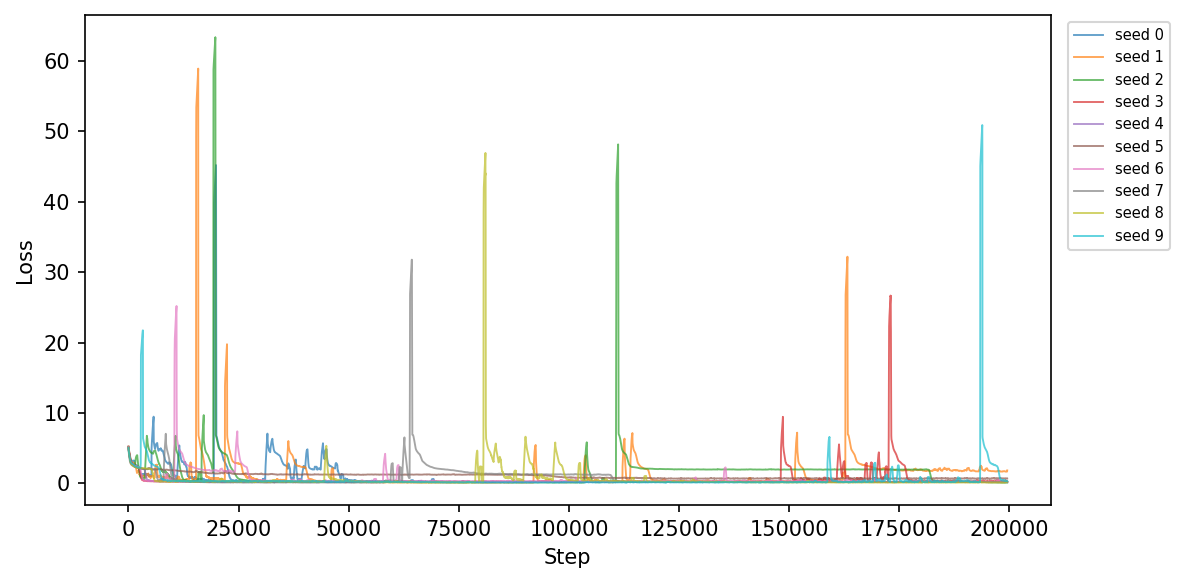}
        \end{subfigure}
        \\[1em]
    \end{tabular}
    \caption{$\mathrm{M_{F_{\mathrm{in}},\, H_{\mathrm{out}}}}$ - Train Loss across epochs, on different hyperparameters}
    \label{fig:loss_FH}
\end{figure*}

\begin{figure*}[t]
    \centering
    \begin{tabular}{c ccc}
        & \textbf{$lr=0.0001$} & \textbf{$lr=0.001$} & \textbf{$lr=0.003$}\\[0.5em]
        \parbox[c]{1em}{\rotatebox{90}{\textbf{$wd=0.01$}}} &
        \begin{subfigure}[c]{0.29\linewidth}
            \includegraphics[width=\linewidth]{Fig/loss_le/FH/FH_lr0.0001_wd0.01_random_eval_acc.png}
        \end{subfigure}
        &
        \begin{subfigure}[c]{0.29\linewidth}
            \includegraphics[width=\linewidth]{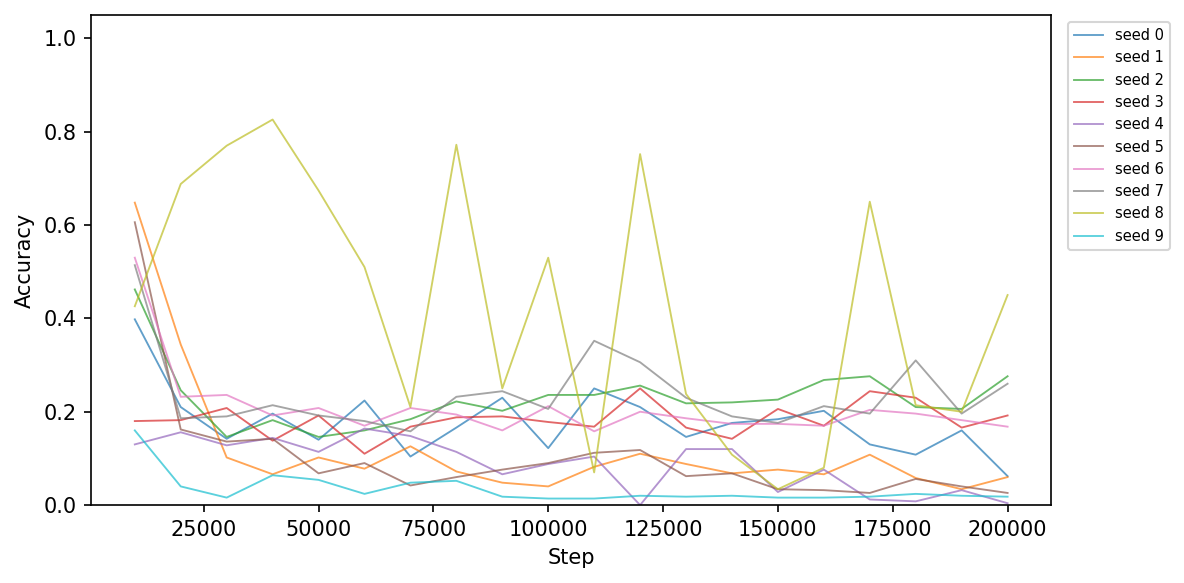}
        \end{subfigure}
        &
        \begin{subfigure}[c]{0.29\linewidth}
            \includegraphics[width=\linewidth]{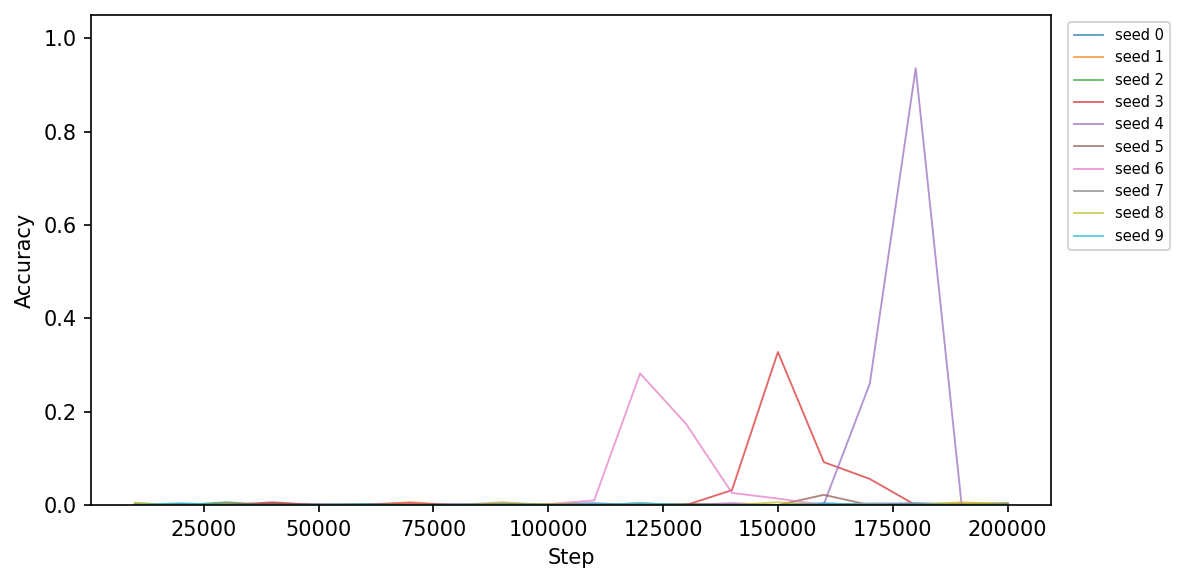}
        \end{subfigure}
        \\[1em]
        \parbox[c]{1em}{\rotatebox{90}{\textbf{$wd=0.1$}}} &
        \begin{subfigure}[c]{0.29\linewidth}
            \includegraphics[width=\linewidth]{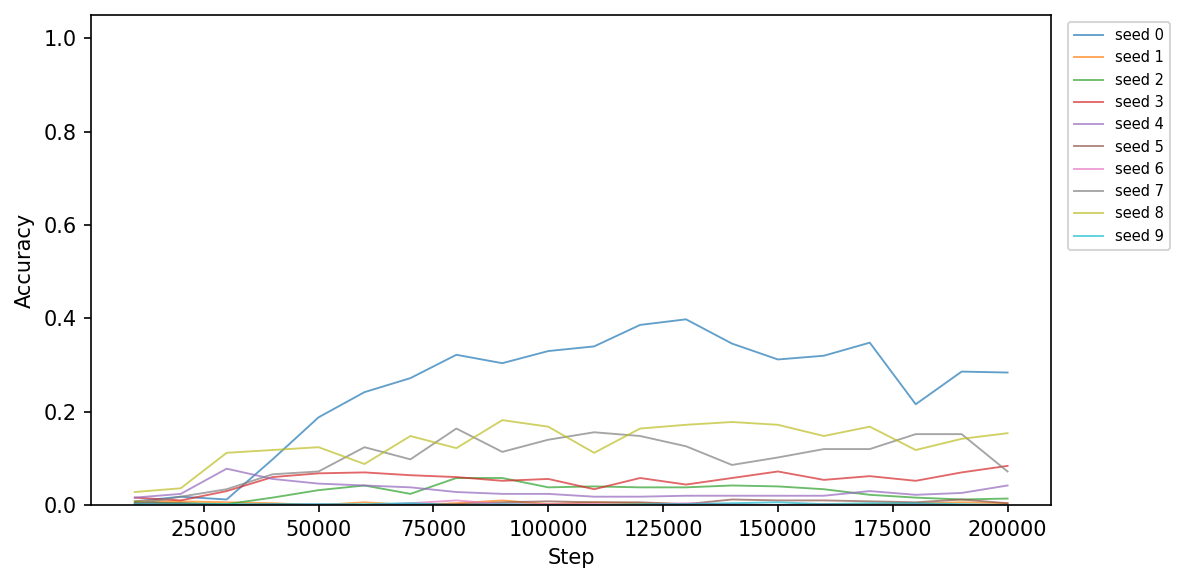}
        \end{subfigure}
        &
        \begin{subfigure}[c]{0.29\linewidth}
            \includegraphics[width=\linewidth]{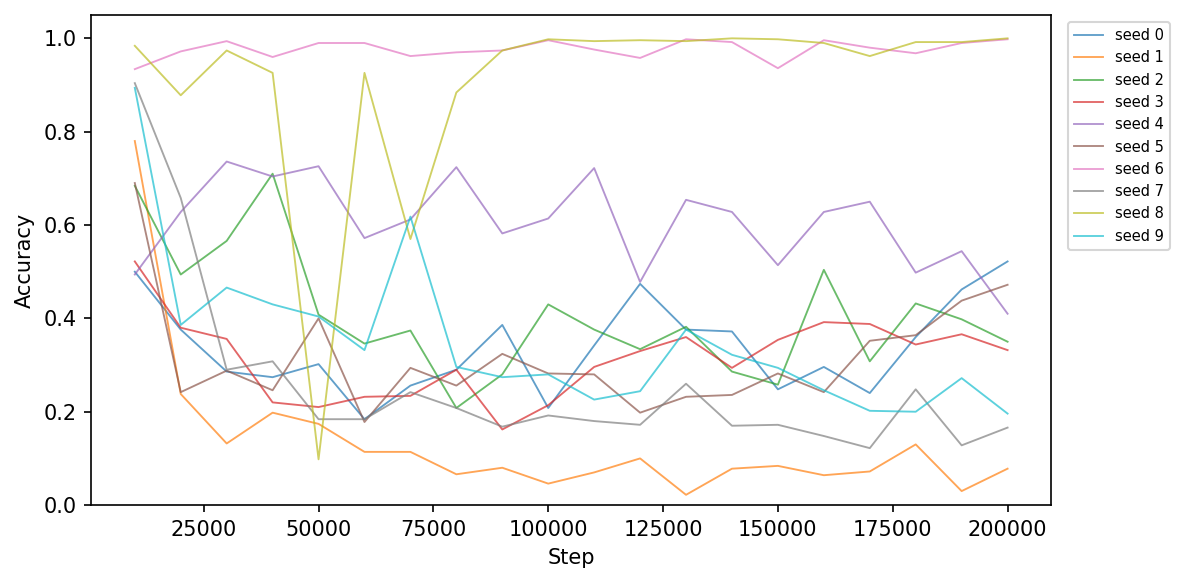}
        \end{subfigure}
        &
        \begin{subfigure}[c]{0.29\linewidth}
            \includegraphics[width=\linewidth]{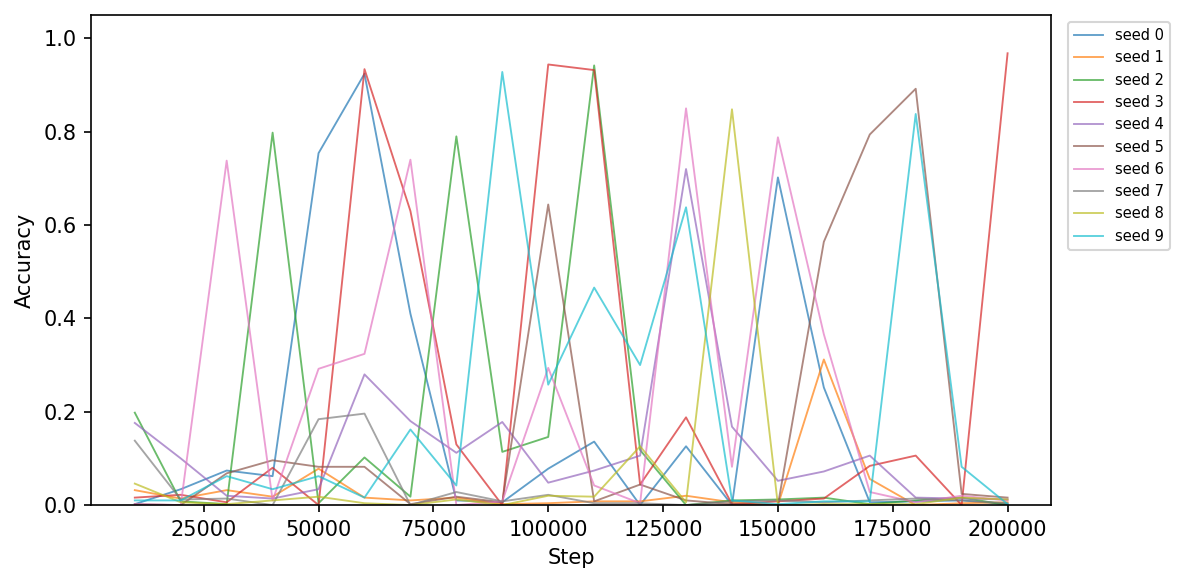}
        \end{subfigure}
        \\[1em]
        \parbox[c]{1em}{\rotatebox{90}{\textbf{$wd=0.2$}}} &
        \begin{subfigure}[c]{0.29\linewidth}
            \includegraphics[width=\linewidth]{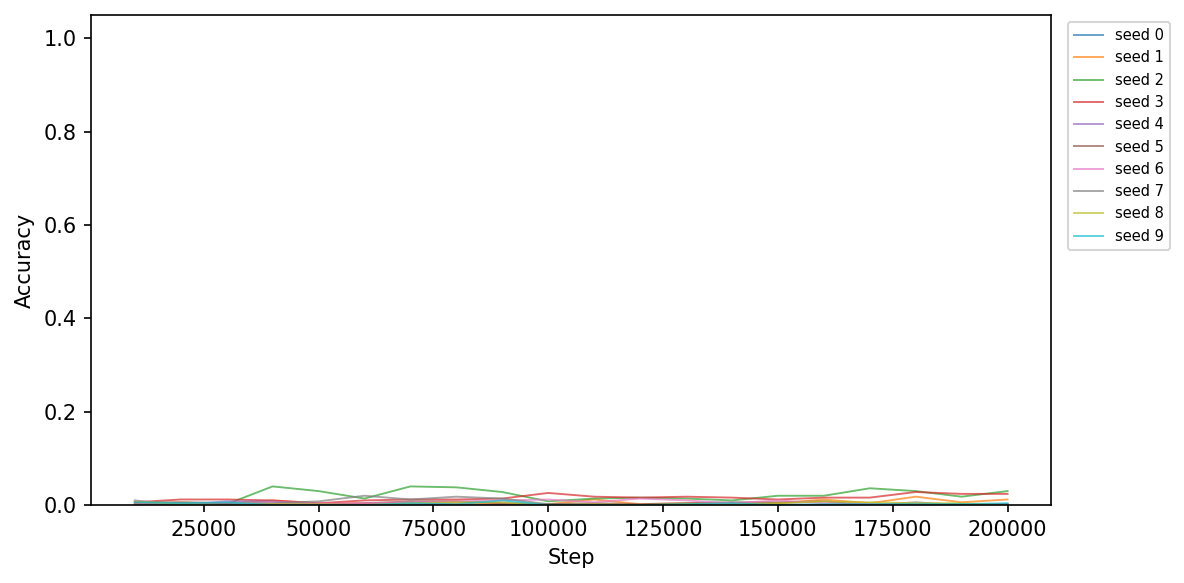}
        \end{subfigure}
        &
        \begin{subfigure}[c]{0.29\linewidth}
            \includegraphics[width=\linewidth]{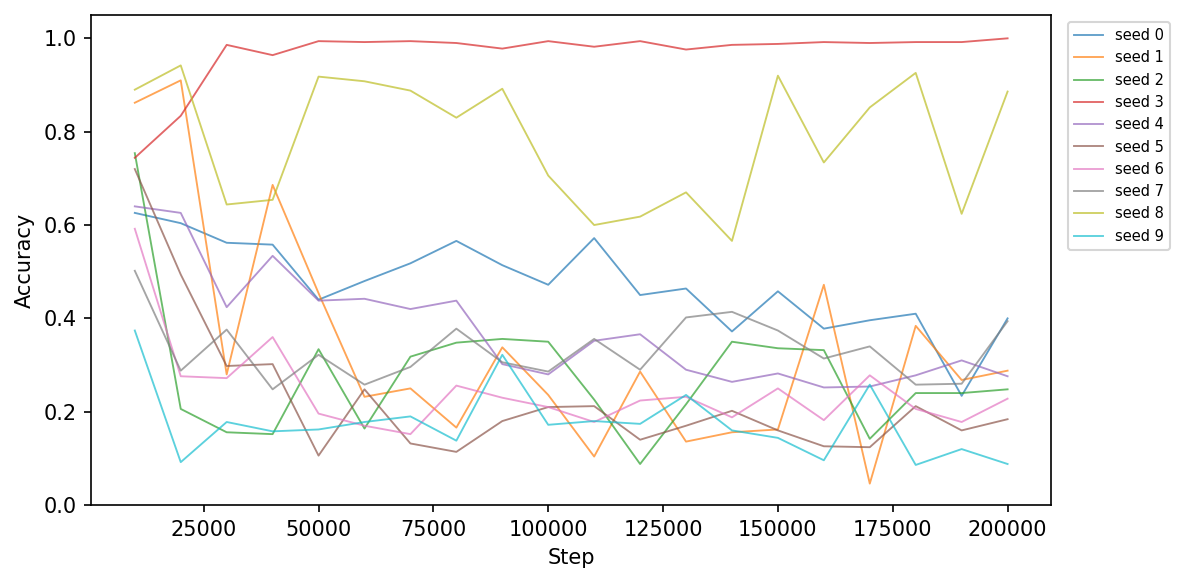}
        \end{subfigure}
        &
        \begin{subfigure}[c]{0.29\linewidth}
            \includegraphics[width=\linewidth]{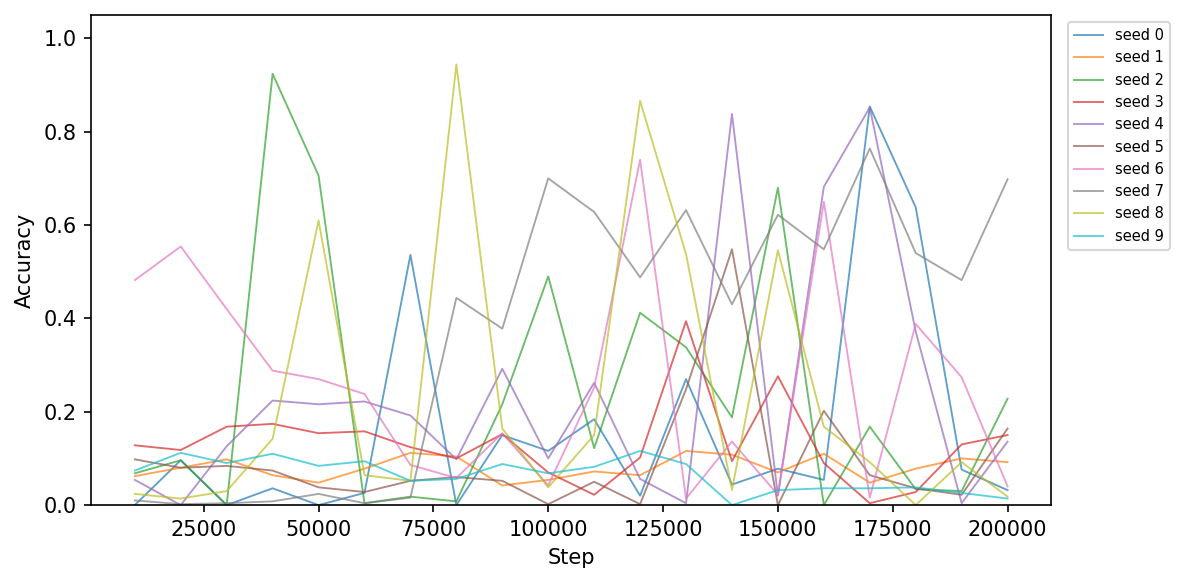}
        \end{subfigure}
        \\[1em]
        \parbox[c]{1em}{\rotatebox{90}{\textbf{$wd=0.3$}}} &
        \begin{subfigure}[c]{0.29\linewidth}
            \includegraphics[width=\linewidth]{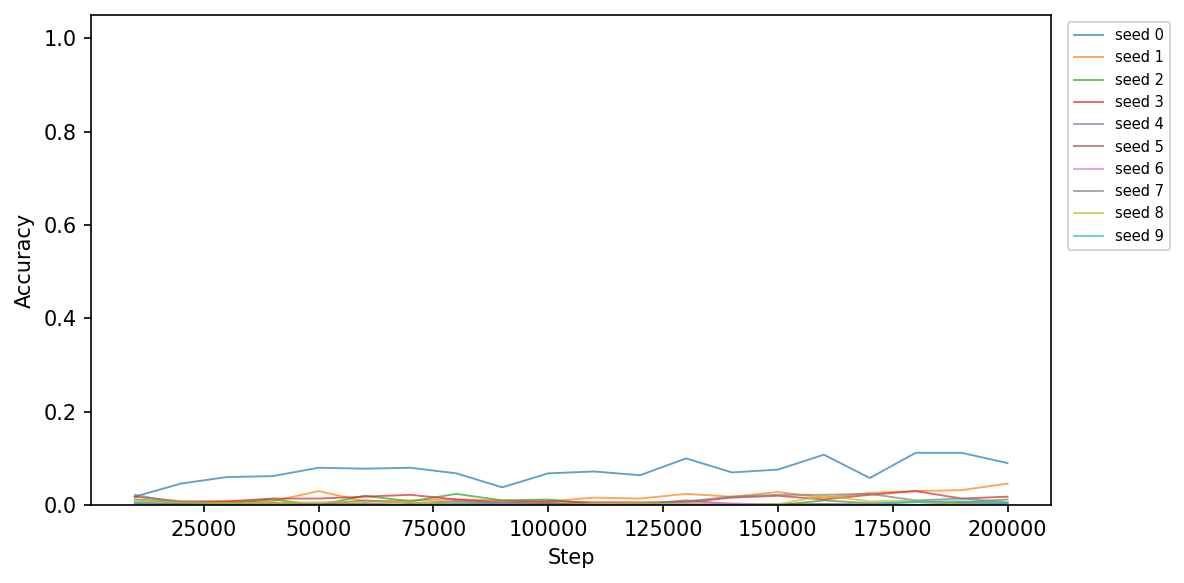}
        \end{subfigure}
        &
        \begin{subfigure}[c]{0.29\linewidth}
            \includegraphics[width=\linewidth]{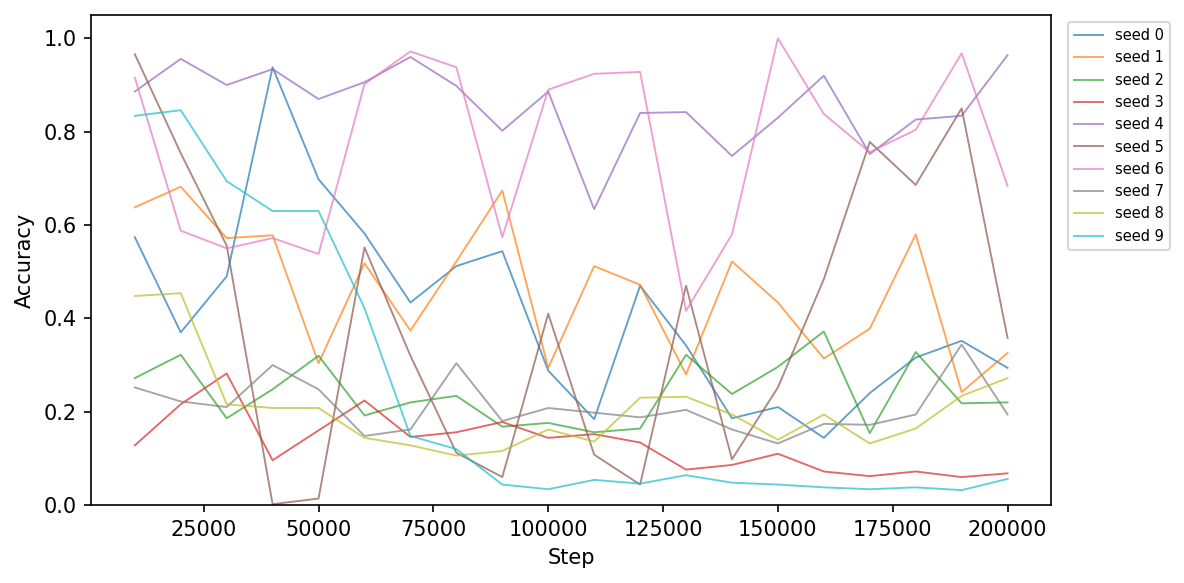}
        \end{subfigure}
        &
        \begin{subfigure}[c]{0.29\linewidth}
            \includegraphics[width=\linewidth]{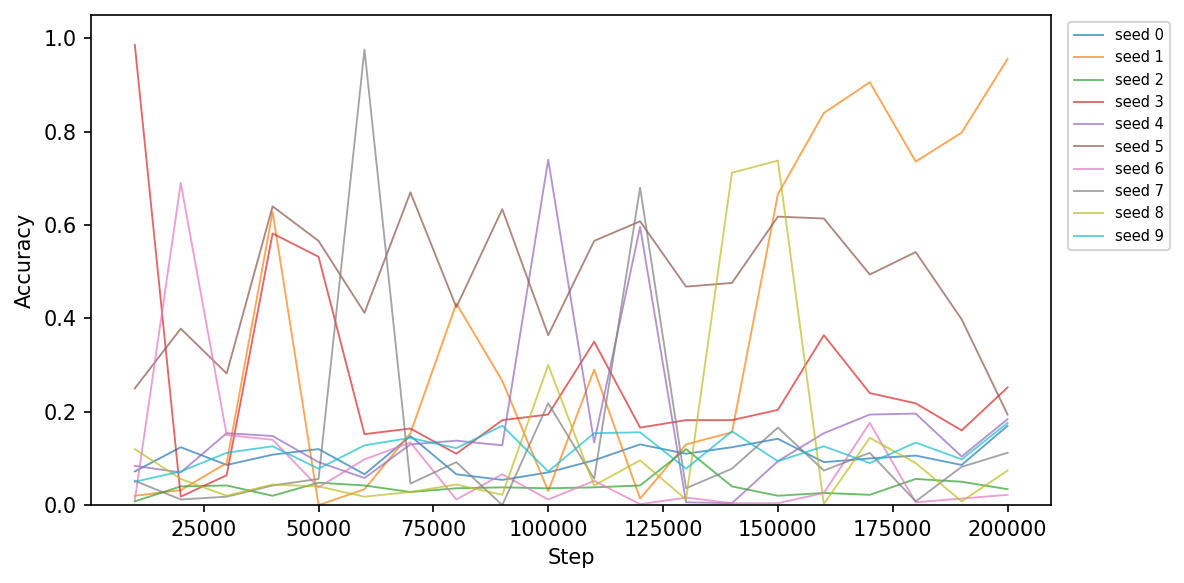}
        \end{subfigure}
        \\[1em]
        \parbox[c]{1em}{\rotatebox{90}{\textbf{$wd=0.4$}}} &
        \begin{subfigure}[c]{0.29\linewidth}
            \includegraphics[width=\linewidth]{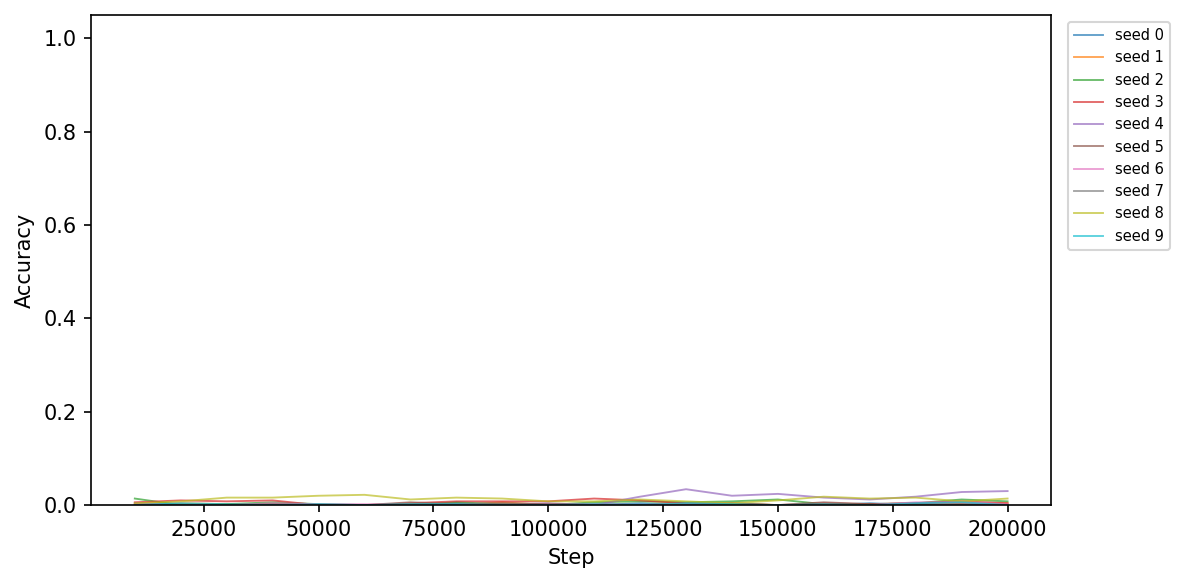}
        \end{subfigure}
        &
        \begin{subfigure}[c]{0.29\linewidth}
            \includegraphics[width=\linewidth]{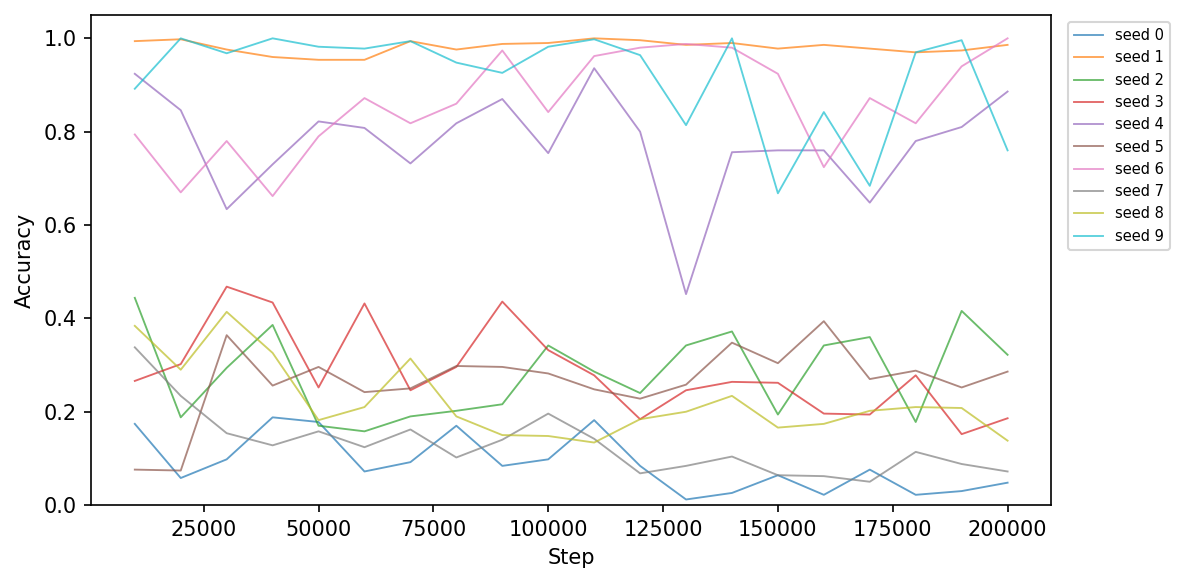}
        \end{subfigure}
        &
        \begin{subfigure}[c]{0.29\linewidth}
            \includegraphics[width=\linewidth]{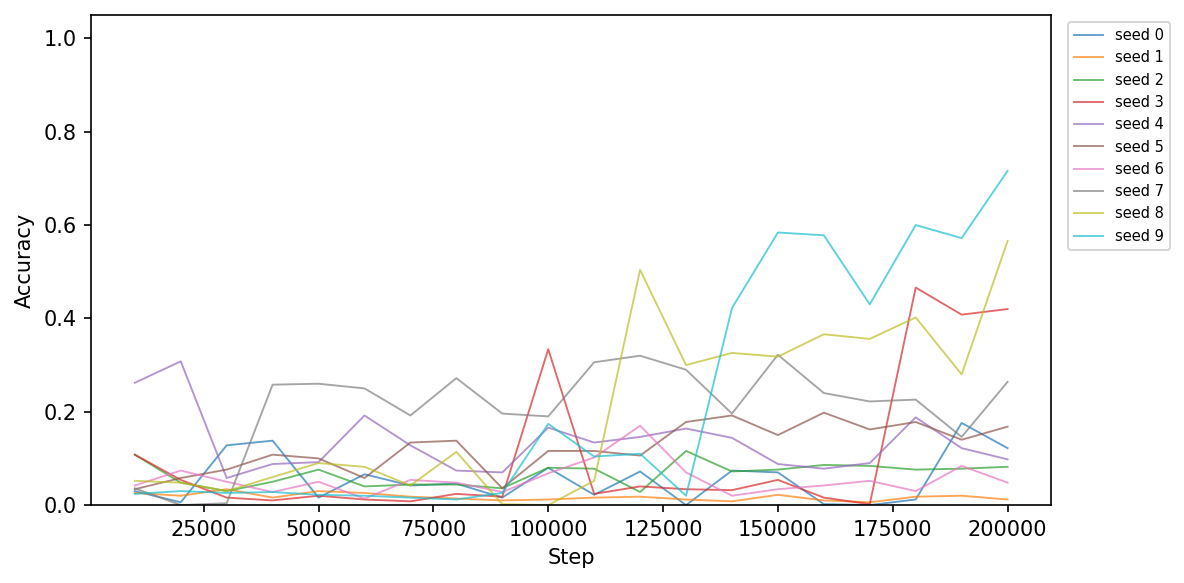}
        \end{subfigure}
        \\[1em]
        \parbox[c]{1em}{\rotatebox{90}{\textbf{$wd=0.5$}}} &
        \begin{subfigure}[c]{0.29\linewidth}
            \includegraphics[width=\linewidth]{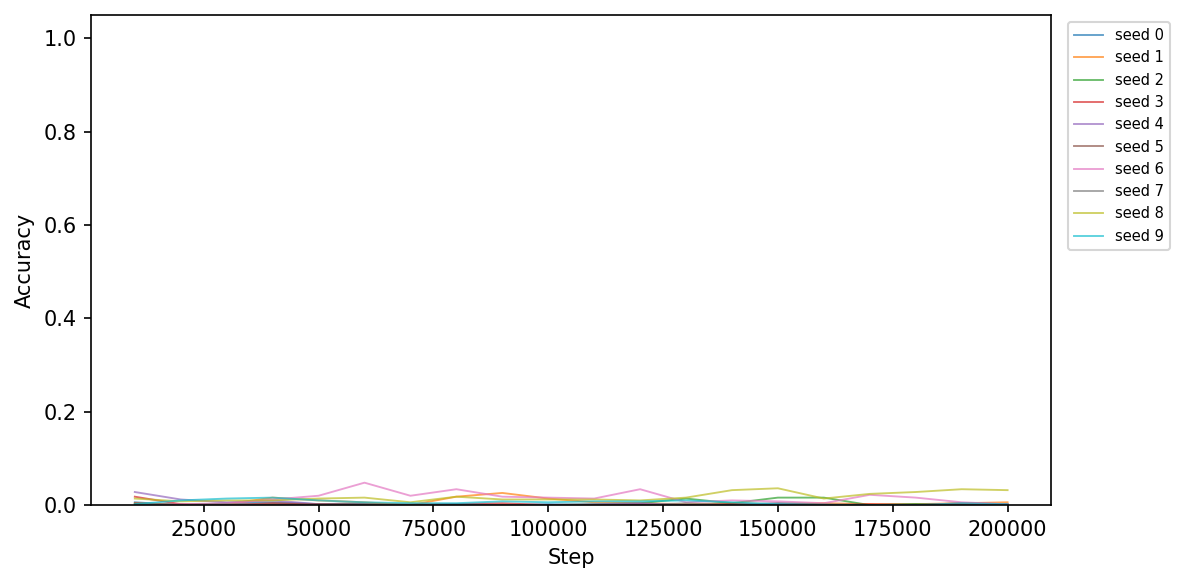}
        \end{subfigure}
        &
        \begin{subfigure}[c]{0.29\linewidth}
            \includegraphics[width=\linewidth]{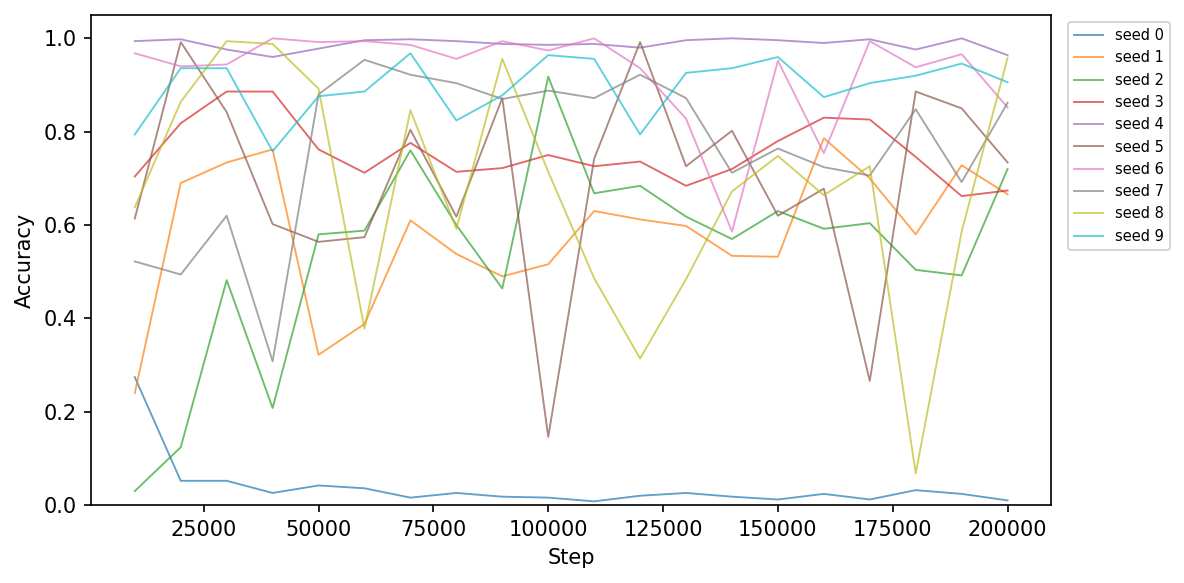}
        \end{subfigure}
        &
        \begin{subfigure}[c]{0.29\linewidth}
            \includegraphics[width=\linewidth]{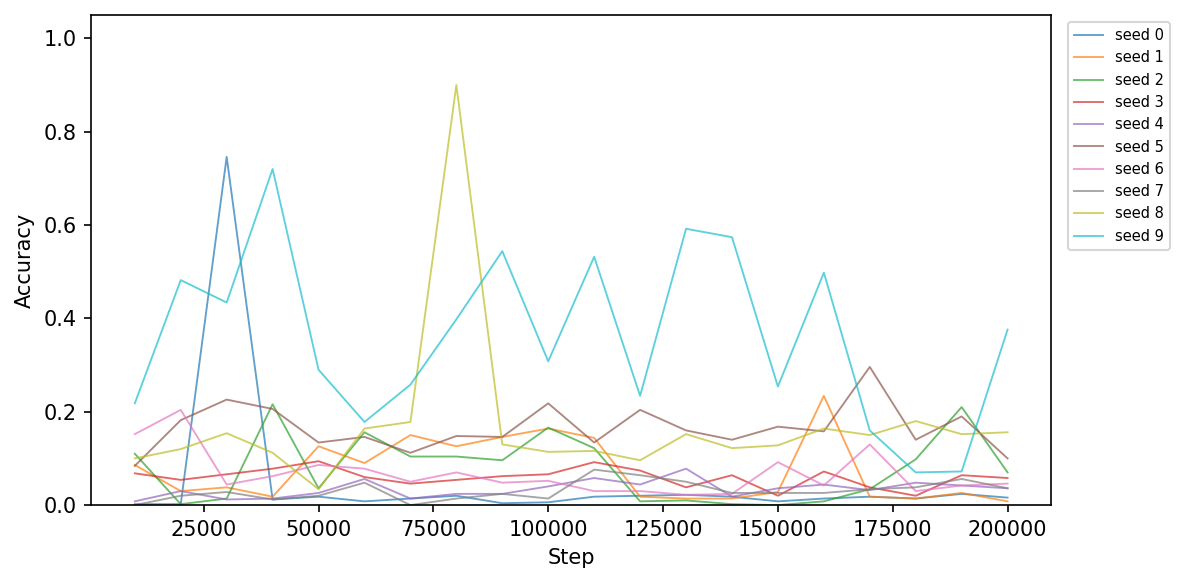}
        \end{subfigure}
        \\[1em]
    \end{tabular}
    \caption{$\mathrm{M_{F_{\mathrm{in}},\, H_{\mathrm{out}}}}$ - Evaluation Accuracy across epochs, on different hyperparameters}
    \label{fig:eval_FH}
\end{figure*}

\begin{figure*}[t]
    \centering
    \begin{tabular}{c ccc}
        & \textbf{$lr=0.0001$} & \textbf{$lr=0.001$} & \textbf{$lr=0.003$}\\[0.5em]
        \parbox[c]{1em}{\rotatebox{90}{\textbf{$wd=0.01$}}} &
        \begin{subfigure}[c]{0.29\linewidth}
            \includegraphics[width=\linewidth]{Fig/loss_le/HF/HF_lr0.0001_wd0.01_train_loss_linear.png}
        \end{subfigure}
        &
        \begin{subfigure}[c]{0.29\linewidth}
            \includegraphics[width=\linewidth]{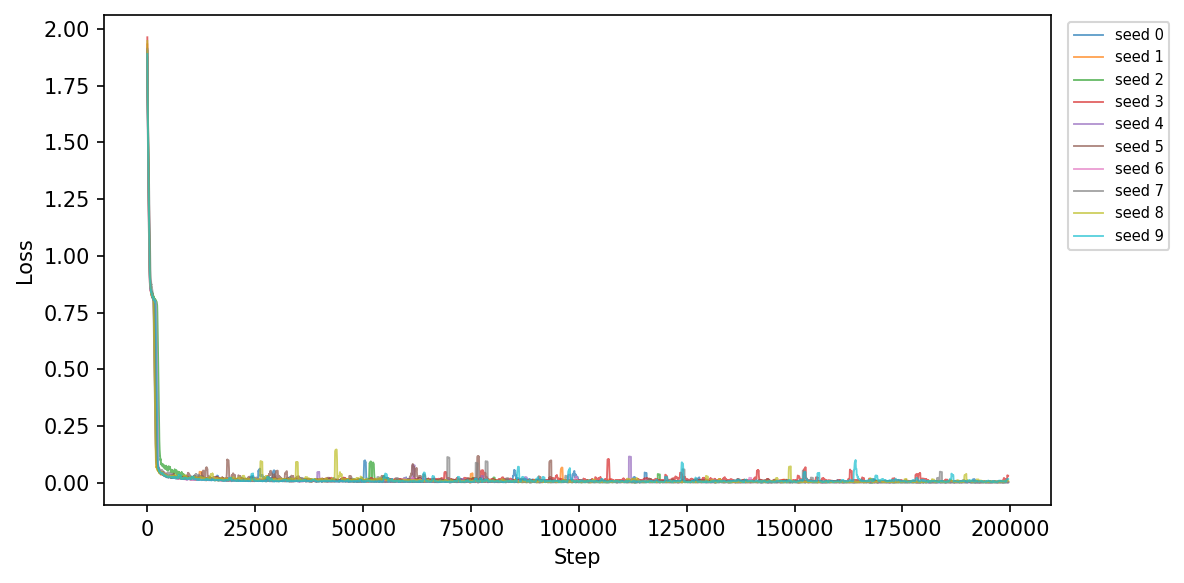}
        \end{subfigure}
        &
        \begin{subfigure}[c]{0.29\linewidth}
            \includegraphics[width=\linewidth]{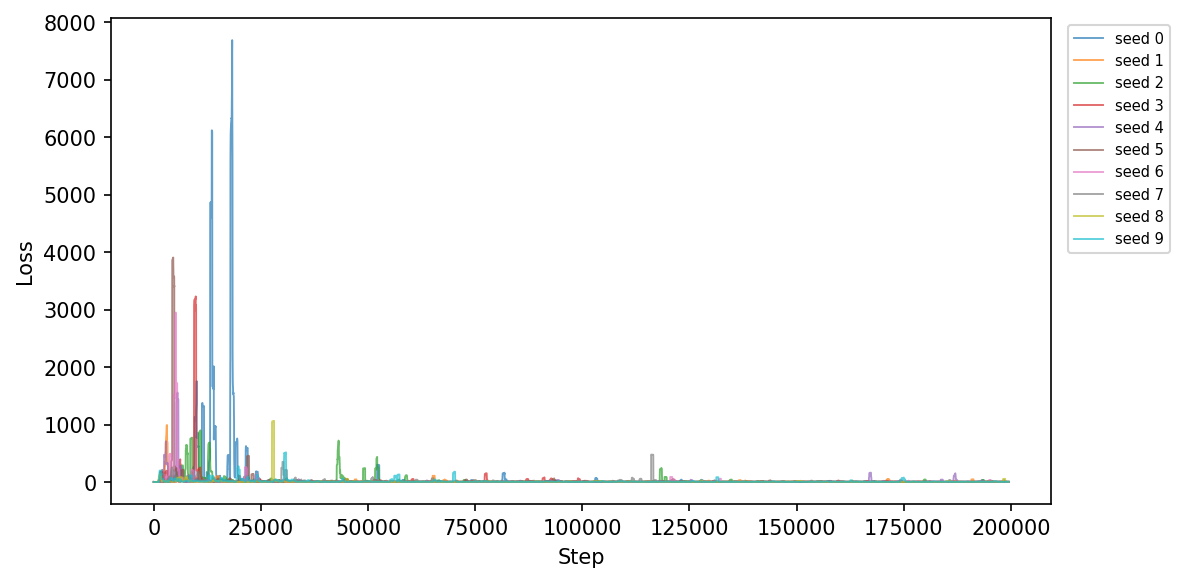}
        \end{subfigure}
        \\[1em]
        \parbox[c]{1em}{\rotatebox{90}{\textbf{$wd=0.1$}}} &
        \begin{subfigure}[c]{0.29\linewidth}
            \includegraphics[width=\linewidth]{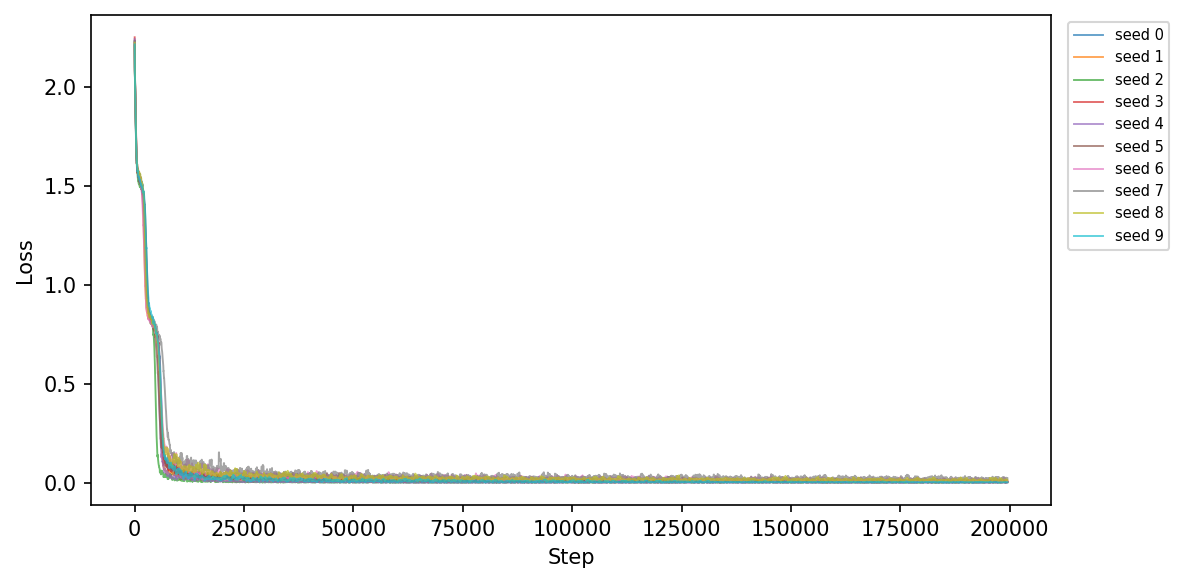}
        \end{subfigure}
        &
        \begin{subfigure}[c]{0.29\linewidth}
            \includegraphics[width=\linewidth]{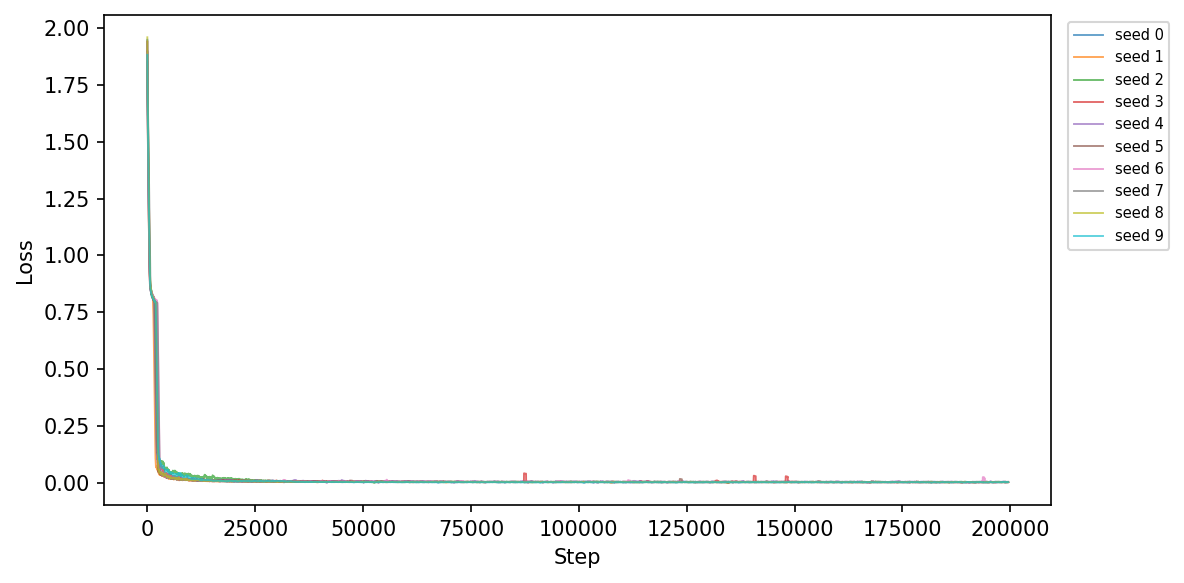}
        \end{subfigure}
        &
        \begin{subfigure}[c]{0.29\linewidth}
            \includegraphics[width=\linewidth]{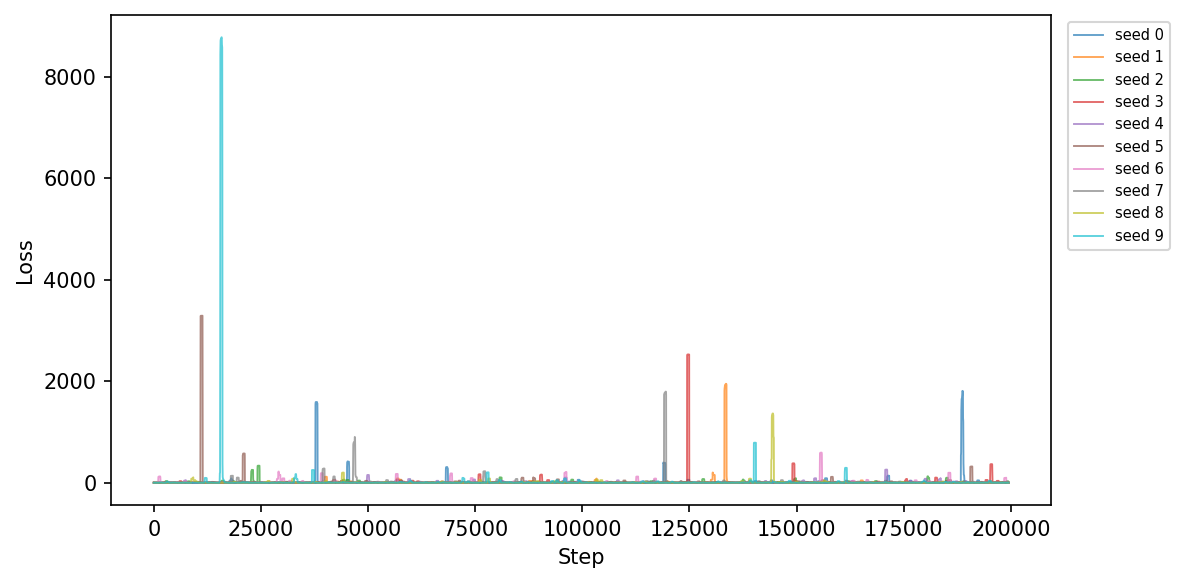}
        \end{subfigure}
        \\[1em]
        \parbox[c]{1em}{\rotatebox{90}{\textbf{$wd=0.2$}}} &
        \begin{subfigure}[c]{0.29\linewidth}
            \includegraphics[width=\linewidth]{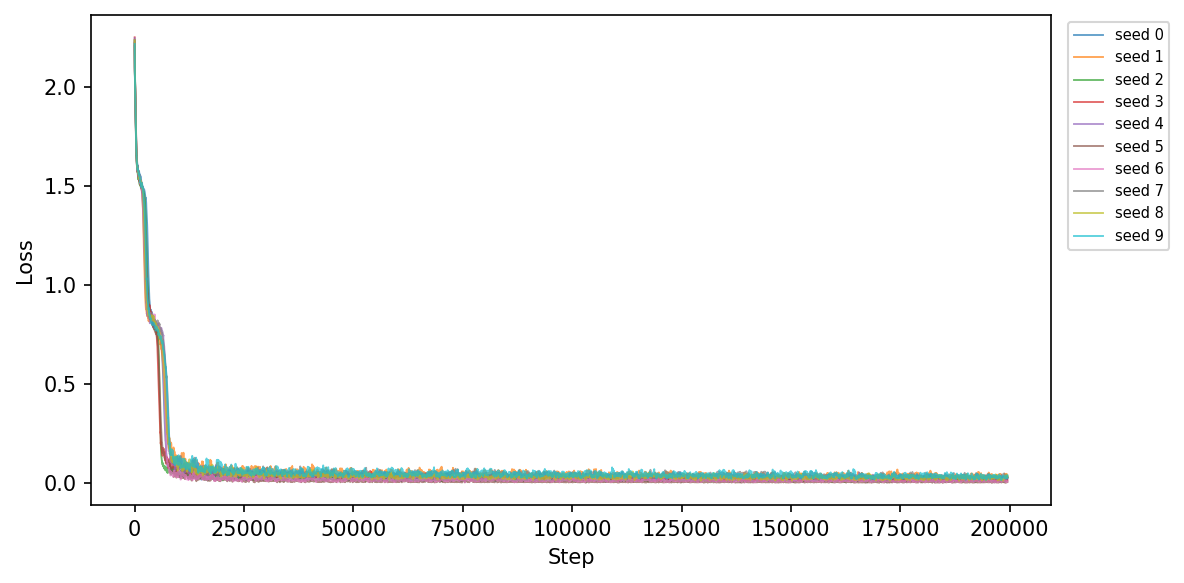}
        \end{subfigure}
        &
        \begin{subfigure}[c]{0.29\linewidth}
            \includegraphics[width=\linewidth]{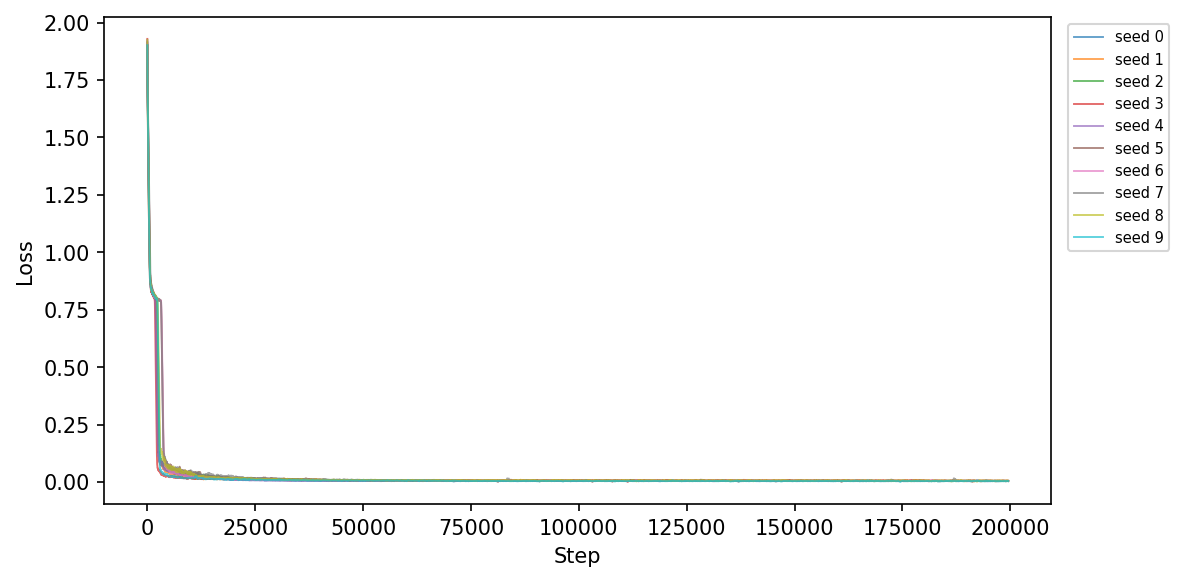}
        \end{subfigure}
        &
        \begin{subfigure}[c]{0.29\linewidth}
            \includegraphics[width=\linewidth]{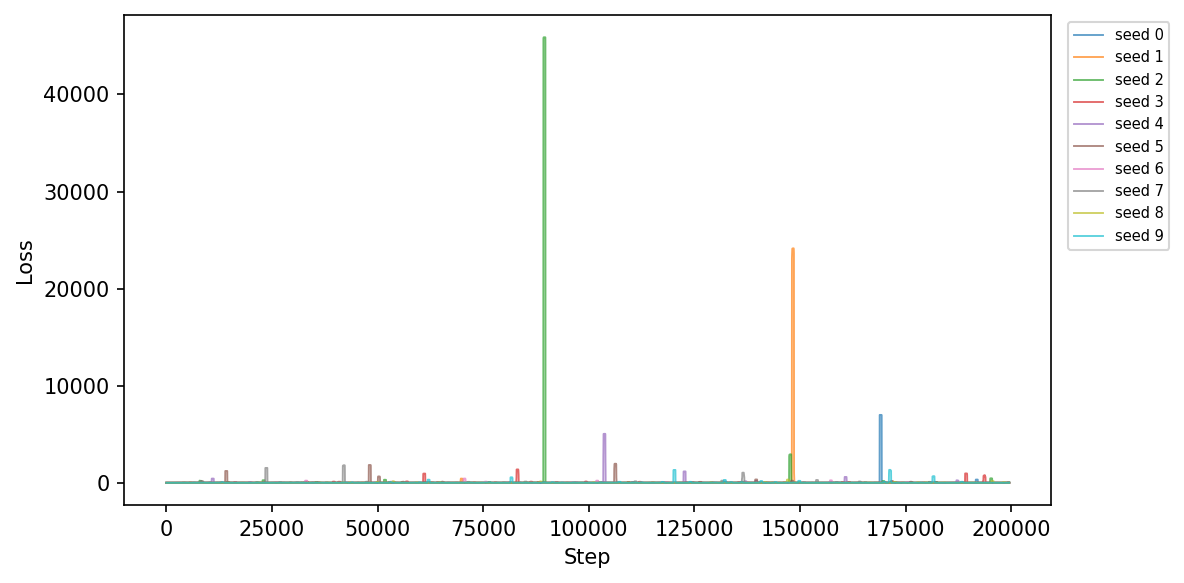}
        \end{subfigure}
        \\[1em]
        \parbox[c]{1em}{\rotatebox{90}{\textbf{$wd=0.3$}}} &
        \begin{subfigure}[c]{0.29\linewidth}
            \includegraphics[width=\linewidth]{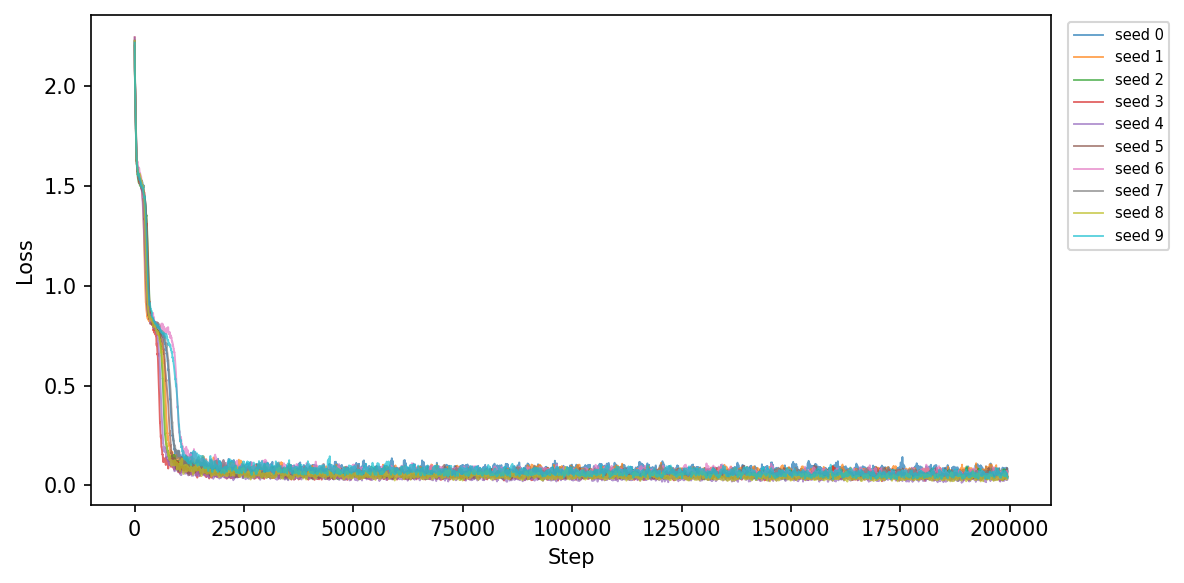}
        \end{subfigure}
        &
        \begin{subfigure}[c]{0.29\linewidth}
            \includegraphics[width=\linewidth]{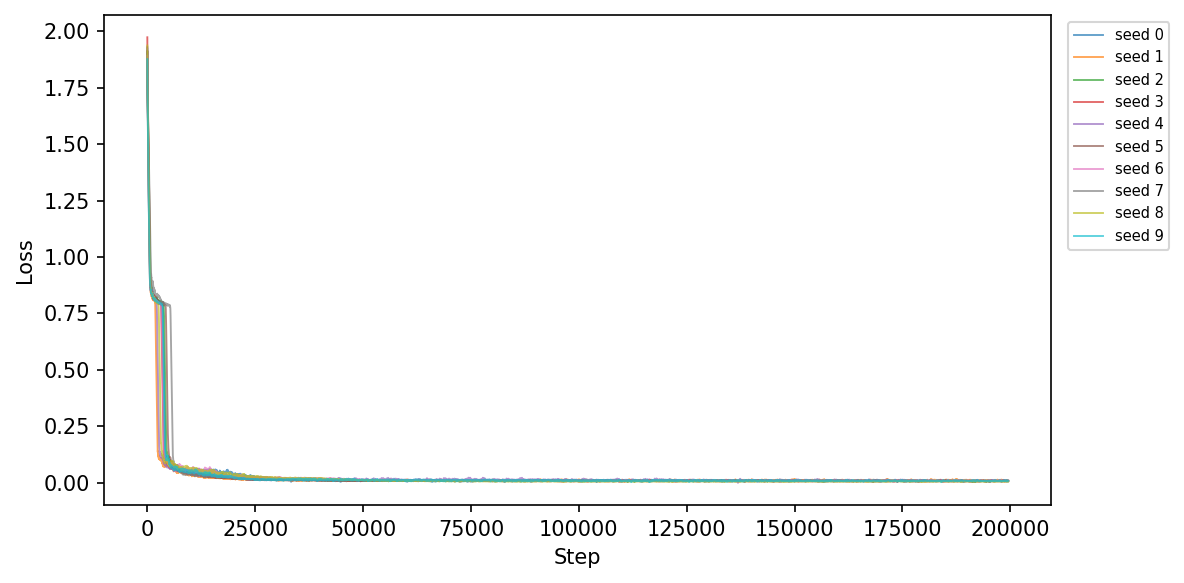}
        \end{subfigure}
        &
        \begin{subfigure}[c]{0.29\linewidth}
            \includegraphics[width=\linewidth]{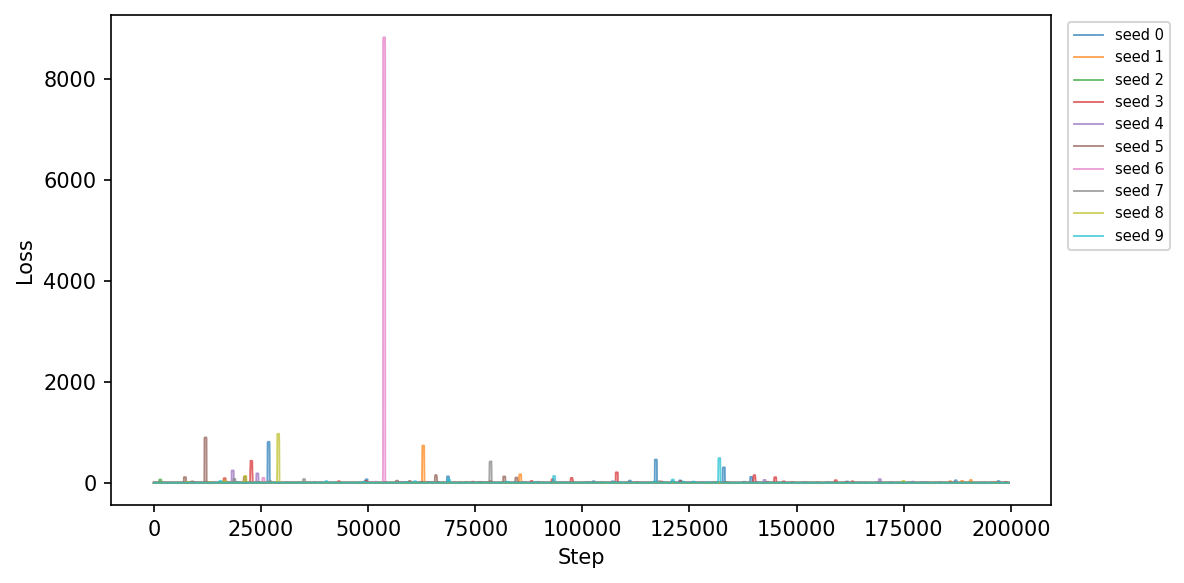}
        \end{subfigure}
        \\[1em]
        \parbox[c]{1em}{\rotatebox{90}{\textbf{$wd=0.4$}}} &
        \begin{subfigure}[c]{0.29\linewidth}
            \includegraphics[width=\linewidth]{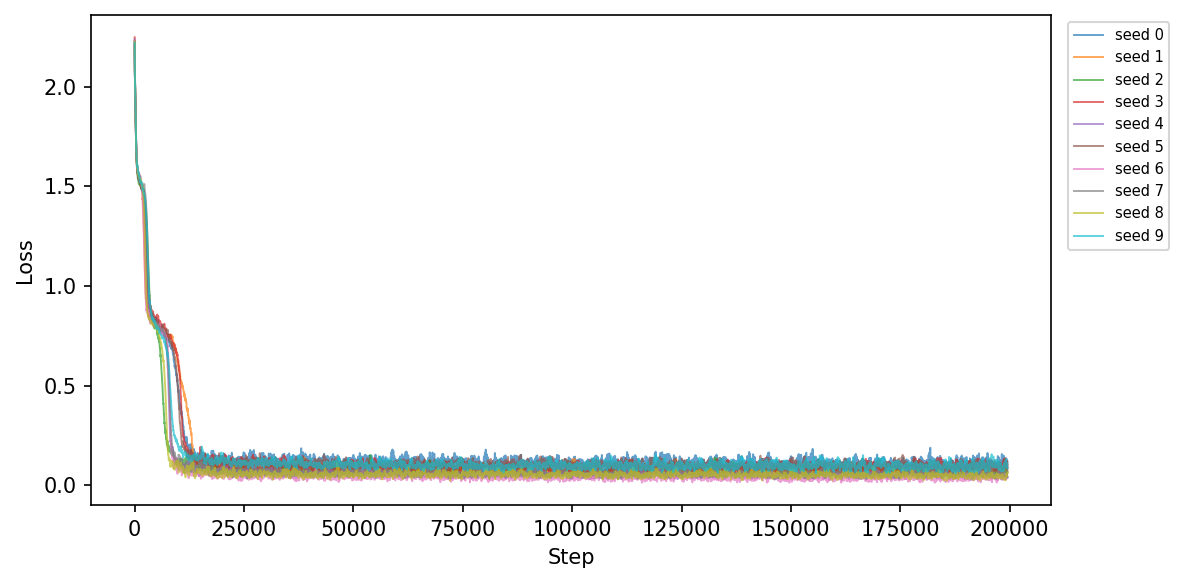}
        \end{subfigure}
        &
        \begin{subfigure}[c]{0.29\linewidth}
            \includegraphics[width=\linewidth]{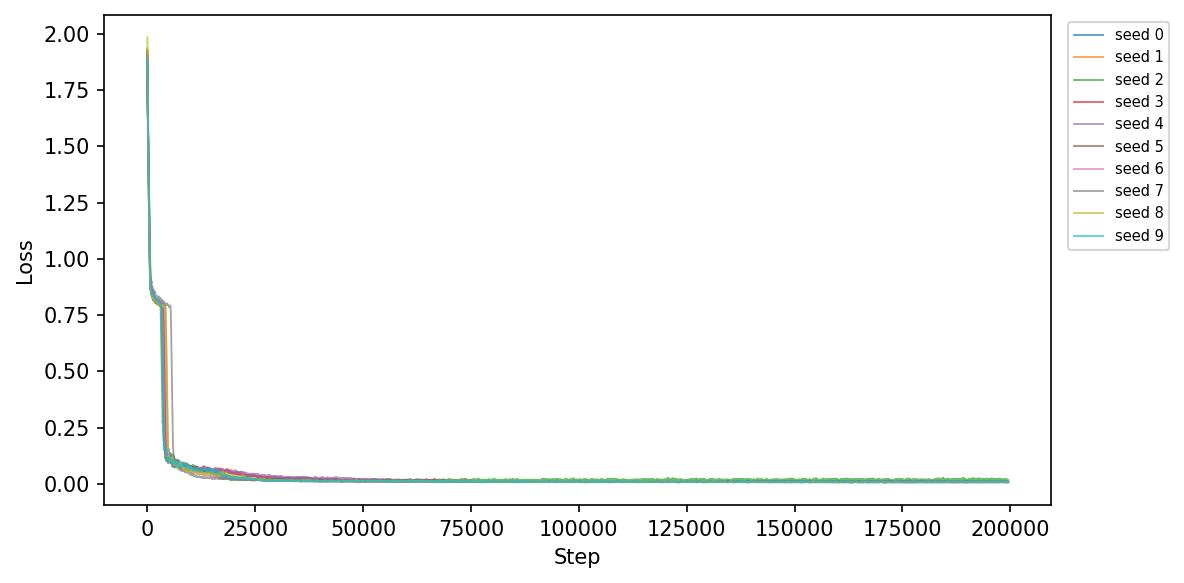}
        \end{subfigure}
        &
        \begin{subfigure}[c]{0.29\linewidth}
            \includegraphics[width=\linewidth]{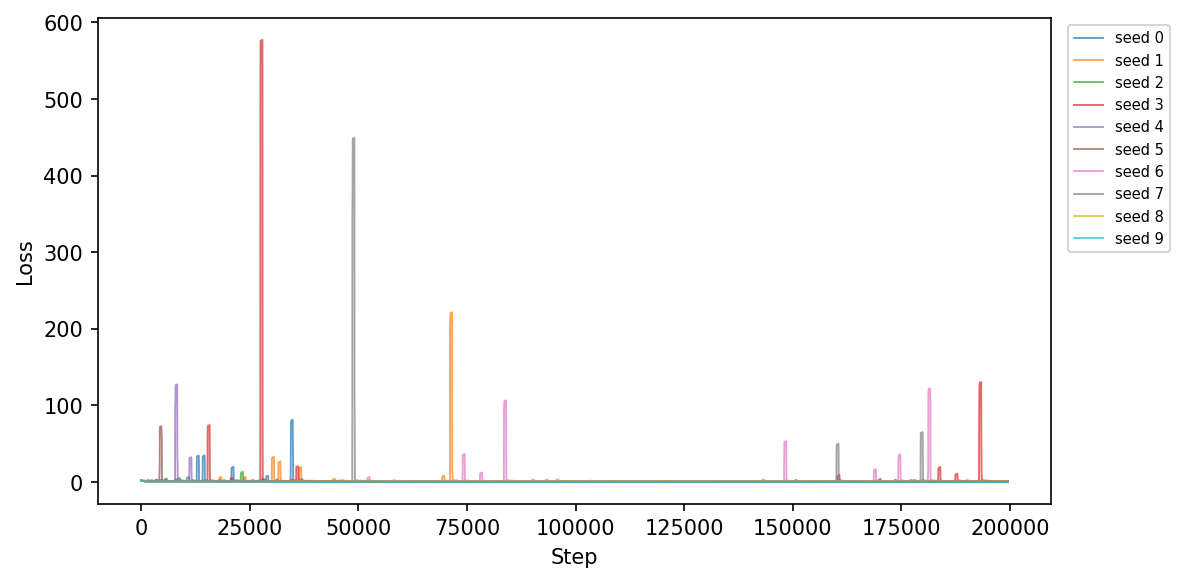}
        \end{subfigure}
        \\[1em]
        \parbox[c]{1em}{\rotatebox{90}{\textbf{$wd=0.5$}}} &
        \begin{subfigure}[c]{0.29\linewidth}
            \includegraphics[width=\linewidth]{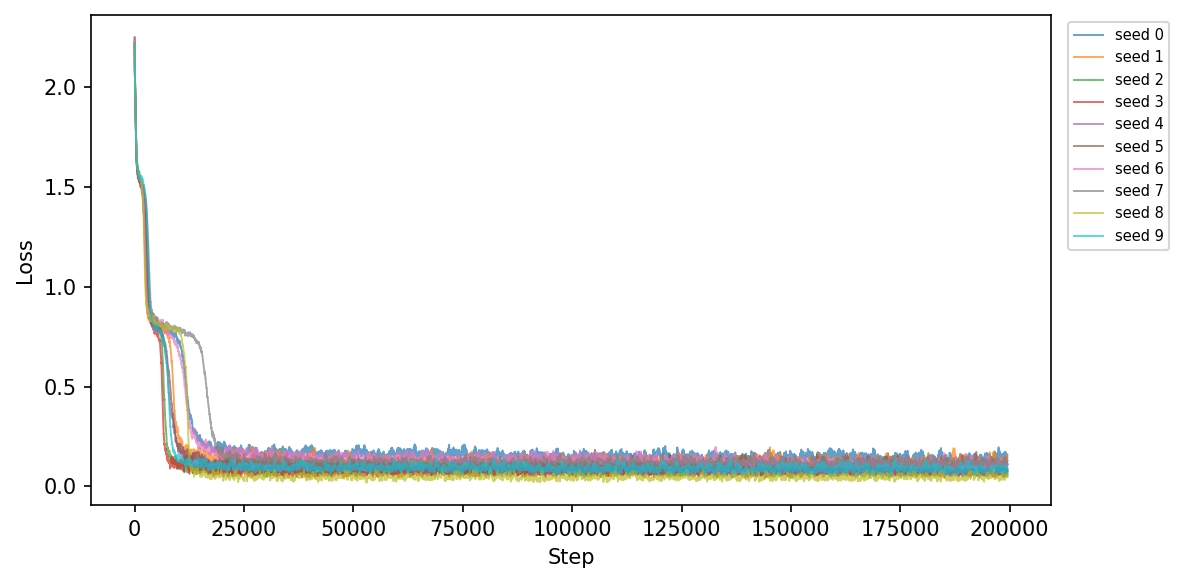}
        \end{subfigure}
        &
        \begin{subfigure}[c]{0.29\linewidth}
            \includegraphics[width=\linewidth]{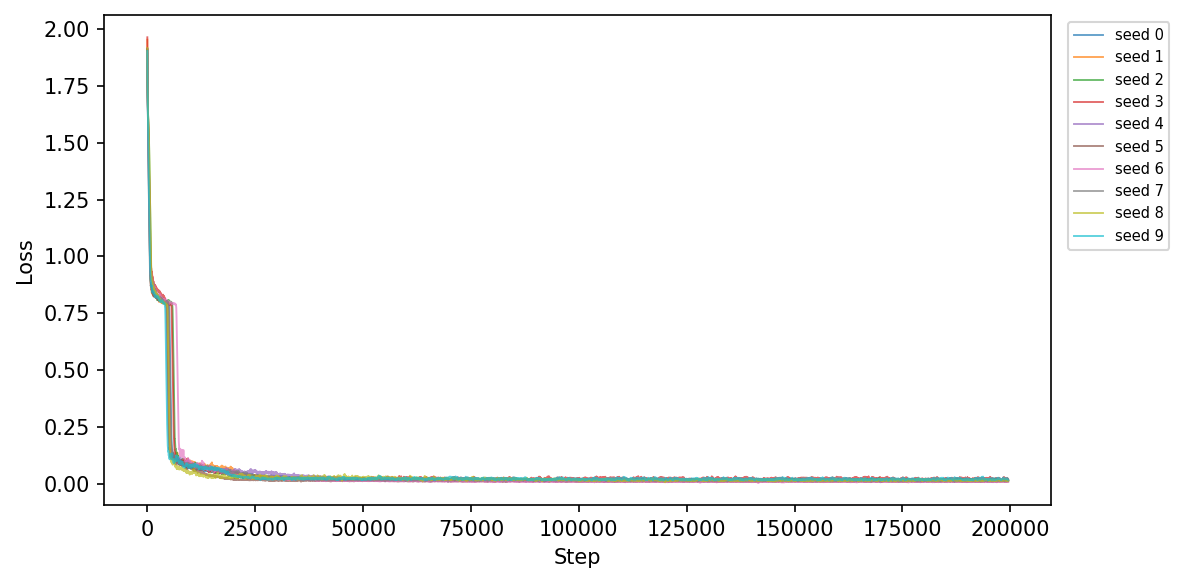}
        \end{subfigure}
        &
        \begin{subfigure}[c]{0.29\linewidth}
            \includegraphics[width=\linewidth]{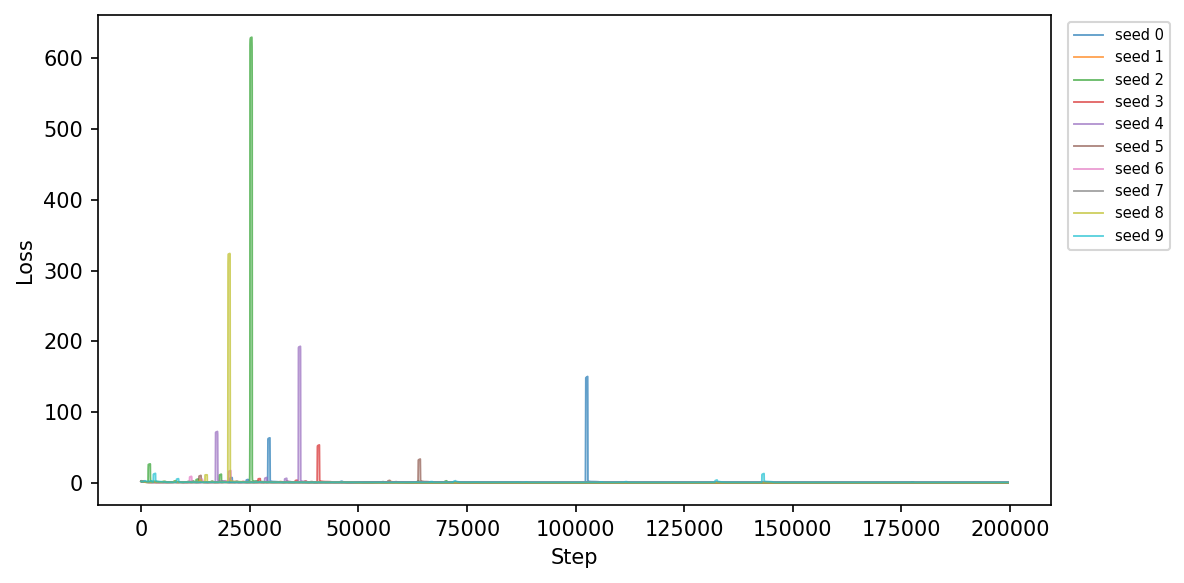}
        \end{subfigure}
        \\[1em]
    \end{tabular}
    \caption{$\mathrm{M_{H_{\mathrm{in}},\, F_{\mathrm{out}}}}$ - Train Loss across epochs, on different hyperparameters}
    \label{fig:loss_HF}
\end{figure*}

\begin{figure*}[t]
    \centering
    \begin{tabular}{c ccc}
        & \textbf{$lr=0.0001$} & \textbf{$lr=0.001$} & \textbf{$lr=0.003$}\\[0.5em]
        \parbox[c]{1em}{\rotatebox{90}{\textbf{$wd=0.01$}}} &
        \begin{subfigure}[c]{0.29\linewidth}
            \includegraphics[width=\linewidth]{Fig/loss_le/HF/HF_lr0.0001_wd0.01_random_eval_acc.png}
        \end{subfigure}
        &
        \begin{subfigure}[c]{0.29\linewidth}
            \includegraphics[width=\linewidth]{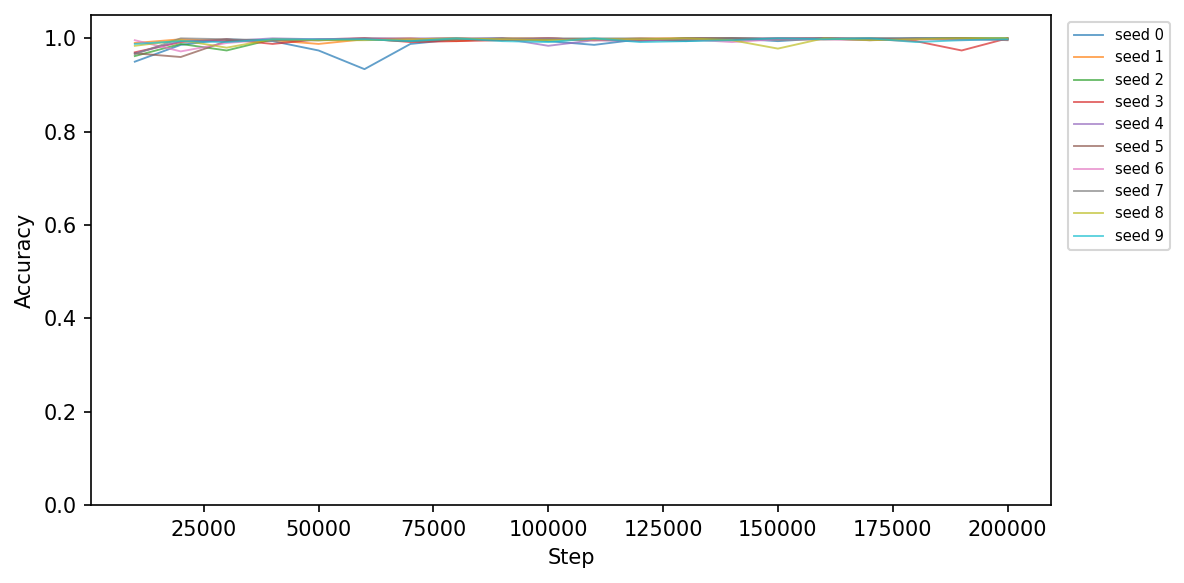}
        \end{subfigure}
        &
        \begin{subfigure}[c]{0.29\linewidth}
            \includegraphics[width=\linewidth]{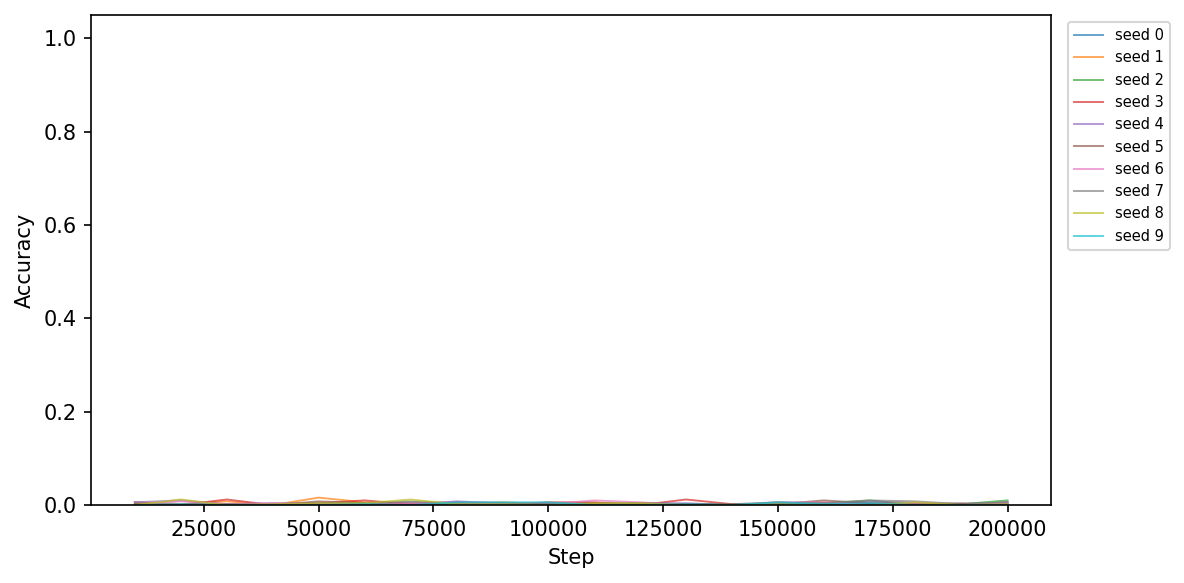}
        \end{subfigure}
        \\[1em]
        \parbox[c]{1em}{\rotatebox{90}{\textbf{$wd=0.1$}}} &
        \begin{subfigure}[c]{0.29\linewidth}
            \includegraphics[width=\linewidth]{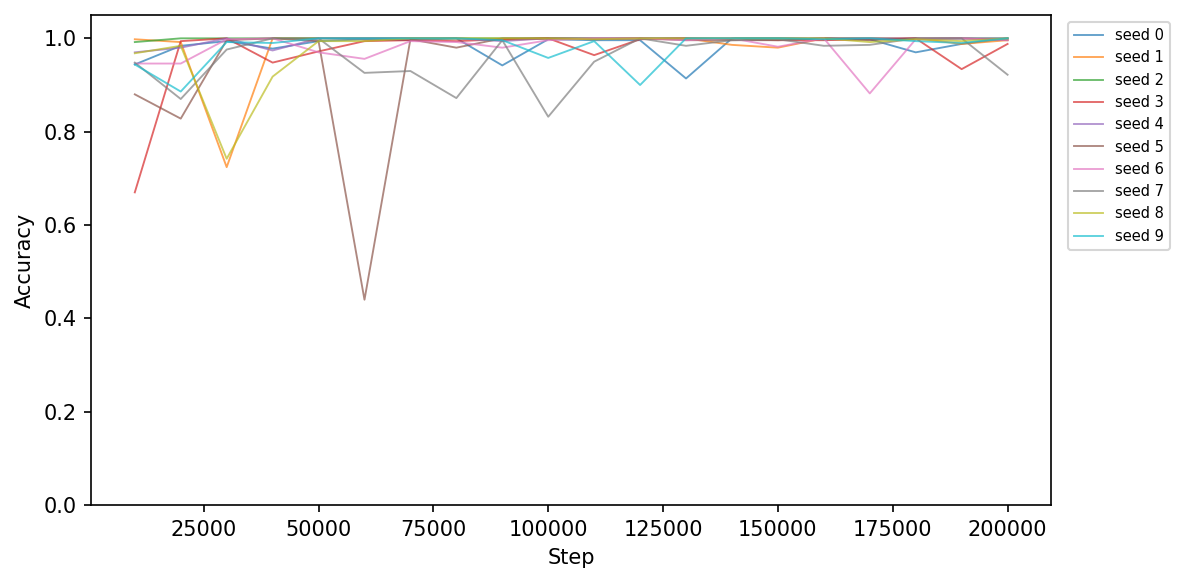}
        \end{subfigure}
        &
        \begin{subfigure}[c]{0.29\linewidth}
            \includegraphics[width=\linewidth]{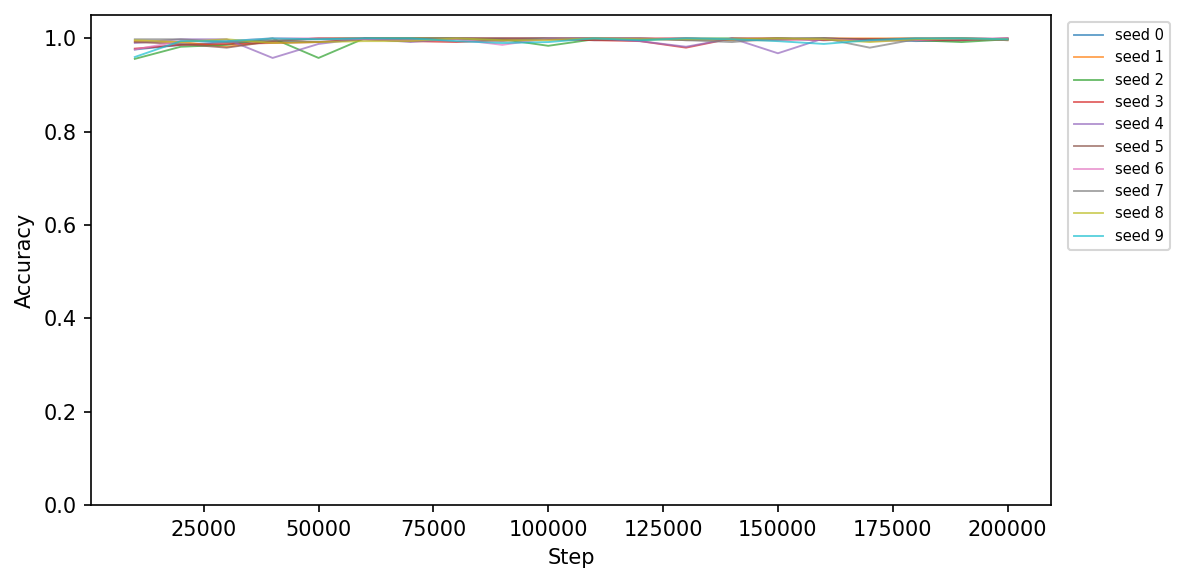}
        \end{subfigure}
        &
        \begin{subfigure}[c]{0.29\linewidth}
            \includegraphics[width=\linewidth]{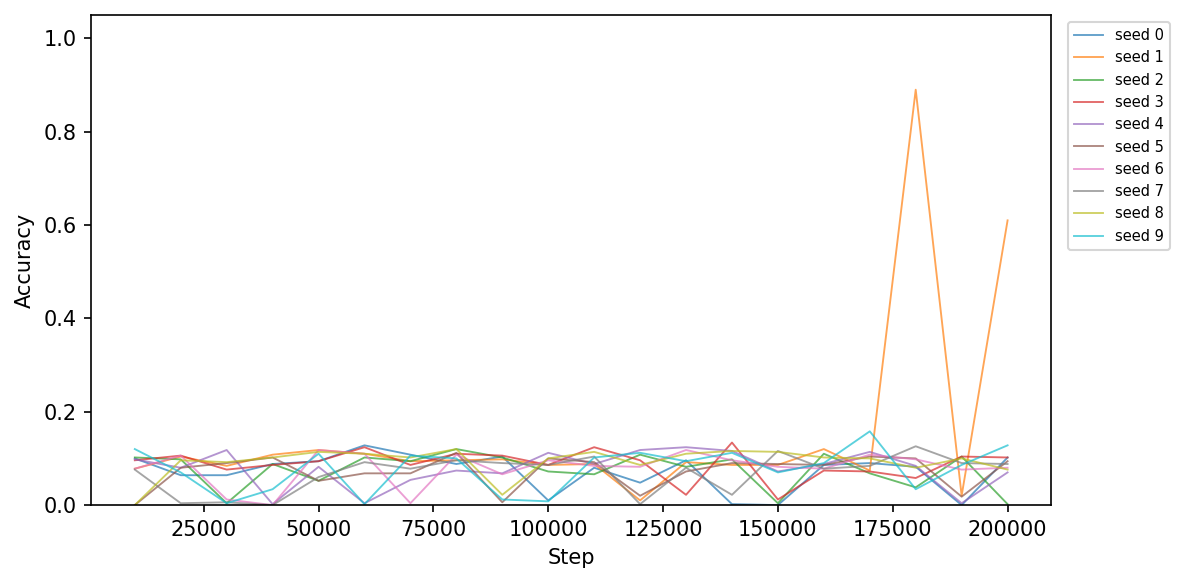}
        \end{subfigure}
        \\[1em]
        \parbox[c]{1em}{\rotatebox{90}{\textbf{$wd=0.2$}}} &
        \begin{subfigure}[c]{0.29\linewidth}
            \includegraphics[width=\linewidth]{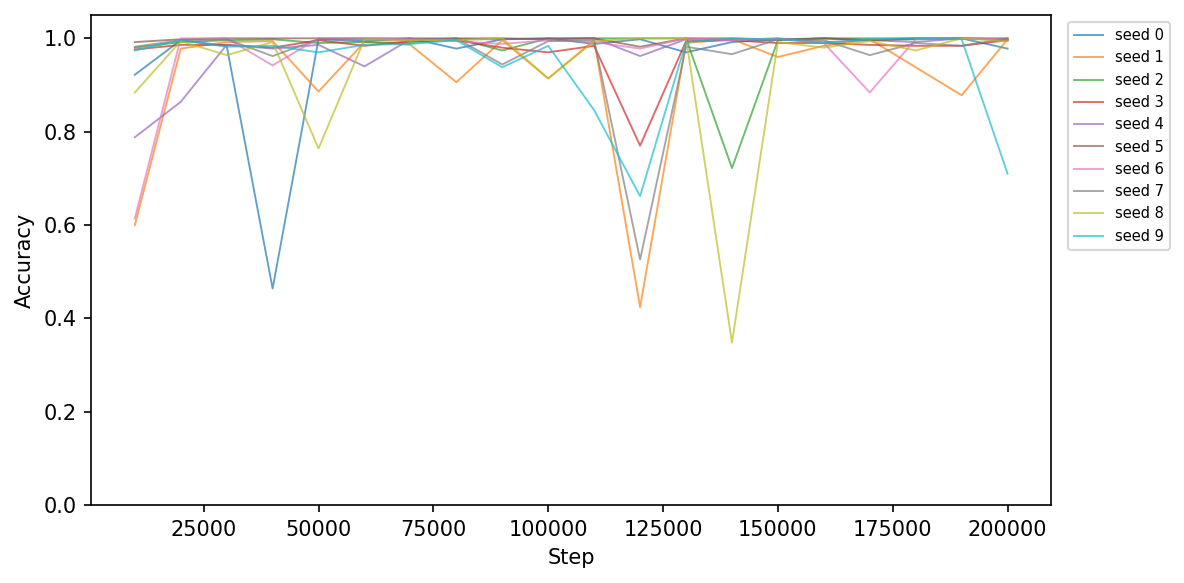}
        \end{subfigure}
        &
        \begin{subfigure}[c]{0.29\linewidth}
            \includegraphics[width=\linewidth]{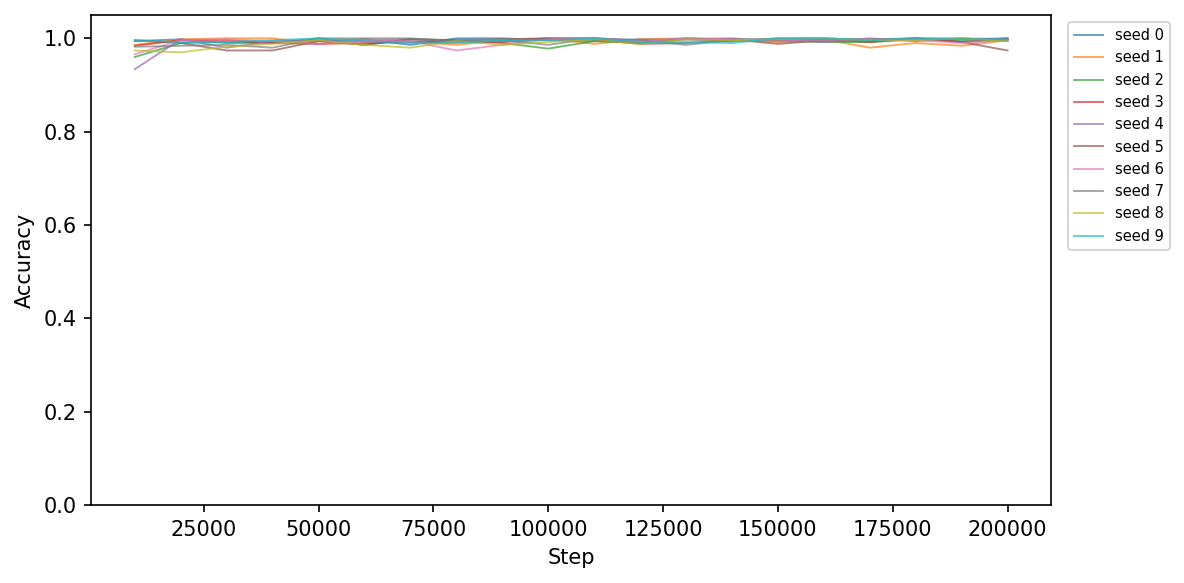}
        \end{subfigure}
        &
        \begin{subfigure}[c]{0.29\linewidth}
            \includegraphics[width=\linewidth]{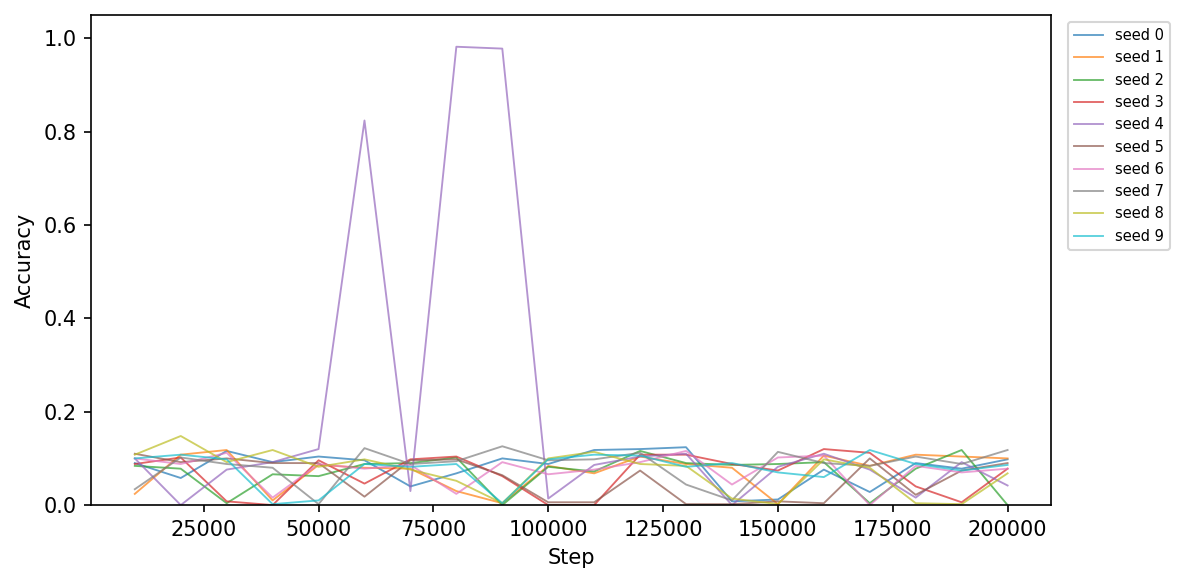}
        \end{subfigure}
        \\[1em]
        \parbox[c]{1em}{\rotatebox{90}{\textbf{$wd=0.3$}}} &
        \begin{subfigure}[c]{0.29\linewidth}
            \includegraphics[width=\linewidth]{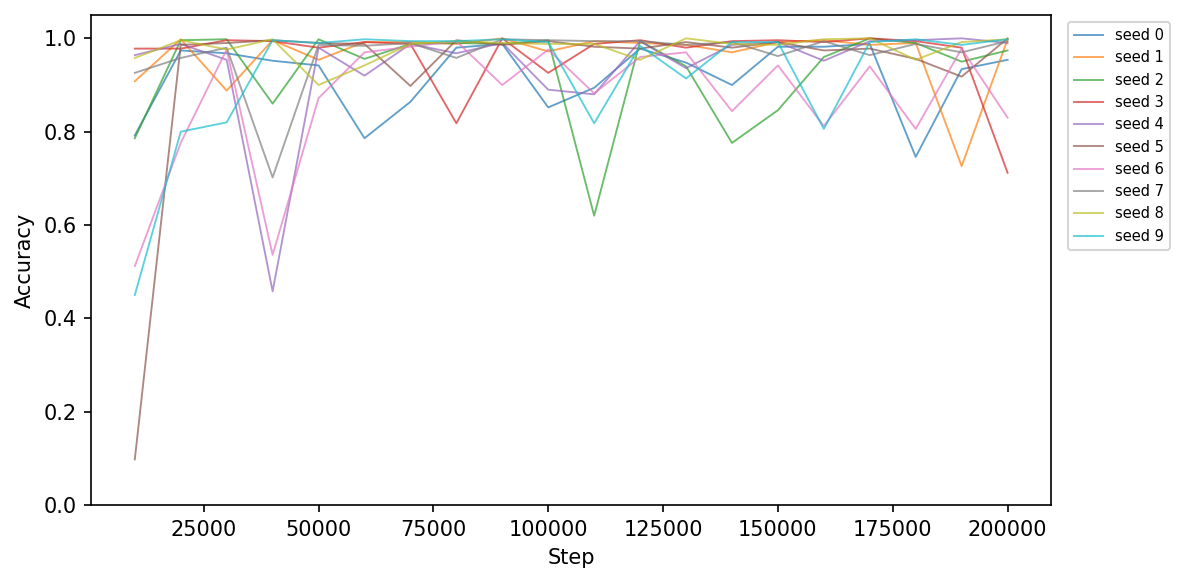}
        \end{subfigure}
        &
        \begin{subfigure}[c]{0.29\linewidth}
            \includegraphics[width=\linewidth]{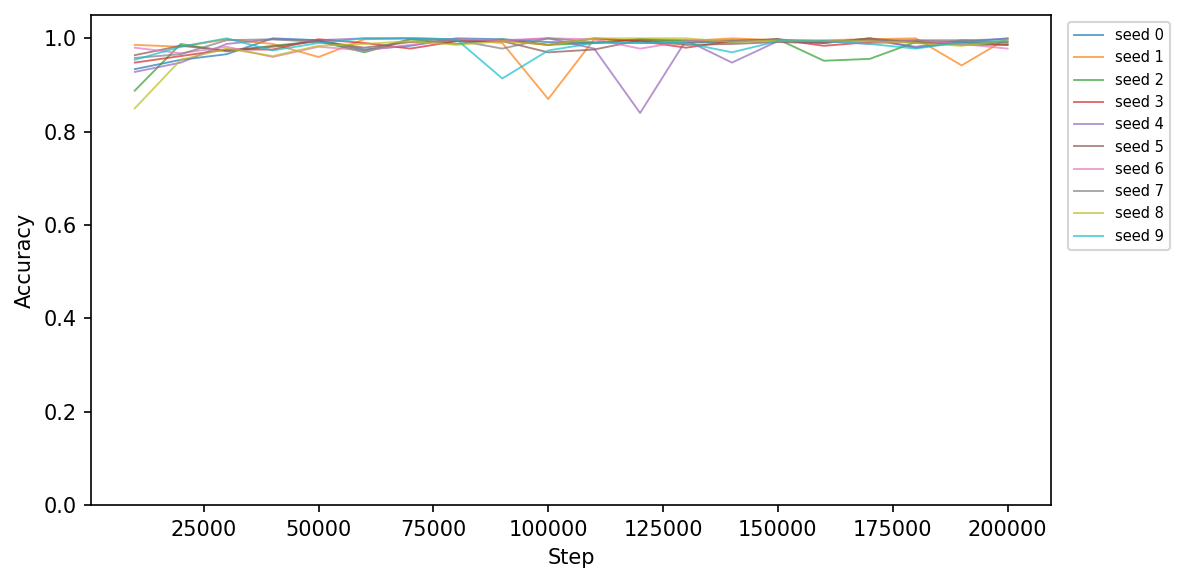}
        \end{subfigure}
        &
        \begin{subfigure}[c]{0.29\linewidth}
            \includegraphics[width=\linewidth]{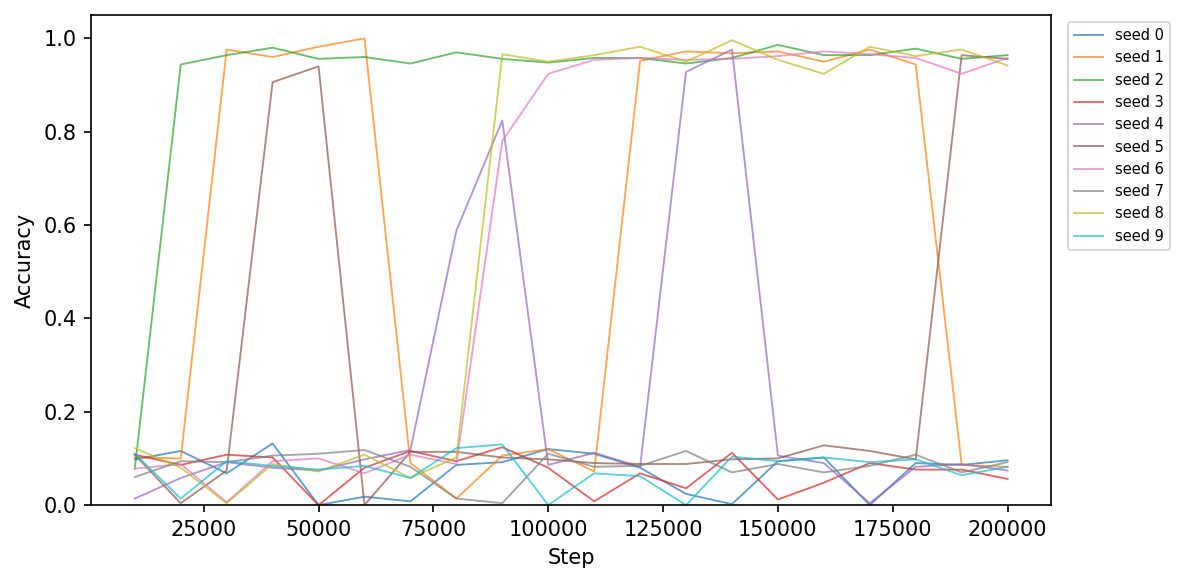}
        \end{subfigure}
        \\[1em]
        \parbox[c]{1em}{\rotatebox{90}{\textbf{$wd=0.4$}}} &
        \begin{subfigure}[c]{0.29\linewidth}
            \includegraphics[width=\linewidth]{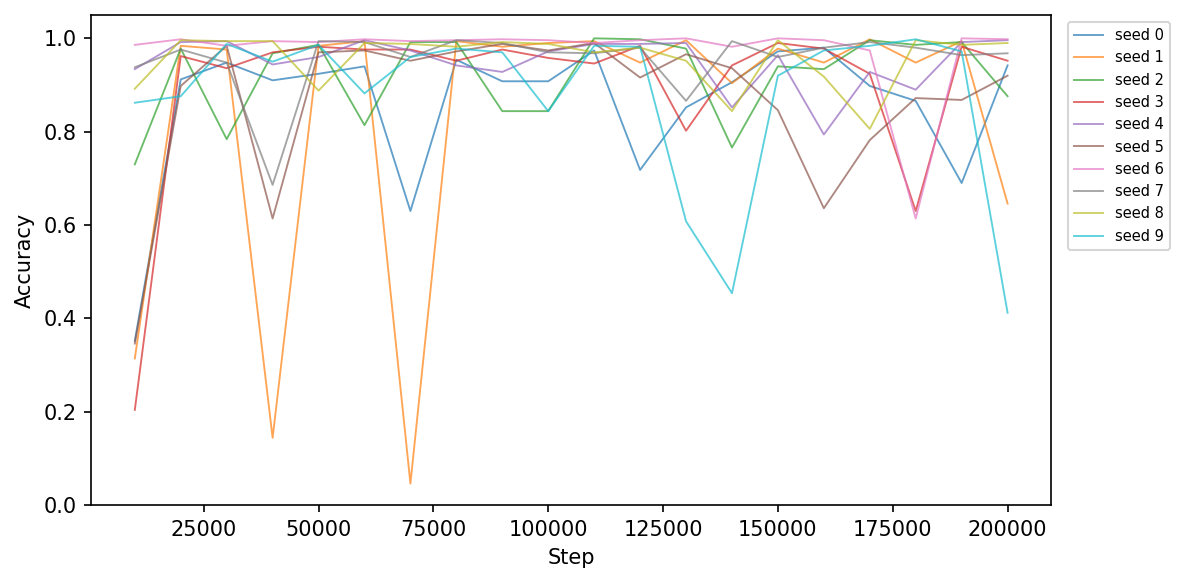}
        \end{subfigure}
        &
        \begin{subfigure}[c]{0.29\linewidth}
            \includegraphics[width=\linewidth]{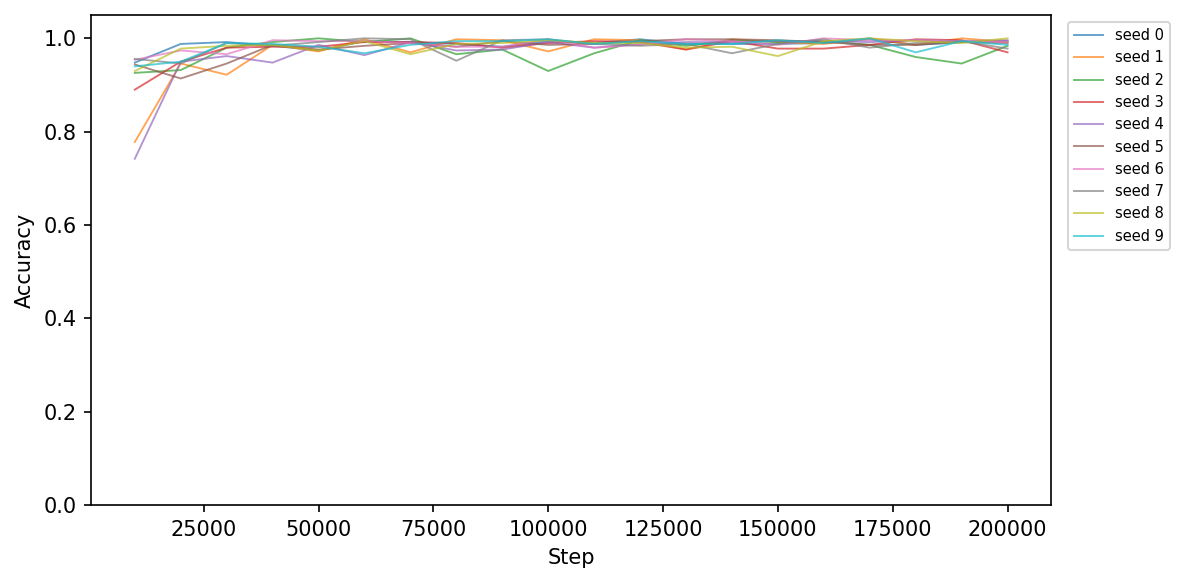}
        \end{subfigure}
        &
        \begin{subfigure}[c]{0.29\linewidth}
            \includegraphics[width=\linewidth]{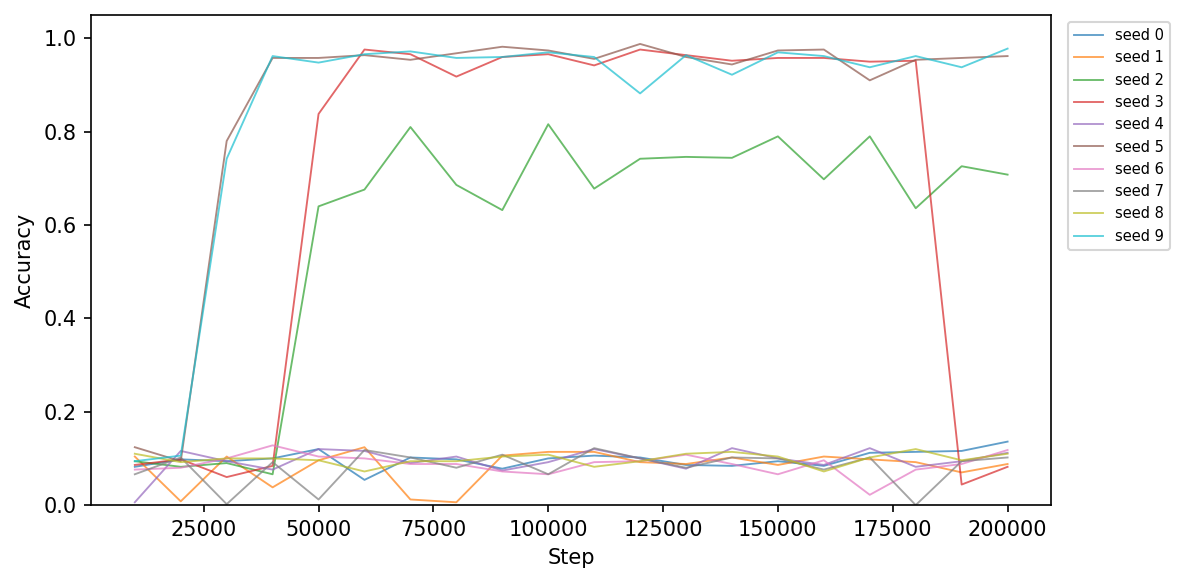}
        \end{subfigure}
        \\[1em]
        \parbox[c]{1em}{\rotatebox{90}{\textbf{$wd=0.5$}}} &
        \begin{subfigure}[c]{0.29\linewidth}
            \includegraphics[width=\linewidth]{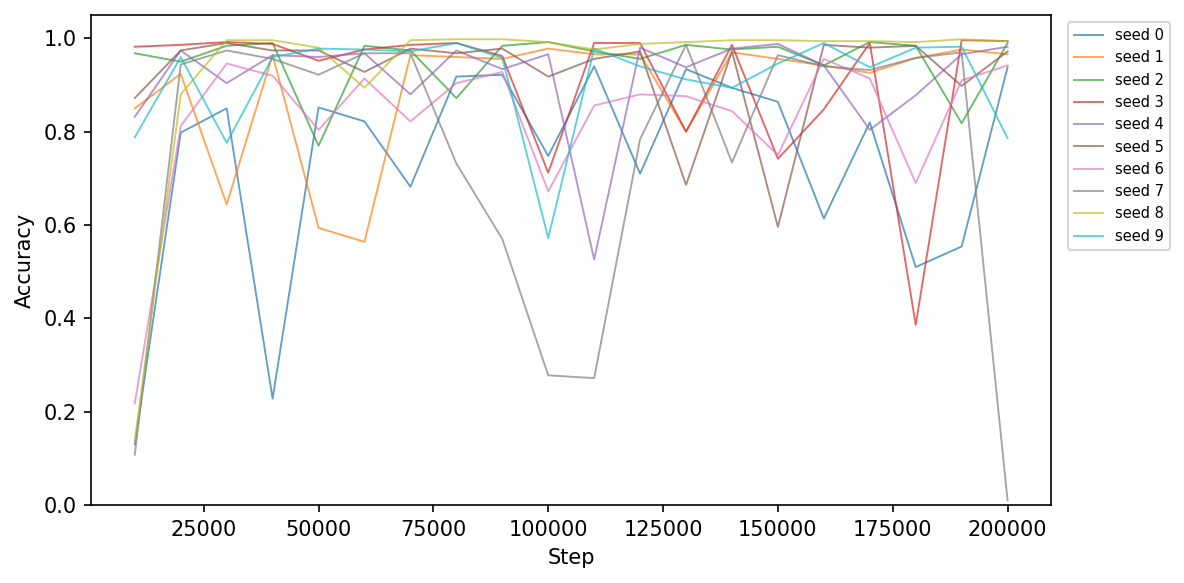}
        \end{subfigure}
        &
        \begin{subfigure}[c]{0.29\linewidth}
            \includegraphics[width=\linewidth]{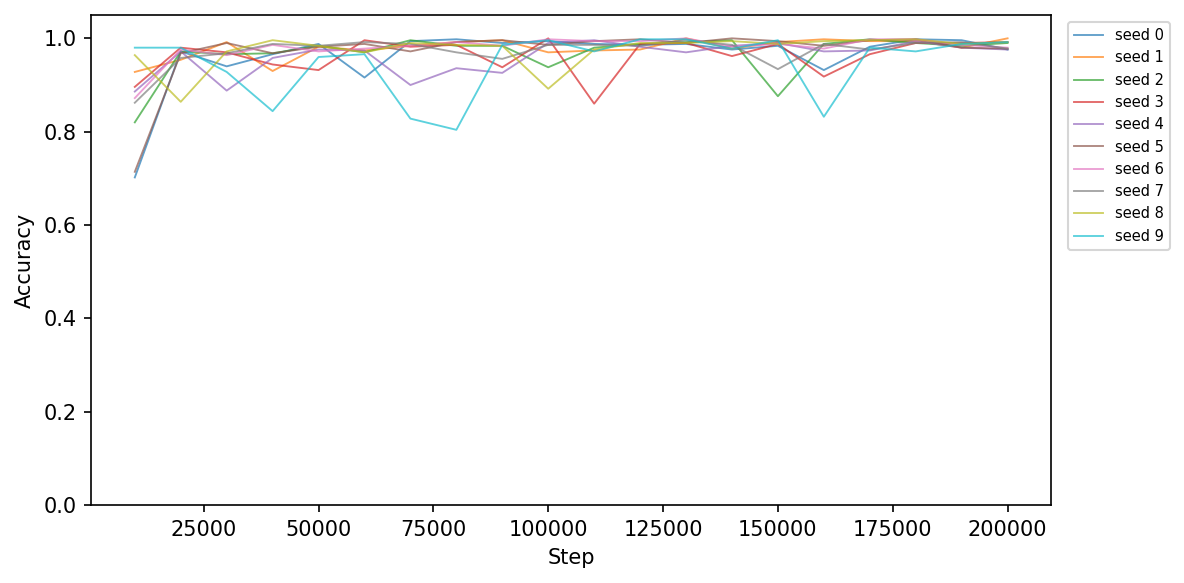}
        \end{subfigure}
        &
        \begin{subfigure}[c]{0.29\linewidth}
            \includegraphics[width=\linewidth]{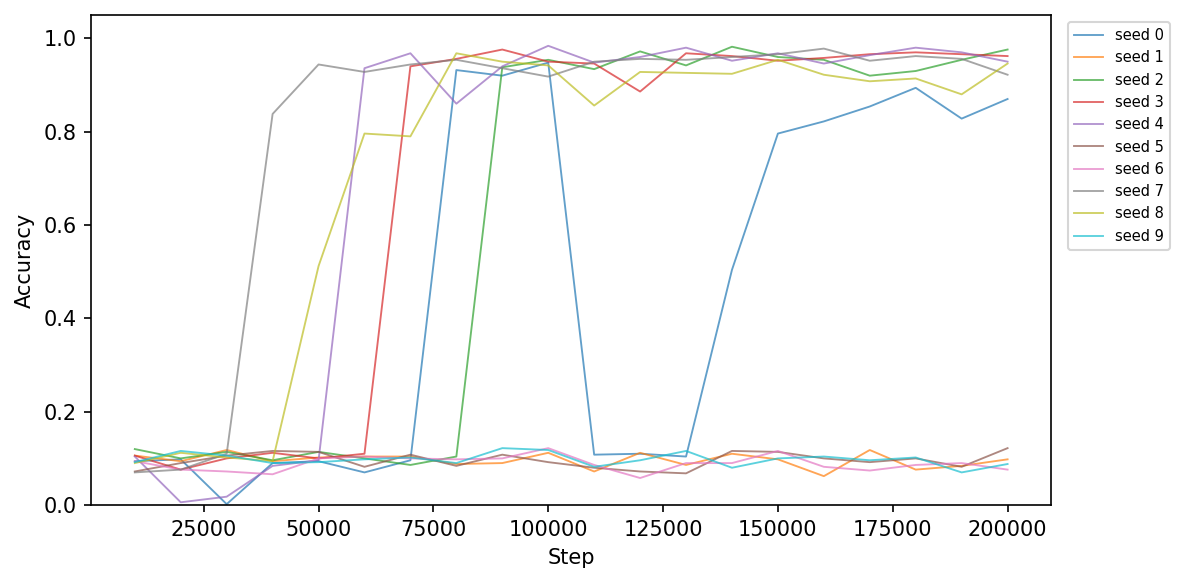}
        \end{subfigure}
        \\[1em]
    \end{tabular}
    \caption{$\mathrm{M_{H_{\mathrm{in}},\, F_{\mathrm{out}}}}$ - Evaluation Accuracy across epochs, on different hyperparameters}
    \label{fig:eval_HF}
\end{figure*}

We provide training loss curves and evaluation accuracy curves for all hyperparameters in Figures \ref{fig:loss_HH}, \ref{fig:eval_HH}, \ref{fig:loss_FF}, \ref{fig:eval_FF}, \ref{fig:loss_FH}, \ref{fig:eval_FH},  \ref{fig:loss_HF}, \ref{fig:eval_HF}.

\section{Probing Classifiers}
\label{app:probes}

To analyse the internal representations of trained models, we train probing classifiers on the residual stream at the '\texttt{=}' token position, i.e., the point at which the model must have computed all information necessary to generate the first output token. Probe training data consists of 5{,}000 randomly sampled addition problems disjoint from the evaluation set. Hidden states are extracted after each of the four transformer blocks, yielding activations of shape $[\text{n\_examples},\ \text{n\_layers},\ d_\text{model}]$.

For each layer and each of the three result digit positions (hundreds, tens, units), we train separate probes to predict the corresponding digit value (0 - 9) from the residual stream. Three probe types are used. The \textbf{linear probe} is a single linear layer optimized with cross-entropy loss. The \textbf{MLP probe} adds one hidden layer of 64 units with ReLU activation. The \textbf{circular probe} maps the residual stream to a 2D sine/cosine embedding and decodes by nearest-class angle \citep{nanda2023progress}, optimized with MSE loss. 
All probes are trained for 1{,}000 epochs with the Adam optimizer at learning rate $10^{-3}$, using an 80/20 train/test split.
In total we thus train probes for 3 digit positions × 4 layers × 3 probe types × 4 input-output tokenizer conditions x 5 model checkpoints per condition 
= 720 probing classifiers, each taking approximately 1 minute, adding approximately 12 GPU hours. Experiments were run on Nvidia RTX 3090.

Probes are trained on the five best model checkpoints across different seeds per condition (input tokenization x output tokenization), see Table~\ref{tab:checkpoints}.

\begin{table*}[t]
\centering
\small
\begin{tabular}{llrrrrrrr}
\toprule
& & & & & & \multicolumn{3}{c}{Per-digit acc.} \\
\cmidrule(lr){7-9}
Condition & lr & wd & Seed & Step & Exact & U & T & H \\
\midrule
\multirow{5}{*}{$\mathrm{M_F}$}
  & 0.0001 & 0.01 & 891 & 60{,}000  & 1.000 & 1.000 & 1.000 & 1.000 \\
  & 0.0001 & 0.01 & 598 & 140{,}000 & 1.000 & 1.000 & 1.000 & 1.000 \\
  & 0.0001 & 0.1  & 193 & 160{,}000 & 1.000 & 1.000 & 1.000 & 1.000 \\
  & 0.0001 & 0.2  & 623 & 80{,}000  & 1.000 & 1.000 & 1.000 & 1.000 \\
  & 0.0001 & 0.4  & 273 & 120{,}000 & 1.000 & 1.000 & 1.000 & 1.000 \\
\midrule
\multirow{5}{*}{$\mathrm{M_{F_{in},\,H_{out}}}$}
  & 0.001 & 0.1 & 623 & 30{,}000  & 0.994 & 0.998 & 0.998 & 0.998 \\
  & 0.001 & 0.1 & 891 & 100{,}000 & 0.998 & 1.000 & 1.000 & 0.998 \\
  & 0.001 & 0.2 & 378 & 50{,}000  & 0.994 & 0.994 & 1.000 & 1.000 \\
  & 0.001 & 0.5 & 623 & 40{,}000  & 0.992 & 0.998 & 0.994 & 1.000 \\
  & 0.001 & 0.5 & 974 & 50{,}000  & 0.968 & 1.000 & 0.972 & 0.990 \\
\midrule
\multirow{5}{*}{$\mathrm{M_{H_{in},\,F_{out}}}$}
  & 0.001 & 0.4 & 416 & 50{,}000  & 0.986 & 0.986 & 0.998 & 0.998 \\
  & 0.001 & 0.4 & 598 & 50{,}000  & 0.976 & 0.986 & 0.988 & 0.998 \\
  & 0.001 & 0.4 & 273 & 200{,}000 & 0.984 & 1.000 & 0.986 & 0.998 \\
  & 0.001 & 0.4 & 891 & 200{,}000 & 1.000 & 1.000 & 1.000 & 1.000 \\
  & 0.001 & 0.4 & 974 & 200{,}000 & 0.986 & 0.988 & 0.996 & 1.000 \\
\midrule
\multirow{5}{*}{$\mathrm{M_H}$}
  & 0.0001 & 0.01 & 891 & 60{,}000  & 0.894 & 0.894 & 0.934 & 0.966 \\
  & 0.0001 & 0.01 & 598 & 140{,}000 & 0.894 & 0.902 & 0.926 & 0.954 \\
  & 0.0001 & 0.1  & 193 & 160{,}000 & 0.878 & 0.880 & 0.902 & 0.952 \\
  & 0.001  & 0.4  & 193 & 170{,}000 & 0.790 & 0.792 & 0.942 & 0.982 \\
  & 0.003  & 0.1  & 416 & 100{,}000 & 0.808 & 0.816 & 0.904 & 0.966 \\
\bottomrule
\end{tabular}
\caption{Checkpoints selected for probing, with exact-match and per-digit
accuracy on the held-out test set. Digit positions are given in
generation order under the little-endian convention: units (U), tens (T),
hundreds (H).}
\label{tab:checkpoints}
\end{table*}

\paragraph{Linear Probe Results.}
Figure~\ref{fig:probing} shows a clear pattern consistent with the minimal computation hypothesis. At Layer 3, models with holistic output tokenization simultaneously encode all three result digit positions; models with fragmented output tokenization encode only the first digit (the units digit under little-endian convention), with subsequent digit positions remaining near chance across all layers.

\begin{figure*}[t]
    \centering
    \begin{tabular}{c cc}
        & \textbf{Fragmented Output} & \textbf{Holistic Output} \\[0.5em]
        \parbox[c]{1em}{\rotatebox{90}{\textbf{Fragmented Input}}} &
        \begin{subfigure}[c]{0.35\linewidth}
            \includegraphics[width=\linewidth]{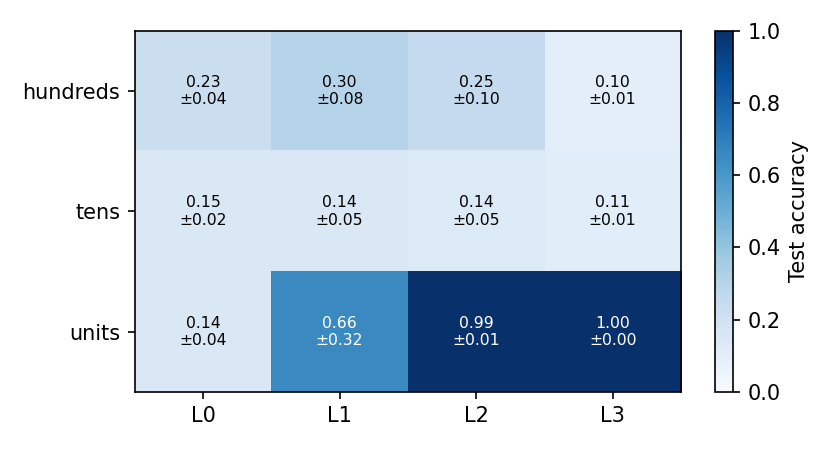
            }
            \caption{$\mathrm{M_F}$}
        \end{subfigure}
        &
        \begin{subfigure}[c]{0.35\linewidth}
            \includegraphics[width=\linewidth]{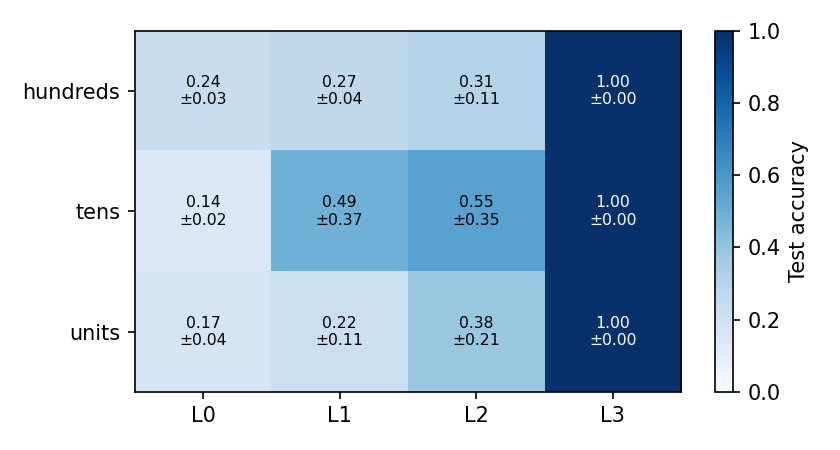
            }
            \caption{$\mathrm{M_{F_{\mathrm{in}},\, H_{\mathrm{out}}}}$}
        \end{subfigure}
        \\[1em]
        \parbox[c]{1em}{\rotatebox{90}{\textbf{Holistic Input}}} &
        \begin{subfigure}[c]{0.35\linewidth}
            \includegraphics[width=\linewidth]{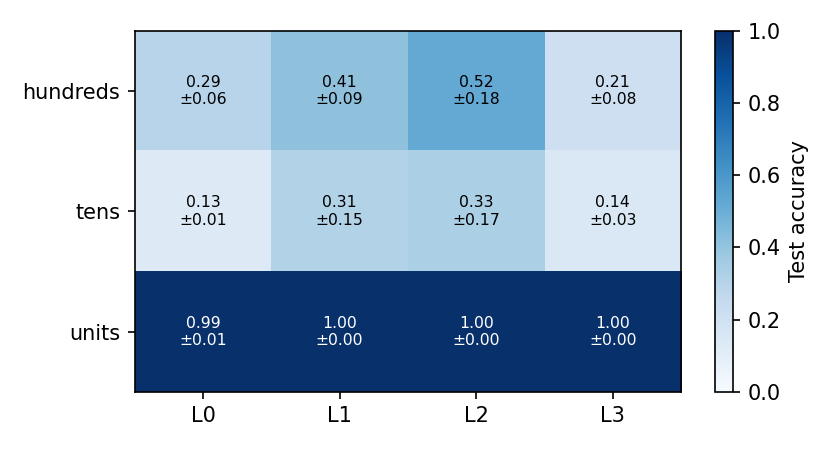
            }
            \caption{$\mathrm{M_{H_{\mathrm{in}},\, F_{\mathrm{out}}}}$}
        \end{subfigure}
        &
        \begin{subfigure}[c]{0.35\linewidth}
            \includegraphics[width=\linewidth]{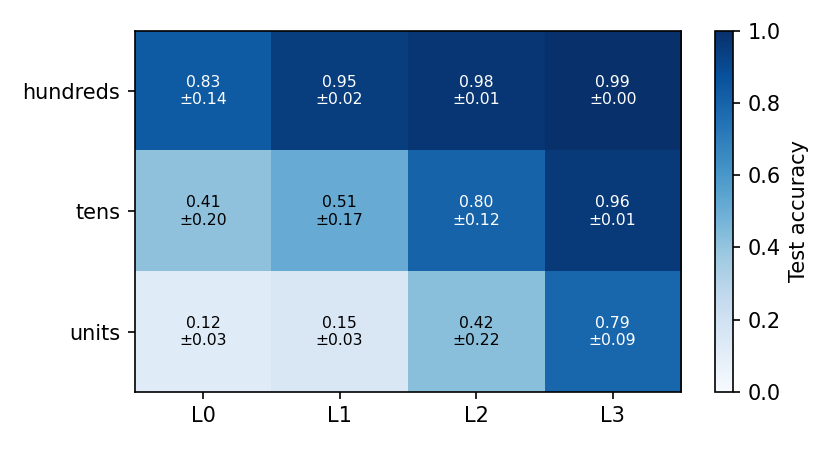
            }
            \caption{$\mathrm{M_H}$}
        \end{subfigure}
    \end{tabular}
    \caption{MLP probe accuracy for individual result digits across layers, in little endian number convention. Values are mean probe accuracy $\pm$ standard deviation across the five probes trained per condition (Table~\ref{tab:checkpoints}).} 
    \label{fig:probing_mlp}
\end{figure*}

\paragraph{MLP Probe Results.} Figure~\ref{fig:probing_mlp} tells the same qualitative story with slightly stronger signal overall. One deviation worth noting is the above-chance accuracy on tens and hundreds digits at layers 1 and 2 for $\mathrm{M_{H_{\mathrm{in}},\, F_{\mathrm{out}}}}$. We attribute this to probe capacity rather than genuine representational content: the pattern does not persist into layer 3, suggesting the probe is partially solving the task itself rather than decoding an internally generated representation. The same pattern is absent with the weaker linear probe, which supports this interpretation and thus does not affect our conclusions.

\begin{figure*}[t]
    \centering
    \begin{tabular}{c cc}
        & \textbf{Fragmented Output} & \textbf{Holistic Output} \\[0.5em]
        \parbox[c]{1em}{\rotatebox{90}{\textbf{Fragmented Input}}} &
        \begin{subfigure}[c]{0.35\linewidth}
            \includegraphics[width=\linewidth]{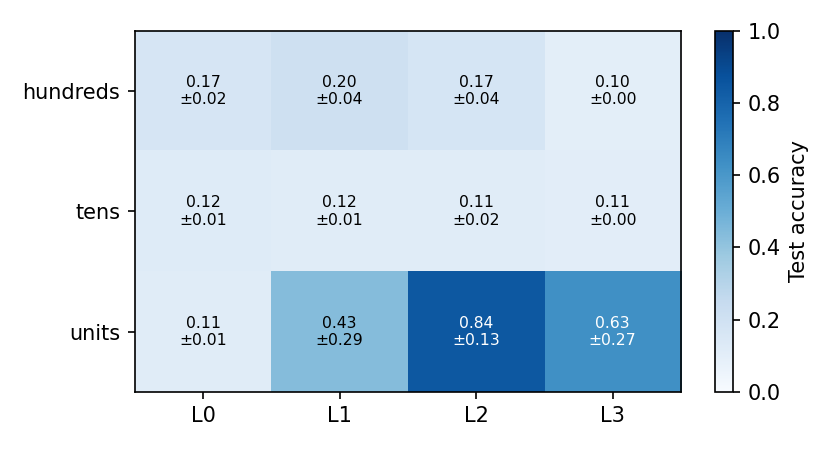}
            \caption{$\mathrm{M_F}$}
        \end{subfigure}
        &
        \begin{subfigure}[c]{0.35\linewidth}
            \includegraphics[width=\linewidth]{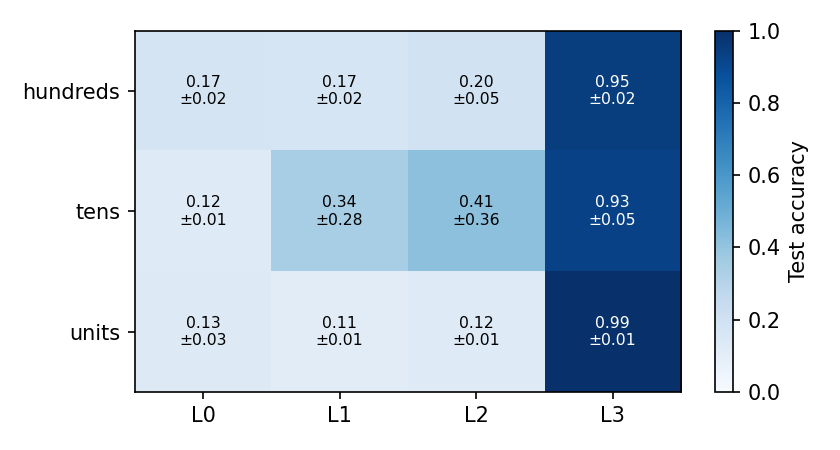}
            \caption{$\mathrm{M_{F_{\mathrm{in}},\, H_{\mathrm{out}}}}$}
        \end{subfigure}
        \\[1em]
        \parbox[c]{1em}{\rotatebox{90}{\textbf{Holistic Input}}} &
        \begin{subfigure}[c]{0.35\linewidth}
            \includegraphics[width=\linewidth]{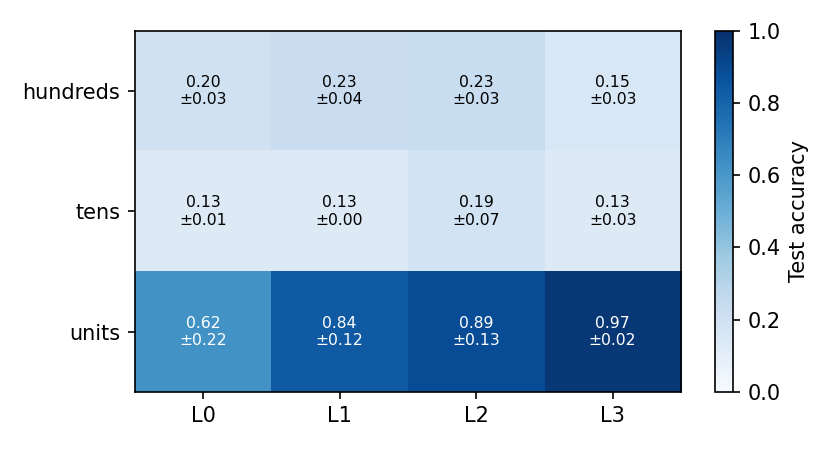}
            \caption{$\mathrm{M_{H_{\mathrm{in}},\, F_{\mathrm{out}}}}$}
        \end{subfigure}
        &
        \begin{subfigure}[c]{0.35\linewidth}
            \includegraphics[width=\linewidth]{
            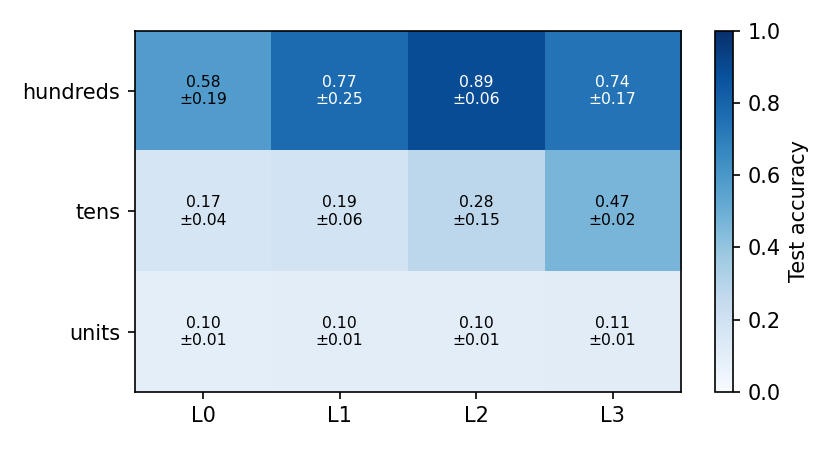 
            }
            \caption{$\mathrm{M_H}$}
        \end{subfigure}
    \end{tabular}
    \caption{Circular probe accuracy for individual result digits across layers, in little endian number convention. Values are mean probe accuracy $\pm$ standard deviation across the five probes trained per condition (Table~\ref{tab:checkpoints}).} 
    \label{fig:probing_circular}
\end{figure*}

\paragraph{Circular Probe Results.} Figure~\ref{fig:probing_circular} is broadly consistent with the linear and MLP results: fragmented output models encode only the units digit, while holistic output models encode all three. One exception is $\mathrm{M_H}$, where the circular probes fail to recover the output digit signal despite the models achieving good task performance. We attribute this to the result digits not being encoded in a circular (modular) format in these models rather than their absence, as the gap between probe accuracy and task accuracy supports this interpretation.

\section{Big-Endian Ablation}
\label{sec:big-endian}

In little-endian digit order the generation order of tokens aligns with the direction of carry propagation, so at every position the information required to emit the current digit is exactly the information the loss at that position supervises. No lookahead into the future is required to solve the task.
We repeat our experiments with standard, big-endian digit order to observe how model internal representation of future result digits changes, when future information has to be resolved in order to correctly solve the task.

\paragraph{Training.} Training is identical to the little-endian setting in every respect except digit order. We select the four highest-accuracy checkpoints per condition for probing (Table~\ref{tab:checkpoints_bigendian}).

\paragraph{Probing.} Probes are trained exactly as described in Appendix~\ref{app:probes}, only under big endian convention the first result digit is labeled as hundreds digit, the second as tens, and the final as units digit.

\begin{figure*}[t]
    \centering
    \begin{tabular}{c cc}
        & \textbf{Fragmented Output} & \textbf{Holistic Output} \\[0.5em]
        \parbox[c]{1em}{\rotatebox{90}{\textbf{Fragmented Input}}} &
        \begin{subfigure}[c]{0.35\linewidth}
            \includegraphics[width=\linewidth]{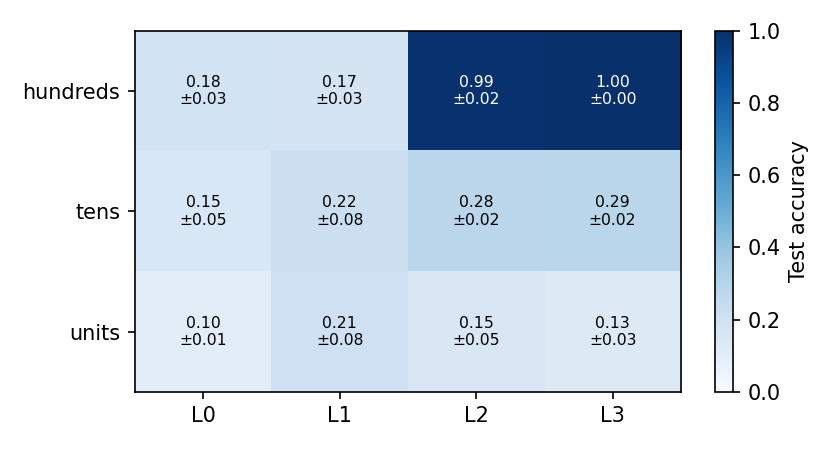}
            \caption{$\mathrm{M_F}$}
        \end{subfigure}
        &
        \begin{subfigure}[c]{0.35\linewidth}
            \includegraphics[width=\linewidth]{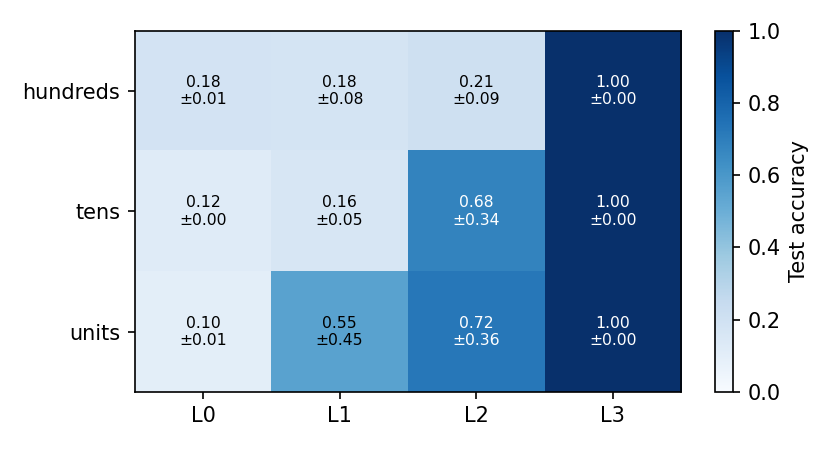}
            \caption{$\mathrm{M_{F_{\mathrm{in}},\, H_{\mathrm{out}}}}$}
        \end{subfigure}
        \\[1em]
        \parbox[c]{1em}{\rotatebox{90}{\textbf{Holistic Input}}} &
        \begin{subfigure}[c]{0.35\linewidth}
            \includegraphics[width=\linewidth]{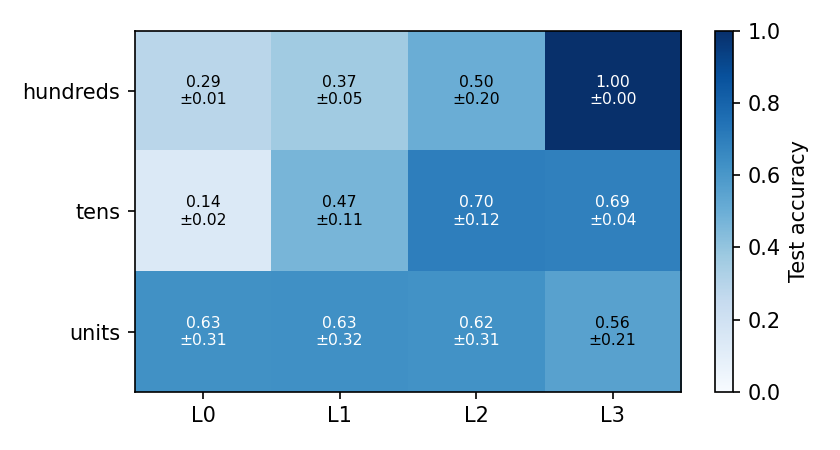}
            \caption{$\mathrm{M_{H_{\mathrm{in}},\, F_{\mathrm{out}}}}$}
        \end{subfigure}
        &
        \begin{subfigure}[c]{0.35\linewidth}
            \includegraphics[width=\linewidth]{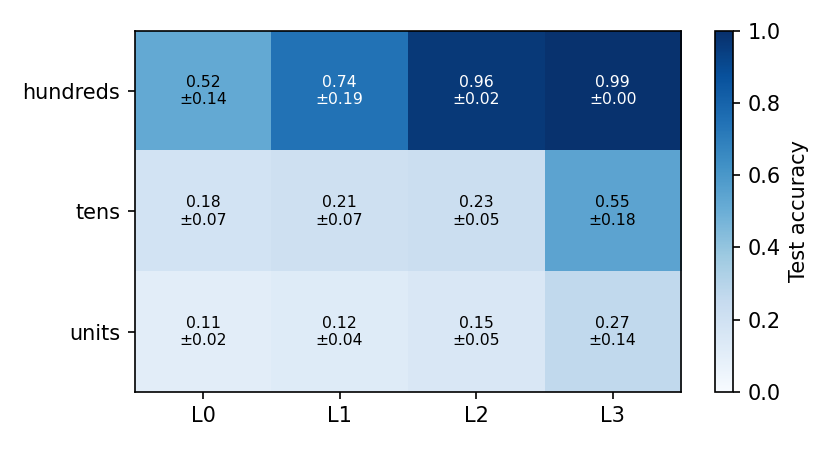}
            \caption{$\mathrm{M_H}$}
        \end{subfigure}
    \end{tabular}
    \caption{Big endian ablation: Linear probe accuracy for individual result digits across layers, in big endian number convention. Values are mean probe accuracy $\pm$ standard deviation across the four probes trained per condition (Table~\ref{tab:checkpoints_bigendian}).} 
    \label{fig:probing_linear_BE}
\end{figure*}

\begin{figure*}[t]
    \centering
    \begin{tabular}{c cc}
        & \textbf{Fragmented Output} & \textbf{Holistic Output} \\[0.5em]
        \parbox[c]{1em}{\rotatebox{90}{\textbf{Fragmented Input}}} &
        \begin{subfigure}[c]{0.35\linewidth}
            \includegraphics[width=\linewidth]{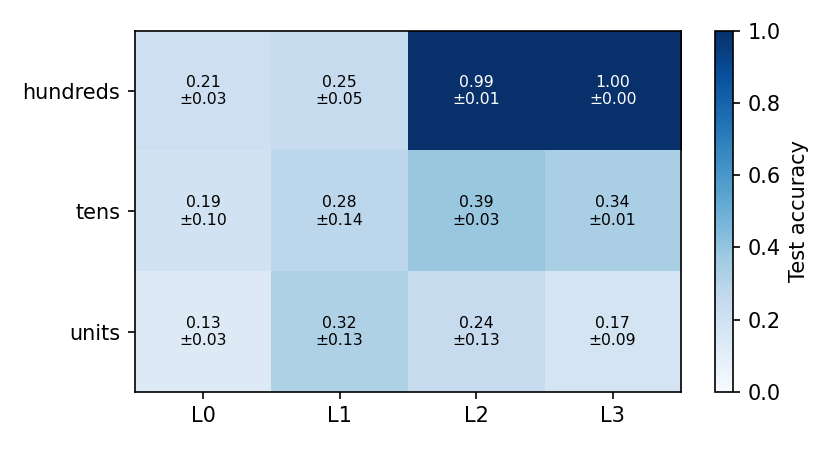}
            \caption{$\mathrm{M_F}$}
        \end{subfigure}
        &
        \begin{subfigure}[c]{0.35\linewidth}
            \includegraphics[width=\linewidth]{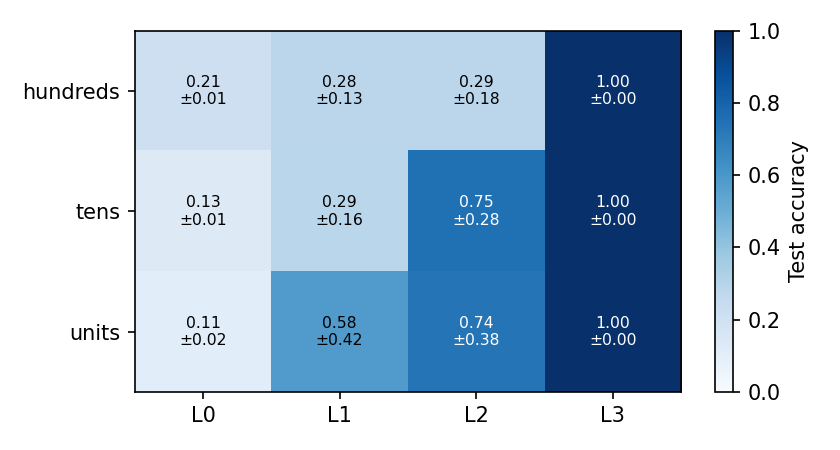}
            \caption{$\mathrm{M_{F_{\mathrm{in}},\, H_{\mathrm{out}}}}$}
        \end{subfigure}
        \\[1em]
        \parbox[c]{1em}{\rotatebox{90}{\textbf{Holistic Input}}} &
        \begin{subfigure}[c]{0.35\linewidth}
            \includegraphics[width=\linewidth]{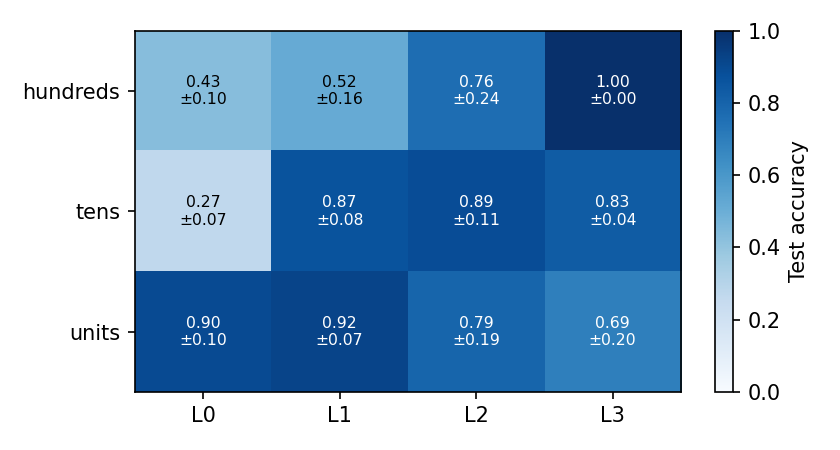}
            \caption{$\mathrm{M_{H_{\mathrm{in}},\, F_{\mathrm{out}}}}$}
        \end{subfigure}
        &
        \begin{subfigure}[c]{0.35\linewidth}
            \includegraphics[width=\linewidth]{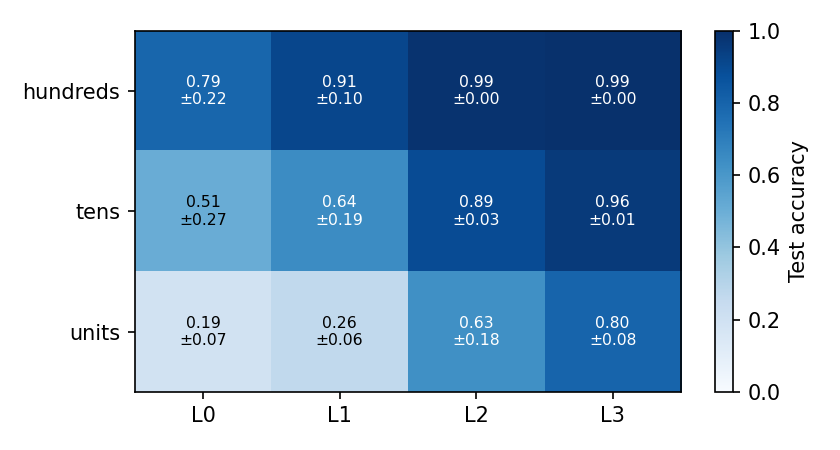}
            \caption{$\mathrm{M_H}$}
        \end{subfigure}
    \end{tabular}
    \caption{Big endian ablation: MLP probe accuracy for individual result digits across layers, in big endian number convention. Values are mean probe accuracy $\pm$ standard deviation across the four probes trained per condition (Table~\ref{tab:checkpoints_bigendian}).} 
    \label{fig:probing_mlp_BE}
\end{figure*}

\begin{figure*}[t]
    \centering
    \begin{tabular}{c cc}
        & \textbf{Fragmented Output} & \textbf{Holistic Output} \\[0.5em]
        \parbox[c]{1em}{\rotatebox{90}{\textbf{Fragmented Input}}} &
        \begin{subfigure}[c]{0.35\linewidth}
            \includegraphics[width=\linewidth]{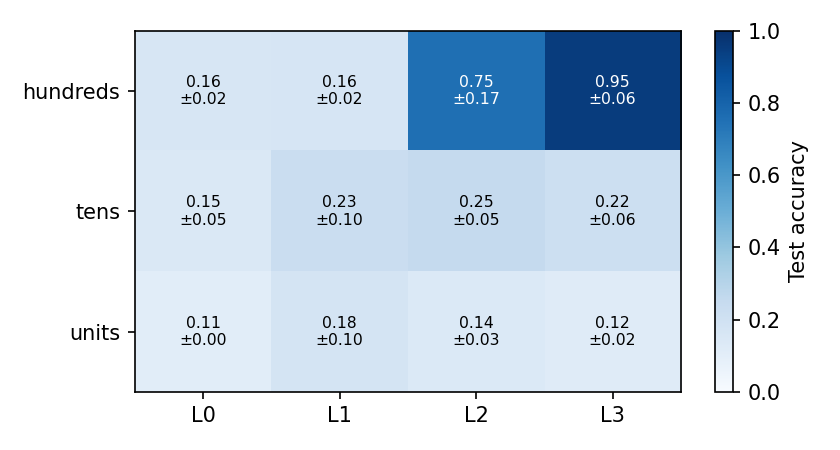}
            \caption{$\mathrm{M_F}$}
        \end{subfigure}
        &
        \begin{subfigure}[c]{0.35\linewidth}
            \includegraphics[width=\linewidth]{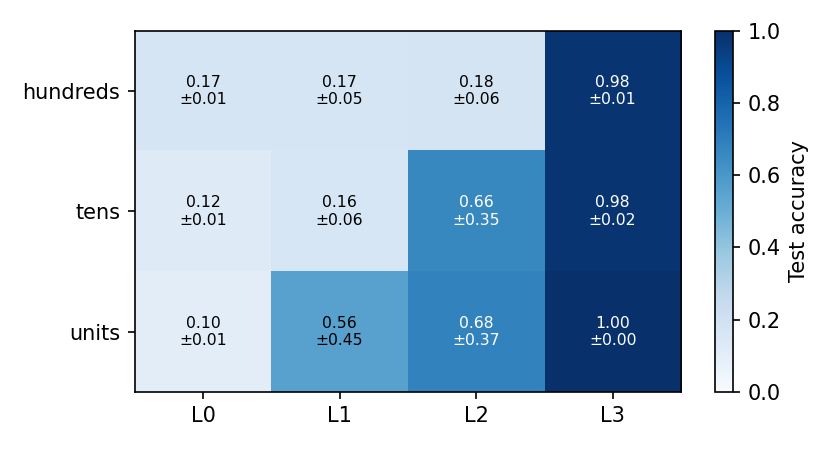}
            \caption{$\mathrm{M_{F_{\mathrm{in}},\, H_{\mathrm{out}}}}$}
        \end{subfigure}
        \\[1em]
        \parbox[c]{1em}{\rotatebox{90}{\textbf{Holistic Input}}} &
        \begin{subfigure}[c]{0.35\linewidth}
            \includegraphics[width=\linewidth]{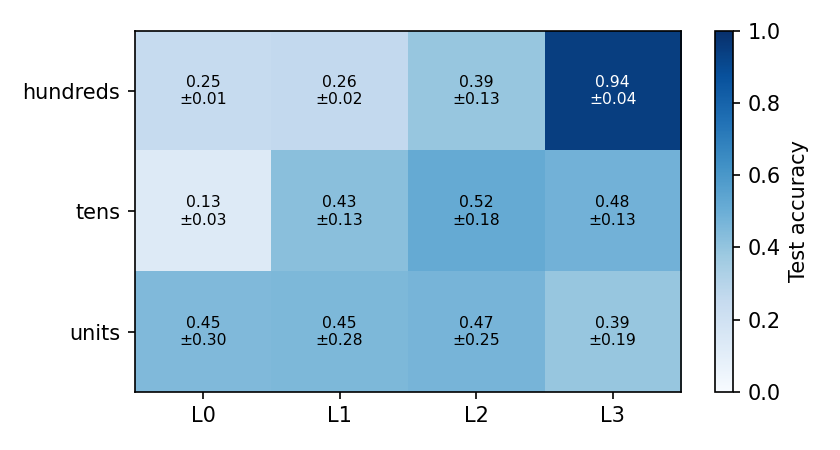}
            \caption{$\mathrm{M_{H_{\mathrm{in}},\, F_{\mathrm{out}}}}$}
        \end{subfigure}
        &
        \begin{subfigure}[c]{0.35\linewidth}
            \includegraphics[width=\linewidth]{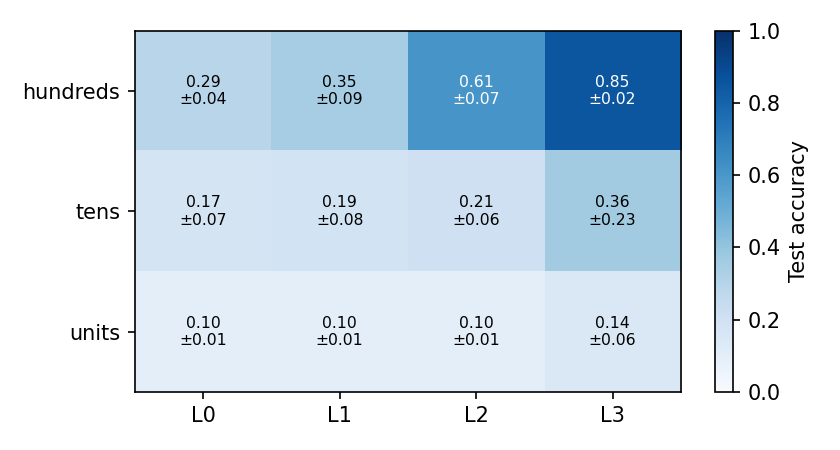}
            \caption{$\mathrm{M_H}$}
        \end{subfigure}
    \end{tabular}
    \caption{Big endian ablation: Circular probe accuracy for individual result digits across layers, in big endian number convention. Values are mean probe accuracy $\pm$ standard deviation across the four probes trained per condition (Table~\ref{tab:checkpoints_bigendian}).} 
    \label{fig:probing_circ_BE}
\end{figure*}

\paragraph{Results.} Figures~\ref{fig:probing_linear_BE}, \ref{fig:probing_mlp_BE}, and \ref{fig:probing_circ_BE} show the digit-wise probing results on models trained under big-endian number convention. As expected, decodability of future result digits is above chance in layer 3 for the fragmented output models, showcasing the effect of the model having to at the very least somewhat resolve future information to accurately solve the task, even in the absence of direct gradient pressure.

Decodability is higher for the immediately following result digit than for the third: a carry from the adjacent position affects the current digit more often than one propagating from two positions away. Even at this very small scale, this is reminiscent of the single-digit lookahead heuristic that production-scale LLMs display~\citep{baeumel-etal-2025-lookahead}.

\begin{table*}[t]
\centering
\small
\begin{tabular}{llrrrr}
\toprule
Condition & lr & wd & Seed & Step & Accuracy \\
\midrule
\multirow{5}{*}{$\mathrm{M_F}$}
  & 0.001 & 0.2 & 123 & 70{,}000  & 1.000 \\
  & 0.001 & 0.2 & 155 & 40{,}000  & 1.000 \\
  & 0.001 & 0.2 & 200 & 130{,}000  & 1.000 \\
  & 0.001 & 0.2 & 832 & 90{,}000  & 1.000 \\
\midrule
\multirow{5}{*}{$\mathrm{M_{F_{in},\,H_{out}}}$}
  & 0.001 & 0.2 & 155 & 150{,}000  & 1.000 \\
  & 0.001 & 0.4 & 456 & 40{,}000  & 0.928 \\
  & 0.001 & 0.4 & 832 & 20{,}000  & 0.942 \\
  & 0.001 & 0.5 & 155 & 40{,}000  & 0.956 \\
\midrule
\multirow{5}{*}{$\mathrm{M_{H_{in},\,F_{out}}}$}
  & 0.001 & 0.2 & 123 & 110{,}000  & 1.000 \\
  & 0.001 & 0.2 & 155 & 140{,}000  & 1.000 \\
  & 0.001 & 0.2 & 456 & 170{,}000  & 1.000 \\
  & 0.001 & 0.2 & 832 & 160{,}000  & 1.000 \\
\midrule
\multirow{5}{*}{$\mathrm{M_H}$}
  & 0.001 & 0.5 & 456 & 90{,}000  & 0.930 \\
  & 0.001 & 0.3 & 155 & 80{,}000  & 0.870 \\
  & 0.001 & 0.3 & 193 & 150{,}000  & 0.900 \\
  & 0.001 & 0.3 & 273 & 130{,}000  & 0.896 \\
\bottomrule
\end{tabular}
\caption{Big-endian model checkpoints with highest task accuracy on held-out test set. Used in probing.}
\label{tab:checkpoints_bigendian}
\end{table*}

\section{Survey Details}
\label{app:survey}

This appendix documents the methodology and full results of the survey reported in Section~\ref{sec:survey}.

\subsection{Method}
\paragraph{Paper retrieval.}
We extract all papers published to the main conference or findings tracks of ACL, EMNLP, NAACL, and EACL in 2024--2025 whose titles contain the substrings \textit{math}, \textit{numer} (as in \textit{numeric}), \textit{number}, or \textit{arithmetic}.

\paragraph{Topical filtering.}
We manually inspect each paper for topical fit and apply the following inclusion and exclusion criteria.

\smallskip\noindent\textbf{Included:} papers whose primary experimental contribution involves LLMs being tasked directly with an arithmetic or numeric reasoning problem, specifically:
\begin{itemize}
  \item papers that propose a benchmark or dataset for a math-related ability;
  \item papers that present methods for generating math problems or response
        options;
  \item papers that propose a framework or training method for improving LLM
        performance on math-related tasks;
  \item papers that introduce a new model targeting math-related tasks.
\end{itemize}

The overarching criterion is: \textit{does the paper contain experimental results in which LLMs are tasked with an arithmetic or numeric reasoning problem, and would different tokenization strategies possibly affect those results?}

\smallskip\noindent\textbf{Excluded:}
\begin{itemize}
  \item papers that are not related to arithmetic or numeric reasoning at all, but where the keyword search gave a false positive (e.g., work on task arithmetic)
  \item papers where the math task has no arithmetic component
        (e.g., purely logical or symbolic tasks);
  \item papers whose primary focus is LLMs acting as an intermediate step for
        another tool (e.g., converting problems to code or formal language
        before solving them);
  \item papers focused on pedagogy or educational applications rather than
        on measuring reasoning ability;
  \item papers where the primary task is multimodal in a way that does not
        centre on numeric reasoning ability
        (e.g., converting handwritten equations to text);
  \item papers on numerical \emph{understanding} tasks
        (e.g., numerical claim verification, numerical NLI) where the
        tokenization of integers in isolation would not affect the result.
\end{itemize}

When a paper was borderline, we classified it as not fitting. This filtering leaves \textbf{120 papers} about mathematical or numeric reasoning \cite{002,
gambardella_language_2024, 
006,007,008,009,010,011,012,013,014,015,016,017,018,019,020,021,022,023,027,
xu_principled_2025, 
032,
feng_how_2025, 
034,035,036,037,038,039,040,042,043,044,050,051,052,053,054,057,059,060,062,066,067,068,069,070,071,072,073,074,077,
fei_advancing_2025, 
li_exposing_2025, 
082,083,087,089,091,092,093,094,095,096,097,098,100,
xie_adversarial_2024, 
102,103,104,105,107,109,110,111,
schwartz_numerologic_2024, 
zhou-etal-2024-scaling, 
bertolazzi_validation_2025, 
sun_probing_2025, 
118,119,120,122,123,124,127,
mamidanna_all_2025, 
131,134,136,138,139,140,141,144,145,150,151,152,153,154,155,158,159,161,166,171,173,174,176,177,178,181,185,188,190,191}.

\paragraph{Tokenization annotation.}
For each of the 120 papers we (i) record all model families evaluated, (ii) search the full text for the substring \textit{token}, and (iii) manually inspect every occurrence to determine whether the distinction between digit-level and multi-digit numeric tokenization is explicitly mentioned and whether any cross-model comparison accounts for that distinction.

\paragraph{Model categorisation.}
For open-source models with a publicly available tokenizer file on HuggingFace, we inspect the vocabulary to identify the largest integer encoded as a single token. We classify a model as using multi-digit tokenization if multi-digit integers are single tokens (e.g.\ up to 999), and as using digit-level tokenization if only single digits 0-9 receive dedicated tokens.
For proprietary or closed-source models whose tokenizer is not publicly documented, we do not assign a category. A paper is classified as \emph{comparing across supervision regimes} if at least one evaluated model uses holistic tokenization and at least one uses fragmented tokenization, based on the identifiable models.
Table~\ref{tab:tokenizer_families} lists the tokenization type for the most commonly evaluated model families.

\begin{table}[ht]
\centering
\small
\begin{tabular}{ll}
\toprule
\textbf{Model family} & \textbf{Numeric tokenization} \\
\midrule
GPT-2              & Multi-Digit Tokens \\
GPT-J              & Multi-Digit Tokens \\
GPT-NeoX           & Multi-Digit Tokens \\
GPT-3.5 Turbo      & Multi-Digit Tokens \\
Pythia             & Multi-Digit Tokens \\
Llama 3 / 3.1 / 3.2 & Multi-Digit Tokens \\
OLMo 2             & Multi-Digit Tokens \\
DeepSeek-R1        & Multi-Digit Tokens \\
Phi-4              & Multi-Digit Tokens \\
\midrule
Llama 2 / Llama-7B & Single-Digit Tokens \\
Mistral            & Single-Digit Tokens \\
Gemma / Gemma 2 / Gemma 3 & Single-Digit Tokens \\
Qwen / Qwen 2 / Qwen 2.5 & Single-Digit Tokens \\
Qwen2-VL           & Single-Digit Tokens \\
CodeLlama          & Single-Digit Tokens \\
Deepseek-Coder-6.7B & Single-Digit Tokens \\
DeepSeek-Math-7B   & Single-Digit Tokens \\
Phi-3              & Single-Digit Tokens \\
Llemma             & Single-Digit Tokens \\
\bottomrule
\end{tabular}
\caption{Numeric tokenization type for commonly evaluated model families.
Classification is based on inspection of publicly available tokenizer
vocabularies on HuggingFace. Proprietary models are omitted.}
\label{tab:tokenizer_families}
\end{table}

\subsection{Results}

Of the 120 retained papers, \textbf{11 (9.2\%)} explicitly mention the numeric tokenization strategy of the models they evaluate \cite{mamidanna_all_2025, sun_probing_2025, bertolazzi_validation_2025, zhou-etal-2024-scaling, schwartz_numerologic_2024, xie_adversarial_2024, li_exposing_2025, fei_advancing_2025, feng_how_2025, xu_principled_2025, gambardella_language_2024}.
The remaining \textbf{109} make no mention of numeric tokenization granularity.

Of the 109 papers that do not mention tokenization:
\begin{itemize}
  \item \textbf{83} actively compare models from both holistic and fragmented tokenization families without flagging the difference or controlling for it, treating models operating under structurally different supervision regimes as interchangeable baselines;
  \item \textbf{20} evaluate models from only one tokenization type;
  \item \textbf{6} cannot be resolved because at least one evaluated model is closed-source or uses an undisclosed tokenizer.
\end{itemize}

Of the 11 papers that do mention tokenization, \textbf{none} frames the distinction in terms of output supervision. When tokenization is mentioned at all, it is consistently treated as an input representation choice. Three of the eleven explicitly compare models with different tokenization schemes and discuss implications for comparison; four restrict evaluation to one tokenization type while citing it as a methodological choice; two work with models trained from scratch, so tokenization is discussed as part of the experimental design; and two mention tokenization only in the context of related work or motivation, without evaluating its effect.

\paragraph{Important caveat.}
We do not claim that the supervision-regime confound is decisive in every one of the 83 papers that compare across tokenization types. We claim that its potential impact \emph{cannot be assessed} from the published record, because tokenization strategy is not reported. The practical consequence ranges from negligible to decisive depending on the specific comparison and task, but without reporting it is impossible to know which.

\end{document}